\documentclass[singlecolumn,10pt,twoside,final]{asme2ej_PKMComplexLimbs}
\usepackage{amsfonts}
\usepackage{graphicx}
\usepackage{amsfonts}
\usepackage{graphicx}
\usepackage{amsmath, amssymb, bm}
\usepackage[fixamsmath,disallowspaces]{mathtools}
\usepackage{xcolor}
\usepackage{fixmath}
\usepackage{url}

\newtheorem{remark}{Remark}
\newtheorem{definition}{Definition}

\newtheorem{example}{Example}

\newtheorem{assumption}{Assumption}

\begin{document}

\pagestyle{empty}

\parindent 0pt \parskip 2pt \setcounter{topnumber}{9} %
\setcounter{bottomnumber}{9} \renewcommand{\textfraction}{0.00001}

\renewcommand {\floatpagefraction}{0.999} \renewcommand{\textfraction}{0.01} %
\renewcommand{\topfraction}{0.999} \renewcommand{\bottomfraction}{0.99} %
\renewcommand{\floatpagefraction}{0.99} \setcounter{totalnumber}{9}

\input{opening_PKMComplexLoops_Kinematics_final.inp}

\section{Introduction}

Many of the successfully applied PKM posses limbs with kinematic loops,
which will be referred to as \emph{complex limbs}. The best known example is
the Delta robot introduced by Clavel \cite{Clavel1988,Pierrot1990}, which
generates a Sch\"{o}nflies motion (also called SCARA motion \cite%
{AngelesBook}). The Delta has exactly one kinematic loop per limb. Other PKM
were proposed whose limbs possess several loops, which are called
fundamental cycles (FCs). Almost all such PKM possess complex multi-loop
limbs where the loops are arranged in series within the limbs, i.e. the FC
share at most one common link. Such limbs are called \emph{hybrid}. From a
modeling and analysis perspective this allows treating the loops
independently so that the loop constraints can be solved separately. The
concept of 'locally' solving loop constraints was proposed in \cite{JainBook}
for general multibody system as \emph{constraint embedding}. A similar
approach will be employed in this paper for PKM modeling.

The main feature of Delta robots is that each limb resembles a 4-bar linkage
that is hinged at the base and the moving platform. The parallelogram(s)
within a limb are often regarded as a kinematic joint referred to as $\Pi $%
-joint \cite{AngelesBook} (notice that this compound joint contributes its
own inherent dynamics, which must be accounted for by the dynamics
formulation). Various PKM whose limbs comprise parallelogram linkages were
reported in the literature. A PKM with two parallelogram limbs generating Sch%
\"{o}nflies motions was proposed in \cite%
{Gauthier_JMR2008,TaghvaeipourAngelesLessard2013}, and another Sch\"{o}%
nflies motion PKM with four limbs in \cite{Altuzarra2009}. The Orthoglide, a
3-DOF translational PKM, was proposed in \cite%
{ChablatWengerARK2000,ChablatWengerTRO2003}. Another 3-DOF example is the
CaPaMan robot \cite{Wolf_MMT2004}. A kinematically redundant PKM involving
parallelogram loops was presented in \cite{Wen-TRO2020}. A redundantly
actuated planar PKM with two parallelogram limbs intended as machine tool
was proposed in \cite{Wu-MMT2009}. Parallelogram limbs were used in \cite%
{Zou2020,ShenZhaoLiWuChablat2021} as building blocks for synthesizing
spatial 3-DOF translational PKM actuated with prismatic joints. A planar
3-DOF PKM with a hybrid arrangement of parallelograms is the NaVaRo robot 
\cite{Klimchik2018}. The Par2 robot \cite%
{Par2_Baradat2009,Par2_MECCANICA2011} is an example for a 2-DOF Delta
variant. The 4-DOF H4 robot \cite{PierrotCompany1999} and the Par4 \cite%
{Pierrot2009} are modifications of the Delta where the end-effector rotation
is controlled by an articulated platform. The relevance of a (conceptually
simple) parallelogram arrangement within the limbs is documented by patents
such as \cite{Pierrot-Patent2011} on a 2-DOF PKM and \cite%
{Pierrot-Patent2003} on a 4-DOF Sch\"{o}nflies motion PKM, in addition to
the original patent by Clavel \cite{ClavelPatent}. To exploit the full
spectrum of PKM with complex limbs, however, other robots were proposed. The
IRSBot-2 \cite{Germain2013} is a 2-DOF PKM where each limbs comprises a
planar parallelogram as well as a loop with four U joints. To increase the
workspace, in \cite%
{RakotomangaChablatCaro2008,ChablatRolland2018,Klimchik2018} a 3-DOF
translational PKM was presented whose limbs are constructed from scissor
mechanisms, which leads to a multi-loop complex limb. The systematic design
and modeling of PKM with complex limbs is a relatively new field of
research, and more general designs with closed loops, other than simple
parallelogram linkages, will yield novel PKM. This paper shall provide a
basis for modeling such general PKM.

The type synthesis of PKM with complex limbs is more involved than it is for
simple limbs due to the much larger varieties of closed loop mechanisms that
can generate the platform motion. A synthesis method for 2-DOF PKM
comprising complex limbs with two loops was reported in \cite%
{NurahmiCaro2016}, but a dedicated approach for PKM with complex limbs has
not yet been proposed. In principle, the general method based on linear
transformations \cite{Gogu_EJM2004,Gogu_Book1} and the synthesis based on
the virtual chain concept \cite{KongGosselinBook2007} or screw system based
approaches \cite{HuangLiDing2013} can be applied. The Lie group methods
based on displacement groups should applicable. It could in particular lead
to a modular synthesis method by synthesizing and combining loops with
certain motion space, as it was pursued in \cite{HuangLiDing2013} based on
instantaneous screw systems.

The dynamics modeling is an integral part of the design and analysis process
of PKM in order to exploit their acceleration and load capacity. There are
many publications that addressed the derivation of dynamic EOM for specific
PKM with simple limbs, e.g. \cite%
{LeeShah1988,NakamuraGhodoussi1989,MataProvenzanoCuadradoValero2002,Abedloo2014,TRORedPKM,Dasgupta1998,WangGosselin1998,ZhaoTRO2014,EOMRedCoord1_NLD2012,EOMRedCoord2_RAS2012}%
. The EOM for Delta robots were derived in \cite%
{Pierrot1990,MillerClavel1992,GuglielmettiLongchamp1994,Brinker_ROMANSY2016,Brinker_RAL2017}
starting from the kinetic and potential energy and then analytically
deriving the Lagrange equations. To make this tractable, simplifying
assumptions are made. In \cite{Pierrot1990,ChemoriPierrot2018}, for
instance, the inertia of the lower arm parallelogram is split and is
distributed to the upper arm and the platform, respectively. Lagrange
equations were derived analytically in \cite{Altuzarra2009} to formulate the
dynamic EOM in terms of actuator coordinates for a 4-DOF Sch\"{o}nflies
motion PKM with parallelogram joints. Aiming at general PKM with simple
limbs without the need for model simplifications, systematic modeling
approaches were proposed in \cite%
{AngelesBook,KhalilIbrahim2007,BriotKhalilBook,AbdellatifHeimann_MMT2009,MuellerAMR2020}%
. These methods have in common that they take into account the special
kinematic structure of PKM. To this end, the kinematics and dynamics model
of the individual limbs, and of the platform, are derived and are used to
formulate the overall EOM by imposing the loop closure conditions. The
differences are found mainly in the kinematics modeling. Key element is the
forward and inverse kinematics of the individual limbs. The latter refers to
the relation of the motion of all joints of the limb and the platform
motion, which is used to express the PKM motion in terms of taskspace
coordinates. It should be mentioned that the bond graph modeling technique
was applied to PKM with simple and complex limbs in \cite%
{VjekoslavCohodar2015,Wu-MMT2018}.

This paper presents a systematic generally applicable modeling approach for
rigid body PKM with complex limbs, following the basic concept of the
formulation for PKM with simple limbs in \cite%
{BriotKhalilBook,MuellerAMR2020}. The modularity is exploited for modeling
the kinematics and dynamics of the (possibly structurally identical) limbs.
The approach is independent of the particular formulation used for
kinematics and dynamics modeling. Its application is described in detail
when using the screw and Lie group formulation, which can be directly
related to the formulation in \cite{AngelesBook,MuellerAMR2020} for PKM with
simple limbs. Special emphasize is given to the computational aspects, and
its applicability for parallel computation is explored. The main steps for
deriving the task space formulation of the EOM can be summarized as follows:%
%TCIMACRO{\TeXButton{TeX field}{\vspace{-1ex}}}%
%BeginExpansion
\vspace{-1ex}%
%EndExpansion

\begin{enumerate}
\item Analyze PKM topology, and partition into separate subgraphs
representing the limbs (sec. \ref{secTopology}). Identify FCs, and construct
a tree-topology system (sec. \ref{secTreeTop}).

\item Express the forward kinematics of the tree-topology system (sec. \ref%
{secTreeKin})

\item Separately solve the velocity constraints of each FC in all limbs
(sec. \ref{secCutJoint},\ref{secCutBody})

\item Incorporate the constraint solution into the forward kinematics of the
tree-topology system (sec. \ref{secVelFKLimb}-\ref{secGeomFKLimb})

\item Solve the inverse kinematics of the mechanism (sec. \ref{secInvKinMech}%
).

\item Pursue the kinematics and dynamics modeling as for the PKM with simple
limbs (sec. \ref{secEOM})%
%TCIMACRO{\TeXButton{TeX field}{\vspace{-1ex}}}%
%BeginExpansion
\vspace{-1ex}%
%EndExpansion
\end{enumerate}

The paper is organized as follows. The graph representation of kinematic
topology, its partitioning into subgraphs corresponding to the limbs, and
the introduction of a tree-topology kinematics are recalled in sec. \ref%
{secTopology}. The kinematics modeling of the tree-topology system is
presented in sec. \ref{secTreeKin} in terms of joint screw coordinates and
the product of exponentials (POE). The constraints for kinematic loops
within the limbs are presented in sec. \ref{secCutJoint} and \ref{secCutBody}%
. In sec. \ref{secCutJoint} the cut-joint formulation is presented as the
method of choice, which is widely used in MBS dynamics. The cut-body
formulation is recalled in sec. \ref{secCutBody} for completeness only as it
allows application of the reciprocal screw method to formulate the
constraint Jacobian, which is an established approach to linkage analysis.
It does, however, not directly lead to an efficient implementation. A
formulation for the inverse kinematics of the mechanism is presented in sec. %
\ref{secInvKinMech}, and for the inverse kinematic of the manipulator in
sec. \ref{secInvKin}. The formulation in sec. \ref{secInvKinMech} is then
employed for deriving the dynamic EOM in sec. \ref{secEOM}. A task space
formulation is presented in sec. \ref{secEOMTaskSpace}, and a formulation in
actuator coordinates is derived in sec. \ref{secEOMActCoord}. The inherent
modularity of PKM is exploited in sec. \ref{secModularModeling} for a
modular modeling approach. Applications of the dynamics model are discussed
in sec. \ref{secApplications}. The system of EOM applicable to time
integration (to solve the forward dynamics) is presented in sec. \ref%
{secAppForwDyn}. The inverse dynamics formulation is summarized in \ref%
{secAppInvDyn} and its use for parallel computation is discussed. The method
is demonstrated in detail for the 3\underline{R}R[2RR]R Delta robot in sec. %
\ref{AppendixDelta}. The paper closes with a short conclusion and outlook in
sec. \ref{secConclusion}. For completeness, appendix \ref{AppendixRigMotion}
summarizes the Lie group formulation of linkage kinematics in terms of joint
screws, which is then used in appendix \ref{AppendixEOMTree} to briefly
summarize the Lie group formulation of EOM. The notation and list of symbols
is summarized in appendix \ref{secSymbols}.%
%TCIMACRO{\TeXButton{TeX field}{\vspace{-3ex}}}%
%BeginExpansion
\vspace{-3ex}%
%EndExpansion

\section{Kinematic Topology%
%TCIMACRO{\TeXButton{secTopology}{\label{secTopology}}}%
%BeginExpansion
\label{secTopology}%
%EndExpansion
}

The topological graph, denoted $\Gamma $, describes the arrangement of
bodies and joints \cite{TopologyRobotica,JainMUBO2011_1,WittenburgBook}.
Vertices represent bodies, and edges represent joints. Edges of $\Gamma $
represent joints with general DOF, but are often used to represent 1-DOF
joints, which are used to model multi-DOF joints. The topological graph
possesses $\gamma $ fundamental cycles (FC), i.e. topologically independent
loops.%
%TCIMACRO{\TeXButton{TeX field}{\vspace{-1ex}}}%
%BeginExpansion
\vspace{-1ex}%
%EndExpansion

\begin{definition}
%TCIMACRO{\TeXButton{rm}{\rm}}%
%BeginExpansion
\rm%
%EndExpansion
The system of bodies and joints that corresponds to a connected component of 
$\Gamma $, which connects the ground vertex with the vertex of the moving
platform, is called a \emph{limb}. The subgraph corresponding to limb $%
l=1,\ldots ,L$ is denoted with $\Gamma _{\left( l\right) }$, where $L$ is
the total number of limbs.%
%TCIMACRO{\TeXButton{TeX field}{\vspace{-1ex}}}%
%BeginExpansion
\vspace{-1ex}%
%EndExpansion
\end{definition}

The number of independent kinematic loops of limb $l$, i.e. the number of
FCs of $\Gamma _{\left( l\right) }$, is denoted with $\gamma _{l}$. The FCs
corresponding to the kinematic loops of limb $l$ are denoted with $\Lambda
_{\lambda \left( l\right) },\lambda =1,\ldots ,\gamma _{l}$.%
%TCIMACRO{\TeXButton{TeX field}{\vspace{-1ex}}}%
%BeginExpansion
\vspace{-1ex}%
%EndExpansion

\begin{definition}
%TCIMACRO{\TeXButton{rm}{\rm}}%
%BeginExpansion
\rm%
%EndExpansion
A limb is called \emph{simple} if and only if it is a simple (i.e. serial)
kinematic chain. It is called \emph{complex} if and only if it comprises
kinematic loops. If any two FCs of the topological graph $\Gamma _{\left(
l\right) }$ have at most one common vertex (and are hence edge-disjoint),
the limb is called \emph{hybrid}. Hybrid limbs are also called
serial-parallel limbs as the kinematic loops are arranged in series within
the limb.%
%TCIMACRO{\TeXButton{TeX field}{\vspace{-1ex}}}%
%BeginExpansion
\vspace{-1ex}%
%EndExpansion
\end{definition}

The following assumption \ref{Assumption1} holds true for almost all PKM
with complex limbs.

\begin{assumption}
%TCIMACRO{\TeXButton{Assumption1}{\label{Assumption1}}}%
%BeginExpansion
\label{Assumption1}%
%EndExpansion
It is assumed in the following that the PKM possess hybrid limbs only.%
%TCIMACRO{\TeXButton{TeX field}{\vspace{-1ex}}}%
%BeginExpansion
\vspace{-1ex}%
%EndExpansion
\end{assumption}

\begin{example}[{3\protect\underline{R}R[2RR]R Delta}]
The Delta robot is arguably the best-known and most successful PKM with
complex limbs. The kinematic design of the Delta robot, as reported in the
patent \cite{ClavelPatent}, contains revolute joints only, as shown in fig. %
\ref{figRR2RRRDelta}a). The four revolute joints 3,4,5,7, forming the
parallelogram loop within a limb, resemble a planar 4-bar linkage (see fig. %
\ref{figRR2RRRDelta}b). The axes of joints 3 and 5 intersect the axis of
joint 2, and the axes of joints 4 and 7 intersect the axis of joint 6. A
similar design was reported in \cite{TsaiStamper1996,TsaiBook1999} with the
difference that the axes do not intersect. Both designs are denoted with 3%
\underline{R}R[2RR]R, which indicates that each of the three limbs comprises
an actuated R joint (joint 1) followed by a passive R joint (joint 2), which
connects to the loop formed by parallel arrangement of two RR chains (joints
3 and 4 respectively 5 and 7), indicated by the bracket, that is connected
to the platform by another R joint (joint 6). Frequently, this design is
simply denoted 3\underline{R}UU as the R joints with intersecting axes
kinematically function as U joints (neglecting bodies 2 and 4). The
topological graph is shown in fig. \ref{figRR2RRRDeltaGraph}a). Each of the $%
L=3$ limbs possesses $\gamma _{l}=1$ FC, as shown in fig. \ref%
{figRR2RRRDeltaGraph}b). 
\begin{figure}[h]
\vspace{-4ex} 
\centerline{
a)~\includegraphics[height=6.3cm]{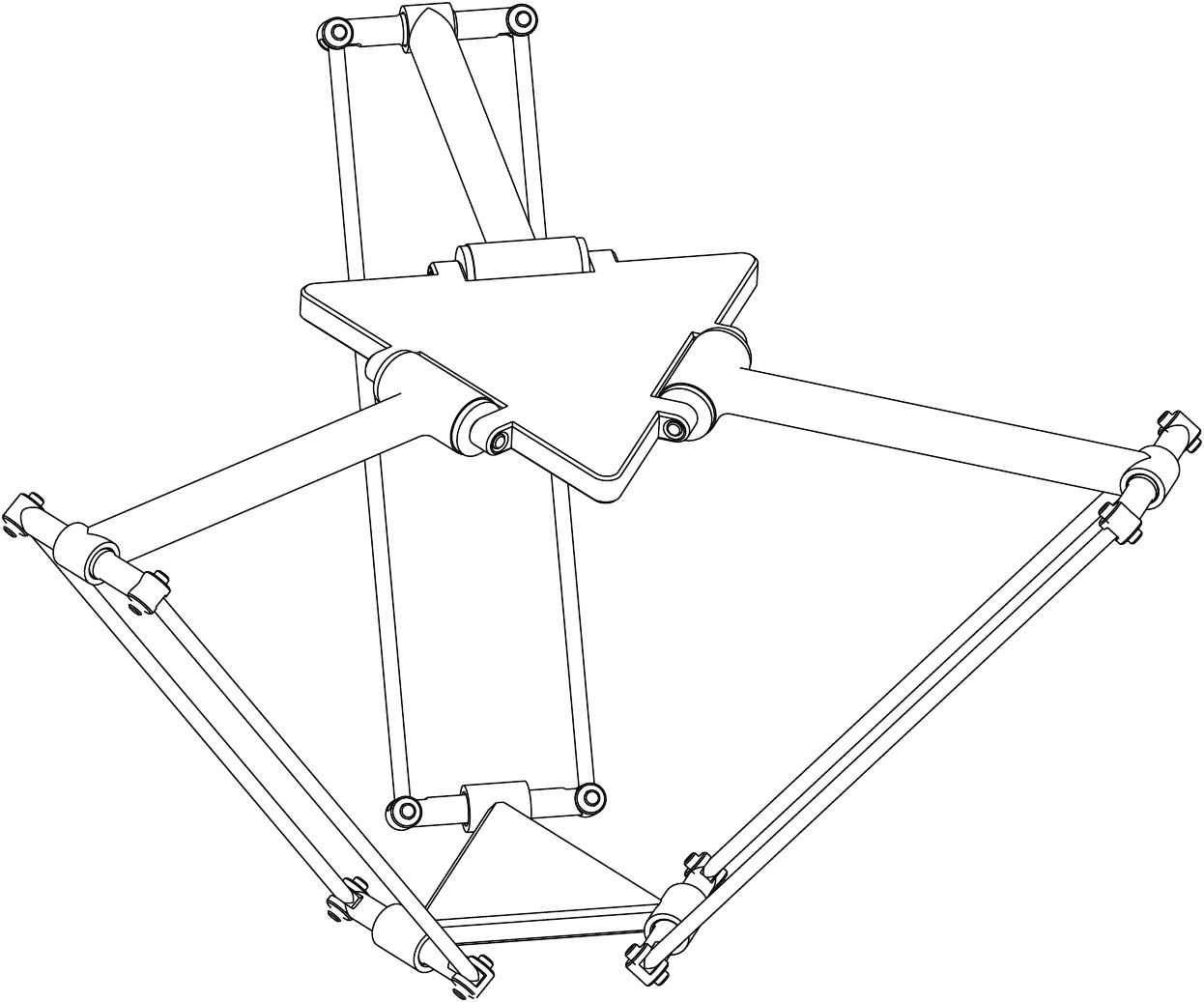}\hspace{8ex} b)~\includegraphics[height=6.9cm]{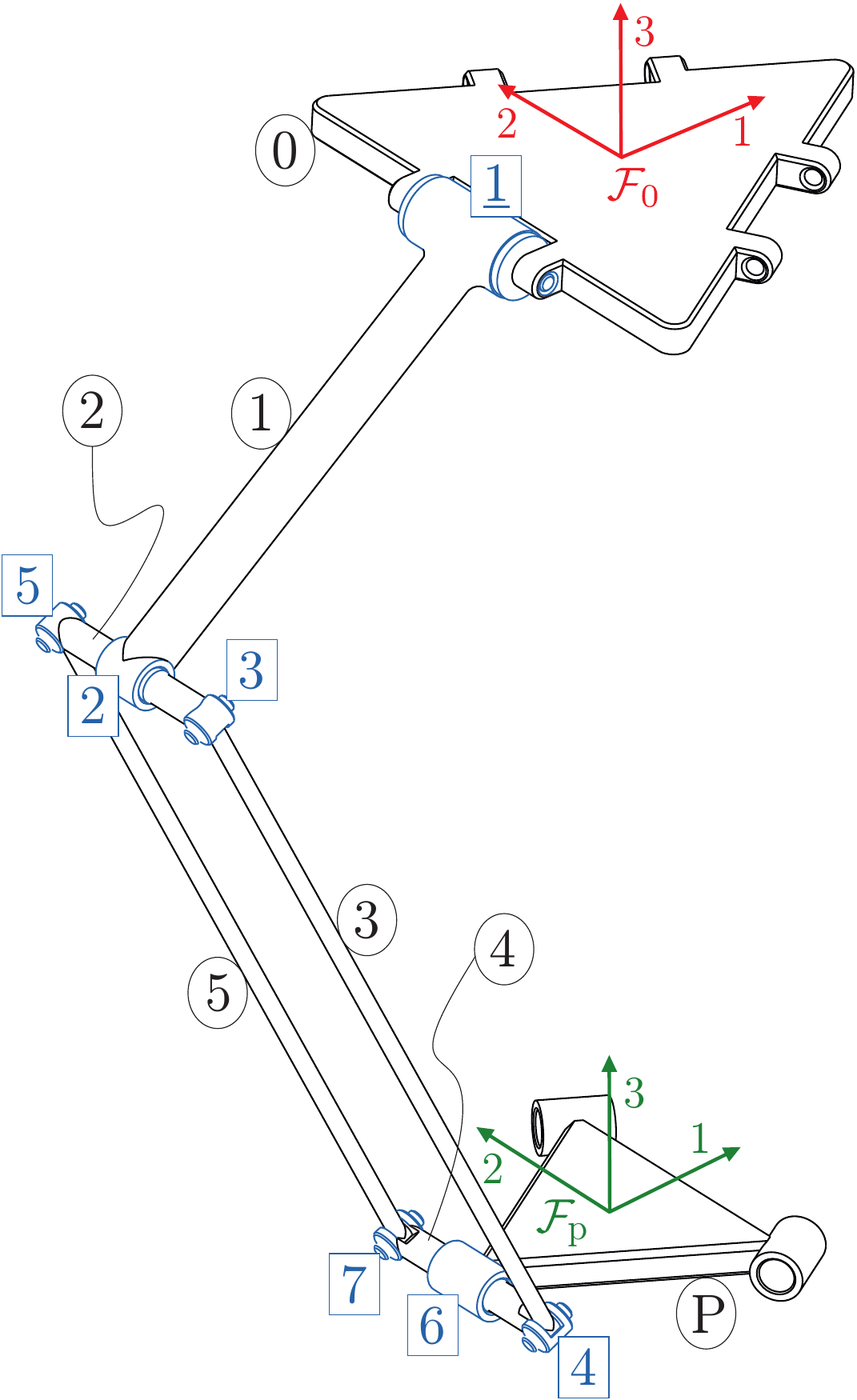}
}
\caption{a) Drawing of a 3\protect\underline{R}R[2RR]R Delta robot. b)
Representative limb of this Delta robot}
\label{figRR2RRRDelta}
\end{figure}
\vspace{-7ex} 
\begin{figure}[tbh]
\centerline{a)~\includegraphics[height=5.5cm]{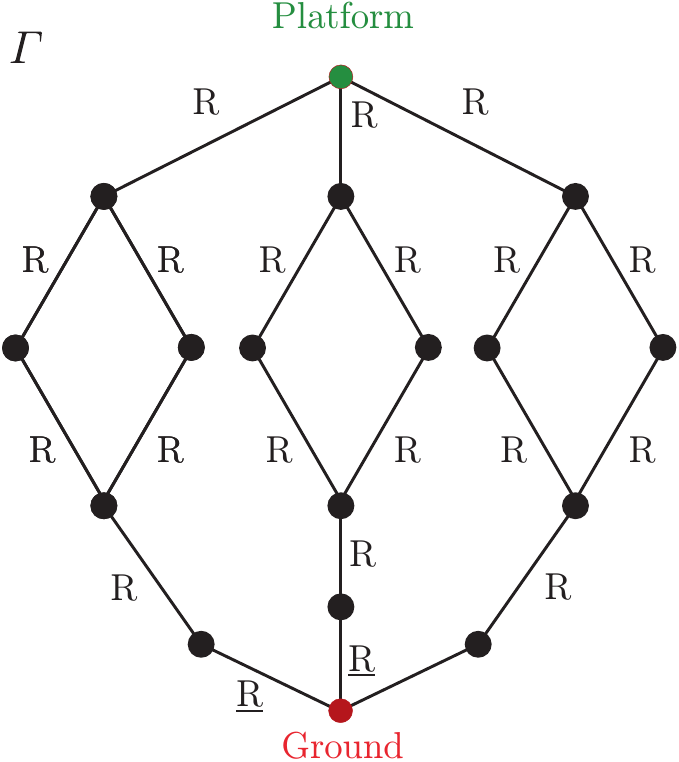}\ \ \ \ \ \ b)~\includegraphics[height=5.5cm]{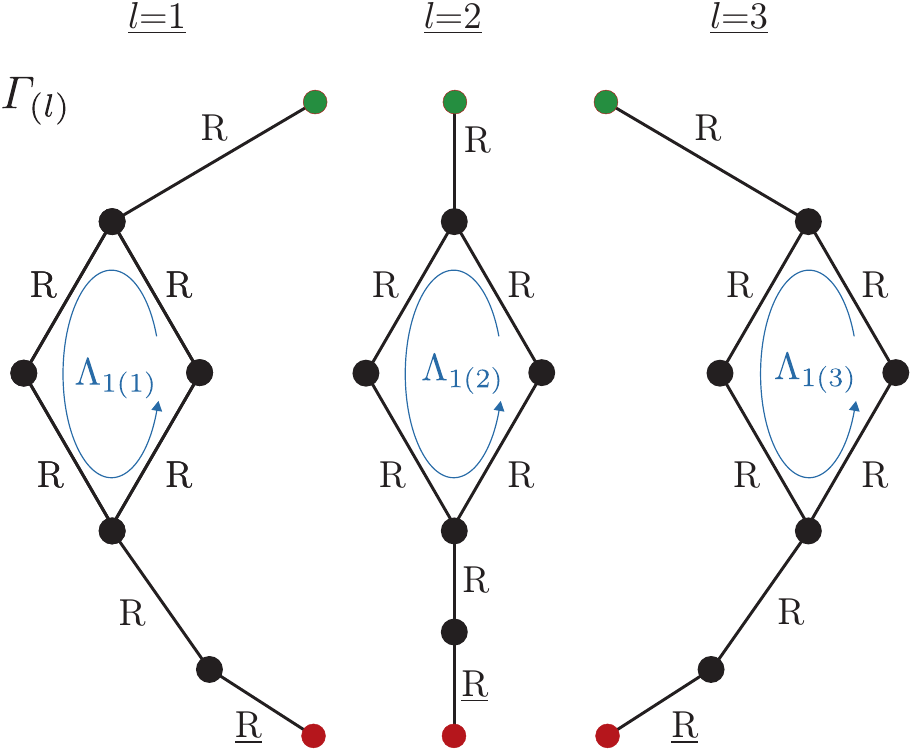}
}
\caption{a) Topological graph of the 3\protect\underline{R}R[2RR]R Delta. b)
Subgraphs $\Gamma _{\left( l\right) }$ representing the $L=3$ limbs of this
Delta.}
\label{figRR2RRRDeltaGraph}
\end{figure}
\end{example}

\begin{example}[{3\protect\underline{R}[2US] Delta}]
From a practical perspective, it is difficult to ensure that the axes of the
R joints forming the 4-bar are parallel. Moreover, as a spatial mechanism,
the parallelogram in the 3\underline{R}[2SS] is overconstrained. In an
alternative design, which is adopted in most commercial implementations of
the Delta concept, spherical joints are used at both ends of the rods to
connected them directly to platform and base, respectively. This is
generally considered to be the original Delta design \cite%
{Clavel1988,Pierrot1990}. It is referred to as 3\underline{R}[2SS] design
(often simply denoted as 3\underline{R}SS). This kinematics would allow
spinning of the rods about their longitudinal axes. In order to avoid this
spinning (or rather slipping), most Delta robots are equipped with pinned
braces (e.g. Yaskawa and Motoman Delta robot has braces at both ends), which
restricts the S joints to function as U joints. Thus this Delta design
becomes a 3\underline{R}[2UU] kinematics (often simply denoted as 3%
\underline{R}UU), where the axes of the two U joints that are fixed to the
ground are parallel, and so are those fixed to the platform. The loop formed
by the four U joints would again be overconstrained, but the braces are
introduced to only restrain the bars rather than to geometrically constraint
them. A model that describes this kinematics is to use S joints at the
platform and U joints to connect the rods to the articulated arm. This model
will be referred to as 3\underline{R}[2US] design in the following. Its
topological graph is shown in fig. \ref{figR2USDeltaGraph}. The
parallelogram loop is now formed by the two U and the two S joints. 
\begin{figure}[tbh]
\centerline{
a)~\includegraphics[height=5.2cm]{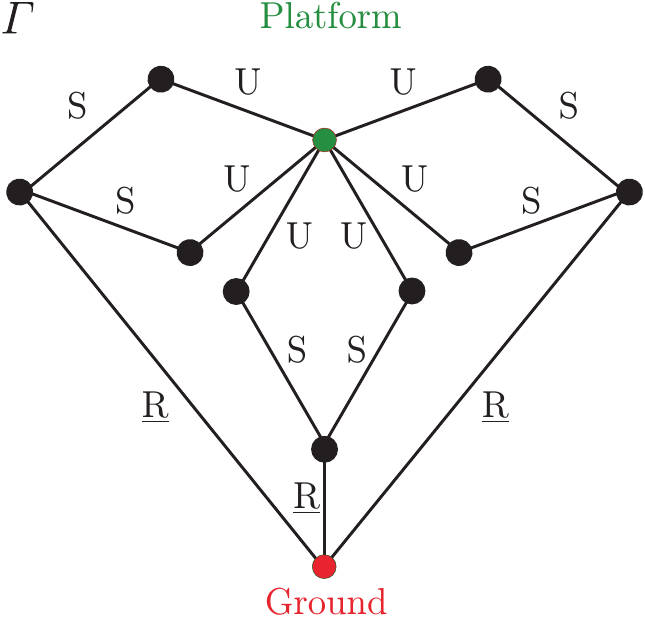}\ \ \ \ \ \ b)~\includegraphics[height=5.2cm]{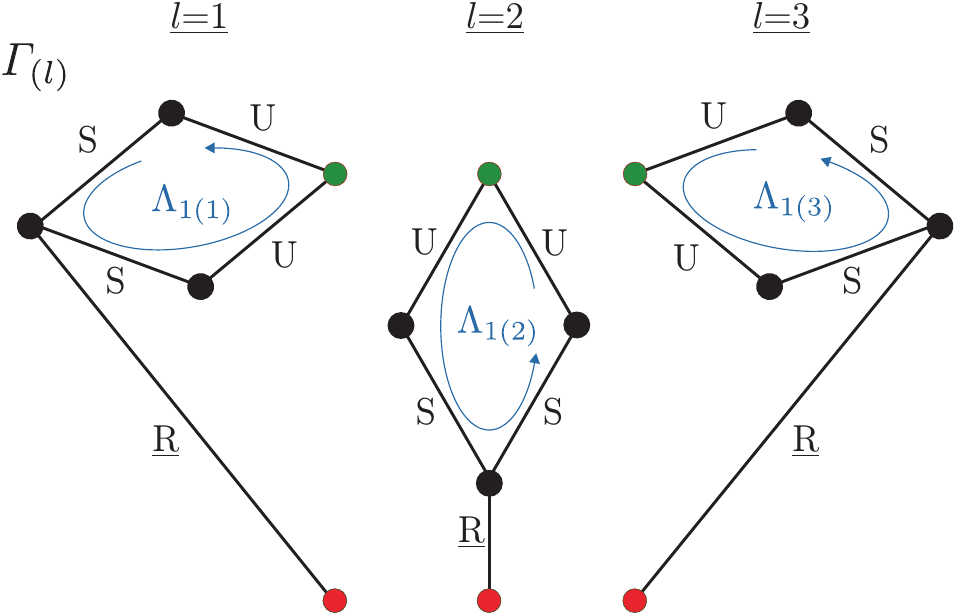}
}
\caption{a) Topological graph of the 3\protect\underline{R}[2US] Delta
robot. b) Subgraphs $\Gamma _{\left( l\right) }$ representing the $L=3$
limbs of this design.}
\label{figR2USDeltaGraph}
\end{figure}
%TCIMACRO{\TeXButton{clearpage}{\clearpage}}%
%BeginExpansion
\clearpage%
%EndExpansion
\end{example}

\begin{example}[IRSBot-2]
Fig. \ref{figIRSBot} shows a sketch of the IRSBot-2, which was presented in 
\cite{Germain2013}, and fig. \ref{figIRSBotGraph} shows its topological
graph. Each of the $L=2$ limbs possesses $\gamma _{l}=2$ kinematic loops, $%
\Lambda _{1\left( l\right) }$ and $\Lambda _{2\left( l\right) }$. Due to the
serial arrangement of these FCs, the limbs possess a hybrid topology. 
\begin{figure}[h]
\centerline{a)\includegraphics[height=5.6cm]{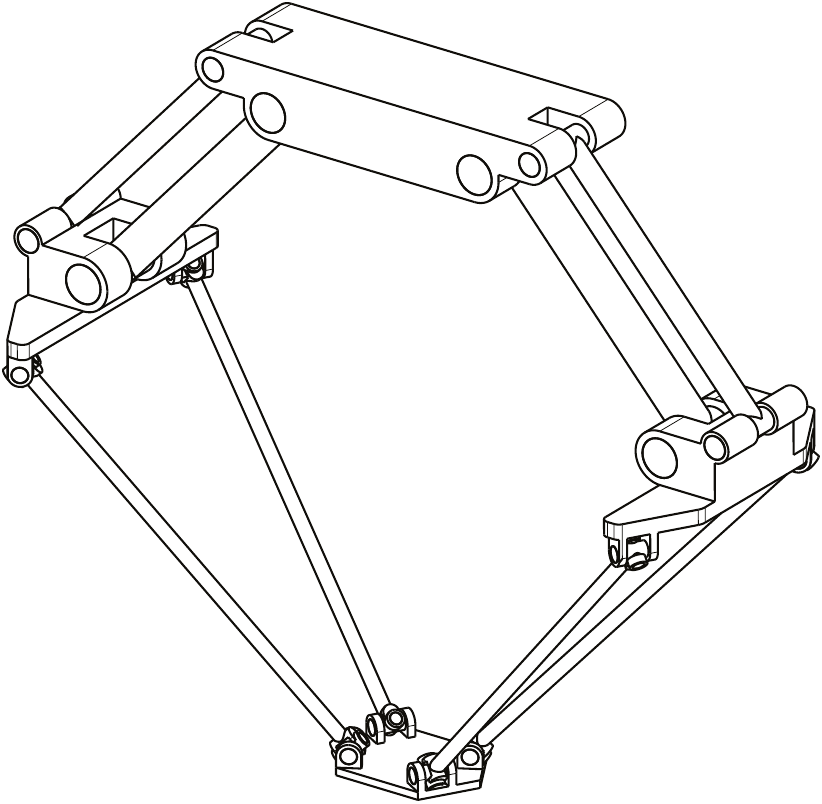}
\ \ \ \ \ \ \ \ \ \ \ \
b)\includegraphics[height=5.6cm]{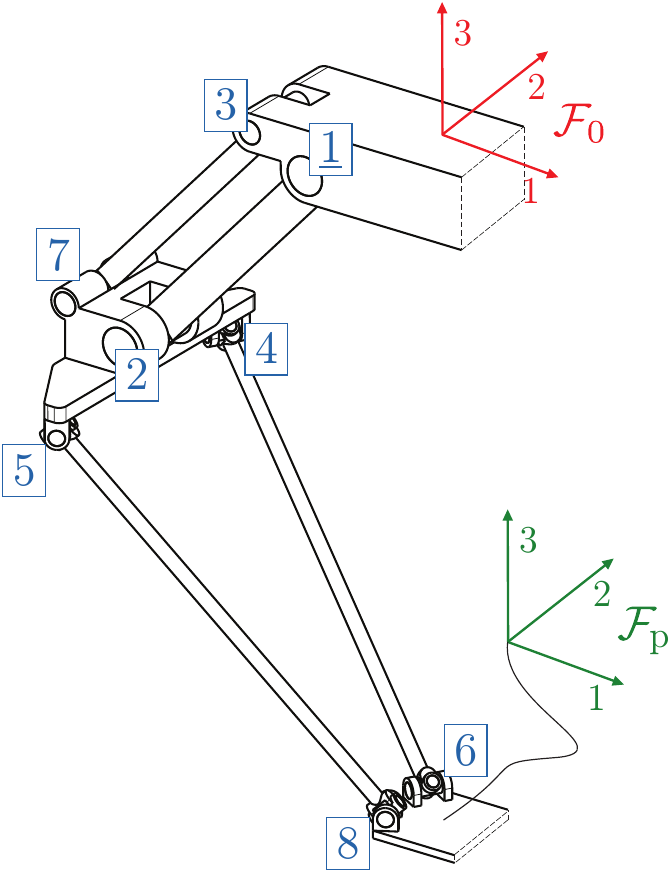}
}
\caption{a) Drawing of the IRSBot-2 presented in \protect\cite{Germain2013}
(courtesy of S\'{e}bastien Briot, Laboratoire des Sciences du Num\'{e}rique
de Nantes). b) Representative limb comprising two loops.}
\label{figIRSBot}
\end{figure}
\begin{figure}[h]
\centerline{
a)~\includegraphics[height=5.6cm]{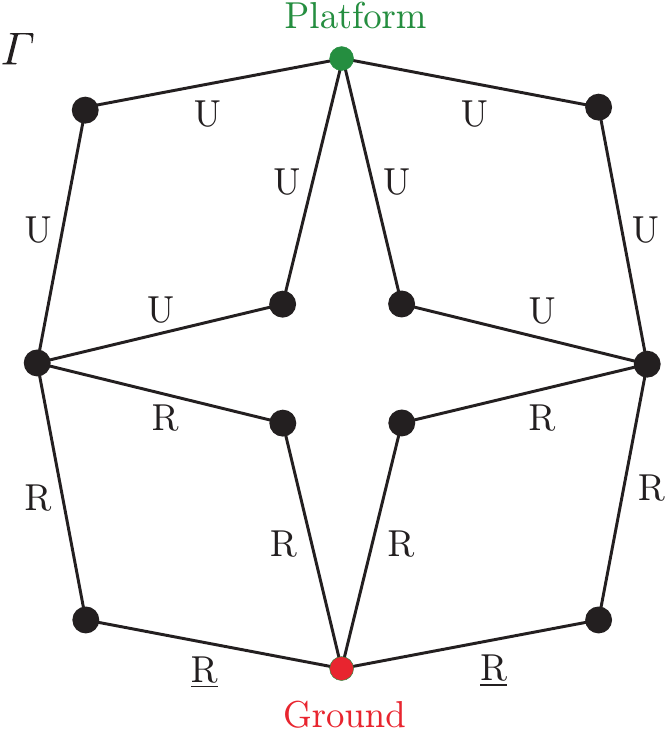}\ \ \ \ \ \ \ \ \ \ \ \ b)~\includegraphics[height=5.6cm]{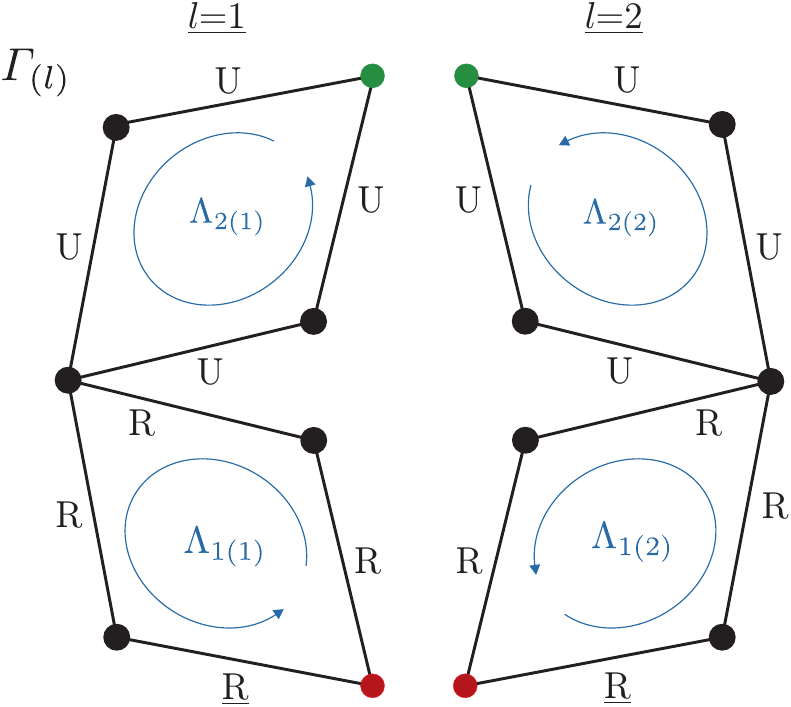}
}
\caption{a) Topological graph of the IRSBot-2. b) Subgraphs $\Gamma _{\left(
l\right) }$ representing the $L=2$ limbs. The hybrid topology of the limbs
is clearly visible as $\Lambda _{1\left( l\right) }$ and $\Lambda _{2\left(
l\right) }$ are arranged in series.}
\label{figIRSBotGraph}
\end{figure}
\end{example}

The topological graph $\Gamma _{\left( l\right) }$ of limb $l$ possesses $%
\gamma _{l}$ fundamental cycles itself. Further, any two limbs, being
connected to the platform, form a FC. In total the PKM hence comprises $%
\gamma =L\cdot \gamma _{l}+L-1=L\left( \gamma _{l}+1\right) -1$ FCs
(assuming identical limbs). Indeed, for PKM with simple limbs there are $L-1$
FCs \cite{MuellerAMR2020}. For structurally identical limbs, the subgraphs
are congruent. Following the standard approach of modeling multibody systems
in terms of relative coordinates (joint variable), a system of loop closure
constraints would be introduced for each of these $\gamma $ FCs. A tailored
formulation for such PKM can be introduced exploiting the special topology
as presented in this paper.

\section{Forward Kinematics of a Complex Limb}

The forward kinematics encompasses the determination of the motion of all
bodies of the limb, including the platform, for given motion of a set of
independent joints. For a simple limb, when separated from the PKM, all
joints can move independently, whereas for a PKM with complex limbs, the
joints must satisfy certain loop constraints. Explicit solution of the
geometric loop constraints in closed form is possible for particular PKM
only. This is therefore not assumed in the following. Instead, the solution
of the velocity constraints will be used to formulate the differential
kinematics limb, and for solving the geometric constraints.%
%TCIMACRO{\TeXButton{TeX field}{\vspace{-3ex}}}%
%BeginExpansion
\vspace{-3ex}%
%EndExpansion

\subsection{Associated Tree-Topology System, Graph Labeling%
%TCIMACRO{\TeXButton{secTreeTop}{\label{secTreeTop}}}%
%BeginExpansion
\label{secTreeTop}%
%EndExpansion
}

Vertices (bodies) of the topological graph $\Gamma _{\left( l\right) }$ of
limb $l$ are numbered with $i=0,1,2,\ldots \mathfrak{n}_{l}$, where index 0
refers to the ground, and the platform is labeled with P. Edges (joints) are
indexed with $i=1,\ldots ,\mathfrak{N}_{l}$. $\Gamma _{\left( l\right) }$
possesses $\gamma _{l}$ fundamental cycles. A spanning tree $G_{\left(
l\right) }$ on $\Gamma _{\left( l\right) }$ is obtained by removing exactly
one edge (cut-edge) of each FC, which defines a tree-topology system
comprising $\mathfrak{n}_{l}$ moving bodies and $\mathfrak{n}_{l}$ joints
(tree-edges). There is a unique path in $G_{\left( l\right) }$ from any
vertex (body) to the root (ground). A \emph{ground-directed spanning tree} $%
\vec{G}_{\left( l\right) }$ is then introduced by directing all edges of $%
G_{\left( l\right) }$ so to point toward the ground within this path. In the
so constructed $\vec{G}_{\left( l\right) }$, there is a unique directed path
from any vertex (moving body) to the ground. In particular, there is a path
from platform to ground, and the platform motion is determined by the motion
of the corresponding kinematic chain. The latter must indeed respect the
loop constraints imposed by the kinematic loops.

The root-directed tree $\vec{G}_{\left( l\right) }$ induces a partial order
of bodies: Body $j$ is a \emph{predecessor} of $i$ if $j$ comes after $i$ in
the directed path from $i$ to 0. This is denoted with $j\prec _{l}i$ (or
simply with $j\prec i$ if is clear that it refers to limb $l$). The direct
predecessor $j$ of $i$ is indicated with $j=i-_{l}1$ (or simply $j=i-1$).
The tree is \emph{canonical} if $j\prec _{l}i$ implies $j<i$. For sake of
simplicity, the $\mathfrak{n}_{l}$ tree-edges are numbered with body indices 
$i=1,\ldots ,\mathfrak{n}_{l}$, so that the tree-edge connecting vertex $i$
with its predecessor is labeled with $i$. Cut-edges are thus indexed with $i=%
\mathfrak{n}_{l}+1,\ldots ,\mathfrak{N}_{l}$.

\begin{definition}
%TCIMACRO{\TeXButton{rm}{\rm}}%
%BeginExpansion
\rm%
%EndExpansion
Joints that correspond to edges in $G_{\left( l\right) }$ are called \emph{%
tree-joints} of the limb. Joints that correspond to the cut-edges in the
co-tree $\Gamma _{\left( l\right) }\backslash G_{\left( l\right) }$ are
called \emph{cut-joints}.
\end{definition}

Body (and thus joint) indices can be taken arbitrarily from the index set $%
\left\{ 1,\ldots ,\mathfrak{N}_{l}\right\} $. However, to aid the matrix
formulation of the kinematics and the EOM (appendix \ref{AppendixRigMotion}%
), the following assumption is made.

\begin{assumption}
%TCIMACRO{\TeXButton{assCanonicalTree}{\label{assCanonicalTree}}}%
%BeginExpansion
\label{assCanonicalTree}%
%EndExpansion
The body indices of each limb are assigned so that the spanning tree $\vec{G}%
_{\left( l\right) }$ is canonical \cite{JainMUBO2011_1,JainBook}.
\end{assumption}

Denote with $%
%TCIMACRO{\TeXButton{red}{}}%
%BeginExpansion
%
%EndExpansion
%TCIMACRO{\TeXButton{eta}{\mathbold{\eta}}}%
%BeginExpansion
\mathbold{\eta}%
%EndExpansion
_{\left( l\right) }%
%TCIMACRO{\TeXButton{black}{\color{black}}}%
%BeginExpansion
\color{black}%
%EndExpansion
:=(\vartheta _{1\left( l\right) },\ldots ,\vartheta _{N_{l}\left( l\right)
})^{T}\in {\mathbb{V}}^{N_{l}}$ the overall vector of $N_{l}$ joint
variables of limb $l$, and with $%
%TCIMACRO{\TeXButton{red}{}}%
%BeginExpansion
%
%EndExpansion
%TCIMACRO{\TeXButton{vartheta}{\mathbold{\vartheta}}}%
%BeginExpansion
\mathbold{\vartheta}%
%EndExpansion
_{\left( l\right) }%
%TCIMACRO{\TeXButton{black}{\color{black}}}%
%BeginExpansion
\color{black}%
%EndExpansion
:=(\vartheta _{1\left( l\right) },\ldots ,\vartheta _{n_{l}\left( l\right)
})\in {\mathbb{V}}^{n_{l}}$ the vector of $n_{l}$ tree-joint variables (i.e.
vector $%
%TCIMACRO{\TeXButton{red}{}}%
%BeginExpansion
%
%EndExpansion
%TCIMACRO{\TeXButton{eta}{\mathbold{\eta}}}%
%BeginExpansion
\mathbold{\eta}%
%EndExpansion
_{\left( l\right) }%
%TCIMACRO{\TeXButton{black}{\color{black}}}%
%BeginExpansion
\color{black}%
%EndExpansion
$ with cut-joint variables removed). If the PKM comprises 1-DOF joints only,
then $N_{l}=\mathfrak{N}_{l}$ and $n_{l}=\mathfrak{n}_{l}$. A general PKM
comprises multi-DOF joints, so that $N_{l}\leq \mathfrak{N}_{l}$ and $%
n_{l}\leq \mathfrak{n}_{l}$.

\begin{example}[{3\protect\underline{R}R[2RR]R Delta --cont.}]
%TCIMACRO{\TeXButton{TeX field}{\label{exDeltaVariables}}}%
%BeginExpansion
\label{exDeltaVariables}%
%EndExpansion
Fig. \ref{figDeltaLimbGraph}a) shows the labeled topological graph of limb $%
l $. Removing, for instance, edge 7 from the FC $\Lambda _{1\left( l\right) }
$ yields the directed spanning tree shown in fig. \ref{figDeltaLimbGraph}b).
Joint 7 is the cut-joint of this FC. The path from platform to ground is
shown in blue color. The platform motion is thus determined by joints
1,2,3,4, and 6. The seven 1-DOF R joints give rise to the vector of $N_{l}=7$
joint variables $%
%TCIMACRO{\TeXButton{red}{}}%
%BeginExpansion
%
%EndExpansion
%TCIMACRO{\TeXButton{eta}{\mathbold{\eta}}}%
%BeginExpansion
\mathbold{\eta}%
%EndExpansion
_{\left( l\right) }%
%TCIMACRO{\TeXButton{black}{\color{black}}}%
%BeginExpansion
\color{black}%
%EndExpansion
=\left( \vartheta _{1\left( l\right) },\ldots ,\vartheta _{7\left( l\right)
}\right) ^{T}$. The vector of $n_{l}=6$ tree-joint variables is $%
%TCIMACRO{\TeXButton{red}{}}%
%BeginExpansion
%
%EndExpansion
%TCIMACRO{\TeXButton{vartheta}{\mathbold{\vartheta}}}%
%BeginExpansion
\mathbold{\vartheta}%
%EndExpansion
_{\left( l\right) }%
%TCIMACRO{\TeXButton{black}{\color{black}}}%
%BeginExpansion
\color{black}%
%EndExpansion
=\left( \vartheta _{1\left( l\right) },\ldots ,\vartheta _{6\left( l\right)
}\right) ^{T}$. 
\begin{figure}[h]
\caption{a) Labeled subgraph $\Gamma _{\left( l\right) }$ of a limb of the 3%
\protect\underline{R}R[2RR]R Delta ($\mathfrak{N}_{l}=7$), and b) directed
spanning tree $\vec{G}_{\left( l\right) }$ ($\mathfrak{n}_{l}=6$) obtained
by removing edge 5 from the FC $\Lambda _{1\left( l\right) }$. c) Subgraph $%
\Gamma _{\left( l\right) }$ of a limb of the 3\protect\underline{R}[2US]
Delta ($\mathfrak{N}_{l}=5$), and d) directed spanning tree $\protect%
\overrightarrow{G}_{\left( l\right) }$ ($\mathfrak{n}_{l}=4$) obtained by
removing edge 4 from the FC $\Lambda _{1\left( l\right) }$.}
\label{figDeltaLimbGraph}%
\centerline{
a)~\includegraphics[height=5.2cm]{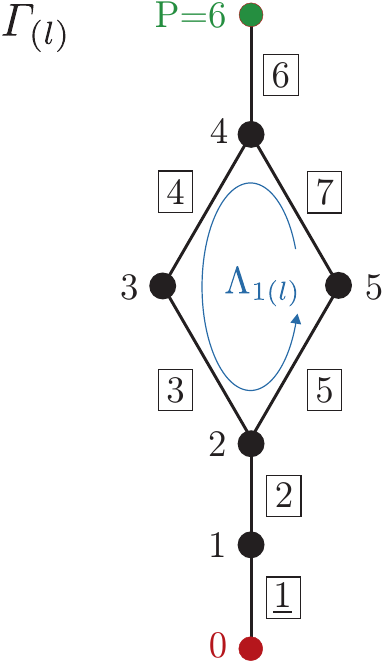}\ \ \ \ \ \ b)~\includegraphics[height=5.2cm]{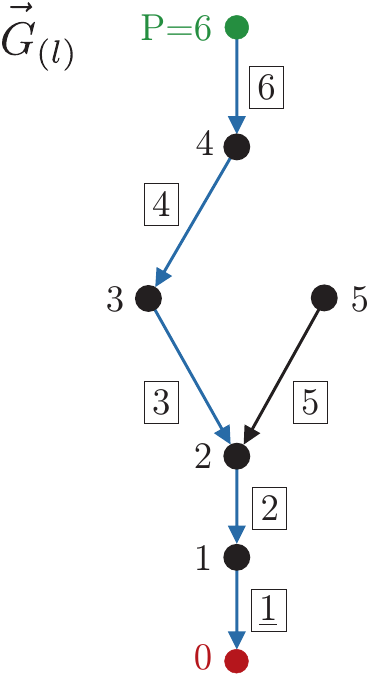}
\ \ \ \ \ \ \ \ \ \ \ \ \ \ \ \ 
c)~\includegraphics[height=4.0cm]{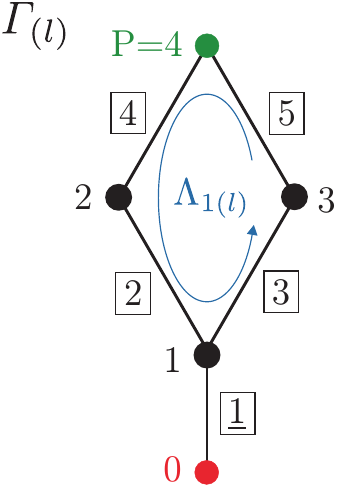}\ \ \ \ \ \ d)~\includegraphics[height=4.0cm]{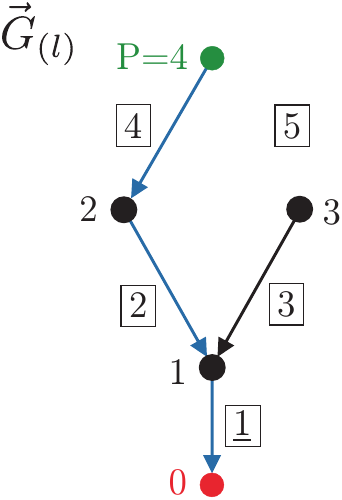}
}
\end{figure}
\end{example}

\begin{example}[{3\protect\underline{R}[2US] Delta --cont}]
The labeled topological graph of limb $l$ is shown in fig. \ref%
{figDeltaLimbGraph}c). Removing, for instance, edge 5 (representing a U
joint) from the FC $\Lambda _{1\left( l\right) }$ yields the directed
spanning tree, with $\mathfrak{n}_{l}=4$ tree-jointsshown in fig. \ref%
{figDeltaLimbGraph}d). The platform motion is determined by joints 1,2, and
4. Using two angles as joint variables for a U joint and three angles for an
S joint leads to the vector $%
%TCIMACRO{\TeXButton{red}{}}%
%BeginExpansion
%
%EndExpansion
%TCIMACRO{\TeXButton{eta}{\mathbold{\eta}}}%
%BeginExpansion
\mathbold{\eta}%
%EndExpansion
_{\left( l\right) }%
%TCIMACRO{\TeXButton{black}{\color{black}}}%
%BeginExpansion
\color{black}%
%EndExpansion
$ of $N_{l}=11$ joint variables, and the vector $%
%TCIMACRO{\TeXButton{red}{}}%
%BeginExpansion
%
%EndExpansion
%TCIMACRO{\TeXButton{vartheta}{\mathbold{\vartheta}}}%
%BeginExpansion
\mathbold{\vartheta}%
%EndExpansion
_{\left( l\right) }%
%TCIMACRO{\TeXButton{black}{\color{black}}}%
%BeginExpansion
\color{black}%
%EndExpansion
$ of $n_{l}=9$ tree-joint variables.
\end{example}

\begin{example}[IRSBot-2 --cont.]
For each of the two FCs of the IRSBot-2, one edge must be removed.
Eliminating edge 7 from $\Lambda _{1\left( l\right) }$ and edge 8 from $%
\Lambda _{2\left( l\right) }$ yields the directed tree shown in fig. \ref%
{figIRSBotLimbGraph}b). Joints 7 and 8 are the cut-joints. The platform
motion is determined by joints 1,2,4, and 6. The system is parameterized by $%
N_{l}=12$ joint variables in $%
%TCIMACRO{\TeXButton{red}{}}%
%BeginExpansion
%
%EndExpansion
%TCIMACRO{\TeXButton{eta}{\mathbold{\eta}}}%
%BeginExpansion
\mathbold{\eta}%
%EndExpansion
_{\left( l\right) }%
%TCIMACRO{\TeXButton{black}{\color{black}}}%
%BeginExpansion
\color{black}%
%EndExpansion
$, and the $n_{l}=9$ tree-joint variables constitute the vector $%
%TCIMACRO{\TeXButton{red}{}}%
%BeginExpansion
%
%EndExpansion
%TCIMACRO{\TeXButton{vartheta}{\mathbold{\vartheta}}}%
%BeginExpansion
\mathbold{\vartheta}%
%EndExpansion
_{\left( l\right) }%
%TCIMACRO{\TeXButton{black}{\color{black}}}%
%BeginExpansion
\color{black}%
%EndExpansion
$. 
\begin{figure}[h]
\caption{a) Labeled subgraph $\Gamma _{\left( l\right) }$ of a limb of the
IRSBot-2 ($\mathfrak{N}_{l}=8$). b) directed spanning tree $\protect%
\overrightarrow{G}_{\left( l\right) }$ ($\mathfrak{n}_{l}=6$) obtained by
removing edge 3 from the FC $\Lambda _{1\left( l\right) }$ and edge 7 from $%
\Lambda _{2\left( l\right) }$.}
\label{figIRSBotLimbGraph}%
\centerline{
a)~\includegraphics[height=5.2cm]{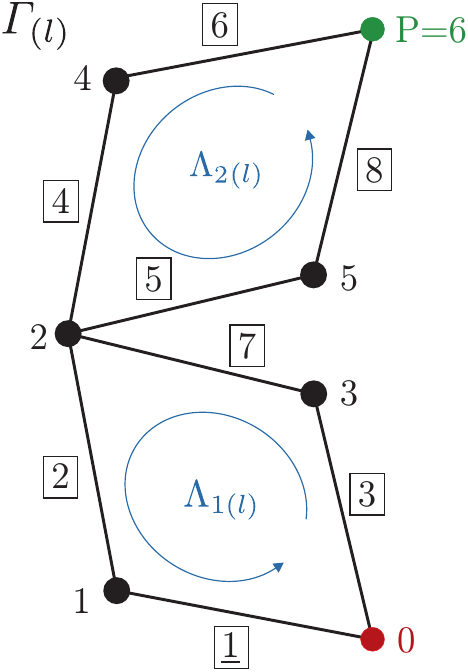}\ \
\ \ \ \ b)~\includegraphics[height=5.2cm]{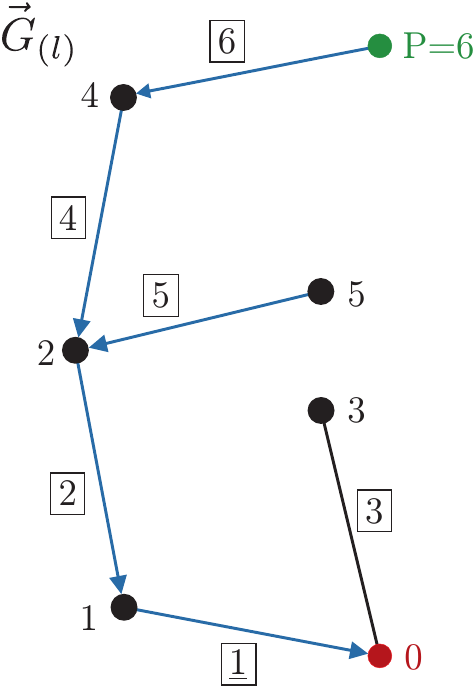}
}
\end{figure}
\end{example}

\subsection{Kinematics of the Associated Tree-Topology System%
%TCIMACRO{\TeXButton{secTreeKin}{\label{secTreeKin}}}%
%BeginExpansion
\label{secTreeKin}%
%EndExpansion
}

Once a spanning tree $G_{\left( l\right) }$ is introduced, the limb can be
treated as a tree-topology system. The pose of body $k$ of the limb is
determined by the joints in the kinematic chain from that body toward the
ground corresponding to the directed path from vertex $k$ to 0 in $\vec{G}%
_{\left( l\right) }$. The direction of edges has a kinematic meaning. The
joint represented by the directed edge is regarded as enabling the motion of
a body relative to its predecessor (defined by the target vertex of the
edge). This is important for interpreting the joint variables. For example,
the angle of revolute joint 5 of the 3\underline{R}R[2RR]R Delta describes
the rotation of body 5 relative to body 2 (not that of body 2 relative to
body 5), according to the directed tree in fig. \ref{figDeltaLimbGraph}b).

The absolute configuration (or pose) of body $k$ w.r.t. the inertial frame
(IFR) is denoted with $\mathbf{C}_{k}\in SE\left( 3\right) $ in (\ref{Ck}).
For sake of simplicity, and without loss of generality, in the following,
the kinematics is described assuming 1-DOF joints (so that $\mathfrak{n}%
_{l}=n_{l}$), which admits describing joint motions by a 1-parameter screw
motion. Notice that any mechanism can be modeled with 1-DOF joints by
modeling multi-DOF joints as series of 1-DOF joints. Denote with $\underline{%
k}$ the index of the last body in the path from body $k$ to the ground, i.e. 
$0=\underline{k}-1$. The pose of body $k$ in limb $l$ is determined by the
product of exponentials (\ref{POE}) as%
\begin{equation}
\mathbf{C}_{k\left( l\right) }(%
%TCIMACRO{\TeXButton{red}{}}%
%BeginExpansion
%
%EndExpansion
%TCIMACRO{\TeXButton{vartheta}{\mathbold{\vartheta}}}%
%BeginExpansion
\mathbold{\vartheta}%
%EndExpansion
_{\left( l\right) }%
%TCIMACRO{\TeXButton{black}{\color{black}}}%
%BeginExpansion
\color{black}%
%EndExpansion
)=f_{k\left( l\right) }(%
%TCIMACRO{\TeXButton{red}{}}%
%BeginExpansion
%
%EndExpansion
%TCIMACRO{\TeXButton{vartheta}{\mathbold{\vartheta}}}%
%BeginExpansion
\mathbold{\vartheta}%
%EndExpansion
_{\left( l\right) }%
%TCIMACRO{\TeXButton{black}{\color{black}}}%
%BeginExpansion
\color{black}%
%EndExpansion
)\mathbf{A}_{k\left( l\right) }  \label{Ckl}
\end{equation}%
with $\mathbf{A}_{k\left( l\right) }=\mathbf{C}_{k\left( l\right) }(\mathbf{0%
})$ being the zero reference configuration of the body, and (omitting index $%
\left( l\right) $)%
\begin{equation}
f_{k}(%
%TCIMACRO{\TeXButton{red}{}}%
%BeginExpansion
%
%EndExpansion
%TCIMACRO{\TeXButton{vartheta}{\mathbold{\vartheta}}}%
%BeginExpansion
\mathbold{\vartheta}%
%EndExpansion
%TCIMACRO{\TeXButton{black}{\color{black}}}%
%BeginExpansion
\color{black}%
%EndExpansion
)=\exp \left( \vartheta _{\underline{k}}\mathbf{Y}_{\underline{k}}\right)
\cdot \ldots \cdot \exp \left( \vartheta _{k-1}\mathbf{Y}_{k-1}\right) \exp
\left( \vartheta _{k}\mathbf{Y}_{k}\right)  \label{fk}
\end{equation}%
where $\mathbf{Y}_{i\left( l\right) }$ denotes the screw coordinate vector
of tree-joint $i$ at the zero reference configuration $%
%TCIMACRO{\TeXButton{eta}{\mathbold{\eta}}}%
%BeginExpansion
\mathbold{\eta}%
%EndExpansion
_{\left( l\right) }=\mathbf{0}$ of the tree-topology system, represented in
IFR. The relative configuration of body $r$ w.r.t. body $k$ is $\mathbf{C}%
_{k,r}:=\mathbf{C}_{k}^{-1}\mathbf{C}_{r}$.

The twist of body $k$ of limb $l$ in body-fixed representation is determined
by (\ref{VkJ}) in appendix \ref{AppendixRigMotion} as%
\begin{equation}
\mathbf{V}_{k}=\mathbf{J}_{k,\underline{k}}(%
%TCIMACRO{\TeXButton{red}{}}%
%BeginExpansion
%
%EndExpansion
%TCIMACRO{\TeXButton{vartheta}{\mathbold{\vartheta}}}%
%BeginExpansion
\mathbold{\vartheta}%
%EndExpansion
%TCIMACRO{\TeXButton{black}{\color{black}}}%
%BeginExpansion
\color{black}%
%EndExpansion
)\dot{\vartheta}_{\underline{k}}+\ldots +\mathbf{J}_{k,k-1}(%
%TCIMACRO{\TeXButton{red}{}}%
%BeginExpansion
%
%EndExpansion
%TCIMACRO{\TeXButton{vartheta}{\mathbold{\vartheta}}}%
%BeginExpansion
\mathbold{\vartheta}%
%EndExpansion
%TCIMACRO{\TeXButton{black}{\color{black}}}%
%BeginExpansion
\color{black}%
%EndExpansion
)\dot{\vartheta}_{k-1}+\mathbf{J}_{k,k}(%
%TCIMACRO{\TeXButton{red}{}}%
%BeginExpansion
%
%EndExpansion
%TCIMACRO{\TeXButton{vartheta}{\mathbold{\vartheta}}}%
%BeginExpansion
\mathbold{\vartheta}%
%EndExpansion
%TCIMACRO{\TeXButton{black}{\color{black}}}%
%BeginExpansion
\color{black}%
%EndExpansion
)\dot{\vartheta}_{k}  \label{Vk}
\end{equation}%
where the $6\times n_{l}$ matrix%
\begin{equation}
\mathbf{J}_{k}=\left( \mathbf{J}_{k,\underline{k}},\cdots ,\mathbf{J}%
_{k,k-1},\mathbf{J}_{k,k},\mathbf{0},\ldots \mathbf{0}\right)  \label{Jk}
\end{equation}%
is the \emph{geometric Jacobian} of body $k$ as part of the tree-topology
system. The non-zero columns of $\mathbf{J}_{k}$ are the instantaneous joint
screw coordinates represented in $\mathcal{F}_{k}$ as shown in (\ref{Jki})
in appendix \ref{AppendixRigMotion}. If tree-joint $i$ is not contained in
the directed path in $\vec{G}$ from body $k$ to ground, then $\mathbf{J}%
_{k,i}=\mathbf{0}$.

The platform pose and twist are determined by the tree-joint variables of
limb $l$ and their velocities as (%
%TCIMACRO{\TeXButton{red}{}}%
%BeginExpansion
%
%EndExpansion
Note that the reference configuration $\mathbf{A}_{\mathrm{p}}$ is the same
for all limbs.)%
%TCIMACRO{\TeXButton{black}{\color{black}}}%
%BeginExpansion
\color{black}%
%EndExpansion
\begin{eqnarray}
\mathbf{C}_{\mathrm{p}}(%
%TCIMACRO{\TeXButton{red}{}}%
%BeginExpansion
%
%EndExpansion
%TCIMACRO{\TeXButton{vartheta}{\mathbold{\vartheta}}}%
%BeginExpansion
\mathbold{\vartheta}%
%EndExpansion
_{\left( l\right) }%
%TCIMACRO{\TeXButton{black}{\color{black}}}%
%BeginExpansion
\color{black}%
%EndExpansion
) &=&f_{\mathrm{p}}(%
%TCIMACRO{\TeXButton{red}{}}%
%BeginExpansion
%
%EndExpansion
%TCIMACRO{\TeXButton{vartheta}{\mathbold{\vartheta}}}%
%BeginExpansion
\mathbold{\vartheta}%
%EndExpansion
_{\left( l\right) }%
%TCIMACRO{\TeXButton{black}{\color{black}}}%
%BeginExpansion
\color{black}%
%EndExpansion
)\mathbf{A}_{\mathrm{p}}  \label{Cp} \\
%TCIMACRO{\TeXButton{red}{}}%
%BeginExpansion
%
%EndExpansion
\mathbf{V}_{\mathrm{p}\left( l\right) }%
%TCIMACRO{\TeXButton{black}{\color{black}} }%
%BeginExpansion
\color{black}
%EndExpansion
&=&%
%TCIMACRO{\TeXButton{red}{}}%
%BeginExpansion
%
%EndExpansion
\mathbf{J}_{\mathrm{p}\left( l\right) }\dot{%
%TCIMACRO{\TeXButton{vartheta}{\mathbold{\vartheta}}}%
%BeginExpansion
\mathbold{\vartheta}%
%EndExpansion
}_{\left( l\right) }%
%TCIMACRO{\TeXButton{black}{\color{black}}}%
%BeginExpansion
\color{black}%
%EndExpansion
.  \label{Jp}
\end{eqnarray}%
The Jacobian $%
%TCIMACRO{\TeXButton{red}{}}%
%BeginExpansion
%
%EndExpansion
\mathbf{J}_{\mathrm{p}\left( l\right) }(%
%TCIMACRO{\TeXButton{vartheta}{\mathbold{\vartheta}}}%
%BeginExpansion
\mathbold{\vartheta}%
%EndExpansion
_{\left( l\right) })%
%TCIMACRO{\TeXButton{black}{\color{black}}}%
%BeginExpansion
\color{black}%
%EndExpansion
$ comprises the instantaneous screws of the joints in the path from platform
to ground in $\vec{G}_{\left( l\right) }$.

The platform pose is usually parameterized in terms of \emph{task-space
coordinates} $\mathbf{x}\in {\mathbb{V}}^{\delta _{\mathrm{p}}}$. With
slight abuse of notation, the corresponding mapping is also denoted with $f_{%
\mathrm{p}}$, so that $\mathbf{x}=f_{\mathrm{p}}(%
%TCIMACRO{\TeXButton{eta}{\mathbold{\eta}}}%
%BeginExpansion
\mathbold{\eta}%
%EndExpansion
)$. The time derivatives of the taskspace coordinates are related to the
platform twist by a relation of the form%
\begin{equation}
\mathbf{V}_{\mathrm{p}}=\mathbf{H}_{\mathrm{p}}\left( \mathbf{x}\right) \dot{%
\mathbf{x}}.  \label{Vpx}
\end{equation}%
Typical choices for $\mathbf{x}$ are components of the position vector
vector $\mathbf{r}_{\mathrm{p}}$ of the platform in combination with three
rotation parameters, e.g. Euler-/Cardan-angles or rotation axis/angle. In
the latter case, the mapping $\mathbf{H}_{\mathrm{p}}$ is the
left-trivialized differential of the exponential map on $SO\left( 3\right) $%
. Another choice for canonical coordinates is the use of screw coordinates,
in which case $\mathbf{H}_{\mathrm{p}}$ is the left-trivialized differential
of the exponential map on $SE\left( 3\right) $ \cite{RSPA2021}.

The platform acceleration is determined by%
\begin{equation}
%TCIMACRO{\TeXButton{red}{}}%
%BeginExpansion
%
%EndExpansion
\dot{\mathbf{V}}_{\mathrm{p}\left( l\right) }=\mathbf{J}_{\mathrm{p}\left(
l\right) }\ddot{%
%TCIMACRO{\TeXButton{vartheta}{\mathbold{\vartheta}}}%
%BeginExpansion
\mathbold{\vartheta}%
%EndExpansion
}+\dot{\mathbf{J}}_{\mathrm{p}\left( l\right) }\dot{%
%TCIMACRO{\TeXButton{vartheta}{\mathbold{\vartheta}}}%
%BeginExpansion
\mathbold{\vartheta}%
%EndExpansion
}%
%TCIMACRO{\TeXButton{black}{\color{black}}}%
%BeginExpansion
\color{black}%
%EndExpansion
.
\end{equation}%
The time derivative of the joint screw coordinates $\mathbf{J}_{k,a}$ are
given in closed form as in (\ref{Jdot}).

\begin{example}[{3\protect\underline{R}R[2RR]R Delta --cont.}]
According to the directed path in $\vec{G}_{\left( l\right) }$ in fig. \ref%
{figDeltaLimbGraph}b), the platform pose is $\mathbf{C}_{\mathrm{p}}=f_{%
\mathrm{p}}(%
%TCIMACRO{\TeXButton{red}{}}%
%BeginExpansion
%
%EndExpansion
%TCIMACRO{\TeXButton{vartheta}{\mathbold{\vartheta}}}%
%BeginExpansion
\mathbold{\vartheta}%
%EndExpansion
%TCIMACRO{\TeXButton{black}{\color{black}}}%
%BeginExpansion
\color{black}%
%EndExpansion
)\mathbf{A}_{\mathrm{p}}$ with ($\underline{k}=1,p=6$)%
\begin{equation}
f_{\mathrm{p}}(%
%TCIMACRO{\TeXButton{red}{}}%
%BeginExpansion
%
%EndExpansion
%TCIMACRO{\TeXButton{vartheta}{\mathbold{\vartheta}}}%
%BeginExpansion
\mathbold{\vartheta}%
%EndExpansion
%TCIMACRO{\TeXButton{black}{\color{black}}}%
%BeginExpansion
\color{black}%
%EndExpansion
)=\exp (\vartheta _{1}\mathbf{Y}_{1})\exp (\vartheta _{2}\mathbf{Y}_{2})\exp
(\vartheta _{3}\mathbf{Y}_{3})\exp (\vartheta _{4}\mathbf{Y}_{4})\exp
(\vartheta _{6}\mathbf{Y}_{6}).
\end{equation}%
and its geometric Jacobian is, with tree-joint variables $%
%TCIMACRO{\TeXButton{red}{}}%
%BeginExpansion
%
%EndExpansion
%TCIMACRO{\TeXButton{vartheta}{\mathbold{\vartheta}}}%
%BeginExpansion
\mathbold{\vartheta}%
%EndExpansion
_{\left( l\right) }%
%TCIMACRO{\TeXButton{black}{\color{black}}}%
%BeginExpansion
\color{black}%
%EndExpansion
=\left( \vartheta _{1},\ldots ,\vartheta _{6}\right) ^{T}$,%
\begin{equation}
\mathbf{J}_{\mathrm{p}}=%
%TCIMACRO{\TeXButton{Big}{\Big}}%
%BeginExpansion
\Big%
%EndExpansion
(\mathbf{J}_{\mathrm{p},1},\ \mathbf{J}_{\mathrm{p},2},~\mathbf{J}_{\mathrm{p%
},3},~\mathbf{J}_{\mathrm{p},4},~\mathbf{0},~\mathbf{J}_{\mathrm{p},6}%
%TCIMACRO{\TeXButton{Big}{\Big}}%
%BeginExpansion
\Big%
%EndExpansion
).  \label{JpDelta}
\end{equation}%
The chain connecting body $k=5$, for instance, to the ground contains joints
1,2, and 5. The body-fixed geometric Jacobian of body 5 is thus%
\begin{equation}
\mathbf{J}_{5}=%
%TCIMACRO{\TeXButton{Big}{\Big}}%
%BeginExpansion
\Big%
%EndExpansion
(\mathbf{J}_{5,1},~\mathbf{J}_{5,2},~\mathbf{0},~\mathbf{0},~\mathbf{J}%
_{5,5},~\mathbf{0}%
%TCIMACRO{\TeXButton{Big}{\Big}}%
%BeginExpansion
\Big%
%EndExpansion
).
\end{equation}%
The platform can only translate, and the platform position vector delivers
the taskspace coordinates $\mathbf{x}:=\mathbf{r}_{\mathrm{p}}$.\newline
The explicit screw coordinate vectors $\mathbf{Y}_{i}$ and reference
configurations are presented in section \ref{secDeltaKin}.
\end{example}

\begin{remark}
The product of exponentials (\ref{POE}) is a powerful method that enables
describing the motion of kinematic chain in terms of simple geometric
parameters. The body poses can, of course, be determined using any other
classical method. Using the Denavit-Hartenberg convention, for instance, it
would be expressed as the product of homogenous transformation matrices,
which take the place of the exponential maps \cite%
{AsadaSlotine1986,UickerRavaniSheth2013}.
\end{remark}

\subsection{Cut-Joint Formulation of Loop Constraints%
%TCIMACRO{\TeXButton{secCutJoint}{\label{secCutJoint}}}%
%BeginExpansion
\label{secCutJoint}%
%EndExpansion
}

The kinematic loops for which constraints are introduced are defined by the
FCs. Since the FCs are topologically independent and edge-disjoint
(assumption \ref{Assumption1}), the loop constraints can be solved
independently. The solution of the overall system of constraints in limb $l$
is hence determined by the solution for the $\gamma _{l}$ individual loops.
A FC is defined by the cut-edge. There are two conceptually different
approaches to formulate the loop constraints: the \emph{cut-joint}
formulation and the \emph{cut-body} formulation. The first method involves
constraints specific to the cut-joint, whereas the latter imposes a generic
system of loop constraints.

\subsubsection{Cut-Joint constraints}

In the cut-joint formulation, the cut-joints, corresponding to the cut-edges
of the FCs $\Lambda _{\lambda \left( l\right) },\lambda =1,\ldots ,\gamma
_{l}$, are removed from the model. This yields a tree-topology system with $%
n_{l}$ tree-joint variables whose topology is represented by the spanning
tree $G_{\left( l\right) }$. This tree-system is then subjected to a system
of cut-joint constraints in order satisfy loop closure.

The configuration of the tree-topology system associated to limb $l$ is
described by the $n_{l}$ joint variables in $%
%TCIMACRO{\TeXButton{red}{}}%
%BeginExpansion
%
%EndExpansion
%TCIMACRO{\TeXButton{vartheta}{\mathbold{\vartheta}}}%
%BeginExpansion
\mathbold{\vartheta}%
%EndExpansion
%TCIMACRO{\TeXButton{black}{\color{black}}}%
%BeginExpansion
\color{black}%
%EndExpansion
_{\left( l\right) }$. Denote with $%
%TCIMACRO{\TeXButton{red}{}}%
%BeginExpansion
%
%EndExpansion
%TCIMACRO{\TeXButton{vartheta}{\mathbold{\vartheta}}}%
%BeginExpansion
\mathbold{\vartheta}%
%EndExpansion
%TCIMACRO{\TeXButton{black}{\color{black}}}%
%BeginExpansion
\color{black}%
%EndExpansion
_{\left( \lambda ,l\right) }$ the vector comprising the $n_{\lambda ,l}$
joint variables of the joints in FC $\Lambda _{\lambda \left( l\right) }$
except those of the cut joint. The system of $m_{\lambda ,l}$ geometric,
velocity and acceleration constraints are respectively expressed as%
\begin{eqnarray}
g_{\left( \lambda ,l\right) }(%
%TCIMACRO{\TeXButton{red}{}}%
%BeginExpansion
%
%EndExpansion
%TCIMACRO{\TeXButton{vartheta}{\mathbold{\vartheta}}}%
%BeginExpansion
\mathbold{\vartheta}%
%EndExpansion
%TCIMACRO{\TeXButton{black}{\color{black}}}%
%BeginExpansion
\color{black}%
%EndExpansion
_{\left( \lambda ,l\right) }) &=&\mathbf{0}  \label{GeomConsLoopCutJoint} \\
\mathbf{G}_{\left( \lambda ,l\right) }%
%TCIMACRO{\TeXButton{red}{}}%
%BeginExpansion
%
%EndExpansion
\dot{%
%TCIMACRO{\TeXButton{vartheta}{\mathbold{\vartheta}}}%
%BeginExpansion
\mathbold{\vartheta}%
%EndExpansion
}%
%TCIMACRO{\TeXButton{black}{\color{black}}}%
%BeginExpansion
\color{black}%
%EndExpansion
_{\left( \lambda ,l\right) } &=&\mathbf{0}  \label{VelConsLoopCutJoint} \\
\mathbf{G}_{\left( \lambda ,l\right) }%
%TCIMACRO{\TeXButton{red}{}}%
%BeginExpansion
%
%EndExpansion
\ddot{%
%TCIMACRO{\TeXButton{vartheta}{\mathbold{\vartheta}}}%
%BeginExpansion
\mathbold{\vartheta}%
%EndExpansion
}%
%TCIMACRO{\TeXButton{black}{\color{black}}}%
%BeginExpansion
\color{black}%
%EndExpansion
_{\left( \lambda ,l\right) }+\dot{\mathbf{G}}_{\left( \lambda ,l\right) }%
%TCIMACRO{\TeXButton{red}{}}%
%BeginExpansion
%
%EndExpansion
\dot{%
%TCIMACRO{\TeXButton{vartheta}{\mathbold{\vartheta}}}%
%BeginExpansion
\mathbold{\vartheta}%
%EndExpansion
}%
%TCIMACRO{\TeXButton{black}{\color{black}}}%
%BeginExpansion
\color{black}%
%EndExpansion
_{\left( \lambda ,l\right) } &=&\mathbf{0}  \label{AccConsLoopCutJoint}
\end{eqnarray}%
A joint with $\delta $ DOF imposes $6-\delta $ constraints. If all
constraints are independent, then $m_{\lambda ,l}=6-\delta $, otherwise they
are reduced to a set of $m_{\lambda ,l}<6-\delta $ independent constraints,
which is not a topic of this paper. It is assumed that the constraint
Jacobian $\mathbf{G}_{\left( \lambda ,l\right) }(%
%TCIMACRO{\TeXButton{eta}{\mathbold{\eta}}}%
%BeginExpansion
\mathbold{\eta}%
%EndExpansion
_{\left( l\right) })$ is a regular $m_{\lambda ,l}\times n_{\lambda ,l}$
matrix. The generic DOF of the separated limb $l$ is then $\delta
_{l}:=n_{l}-(m_{1,l}+\ldots +m_{\gamma _{l},l})$.

The formulation of cut-joint constraints for various technical joints (lower
and higher pairs) are well-known in the field of multibody system dynamics 
\cite{NikraveshBook1988,HaugBook1989,ShabanaBook}. For completeness the
formulation of \emph{elementary cut-joint constraints}, which can be
combined to the constraints for specific technical joints, is briefly
presented next.

Assume the cut-joint of FC $\Lambda _{\lambda \left( l\right) }$ connects
body $k$ and body $r$. A cut-joint frame is introduced at either body,
denoted with $\mathcal{J}_{k,\lambda \left( l\right) }$ and $\mathcal{J}%
_{r,\lambda \left( l\right) }$, respectively. These are usually located at
the rotation center of a spherical or universal joint, or at the joint axis
of a revolute or cylindrical joints, for instance. The configuration of the
cut-joint frame on body $k$ and $r$ relative to $\mathcal{F}_{k}$ and $%
\mathcal{F}_{r}$ is respectively%
\begin{equation}
\mathbf{S}_{k,\lambda \left( l\right) }=\left( 
\begin{array}{cc}
\mathbf{R}_{k,\lambda \left( l\right) } & {^{k}}\mathbf{d}_{k,\lambda \left(
l\right) } \\ 
0 & 1%
\end{array}%
\right) ,\ \mathbf{S}_{r,\lambda \left( l\right) }=\left( 
\begin{array}{cc}
\mathbf{R}_{r,\lambda \left( l\right) } & {^{r}}\mathbf{d}_{r,\lambda \left(
l\right) } \\ 
0 & 1%
\end{array}%
\right)
\end{equation}%
where ${^{k}}\mathbf{d}_{k,\lambda \left( l\right) }\in {\mathbb{R}}^{3}$ is
the position vector from the body-fixed reference frame $\mathcal{F}_{k}$ to
the origin of joint frame $\mathcal{J}_{k,\lambda \left( l\right) }$,
resolved in $\mathcal{F}_{k}$, and analogously ${^{r}}\mathbf{d}_{r,\lambda
\left( l\right) }\in {\mathbb{R}}^{3}$ at body $r$. Further, $\mathbf{R}%
_{k,\lambda \left( l\right) }$ and $\mathbf{R}_{r,\lambda \left( l\right) }$
is the rotation matrix from $\mathcal{J}_{k,\lambda \left( l\right) }$ to $%
\mathcal{F}_{k}$, and from $\mathcal{J}_{r,\lambda \left( l\right) }$ to $%
\mathcal{F}_{r}$. Since in the following all derivations refer to limb $l$,
the index $\left( l\right) $ will be omitted.

\paragraph{Distance constraints:}

The distance vector of the origin of the two cut-joint frames resolved in $%
\mathcal{J}_{k,\lambda }$ is%
\begin{eqnarray}
{^{k,\lambda }}\bm{\Delta }\mathbf{r}_{\lambda } &=&\mathbf{R}_{k,\lambda
}^{T}\left( \mathbf{R}_{k,r}{^{r}}\mathbf{d}_{r,\lambda }-{^{k}}\mathbf{d}%
_{k,\lambda }+\mathbf{R}_{k}^{T}\left( \mathbf{r}_{r}-\mathbf{r}_{k}\right)
\right)  \notag \\
&=&{^{k,\lambda }}\mathbf{r}_{r}-{^{k,\lambda }}\mathbf{r}_{k}+{^{k,\lambda }%
}\mathbf{d}_{r,\lambda }-{^{k,\lambda }}\mathbf{d}_{k,\lambda }
\end{eqnarray}%
with relative rotation matrix $\mathbf{R}_{k,r}:=\mathbf{R}_{k}^{T}\mathbf{R}%
_{r}$ of body $r$ relative body $k$. This relative translation is restricted
according to the cut-joint motion. For most technical joints, the cut-joint
frame $\mathcal{J}_{k}$ can be introduced so that some components of ${%
^{k,\lambda }}\bm{\Delta }\mathbf{r}_{\lambda }$ must vanish (or depend on
the joint rotation, e.g. screw joints).

The relative velocity is readily found in terms of the body-fixed twists of
the connected bodies $k$ and $r$, and their geometric Jacobians, as%
\begin{align}
{^{k,\lambda }}\bm{\Delta }\dot{\mathbf{r}}_{\lambda }& =%
%TCIMACRO{\TeXButton{TeX field}{\Bigg}}%
%BeginExpansion
\Bigg%
%EndExpansion
(%
\begin{array}{cccc}
\mathbf{R}_{k,\lambda }^{T}({^{k}\widetilde{\mathbf{r}}}_{r}-{^{k}\widetilde{%
\mathbf{r}}}_{k}+{^{k}}\widetilde{\mathbf{d}}_{r,\lambda }) & \ -\mathbf{R}%
_{k,\lambda }^{T} & \ -\mathbf{R}_{\lambda ,r}{^{r}\widetilde{\mathbf{d}}}%
_{r,\lambda } & \ \mathbf{R}_{\lambda ,r}%
\end{array}%
%TCIMACRO{\TeXButton{TeX field}{\Bigg}}%
%BeginExpansion
\Bigg%
%EndExpansion
)\left( 
\begin{array}{c}
\bm{\omega}_{k} \\ 
\mathbf{v}_{k} \\ 
\bm{\omega}_{r} \\ 
\mathbf{v}_{r}%
\end{array}%
\right)  \notag \\
& =%
%TCIMACRO{\TeXButton{TeX field}{\Bigg}}%
%BeginExpansion
\Bigg%
%EndExpansion
(%
\begin{array}{cccc}
{^{k,\lambda }}\bm{\Delta }\mathbf{r}_{\lambda }+{^{k}}\widetilde{\mathbf{d}}%
_{r,\lambda } & \ -\mathbf{R}_{k,\lambda }^{T} & \ -\mathbf{R}_{\lambda ,r}{%
^{r}\widetilde{\mathbf{d}}}_{r,\lambda } & \ \mathbf{R}_{\lambda ,r}%
\end{array}%
%TCIMACRO{\TeXButton{TeX field}{\Bigg}}%
%BeginExpansion
\Bigg%
%EndExpansion
)\left( 
\begin{array}{c}
\bm{\omega}_{k} \\ 
\mathbf{v}_{k} \\ 
\bm{\omega}_{r} \\ 
\mathbf{v}_{r}%
\end{array}%
\right)  \notag \\
& =%
%TCIMACRO{\TeXButton{TeX field}{\Bigg}}%
%BeginExpansion
\Bigg%
%EndExpansion
(%
\begin{array}{cccc}
{^{k,\lambda }}\bm{\Delta }\mathbf{r}_{\lambda }+{^{k}}\widetilde{\mathbf{d}}%
_{r,\lambda } & \ -\mathbf{R}_{k,\lambda }^{T} & \ -\mathbf{R}_{\lambda ,r}{%
^{r}\widetilde{\mathbf{d}}}_{r,\lambda } & \ \mathbf{R}_{\lambda ,r}%
\end{array}%
%TCIMACRO{\TeXButton{TeX field}{\Bigg}}%
%BeginExpansion
\Bigg%
%EndExpansion
)\left( 
\begin{array}{c}
\mathbf{J}_{k} \\ 
\mathbf{J}_{r}%
\end{array}%
\right) 
%TCIMACRO{\TeXButton{red}{}}%
%BeginExpansion
%
%EndExpansion
\dot{%
%TCIMACRO{\TeXButton{vartheta}{\mathbold{\vartheta}}}%
%BeginExpansion
\mathbold{\vartheta}%
%EndExpansion
}%
%TCIMACRO{\TeXButton{black}{\color{black}}}%
%BeginExpansion
\color{black}%
%EndExpansion
_{\left( l\right) }=:\mathbf{G}_{\left( \lambda ,l\right) }^{\mathrm{dist}}%
%TCIMACRO{\TeXButton{red}{}}%
%BeginExpansion
%
%EndExpansion
\dot{%
%TCIMACRO{\TeXButton{vartheta}{\mathbold{\vartheta}}}%
%BeginExpansion
\mathbold{\vartheta}%
%EndExpansion
}%
%TCIMACRO{\TeXButton{black}{\color{black}}}%
%BeginExpansion
\color{black}%
%EndExpansion
_{\left( \lambda ,l\right) }
\end{align}%
with rotation matrix $\mathbf{R}_{\lambda ,r}:=\mathbf{R}_{k,\lambda }^{T}%
\mathbf{R}_{k,r}$ from body frame $\mathcal{F}_{r}$ to cut-joint frame $%
\mathcal{J}_{k,\lambda \left( l\right) }$. The last term follows noting that
the relative velocity only depends on motions of joints within the FC $%
\Lambda _{\lambda \left( l\right) }$. The velocity constraints are are
introduced by equating the relevant components of ${^{k,\lambda }}%
\bm{\Delta
}\dot{\mathbf{r}}_{\lambda }$ to zero.

\paragraph{Orientation constraints:}

Denote with ${^{k,\lambda }}\mathbf{e}_{k}$ a constant unit vector expressed
in $\mathcal{J}_{k,\lambda }$, and with ${^{r,\lambda }}\mathbf{e}_{r}$ one
expressed in $\mathcal{J}_{r,\lambda }$ (omitting subscript $\left( l\right) 
$). A constraint on the relative orientation of body $k$ and $r$ can be
described by enforcing that these two vectors remain perpendicular, i.e. 
\begin{equation}
0={^{k,\lambda }}\mathbf{e}_{k}^{T}\Delta \mathbf{R}_{\lambda }{^{r,\lambda }%
}\mathbf{e}_{r}  \label{OriConstr}
\end{equation}%
with $\Delta \mathbf{R}_{\lambda }:=\mathbf{R}_{k,\lambda }^{T}\mathbf{R}%
_{k,r}\mathbf{R}_{r,\lambda }$. The velocity constraints are readily
obtained with $\Delta \dot{\mathbf{R}}_{\lambda }=\mathbf{R}_{k,r}^{T}\left( 
\mathbf{R}_{k,r}\widetilde{\bm{\omega}}_{r}-\widetilde{\bm{\omega}}_{k}%
\mathbf{R}_{k,r}\right) \mathbf{R}_{r,\lambda }$ as%
\begin{align}
0& ={^{k}}\mathbf{e}_{k}^{T}{^{k}\widetilde{\mathbf{e}}}_{r}\widetilde{%
\bm{\omega}}_{k}-{^{r}}\mathbf{e}_{k}^{T}{^{r}\widetilde{\mathbf{e}}}_{r}%
\widetilde{\bm{\omega}}_{r}  \notag \\
& =%
%TCIMACRO{\TeXButton{TeX field}{\Bigg}}%
%BeginExpansion
\Bigg%
%EndExpansion
(%
\begin{array}{cc}
{^{k}}\mathbf{e}_{k}^{T}{^{k}\widetilde{\mathbf{e}}}_{r} & \ \ \ -{^{r}}%
\mathbf{e}_{k}^{T}{^{r}\widetilde{\mathbf{e}}}_{r}%
\end{array}%
%TCIMACRO{\TeXButton{TeX field}{\Bigg}}%
%BeginExpansion
\Bigg%
%EndExpansion
)\left( 
\begin{array}{c}
\bm{\omega}_{k} \\ 
\bm{\omega}_{r}%
\end{array}%
\right)  \notag \\
& =%
%TCIMACRO{\TeXButton{TeX field}{\Bigg}}%
%BeginExpansion
\Bigg%
%EndExpansion
(%
\begin{array}{cccc}
{^{k}}\mathbf{e}_{k}^{T}{^{k}\widetilde{\mathbf{e}}}_{r} & \ \ \ \mathbf{0}
& \ \ \ -{^{r}}\mathbf{e}_{k}^{T}{^{r}\widetilde{\mathbf{e}}}_{r} & \ \ \ 
\mathbf{0}%
\end{array}%
%TCIMACRO{\TeXButton{TeX field}{\Bigg}}%
%BeginExpansion
\Bigg%
%EndExpansion
)\left( 
\begin{array}{c}
\mathbf{J}_{k} \\ 
\mathbf{J}_{r}%
\end{array}%
\right) 
%TCIMACRO{\TeXButton{red}{}}%
%BeginExpansion
%
%EndExpansion
\dot{%
%TCIMACRO{\TeXButton{vartheta}{\mathbold{\vartheta}}}%
%BeginExpansion
\mathbold{\vartheta}%
%EndExpansion
}%
%TCIMACRO{\TeXButton{black}{\color{black}}}%
%BeginExpansion
\color{black}%
%EndExpansion
_{\left( l\right) }=:\mathbf{G}_{\left( \lambda ,l\right) }^{\mathrm{ori}}%
%TCIMACRO{\TeXButton{red}{}}%
%BeginExpansion
%
%EndExpansion
\dot{%
%TCIMACRO{\TeXButton{vartheta}{\mathbold{\vartheta}}}%
%BeginExpansion
\mathbold{\vartheta}%
%EndExpansion
}%
%TCIMACRO{\TeXButton{black}{\color{black}}}%
%BeginExpansion
\color{black}%
%EndExpansion
_{\left( \lambda ,l\right) }
\end{align}%
with the constant vectors ${^{k}}\mathbf{e}_{k}:=\mathbf{R}_{k,\lambda }{%
^{k,\lambda }}\mathbf{e}_{k}$ and ${^{r}}\mathbf{e}_{r}:=\mathbf{R}%
_{r,\lambda }{^{r,\lambda }}\mathbf{e}_{r}$ resolved in $\mathcal{F}_{k}$
and $\mathcal{F}_{r}$, respectively. For a joint with $\delta ^{\mathrm{rot}%
} $ rotary DOFs, $6-\delta ^{\mathrm{rot}}$ of such orientation constraints
are introduced. The joint frames are usually introduced so that $\mathbf{e}%
_{k}$ and $\mathbf{e}_{r}$ are aligned with the coordinate axes. Then ${%
^{r,\lambda }}\mathbf{e}_{r}$ and ${^{k,\lambda }}\mathbf{e}_{k}$ are one of
the unit vectors $\mathbf{u}_{1}=\left( 1,0,0\right) ^{T}$ etc., and (\ref%
{OriConstr}) simply requires that one component of the relative rotation
matrix $\Delta \mathbf{R}_{\lambda }$ must be zero.

\paragraph{Constraints for technical joints:}

The above two types of elementary constraints can be combined to the
constraint system of particular technical joints. Spherical and universal
joints, for example, do not allow for relative translations so that the
distance constraints are%
\begin{equation}
{^{k,\lambda }}\bm{\Delta }\mathbf{r}_{\lambda }=\mathbf{0},\ {^{k,\lambda }}%
\bm{\Delta }\dot{\mathbf{r}}_{\lambda }=\mathbf{0}.  \label{Sconstr}
\end{equation}%
A universal joint allows for independent rotations about two perpendicular
axes, and thus imposes one orientation constraint. Introducing the joint
frames so that one rotation axes is along the 1-axis of $\mathcal{J}%
_{k,\lambda }$, and the other rotation axis is along the 2-axis of $\mathcal{%
J}_{r,\lambda }$ (which is common practice \cite%
{NikraveshBook1988,HaugBook1989,Meccanica2016}), the constant vectors in (%
\ref{OriConstr}) are ${^{r,\lambda }}\mathbf{e}_{r}=\mathbf{u}_{1}=\left(
1,0,0\right) ^{T}$ and ${^{k,\lambda }}\mathbf{e}_{k}=\mathbf{u}_{2}=\left(
0,1,0\right) ^{T}$. The orientation constraint is that the $\left(
1,2\right) $-element of $\Delta \mathbf{R}_{\lambda }$ must be zero.

A revolute joint imposes two rotation constraints. If the rotation axis of a
revolute joint is aligned with the 3-axes of both joint frames (remaining
parallel), then one constraint is defined by the vectors ${^{r,\lambda }}%
\mathbf{e}_{r}=\mathbf{u}_{1}=\left( 0,0,1\right) ^{T}$ and ${^{k,\lambda }}%
\mathbf{e}_{k}=\mathbf{u}_{2}=\left( 1,0,0\right) ^{T}$, and the second
constrained by ${^{r,\lambda }}\mathbf{e}_{r}=\mathbf{u}_{1}=\left(
0,0,1\right) ^{T}$ and ${^{k,\lambda }}\mathbf{e}_{k}=\mathbf{u}_{2}=\left(
0,1,0\right) ^{T}$.

\subsubsection{Resolution of velocity constraints via block partitioning of
constraint Jacobian}

It is assumed in the following that the $m_{\lambda ,l}$ cut-joint
constraints (\ref{GeomConsLoopCutJoint}) are independent, so that $\mathbf{G}%
_{\left( \lambda ,l\right) }$ is a full rank $m_{\lambda ,l}\times
n_{\lambda ,l}$ matrix. 
%TCIMACRO{\TeXButton{red}{}}%
%BeginExpansion
%
%EndExpansion
The constraints (\ref{VelConsLoopCutJoint}) can be written as%
\begin{equation}
\mathbf{G}_{\mathbf{y}\left( \lambda ,l\right) }\dot{\mathbf{y}}_{\left(
\lambda ,l\right) }+\mathbf{G}_{\mathbf{q}\left( \lambda ,l\right) }\dot{%
\mathbf{q}}_{\left( \lambda ,l\right) }=\mathbf{0}  \label{GyGq}
\end{equation}%
where $\dot{\mathbf{q}}_{\left( \lambda ,l\right) }$ comprises $\delta
_{\lambda ,l}=n_{\lambda ,l}-m_{\lambda ,l}$ independent velocity
coordinates, and $\dot{\mathbf{y}}_{\left( \lambda ,l\right) }$ consists of $%
m_{\lambda ,l}$ dependent joint rates. Accordingly, $\mathbf{G}_{\mathbf{y}%
\left( \lambda ,l\right) }$ is a regular $m_{l}\times m_{l}$ submatrix, and $%
\mathbf{G}_{\mathbf{q}\left( \lambda ,l\right) }$ is a $m_{\lambda ,l}\times
\delta _{\lambda ,l}$ matrix. The solution of the velocity constraints (\ref%
{VelConsLoopCutJoint}) is then given in terms of the independent velocities
as%
\begin{equation}
\left( 
\begin{array}{c}
\dot{\mathbf{y}}_{\left( \lambda ,l\right) } \\ 
\dot{\mathbf{q}}_{\left( \lambda ,l\right) }%
\end{array}%
\right) =\mathbf{H}_{\left( \lambda ,l\right) }\dot{\mathbf{q}}_{\left(
\lambda ,l\right) }  \label{Hlambdal2}
\end{equation}%
with the $n_{\lambda ,l}\times \delta _{\lambda ,l}$ matrix $\mathbf{H}%
_{\left( \lambda ,l\right) }$ given explicitly as%
\begin{equation}
\mathbf{H}_{\left( \lambda ,l\right) }:=\left( 
\begin{array}{c}
-\mathbf{G}_{\mathbf{y}}^{-1}\mathbf{G}_{\mathbf{q}} \\ 
\mathbf{I}%
\end{array}%
\right) _{\left( \lambda ,l\right) }.  \label{Hlambda2}
\end{equation}%
With the block partitioning in (\ref{GyGq}), a solution of the acceleration
constraints (\ref{AccConsLoopCutJoint}) is%
\begin{equation}
\left( 
\begin{array}{c}
\ddot{\mathbf{y}}_{\left( \lambda ,l\right) } \\ 
\ddot{\mathbf{q}}_{\left( \lambda ,l\right) }%
\end{array}%
\right) =\mathbf{H}_{\left( \lambda ,l\right) }\ddot{\mathbf{q}}_{\left(
\lambda ,l\right) }+\dot{\mathbf{H}}_{\left( \lambda ,l\right) }\dot{\mathbf{%
q}}_{\left( \lambda ,l\right) }  \label{eta2dH}
\end{equation}%
with%
\begin{equation}
\dot{\mathbf{H}}_{\left( \lambda ,l\right) }(%
%TCIMACRO{\TeXButton{eta}{\mathbold{\eta}}}%
%BeginExpansion
\mathbold{\eta}%
%EndExpansion
_{\left( \lambda ,l\right) },\dot{%
%TCIMACRO{\TeXButton{eta}{\mathbold{\eta}}}%
%BeginExpansion
\mathbold{\eta}%
%EndExpansion
}_{\left( \lambda ,l\right) })=\left( 
\begin{array}{c}
\mathbf{G}_{\mathbf{y}}^{-1}(\dot{\mathbf{G}}_{\mathbf{y}}\mathbf{G}_{%
\mathbf{y}}^{-1}\mathbf{G}_{\mathbf{q}}-\dot{\mathbf{G}}_{\mathbf{q}}) \\ 
\mathbf{0}%
\end{array}%
\right) _{\left( \lambda ,l\right) }.  \label{Hldot}
\end{equation}%
It should be noticed, that the set of independent velocities can by
determined numerically by computing the null-space of $\mathbf{G}_{\left(
\lambda ,l\right) }$ using SVD or QR decompositions. However, a
predetermined set of independent coordinates $\mathbf{q}_{\left( \lambda
,l\right) }$ is often available, which determined the block partitioning.%
%TCIMACRO{\TeXButton{black}{\color{black}}}%
%BeginExpansion
\color{black}%
%EndExpansion

\begin{example}[{3\protect\underline{R}R[2RR]R Delta --cont.}]
%TCIMACRO{\TeXButton{exDeltaLoopSol}{\label{exDeltaLoopSol}}}%
%BeginExpansion
\label{exDeltaLoopSol}%
%EndExpansion
Joint 7 is used as cut-joint of the 4-bar parallelogram forming the FC $%
\Lambda _{1\left( l\right) }$. The three remaining tree-joints are subjected
to the revolute joint constraints. Selecting $\dot{\vartheta}_{4}$ as
independent velocity coordinate for the 4-bar loop yields%
\begin{equation*}
%TCIMACRO{\TeXButton{red}{}}%
%BeginExpansion
%
%EndExpansion
%TCIMACRO{\TeXButton{vartheta}{\mathbold{\vartheta}}}%
%BeginExpansion
\mathbold{\vartheta}%
%EndExpansion
%TCIMACRO{\TeXButton{black}{\color{black}}}%
%BeginExpansion
\color{black}%
%EndExpansion
_{\left( 1,l\right) }=\left( \vartheta _{3},\vartheta _{4},\vartheta
_{5}\right) ^{T},\ \mathbf{y}_{\left( 1,l\right) }=\left( \vartheta
_{3},\vartheta _{5}\right) ^{T},\ \mathbf{q}_{\left( 1,l\right) }=\left(
\vartheta _{4}\right) .
\end{equation*}%
Joint frames $\mathcal{J}_{4}$ and $\mathcal{J}_{5}$ are defined at link 4
and 5, respectively, as shown in fig. \ref{figDeltaLimbContraintsCutJoint}.
Then three rows of the constraint Jacobian $\mathbf{G}_{\left( \lambda
,l\right) }$ are identically zero, and it can be reduced to a $3\times 4$
matrix. As long as the 4-bar linkage remains a parallelogram, it is $\mathbf{%
G}_{\mathbf{y}}^{-1}\mathbf{G}_{\mathbf{q}}=\left( 1,1\right) ^{T}$, and the
solution of the velocity constraints is 
%TCIMACRO{\TeXButton{red}{} }%
%BeginExpansion

%EndExpansion
(\ref{Hlambdal2})%
%TCIMACRO{\TeXButton{black}{\color{black}}}%
%BeginExpansion
\color{black}%
%EndExpansion
, with 
\begin{equation}
\mathbf{H}_{\left( 1,l\right) }=\left( 
\begin{array}{c}
-\mathbf{G}_{\mathbf{y}}^{-1}\mathbf{G}_{\mathbf{q}} \\ 
1%
\end{array}%
\right) =\left( 
\begin{array}{c}
-1 \\ 
-1 \\ 
1%
\end{array}%
\right) .  \label{H1lDelta}
\end{equation}%
Accordingly, the solution of the geometric loop constraints is $\vartheta
_{3}=\vartheta _{5}=-\vartheta _{4}$, with $%
%TCIMACRO{\TeXButton{red}{}}%
%BeginExpansion
%
%EndExpansion
%TCIMACRO{\TeXButton{vartheta}{\mathbold{\vartheta}}}%
%BeginExpansion
\mathbold{\vartheta}%
%EndExpansion
%TCIMACRO{\TeXButton{black}{\color{black}}}%
%BeginExpansion
\color{black}%
%EndExpansion
=\mathbf{0}$ corresponding to the reference in fig. \ref{figRR2RRRDelta}. 
%\vspace{-6ex} 
\begin{figure}[h]
\centerline{
\includegraphics[height=6.0cm]{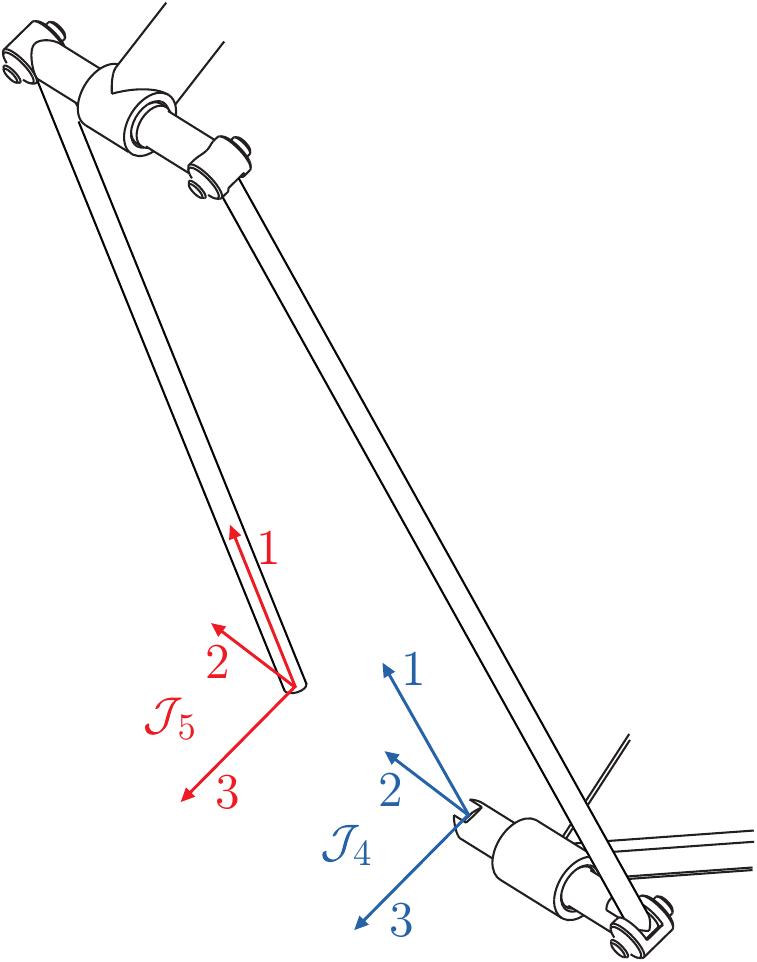}
}
\caption{Parallelogram loop of the Delta. Cut-joint constraints are
introduced between joint frames $\mathcal{J}_{4}$ and $\mathcal{J}_{5}$. The
loop constraints are expressed in cut-joint frame $\mathcal{J}_{4}$ at link
4.}
\label{figDeltaLimbContraintsCutJoint}
\end{figure}
\end{example}

\begin{example}[IRSBot-2 --cont.]
%TCIMACRO{\TeXButton{exIRSBot2-2}{\label{exIRSBot2-2}}}%
%BeginExpansion
\label{exIRSBot2-2}%
%EndExpansion
The constraint resolution of the parallelogram loop $\Lambda _{1\left(
l\right) }$ proceeds as for the Delta. The tree-joint velocities $%
%TCIMACRO{\TeXButton{red}{}}%
%BeginExpansion
%
%EndExpansion
\dot{%
%TCIMACRO{\TeXButton{vartheta}{\mathbold{\vartheta}}}%
%BeginExpansion
\mathbold{\vartheta}%
%EndExpansion
}%
%TCIMACRO{\TeXButton{black}{\color{black}}}%
%BeginExpansion
\color{black}%
%EndExpansion
_{\left( 1,l\right) }=\left( \dot{\vartheta}_{1},\dot{\vartheta}_{2},\dot{%
\vartheta}_{3}\right) ^{T}$ are expressed in terms of the independent
velocity $\dot{\mathbf{q}}_{\left( 1,l\right) }=\left( \vartheta _{1}\right) 
$, with $\mathbf{H}_{\left( 1,l\right) }$ as in (\ref{H1lDelta}). The 4U
loop $\Lambda _{2\left( l\right) }$ is cut open by removing joint 8
according to the spanning tree in fig. \ref{figIRSBotLimbGraph}b). The
remaining $n_{2,l}=6$ joint variables of the three U-joints in $%
%TCIMACRO{\TeXButton{red}{}}%
%BeginExpansion
%
%EndExpansion
%TCIMACRO{\TeXButton{vartheta}{\mathbold{\vartheta}}}%
%BeginExpansion
\mathbold{\vartheta}%
%EndExpansion
%TCIMACRO{\TeXButton{black}{\color{black}}}%
%BeginExpansion
\color{black}%
%EndExpansion
_{\left( 2,l\right) }=\left( \vartheta _{4,1},\vartheta _{4,2},\vartheta
_{5,1},\vartheta _{5,2},\vartheta _{6,1},\vartheta _{6,2}\right) ^{T}$ are
subjected to $m_{2,l}=4$ constraints (three distance and one orientation
constraint). The DOF of the loop is $\delta _{\lambda ,l}=2$, and the joint
variables of joint 5 are selected as independent $\mathbf{q}_{\left(
2,l\right) }=\left( \vartheta _{6,1},\vartheta _{6,2}\right) $, and $\mathbf{%
y}_{\left( 2,l\right) }=\left( \vartheta _{4,1},\vartheta _{4,2},\vartheta
_{5,1},\vartheta _{5,2}\right) ^{T}$. The $4\times 6$ constraint Jacobian $%
\mathbf{G}$ has full rank and so has the submatrix $\mathbf{G}_{\mathbf{y}}$%
, except at a kinematic singularity, which is not critical for the relevant
range of motion. The solution (\ref{Hlambdal2}) of the velocity constraints
is obtained with%
\begin{equation*}
\mathbf{H}_{\left( 2,l\right) }:=\left( 
\begin{array}{c}
-\mathbf{G}_{\mathbf{y}}^{-1}\mathbf{G}_{\mathbf{q}} \\ 
\mathbf{I}_{2,2}%
\end{array}%
\right)
\end{equation*}
\end{example}

\begin{remark}[Computational Aspects]
%TCIMACRO{\TeXButton{RemCompEff}{\label{RemCompEff}}}%
%BeginExpansion
\label{RemCompEff}%
%EndExpansion
The selection of independent coordinates, i.e. partitioning of the matrix $%
\mathbf{G}_{\left( \lambda ,l\right) }$ is not unique, unless these $\mathbf{%
q}_{\left( \lambda ,l\right) }$ correspond to the actuated joints. Moreover,
the selection of a well-conditioned submatrix $\mathbf{G}_{\mathbf{y}\left(
\lambda ,l\right) }$ is crucial for numerically stability. The partitioning
can be carried out by means of numerical methods for matrix decomposition,
such as QR or SVD, which would take into account the selection of a
well-conditioned submatrix to be inverted. This is also addressed in
connection with constraint stabilization of MBS models \cite%
{Blajer2011,Orden2012}. Computational methods for optimal coordinate
partitioning for multibody system models were reported in \cite%
{WehageHaug1982,Wehage2015,Nikravesh1990,Blajer1994,Terze2010}.\newline
The assumption of a full rank constraint Jacobian may not be satisfied if
the loop constraints are redundant, so that the $m_{\lambda ,l}\times
m_{\lambda ,l}$ matrix $\mathbf{G}_{\mathbf{y}\left( \lambda ,l\right) }$ is
singular. The treatment of redundant constraints in multibody system models
has been a research for many years, and numerical approaches were proposed 
\cite{Wojtyra2009,deJalon2013}. Whether and how redundant constraints can be
eliminated analytically depends on the particular mechanism. When the motion
space of a loop is know explicitly, as in case of planar or spherical
linkages, the elimination is straightforward. Clearly, then the choice of
reference frame in which the constraints are expressed is crucial, as
obvious from the above 4-bar parallelogram. A generally applicable
semianalytic method based on analytic identification of motion spaces has
been reported in \cite{MuellerCND2011,MuellerAME2014}. When modeling the
Delta with universal joints at either end of the bars, the constraints are
redundant, for example. Yet, in this case it is possible to remove redundant
constraints. In the general case of overconstrained and so-called
paradoxical \cite{Herve1978,Herve1982} mechanisms, this is not possible and
numerical decomposition must be used.
\end{remark}

\begin{remark}[Selection of cut-joint]
For a cut-joint with $\delta $ DOF, a system of $m_{\lambda ,l}=6-\delta $
cut-joint constraints is imposed to the $n_{l}=N_{l}-\delta $ joint
variables. Consequently, the system of constraints can be minimized by
selecting a cut-joint with high DOF $\delta $ on the expense of a larger DOF
of the tree-system. The latter implies a larger system of of dynamic motion
equations for the tree-topology system (see sec. \ref{secEOMLimb}).
\end{remark}

\subsection{Cut-Body Formulation of Loop Constraints%
%TCIMACRO{\TeXButton{secCutBody}{\label{secCutBody}}}%
%BeginExpansion
\label{secCutBody}%
%EndExpansion
}

In the cut-body formulation, the cut-edge merely determines how a FC is
traversed in order to formulate the loop closure constraints, while the
number of constraints does not depend on the selection of cut-edge. The
cut-body formulation does not introduce a tree-topology system. This is
clearly different from the cut-joint formulation, where the cut-edge
determines the cut-joint to be removed, and thus the tree-topology system as
well as the number of constraints.

\subsubsection{Kinematic loop constraints}

The $N_{l}$ joint variables of limb $l$ are subjected to the constraints due
to the $\gamma _{l}$ loops within the limb. Denote with $%
%TCIMACRO{\TeXButton{red}{}}%
%BeginExpansion
%
%EndExpansion
%TCIMACRO{\TeXButton{eta}{\mathbold{\eta}}}%
%BeginExpansion
\mathbold{\eta}%
%EndExpansion
%TCIMACRO{\TeXButton{black}{\color{black}}}%
%BeginExpansion
\color{black}%
%EndExpansion
_{\left( \lambda ,l\right) }$ the vector of $N_{\lambda ,l}$ joint variables
of all joints in FC $\Lambda _{\lambda \left( l\right) }$. The loop closure
condition for $\Lambda _{\lambda \left( l\right) },\lambda =1,\ldots ,\gamma
_{l}$ gives rise to a system of geometric constraints of the form \cite%
{Herve1982,Mueller_MMT2019}%
\begin{equation}
g_{\left( \lambda ,l\right) }(%
%TCIMACRO{\TeXButton{red}{}}%
%BeginExpansion
%
%EndExpansion
%TCIMACRO{\TeXButton{eta}{\mathbold{\eta}}}%
%BeginExpansion
\mathbold{\eta}%
%EndExpansion
%TCIMACRO{\TeXButton{black}{\color{black}}}%
%BeginExpansion
\color{black}%
%EndExpansion
_{\left( \lambda ,l\right) })=\mathbf{I}  \label{GeomConsLoop}
\end{equation}%
and the corresponding velocity constraints%
\begin{equation}
\mathbf{G}_{\left( \lambda ,l\right) }%
%TCIMACRO{\TeXButton{red}{}}%
%BeginExpansion
%
%EndExpansion
\dot{%
%TCIMACRO{\TeXButton{eta}{\mathbold{\eta}}}%
%BeginExpansion
\mathbold{\eta}%
%EndExpansion
}%
%TCIMACRO{\TeXButton{black}{\color{black}}}%
%BeginExpansion
\color{black}%
%EndExpansion
_{\left( \lambda ,l\right) }=\mathbf{0}.  \label{VelConsLoop}
\end{equation}%
The constraints of the $\gamma _{l}$ individual FCs are independent due to
the loop-partitioned limbs. The generic DOF of the separated limb $l$ is $%
\delta _{l}:=N_{l}-(m_{1,l}+\ldots +m_{\gamma _{l},l})$, with $m_{\lambda
,l} $ being the generic rank of $\mathbf{G}_{\left( \lambda ,l\right) }$.

The constraints can be formulated as a POE, similarly to (\ref{fk}). In
contrast to the forward kinematics of the tree-topology system, the
interpretation of joint motions depends on the direction of the FC, so that
orientation of edges must be taken into account. Denote with $\sigma
_{\left( \lambda ,l\right) }\left( i\right) \in \left\{ -1,0,1\right\} $ the
entries of the fundamental cycle matrix, which indicates whether a directed
edge is part of a FC. If edge $i$ is contained in $\Lambda _{l}$ and has the
same direction as $\Lambda _{\lambda \left( l\right) }$, then $\sigma
_{\left( \lambda ,l\right) }\left( i\right) =1$; if it is directed opposite
to the FC, then $\sigma _{\left( \lambda ,l\right) }\left( i\right) =-1$;
and $\sigma _{\left( \lambda ,l\right) }\left( i\right) =0$ if it is not
part of the FC.

Denote with $\underline{\lambda }$ the cut-edge, i.e. the single edge of $%
\Lambda _{l}$ not belonging to the spanning tree. This edge is used as start
edge when traversing the FC. The last edge visited is denoted with $\bar{%
\lambda}$. The loop orientation thus induces an ordering within the FC. The
geometric loop constraint can now be expressed as%
\begin{equation}
g_{\left( \lambda ,l\right) }(%
%TCIMACRO{\TeXButton{red}{}}%
%BeginExpansion
%
%EndExpansion
%TCIMACRO{\TeXButton{eta}{\mathbold{\eta}}}%
%BeginExpansion
\mathbold{\eta}%
%EndExpansion
%TCIMACRO{\TeXButton{black}{\color{black}}}%
%BeginExpansion
\color{black}%
%EndExpansion
_{\left( \lambda ,l\right) })=\exp \left( \sigma _{\left( \lambda ,l\right)
}(\bar{\lambda})\vartheta _{\bar{\lambda}}\mathbf{Y}_{\bar{\lambda}}\right)
\cdot \ldots \cdot \exp \left( \sigma _{\left( \lambda ,l\right) }(%
\underline{\lambda })\vartheta _{\underline{\lambda }}\mathbf{Y}_{\underline{%
\lambda }}\right) .
\end{equation}%
By convention, the cut-joint has the same orientation as the FC, i.e. $%
\sigma _{\left( \lambda ,l\right) }\left( \underline{\lambda }\right) =1$.

The $6\times N_{\lambda ,l}$ Jacobian $\mathbf{G}_{\left( \lambda ,l\right)
}(%
%TCIMACRO{\TeXButton{red}{}}%
%BeginExpansion
%
%EndExpansion
%TCIMACRO{\TeXButton{eta}{\mathbold{\eta}}}%
%BeginExpansion
\mathbold{\eta}%
%EndExpansion
%TCIMACRO{\TeXButton{black}{\color{black}}}%
%BeginExpansion
\color{black}%
%EndExpansion
_{\left( l\right) })$ is given in terms of the instantaneous screw
coordinate vectors of all joints belonging to the kinematic loop according
to FC $\Lambda _{\lambda \left( l\right) }$. Denote with $\mathbf{S}%
_{i},i\in \Lambda _{\lambda \left( l\right) }$ the instantaneous joint screw
coordinate vector of joint $i$ in the FC $\lambda $ of limb $l$ represented
in a general reference frame (when represented in the body-fixed frame at
body $k$, for instance, then $\mathbf{S}_{i}:=\mathbf{J}_{k,i}$ are the
body-fixed screws (\ref{Jki})). Taking into account the orientation within
the FC, the velocity constraints (\ref{VelConsLoop}) are%
\begin{equation}
\mathbf{0}=\sum_{i\in \Lambda _{\lambda \left( l\right) }}\sigma _{\left(
\lambda ,l\right) }(i)\mathbf{S}_{i}\dot{\vartheta}_{i}  \label{VelConsLoop2}
\end{equation}%
which can be written in the form (\ref{VelConsLoop}) with the constraint
Jacobian%
\begin{equation}
\mathbf{G}_{\left( \lambda ,l\right) }=%
%TCIMACRO{\TeXButton{Big}{\Big}}%
%BeginExpansion
\Big%
%EndExpansion
(\sigma _{\left( \lambda ,l\right) }(1)\mathbf{S}_{1},\cdots ,\sigma
_{\left( \lambda ,l\right) }(N_{l})\mathbf{S}_{N_{l}}%
%TCIMACRO{\TeXButton{Big}{\Big}}%
%BeginExpansion
\Big%
%EndExpansion
)  \label{VelConsLoop3}
\end{equation}%
where column $i$ not corresponding to a joint in $\Lambda _{\lambda \left(
l\right) }$ is zero since then $\sigma _{\left( \lambda ,l\right) }\left(
i\right) =0$.

\begin{example}[{3\protect\underline{R}R[2RR]R Delta --cont.}]
The one kinematic loop is formed by joints $3,4,5,7$ of the parallelogram
linkage (fig. \ref{figRR2RRRDelta}b), and represented by the FC $\Lambda
_{1\left( l\right) }$. The corresponding joint coordinate vector is $%
%TCIMACRO{\TeXButton{red}{}}%
%BeginExpansion
%
%EndExpansion
%TCIMACRO{\TeXButton{eta}{\mathbold{\eta}}}%
%BeginExpansion
\mathbold{\eta}%
%EndExpansion
%TCIMACRO{\TeXButton{black}{\color{black}}}%
%BeginExpansion
\color{black}%
%EndExpansion
_{\left( 1,l\right) }=\left( \vartheta _{3},\vartheta _{4},\vartheta
_{5},\vartheta _{7}\right) $. Edge 7 is used as cut-edge. The orientation of 
$\Lambda _{1\left( l\right) }$ in fig. \ref{figDeltaLimbGraph}b) induces the
ordering 5,3,47. Edge 5 is directed opposite to the FC. The constraint
mapping in (\ref{GeomConsLoop}) and the Jacobian in (\ref{VelConsLoop3}) are
thus (omitting subscript $\left( l\right) $ on the right-hand side)%
\begin{eqnarray}
\mathbf{g}_{\left( 1,l\right) }(%
%TCIMACRO{\TeXButton{red}{}}%
%BeginExpansion
%
%EndExpansion
%TCIMACRO{\TeXButton{eta}{\mathbold{\eta}}}%
%BeginExpansion
\mathbold{\eta}%
%EndExpansion
%TCIMACRO{\TeXButton{black}{\color{black}}}%
%BeginExpansion
\color{black}%
%EndExpansion
_{\left( 1,l\right) }) &=&\exp \left( -\vartheta _{5}\mathbf{Y}_{5}\right)
\exp \left( \vartheta _{3}\mathbf{Y}_{3}\right) \exp \left( \vartheta _{4}%
\mathbf{Y}_{4}\right) \exp \left( \vartheta _{7}\mathbf{Y}_{7}\right)  \notag
\\
\mathbf{G}_{\left( 1,l\right) } &=&%
%TCIMACRO{\TeXButton{Big}{\Big}}%
%BeginExpansion
\Big%
%EndExpansion
(\mathbf{S}_{3},~\mathbf{S}_{4},~-\mathbf{S}_{5},~\mathbf{S}_{7}%
%TCIMACRO{\TeXButton{Big}{\Big}}%
%BeginExpansion
\Big%
%EndExpansion
).
\end{eqnarray}
\end{example}

\begin{example}[IRSBot-2 --cont.]
According to the directed spanning tree in fig. \ref{figIRSBotLimbGraph}b),
the geometric loop constraints for the FCs $\Lambda _{1\left( l\right) }$
and $\Lambda _{2\left( l\right) }$ of the IRSBot-2 are determined by the
constraint maps (omitting again subscript $\left( l\right) $ on the
right-hand side)%
\begin{eqnarray}
\mathbf{g}_{\left( 1,l\right) }\left( 
%TCIMACRO{\TeXButton{red}{}}%
%BeginExpansion
%
%EndExpansion
%TCIMACRO{\TeXButton{eta}{\mathbold{\eta}}}%
%BeginExpansion
\mathbold{\eta}%
%EndExpansion
%TCIMACRO{\TeXButton{black}{\color{black}}}%
%BeginExpansion
\color{black}%
%EndExpansion
_{\left( 1,l\right) }\right) &=&\exp \left( -\vartheta _{3}\mathbf{Y}%
_{3}\right) \exp \left( \vartheta _{1}\mathbf{Y}_{1}\right) \exp \left(
\vartheta _{2}\mathbf{Y}_{2}\right) \exp \left( \vartheta _{7}\mathbf{Y}%
_{7}\right)  \notag \\
\mathbf{g}_{\left( 2,l\right) }\left( 
%TCIMACRO{\TeXButton{red}{}}%
%BeginExpansion
%
%EndExpansion
%TCIMACRO{\TeXButton{eta}{\mathbold{\eta}}}%
%BeginExpansion
\mathbold{\eta}%
%EndExpansion
%TCIMACRO{\TeXButton{black}{\color{black}}}%
%BeginExpansion
\color{black}%
%EndExpansion
_{\left( 2,l\right) }\right) &=&\exp \left( -\vartheta _{5}\mathbf{Y}%
_{5}\right) \exp \left( \vartheta _{4}\mathbf{Y}_{4}\right) \exp \left(
\vartheta _{6}\mathbf{Y}_{6}\right) \exp \left( \vartheta _{8}\mathbf{Y}%
_{8}\right) .
\end{eqnarray}%
The constraint Jacobian in (\ref{VelConsLoop}) for the respective FC is
(omitting subscript $\left( l\right) $ on the right-hand side)%
\begin{eqnarray}
\mathbf{G}_{\left( 1,l\right) } &=&%
%TCIMACRO{\TeXButton{Big}{\Big}}%
%BeginExpansion
\Big%
%EndExpansion
(\mathbf{S}_{1},~\mathbf{S}_{2},~-\mathbf{S}_{3},~\mathbf{S}_{7}%
%TCIMACRO{\TeXButton{Big}{\Big}}%
%BeginExpansion
\Big%
%EndExpansion
) \\
\mathbf{G}_{\left( 2,l\right) } &=&%
%TCIMACRO{\TeXButton{Big}{\Big}}%
%BeginExpansion
\Big%
%EndExpansion
(\mathbf{S}_{4},~-\mathbf{S}_{5},~\mathbf{S}_{6},~\mathbf{S}_{8}%
%TCIMACRO{\TeXButton{Big}{\Big}}%
%BeginExpansion
\Big%
%EndExpansion
).
\end{eqnarray}%
The corresponding joint coordinate vectors are $%
%TCIMACRO{\TeXButton{red}{}}%
%BeginExpansion
%
%EndExpansion
%TCIMACRO{\TeXButton{eta}{\mathbold{\eta}}}%
%BeginExpansion
\mathbold{\eta}%
%EndExpansion
%TCIMACRO{\TeXButton{black}{\color{black}}}%
%BeginExpansion
\color{black}%
%EndExpansion
_{\left( 1,l\right) }=\left( \vartheta _{1},\vartheta _{2},\vartheta
_{3},\vartheta _{7}\right) $ and $%
%TCIMACRO{\TeXButton{red}{}}%
%BeginExpansion
%
%EndExpansion
%TCIMACRO{\TeXButton{eta}{\mathbold{\eta}}}%
%BeginExpansion
\mathbold{\eta}%
%EndExpansion
%TCIMACRO{\TeXButton{black}{\color{black}}}%
%BeginExpansion
\color{black}%
%EndExpansion
_{\left( 2,l\right) }=\left( \vartheta _{4},\vartheta _{5},\vartheta
_{6},\vartheta _{8}\right) $.
\end{example}

\subsubsection{Constraint resolution using reciprocal screws}

The reciprocal screw approach is widely used for deriving the inverse
kinematics Jacobian of a PKM \cite%
{Tsai1998,TsaiBook1999,JoshiTsai2002,HuangLiuChetwyndMMT2011}. It also
provides a means to analytically solve the velocity constraints. Although
for general complex mechanisms this approach is difficult to pursue, it is
described in the following as it offers inside into the PKM kinematics.

The rank of the $6\times N_{\lambda ,l}$ Jacobian $\mathbf{G}_{\left(
\lambda ,l\right) }$, denoted with $m_{\lambda ,l}$, is the dimension of the
screw system defined by the $\mathbf{S}_{i}$. A set of $m_{\lambda ,l}$
linearly independent screws can be selected and used to form a $6\times
m_{\lambda ,l}$ submatrix $\mathbf{G}_{\mathbf{z}\left( \lambda ,l\right) }$%
. The remaining $\delta _{\lambda ,l}:=N_{\lambda ,l}-m_{\lambda ,l}$ screw
coordinates, which are linearly dependent to the former, provide the columns
of a $6\times \left( N_{\lambda ,l}-m_{\lambda ,l}\right) $ submatrix $%
\mathbf{G}_{\mathbf{q}\left( \lambda ,l\right) }$. Here, $\delta _{\lambda
,l}$ is the generic DOF of the FC $\Lambda _{\lambda \left( l\right) }$ when
considered separated from the mechanism. With this partitioning of the
constraint Jacobian, the constraint (\ref{VelConsLoop}) can be written as%
\begin{equation}
\mathbf{G}_{\mathbf{z}\left( \lambda ,l\right) }\dot{\mathbf{z}}_{\left(
\lambda ,l\right) }+\mathbf{G}_{\mathbf{q}\left( \lambda ,l\right) }\dot{%
\mathbf{q}}_{\left( \lambda ,l\right) }=\mathbf{0}  \label{G1G2}
\end{equation}%
where $\dot{\mathbf{z}}_{\left( \lambda ,l\right) }$ is the vector of
dependent, and $\dot{\mathbf{q}}_{\left( \lambda ,l\right) }$ that of the $%
N_{\lambda ,l}-m_{\lambda ,l}$ independent joint velocity variable of the
loop.

There is a screw that is reciprocal to all $m_{\lambda ,l}$ screws forming
the columns of $\mathbf{G}_{\mathbf{z}\left( \lambda ,l\right) }$ but not to
the screw forming its $j$th column and not to the screws forming $\mathbf{G}%
_{\mathbf{q}}$. There is one such screw for each of the $m_{\lambda ,l}$
columns of $\mathbf{G}_{\mathbf{q}}$. A $m_{\lambda ,l}\times 6$ matrix $%
\mathbf{W}_{\left( \lambda ,l\right) }$ is constructed whose rows are the
axes coordinate vectors of these reciprocal screws. Premultiplication of (%
\ref{G1G2}) with this matrix yields%
\begin{equation}
\mathbf{W}_{\left( \lambda ,l\right) }\mathbf{G}_{\mathbf{z}\left( \lambda
,l\right) }\dot{\mathbf{z}}_{\left( \lambda ,l\right) }+\mathbf{W}_{\left(
\lambda ,l\right) }\mathbf{G}_{\mathbf{q}\left( \lambda ,l\right) }\dot{%
\mathbf{q}}_{\left( \lambda ,l\right) }=\mathbf{0}.  \label{WG}
\end{equation}%
The term $\mathbf{W}_{\left( \lambda ,l\right) }\mathbf{G}_{\mathbf{z}\left(
\lambda ,l\right) }$ is a diagonal $m_{\lambda ,l}\times m_{\lambda ,l}$
matrix, and (\ref{WG}) can be solved as%
\begin{equation}
\dot{\mathbf{z}}_{\left( \lambda ,l\right) }=-\left( \mathbf{W}_{\left(
\lambda ,l\right) }\mathbf{G}_{\mathbf{z}\left( \lambda ,l\right) }\right)
^{-1}\mathbf{W}_{\left( \lambda ,l\right) }\mathbf{G}_{\mathbf{q}\left(
\lambda ,l\right) }\dot{\mathbf{q}}_{\left( \lambda ,l\right) }.  \label{WG3}
\end{equation}%
For hybrid limbs these solutions can be derived independently for all $%
\gamma _{l}$ FCs giving rise to a solution of the overall velocity loop
constraints of FC $\Lambda _{\lambda \left( l\right) }$ in terms of $\dot{%
\mathbf{q}}_{\left( l\right) }$. The main task, and principle challenge, of
this approach is the determination of the reciprocal screws. They can be
constructed geometrically for specific linkages, but this becomes cumbersome
for more complex robots. The method for constructing reciprocal screws
presented in \cite{KimChung2003} may alleviate this difficulty.

\begin{example}[{3\protect\underline{R}R[2RR]R Delta --cont.}]
Joints $3,4,5,7$ of the 3\underline{R}R[2RR]R Delta constitute a planar
4-bar parallelogram linkage (fig. \ref{figRR2RRRDelta}b), which defines the
only FC $\Lambda _{\left( 1,l\right) }$ of limb $l=1,2,3$. The subscript $%
\left( 1,l\right) $ is omitted in the following. Denote with $\mathbf{p}_{i}$
the position vector of a point on the axis of joint $i$ relative to an
arbitrary reference frame, and with the $\mathbf{e}_{i}$ the unit vector
along this joint axis, as shown in fig. \ref{figRR2RRRDeltaLimbScrews}. The
joint screws are then $\mathbf{S}_{i}=\left( \mathbf{e}_{i},\mathbf{p}%
_{i}\times \mathbf{e}_{i}\right) ^{T}$. The velocity loop constraints (\ref%
{VelConsLoop}) are%
\begin{equation}
\mathbf{S}_{3}\dot{\vartheta}_{3}+\mathbf{S}_{4}\dot{\vartheta}_{4}-\mathbf{S%
}_{5}\dot{\vartheta}_{5}+\mathbf{S}_{7}\dot{\vartheta}_{7}=\mathbf{0}\text{.}
\label{VelCons4Bar}
\end{equation}%
Since all $\mathbf{e}_{i}$ are identical, the dimension of the screw system
is $\mathrm{rank}~(\mathbf{S}_{3},\mathbf{S}_{4},-\mathbf{S}_{5},\mathbf{S}%
_{7})=3$. The screws of joints 3,5,7 are used as independent columns, and
the constraints (\ref{VelCons4Bar}) are written as in (\ref{G1G2}) with 
\begin{equation}
\mathbf{G}_{\mathbf{z}}=%
%TCIMACRO{\TeXButton{Big}{\Big}}%
%BeginExpansion
\Big%
%EndExpansion
(\mathbf{S}_{3},-\mathbf{S}_{5},\mathbf{S}_{7}%
%TCIMACRO{\TeXButton{Big}{\Big}}%
%BeginExpansion
\Big%
%EndExpansion
),\ \ \mathbf{G}_{\mathbf{q}}=%
%TCIMACRO{\TeXButton{Big}{\Big}}%
%BeginExpansion
\Big%
%EndExpansion
(\mathbf{S}_{4}%
%TCIMACRO{\TeXButton{Big}{\Big}}%
%BeginExpansion
\Big%
%EndExpansion
)
\end{equation}%
where $\mathbf{z}=\left( \vartheta _{3},\vartheta _{5},\vartheta _{7}\right)
^{T}$ and $\mathbf{q}=\left( \vartheta _{4}\right) $. The 4-bar motion is
thus parameterized by the angle of joint 4, as in example \ref%
{exDeltaLoopSol}. The axis coordinates of the (unit) screws that are
reciprocal all joint screws, but not to the screw of joint $j$ and joint 4,
can be introduced as%
\begin{equation}
\mathbf{W}_{3}=\left( 
\begin{array}{c}
\mathbf{p}_{5}\times \mathbf{u}_{5,7} \\ 
\mathbf{u}_{5,7}%
\end{array}%
\right) ,\ \mathbf{W}_{5}=\left( 
\begin{array}{c}
\mathbf{p}_{3}\times \mathbf{u}_{3,7} \\ 
\mathbf{u}_{3,7}%
\end{array}%
\right) ,\ \mathbf{W}_{7}=\left( 
\begin{array}{c}
\mathbf{p}_{3}\times \mathbf{u}_{3,5} \\ 
\mathbf{u}_{3,5}%
\end{array}%
\right)
\end{equation}%
where $\mathbf{u}_{i,j}:=\left( \mathbf{p}_{i}-\mathbf{p}_{j}\right)
/\left\Vert \mathbf{p}_{i}-\mathbf{p}_{j}\right\Vert $ is the unit vector
along the line passing through joints $i$ and $j$. $\mathbf{W}_{3}$ and $%
\mathbf{W}_{7}$ represent a force along link 5 and 2, respectively, while $%
\mathbf{W}_{5}$ represents a force along the line through points $\mathbf{p}%
_{7}$ and $\mathbf{p}_{3}$. Notice that, $\left\Vert \mathbf{p}_{7}-\mathbf{p%
}_{4}\right\Vert =L_{4},\left\Vert \mathbf{p}_{3}-\mathbf{p}_{4}\right\Vert
=L_{3},\left\Vert \mathbf{p}_{5}-\mathbf{p}_{7}\right\Vert =L_{5},\left\Vert 
\mathbf{p}_{5}-\mathbf{p}_{3}\right\Vert =L_{2}$, where $L_{i}$ is the
length of link $i$. Denote with $\left[ \mathbf{a},\mathbf{b},\mathbf{c}%
\right] =\mathbf{a}^{T}\left( \mathbf{b}\times \mathbf{c}\right) $ the wedge
product. Premultiplication of $\mathbf{G}_{\mathbf{z}}$ and $\mathbf{G}_{%
\mathbf{q}}$ with $\mathbf{W}=%
%TCIMACRO{\TeXButton{Big}{\Big}}%
%BeginExpansion
\Big%
%EndExpansion
(\mathbf{W}_{3},\mathbf{W}_{5},\mathbf{W}_{7}%
%TCIMACRO{\TeXButton{Big}{\Big}}%
%BeginExpansion
\Big%
%EndExpansion
)^{T}$ yields%
\begin{eqnarray}
\mathbf{WG}_{\mathbf{z}} &=&\mathrm{diag}\,\left( \mathbf{W}_{3}^{T}\mathbf{S%
}_{3},-\mathbf{W}_{5}^{T}\mathbf{S}_{5},\mathbf{W}_{7}^{T}\mathbf{S}%
_{7}\right) =\mathrm{diag}\,\left( L_{2}\left[ \mathbf{e}_{3},\mathbf{u}%
_{5,3},\mathbf{u}_{5,7}\right] ,-L_{2}\left[ \mathbf{e}_{5},\mathbf{u}_{3,5},%
\mathbf{u}_{3,7}\right] ,L_{5}\left[ \mathbf{e}_{7},\mathbf{u}_{5,7},\mathbf{%
u}_{3,5}\right] \right) \\
\mathbf{WG}_{\mathbf{q}} &=&\left( 
\begin{array}{c}
\mathbf{W}_{3}^{T}\mathbf{S}_{4} \\ 
\mathbf{W}_{5}^{T}\mathbf{S}_{4} \\ 
\mathbf{W}_{7}^{T}\mathbf{S}_{4}%
\end{array}%
\right) =\left( 
\begin{array}{c}
L_{4}\left[ \mathbf{e}_{4},\mathbf{u}_{7,4},\mathbf{u}_{5,7}\right] \\ 
L_{3}\left[ \mathbf{e}_{4},\mathbf{u}_{3,4},\mathbf{u}_{3,7}\right] \\ 
L_{3}\left[ \mathbf{e}_{4},\mathbf{u}_{3,4},\mathbf{u}_{3,5}\right]%
\end{array}%
\right)
\end{eqnarray}%
and thus%
\begin{equation}
\left( \mathbf{WG}_{\mathbf{z}}\right) ^{-1}\mathbf{WG}_{\mathbf{q}}=\left( 
\begin{array}{c}
\frac{L_{4}}{L_{2}}\frac{\left[ \mathbf{e}_{4},\mathbf{u}_{7,4},\mathbf{u}%
_{5,7}\right] }{\left[ \mathbf{e}_{3},\mathbf{u}_{5,3},\mathbf{u}_{5,7}%
\right] }%
%TCIMACRO{\TeXButton{TeX field}{\vspace{0.5ex}} }%
%BeginExpansion
\vspace{0.5ex}
%EndExpansion
\\ 
-\frac{L_{3}}{L_{2}}\frac{\left[ \mathbf{e}_{4},\mathbf{u}_{3,4},\mathbf{u}%
_{3,7}\right] }{\left[ \mathbf{e}_{5},\mathbf{u}_{3,5},\mathbf{u}_{3,7}%
\right] }%
%TCIMACRO{\TeXButton{TeX field}{\vspace{0.5ex}} }%
%BeginExpansion
\vspace{0.5ex}
%EndExpansion
\\ 
\frac{L_{3}}{L_{5}}\frac{\left[ \mathbf{e}_{4},\mathbf{u}_{3,4},\mathbf{u}%
_{3,5}\right] }{\left[ \mathbf{e}_{7},\mathbf{u}_{5,7},\mathbf{u}_{3,5}%
\right] }%
\end{array}%
\right) =\left( 
\begin{array}{c}
1 \\ 
1 \\ 
1%
\end{array}%
\right) .  \label{WG2}
\end{equation}%
The last term is obtained with the special geometry, according to which $%
\mathbf{e}_{3}=\mathbf{e}_{4}=\mathbf{e}_{5}=\mathbf{e}_{6}$, $\mathbf{u}%
_{5,3}=\mathbf{u}_{7,4},\mathbf{u}_{3,4}=\mathbf{u}_{5,7}$, and $%
L_{2}=L_{4},L_{3}=L_{5}$, and with $L_{3}\left[ \mathbf{e}_{4},\mathbf{u}%
_{3,4},\mathbf{u}_{3,7}\right] =-L_{2}\left[ \mathbf{e}_{5},\mathbf{u}_{3,5},%
\mathbf{u}_{3,7}\right] $. Thus (\ref{WG3}) yields the obvious 4-bar
relations $\dot{\vartheta}_{3}=\dot{\vartheta}_{5}=\dot{\vartheta}_{7}=-\dot{%
\vartheta}_{4}$ (ref. example \ref{exDeltaLoopSol}). 
\begin{figure}[h]
\centerline{
\includegraphics[height=8.0cm]{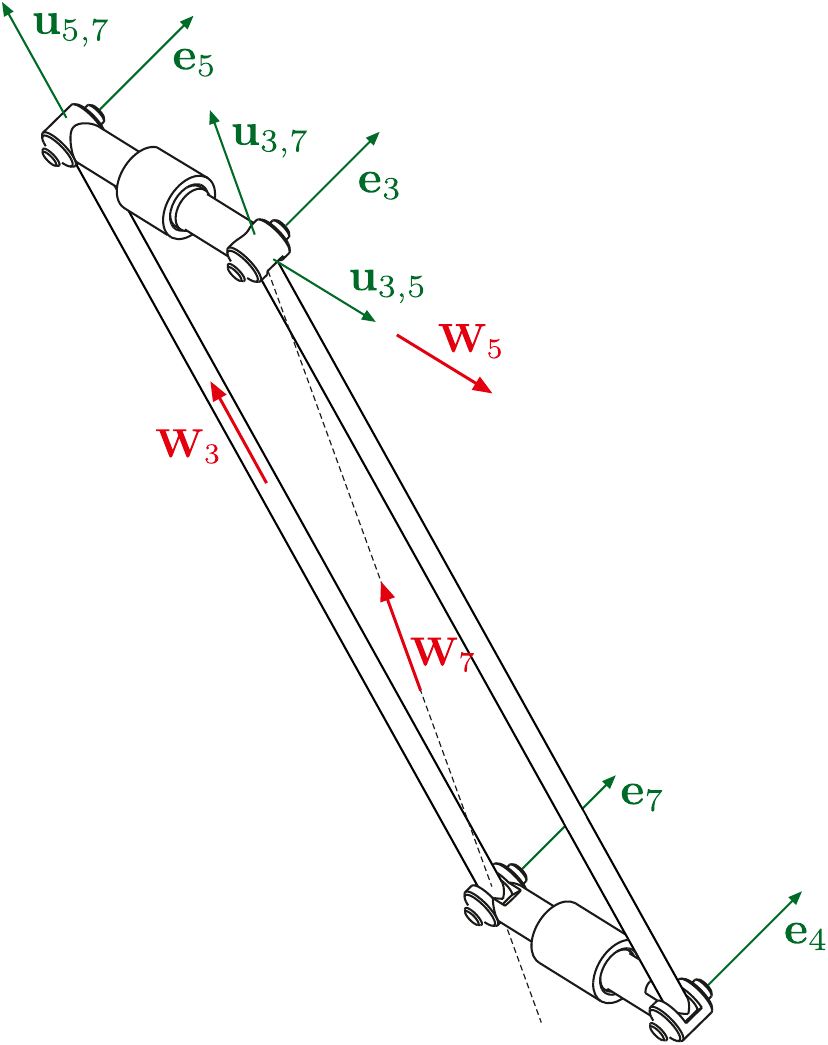}
}
\caption{Screw geometry of the 4-bar parallelogram loop of the Delta. For
joint and body numbering refer to fig. \protect\ref{figRR2RRRDelta}b).}
\label{figRR2RRRDeltaLimbScrews}
\end{figure}
\end{example}

\begin{example}[IRSBot-2 --cont.]
The kinematics of the parallelogram loop $\Lambda _{1\left( l\right) }$ and
the 4U loop $\Lambda _{2\left( l\right) }$ of the IRSBot-2 were investigated
in \cite{Germain2013} using the method of reciprocal screws. In this
publication, the loop constraints were not resolved, rather the
representative screw system (which is a system of screws describing the
motion of a loop, i.e. the constraint solution) was determined in order to
solve the velocity forward kinematics. The parallelogram loop can be treated
as for the 3\underline{R}R[2RR]R Delta. The 4U loop becomes rather
complicated, however, and is not presented here.
\end{example}

\subsubsection{Resolution of velocity constraints via block partitioning of
constraint Jacobian%
%TCIMACRO{\TeXButton{secCutBodySol}{\label{secCutBodySol}}}%
%BeginExpansion
\label{secCutBodySol}%
%EndExpansion
}

In contrast to the above reciprocal screw approach, the constraints (\ref%
{VelConsLoop}) be solved via block partitioning of $\mathbf{G}_{\left(
\lambda ,l\right) }$, as for the cut-joint formulation, which provides a
generally applicable numerical approach. It is assumed that all $m_{\lambda
,l}\leq 6$ constraints are independent, i.e. $\mathbf{G}_{\left( \lambda
,l\right) }$ is a full rank $m_{\lambda ,l}\times N_{\lambda ,l}$ matrix. 
%TCIMACRO{\TeXButton{red}{}}%
%BeginExpansion
%
%EndExpansion
The coordinate vector $%
%TCIMACRO{\TeXButton{red}{}}%
%BeginExpansion
%
%EndExpansion
%TCIMACRO{\TeXButton{eta}{\mathbold{\eta}}}%
%BeginExpansion
\mathbold{\eta}%
%EndExpansion
%TCIMACRO{\TeXButton{black}{\color{black}}}%
%BeginExpansion
\color{black}%
%EndExpansion
$ is split into a vector of $\delta _{\lambda ,l}=N_{\lambda ,l}-m_{\lambda
,l}$ independent coordinates $\mathbf{q}_{\left( \lambda ,l\right) }$, and a
vector $\mathbf{z}_{\left( \lambda ,l\right) }$ consists of the remaining
dependent coordinates of the FC. 
%TCIMACRO{\TeXButton{black}{\color{black}}}%
%BeginExpansion
\color{black}%
%EndExpansion
The velocity constraints are then written as in (\ref{G1G2}) with a regular
square $m_{\lambda ,l}\times m_{\lambda ,l}$ submatrix $\mathbf{G}_{\mathbf{z%
}\left( \lambda ,l\right) }$, and a $m_{\lambda ,l}\times \delta _{\lambda
,l}$ submatrix $\mathbf{G}_{\mathbf{q}\left( \lambda ,l\right) }$. 
%TCIMACRO{\TeXButton{red}{}}%
%BeginExpansion
%
%EndExpansion
The solution of (\ref{VelConsLoop}) is then%
\begin{equation}
\left( 
\begin{array}{c}
\dot{\mathbf{z}}_{\left( \lambda ,l\right) } \\ 
\dot{\mathbf{q}}_{\left( \lambda ,l\right) }%
\end{array}%
\right) =\mathbf{H}_{\left( \lambda ,l\right) }\dot{\mathbf{q}}_{\left(
\lambda ,l\right) }  \label{Hlambdal}
\end{equation}%
with%
\begin{equation}
\mathbf{H}_{\left( \lambda ,l\right) }:=\left( 
\begin{array}{c}
-\mathbf{G}_{\mathbf{z}}^{-1}\mathbf{G}_{\mathbf{q}} \\ 
\mathbf{I}%
\end{array}%
\right) _{\left( \lambda ,l\right) }.  \label{Hlambda}
\end{equation}%
%TCIMACRO{\TeXButton{black}{\color{black}}}%
%BeginExpansion
\color{black}%
%EndExpansion
Notice that the same independent coordinates $\mathbf{q}_{\left( \lambda
,l\right) }$ may be selected as in the cut-joint formulation since the DOF $%
\delta _{\lambda ,l}$ of the FC is indeed the same for both formulations.
Then $\mathbf{y}_{\left( \lambda ,l\right) }$ is the vector $\mathbf{z}%
_{\left( \lambda ,l\right) }$ with the cut-joint variables removed.

\begin{remark}[Cut-Body vs. cut-joint formulation]
The cut-joint formulation has a two-fold advantage over the cut-body
formulation. Firstly, it leads to a smaller system of $m_{\lambda ,l}<6$
constraints for the FC, and a smaller dimension of the matrix to be inverted
in (\ref{Hlambda2}). Secondly, the number $n_{l}$ of tree-joint variables,
i.e. the DOF of the tree-topology system of limb $l$, is smaller than the
total number $N_{l}$ of joint variables. On the other hand, the cut-joint
formulation may introduce artificial singularities.
\end{remark}

\begin{example}[{3\protect\underline{R}R[2RR]R Delta --cont.}]
The 4-bar parallelogram forms the only FC $\Lambda _{1\left( l\right) }$.
The constraints are expressed in a body-fixed frame $\mathcal{F}_{4}$ that
is arbitrarily located at link 4 but oriented as $\mathcal{J}_{4}$ in fig. %
\ref{figDeltaLimbContraintsCutJoint}. Then the rows 1,2, and 6 are
identically zero, and $\mathbf{G}_{\left( 1,l\right) }$ can be reduced to a $%
3\times 4$ matrix. Again, $\dot{\vartheta}_{4}$ is selected as independent
velocity coordinate for the 4-bar loop so that%
\begin{equation*}
%TCIMACRO{\TeXButton{red}{}}%
%BeginExpansion
%
%EndExpansion
%TCIMACRO{\TeXButton{eta}{\mathbold{\eta}}}%
%BeginExpansion
\mathbold{\eta}%
%EndExpansion
%TCIMACRO{\TeXButton{black}{\color{black}}}%
%BeginExpansion
\color{black}%
%EndExpansion
_{\left( 1,l\right) }=\left( \vartheta _{3},\vartheta _{4},\vartheta
_{5},\vartheta _{7}\right) ^{T},\ \mathbf{z}_{\left( 1,l\right) }=\left(
\vartheta _{3},\vartheta _{5},\vartheta _{7}\right) ^{T},\ \mathbf{q}%
_{\left( 1,l\right) }=\left( \vartheta _{4}\right)
\end{equation*}%
where the coordinates have been rearranged in accordance with (\ref{Hlambda}%
). As long as the 4-bar linkage remains a parallelogram, it is $\mathbf{G}_{%
\mathbf{z}}^{-1}\mathbf{G}_{\mathbf{q}}=\left( 1,1,1\right) ^{T}$, and the
solution of the velocity constraints is 
%TCIMACRO{\TeXButton{red}{} }%
%BeginExpansion

%EndExpansion
(\ref{Hlambdal2})%
%TCIMACRO{\TeXButton{black}{\color{black}}}%
%BeginExpansion
\color{black}%
%EndExpansion
, with%
\begin{equation*}
\mathbf{H}_{\left( 1,l\right) }:=\left( 
\begin{array}{c}
-\mathbf{G}_{\mathbf{z}}^{-1}\mathbf{G}_{\mathbf{q}} \\ 
1%
\end{array}%
\right) =\left( 
\begin{array}{c}
-1 \\ 
-1 \\ 
-1 \\ 
1%
\end{array}%
\right) .
\end{equation*}
\end{example}

\subsection{Velocity Forward Kinematics of Limb Mechanism%
%TCIMACRO{\TeXButton{secVelFKLimb}{\label{secVelFKLimb}}}%
%BeginExpansion
\label{secVelFKLimb}%
%EndExpansion
}

The motion of the FC $\Lambda _{\lambda \left( l\right) }$ of limb $l$ is
determined by the $\delta _{\lambda ,l}$ independent joint variables $%
\mathbf{q}_{\left( \lambda ,l\right) }$. In general, some joints of the limb
are not part of a FC. The corresponding $\delta _{0,l}$ joint variables are $%
\vartheta _{i},i\in G_{l}\backslash \Lambda _{\lambda \left( l\right)
},l=1,\ldots ,\gamma _{l}$. Since the $\gamma _{l}$ FCs are serially
arranged within the limb (assumption \ref{Assumption1}), the motion of the
separated limb $l$ (including the platform) is determined by the $\delta
_{l}:=\delta _{0,l}+\delta _{1,l}+\ldots +\delta _{\gamma _{l},l}$ joint
variables, which are summarized in the vector%
\begin{equation}
\mathbf{q}_{\left( l\right) }:=\left( \mathbf{q}_{\left( 1,l\right) },\ldots
,\mathbf{q}_{\left( \gamma _{l},l\right) },\vartheta _{n+1},\ldots
,,\vartheta _{N}\right) ^{T}  \label{ql}
\end{equation}%
which consists of the elements of $\mathbf{q}_{\left( \lambda ,l\right)
},\lambda =1,\ldots ,\gamma _{l}$ followed by variables of the remaining
tree-joint that are not contained in a FC. Elements of $\mathbf{q}_{\left(
l\right) }$ represent generalized coordinates, and $\delta _{l}$ is the DOF
of the separated limb including the platform.

In the following, the cut-joint formulation is employed as this can be
immediately exploited for the dynamics modeling. The tree-joint velocities
of limb $l$ are determined by the generalized velocities of the limb as%
\begin{equation}
%TCIMACRO{\TeXButton{red}{}}%
%BeginExpansion
%
%EndExpansion
\dot{%
%TCIMACRO{\TeXButton{vartheta}{\mathbold{\vartheta}}}%
%BeginExpansion
\mathbold{\vartheta}%
%EndExpansion
}%
%TCIMACRO{\TeXButton{black}{\color{black}}}%
%BeginExpansion
\color{black}%
%EndExpansion
_{\left( l\right) }=\mathbf{H}_{\left( l\right) }\dot{\mathbf{q}}_{\left(
l\right) }  \label{etaq}
\end{equation}%
where the $n_{l}\times \delta _{l}$ matrix is%
\begin{equation}
\mathbf{H}_{\left( l\right) }:=\mathbf{P}_{\left( l\right) }\left( 
\begin{array}{ccccc}
\mathbf{H}_{\left( 1,l\right) } & \mathbf{0} &  & \cdots & \mathbf{0} \\ 
\mathbf{0} & \mathbf{H}_{\left( 2,l\right) } &  &  & \mathbf{0} \\ 
&  & \ddots &  & \vdots \\ 
\vdots &  &  & \mathbf{H}_{\left( \gamma _{l},l\right) } & \mathbf{0} \\ 
\mathbf{0} & \mathbf{0} & \cdots & \mathbf{0} & \mathbf{I}%
\end{array}%
\right)  \label{Hl}
\end{equation}%
constructed from the matrices in (\ref{Hlambdal2}), and $\mathbf{P}_{\left(
l\right) }$ is a permutation matrix that assigns the rows of $\mathbf{H}%
_{\left( \lambda ,l\right) }$ to the corresponding joint variables in $%
%TCIMACRO{\TeXButton{red}{}}%
%BeginExpansion
%
%EndExpansion
%TCIMACRO{\TeXButton{vartheta}{\mathbold{\vartheta}}}%
%BeginExpansion
\mathbold{\vartheta}%
%EndExpansion
%TCIMACRO{\TeXButton{black}{\color{black}}}%
%BeginExpansion
\color{black}%
%EndExpansion
_{\left( l\right) }$.

The body-fixed twist of body $k$ of the separated limb $l$ is determined in
terms of the generalized velocities, by the Jacobian (\ref{Jk}) along with
the solution (\ref{etaq}) as%
\begin{equation}
\mathbf{V}_{k\left( l\right) }=\mathbf{L}_{k\left( l\right) }\dot{\mathbf{q}}%
_{\left( l\right) }.  \label{Vkl}
\end{equation}%
with the \emph{compound geometric Jacobian} of body $k$ in limb $l$ 
\begin{equation}
\mathbf{L}_{k\left( l\right) }:=\mathbf{J}_{k\left( l\right) }\mathbf{H}%
_{\left( l\right) }.  \label{Lkl}
\end{equation}%
In particular,%
\begin{equation}
\mathbf{L}_{\mathrm{p}\left( l\right) }:=\mathbf{J}_{\mathrm{p}\left(
l\right) }\mathbf{H}_{\left( l\right) }  \label{Lpl}
\end{equation}%
is the \emph{compound forward kinematics Jacobian} of limb $l$.

\begin{remark}
To construct the compound Jacobian, instead of using the permutation matrix $%
\mathbf{P}_{\left( l\right) }$ in (\ref{Hl}), the joint variables in $%
%TCIMACRO{\TeXButton{eta}{\mathbold{\eta}}}%
%BeginExpansion
\mathbold{\eta}%
%EndExpansion
_{\left( l\right) }$, and accordingly the columns of $\mathbf{J}_{k\left(
l\right) }$, could simply be rearranged.
\end{remark}

\begin{example}[{3\protect\underline{R}R[2RR]R Delta --cont.}]
%TCIMACRO{\TeXButton{exDeltaq}{\label{exDeltaq}}}%
%BeginExpansion
\label{exDeltaq}%
%EndExpansion
In the separated limb, the $\delta _{0,l}=3$ joints 1,2, and 
%TCIMACRO{\TeXButton{red}{}}%
%BeginExpansion
%
%EndExpansion
5 
%TCIMACRO{\TeXButton{black}{\color{black}}}%
%BeginExpansion
\color{black}%
%EndExpansion
are unconstrained (see tree topology in fig. \ref{figDeltaLimbGraph}b). The
vector $%
%TCIMACRO{\TeXButton{red}{}}%
%BeginExpansion
%
%EndExpansion
%TCIMACRO{\TeXButton{vartheta}{\mathbold{\vartheta}}}%
%BeginExpansion
\mathbold{\vartheta}%
%EndExpansion
%TCIMACRO{\TeXButton{black}{\color{black}}}%
%BeginExpansion
\color{black}%
%EndExpansion
_{1\left( l\right) }=\left( \vartheta _{3},\vartheta _{4},\vartheta
_{5}\right) ^{T}$ comprises the joint variables of joints forming the FC $%
\Lambda _{1\left( l\right) }$, and $%
%TCIMACRO{\TeXButton{red}{}}%
%BeginExpansion
%
%EndExpansion
\dot{%
%TCIMACRO{\TeXButton{vartheta}{\mathbold{\vartheta}}}%
%BeginExpansion
\mathbold{\vartheta}%
%EndExpansion
}%
%TCIMACRO{\TeXButton{black}{\color{black}}}%
%BeginExpansion
\color{black}%
%EndExpansion
_{1\left( l\right) }$ is subjected to the loop constraints. Their solution
is expressed in terms of $\dot{\vartheta}_{4}$, with matrix $\mathbf{H}%
_{\left( 1,l\right) }$ in (\ref{H1lDelta}). The velocity of the limb $l$ is
thus determined by $\dot{\mathbf{q}}_{\left( l\right) }=\left( \dot{\vartheta%
}_{4},\dot{\vartheta}_{1},\dot{\vartheta}_{2},\dot{\vartheta}_{6}\right)
^{T} $ as%
\begin{equation*}
\left( \dot{\vartheta}_{3},\dot{\vartheta}_{5},\dot{\vartheta}_{4},\dot{%
\vartheta}_{1},\dot{\vartheta}_{2},\dot{\vartheta}_{6}\right) ^{T}=\left( 
\begin{array}{cc}
\mathbf{H}_{\left( 1,l\right) } & \mathbf{0}_{3,3} \\ 
\mathbf{0}_{3,1} & \mathbf{I}_{3,3}%
\end{array}%
\right) \dot{\mathbf{q}}_{\left( l\right) }.
\end{equation*}%
The DOF of the limb is $\delta _{l}=4$. The velocity vector $%
%TCIMACRO{\TeXButton{red}{}}%
%BeginExpansion
%
%EndExpansion
\dot{%
%TCIMACRO{\TeXButton{vartheta}{\mathbold{\vartheta}}}%
%BeginExpansion
\mathbold{\vartheta}%
%EndExpansion
}%
%TCIMACRO{\TeXButton{black}{\color{black}}}%
%BeginExpansion
\color{black}%
%EndExpansion
_{\left( l\right) }=\left( \dot{\vartheta}_{1},\ldots ,\dot{\vartheta}%
_{6}\right) ^{T}$ is then obtained be reordering the elements of the vector
on the left-hand side using the permutation matrix%
\begin{equation*}
\mathbf{P}_{\left( l\right) }=\left( 
\begin{array}{cccccc}
0 & 0 & 0 & 1 & 0 & 0 \\ 
0 & 0 & 0 & 0 & 1 & 0 \\ 
1 & 0 & 0 & 0 & 0 & 0 \\ 
0 & 0 & 1 & 0 & 0 & 0 \\ 
0 & 1 & 0 & 0 & 0 & 0 \\ 
0 & 0 & 0 & 0 & 0 & 1%
\end{array}%
\right) .
\end{equation*}%
Along with (\ref{H1lDelta}), the matrix in (\ref{Hl}) is thus%
\begin{equation}
\mathbf{H}_{\left( l\right) }=\mathbf{P}_{\left( l\right) }\left( 
\begin{array}{cc}
\mathbf{H}_{\left( 1,l\right) } & \mathbf{0}_{3,3} \\ 
\mathbf{0}_{3,1} & \mathbf{I}_{3,3}%
\end{array}%
\right) =\left( 
\begin{array}{cccc}
0 & 1 & 0 & 0 \\ 
0 & 0 & 1 & 0 \\ 
-1 & 0 & 0 & 0 \\ 
1 & 0 & 0 & 0 \\ 
-1 & 0 & 0 & 0 \\ 
0 & 0 & 0 & 1%
\end{array}%
\right) .  \label{HlDelta}
\end{equation}%
The solution of velocity constraints is determined by the constant matrix $%
\mathbf{H}_{\left( l\right) }$ due to the special parallelogram geometry so
that the joint variables in the FC are linearly related. The compound
Jacobian $\mathbf{L}_{k\left( l\right) }$ of body $k$ is then obtained with (%
\ref{Lkl}). In particular, the forward kinematics Jacobian $\mathbf{L}_{%
\mathrm{p}\left( l\right) }$ is obtained by premultiplication of $\mathbf{J}%
_{\mathrm{p}\left( l\right) }$ in (\ref{JpDelta}) with $\mathbf{H}_{\left(
l\right) }$. The rank of this matrix 4.
\end{example}

\begin{example}[IRSBot-2 --cont.]
With the joint coordinate vectors $%
%TCIMACRO{\TeXButton{red}{}}%
%BeginExpansion
%
%EndExpansion
%TCIMACRO{\TeXButton{vartheta}{\mathbold{\vartheta}}}%
%BeginExpansion
\mathbold{\vartheta}%
%EndExpansion
%TCIMACRO{\TeXButton{black}{\color{black}}}%
%BeginExpansion
\color{black}%
%EndExpansion
_{1\left( l\right) }=\left( \vartheta _{1},\vartheta _{2},\vartheta
_{3}\right) ^{T}$ and $%
%TCIMACRO{\TeXButton{red}{}}%
%BeginExpansion
%
%EndExpansion
%TCIMACRO{\TeXButton{vartheta}{\mathbold{\vartheta}}}%
%BeginExpansion
\mathbold{\vartheta}%
%EndExpansion
%TCIMACRO{\TeXButton{black}{\color{black}}}%
%BeginExpansion
\color{black}%
%EndExpansion
_{\left( 2,l\right) }=\left( \vartheta _{4,1},\vartheta _{4,2},\vartheta
_{5,1},\vartheta _{5,2},\vartheta _{6,1},\vartheta _{6,2}\right) ^{T}$, and
independent coordinates $\mathbf{q}_{\left( l\right) }=\left( \vartheta
_{1},\vartheta _{6,1},\vartheta _{6,2}\right) ^{T}$ (see example \ref%
{exIRSBot2-2}), the velocity constraints of limb $l$ are resolved as%
\begin{equation*}
\left( \dot{\vartheta}_{1},\dot{\vartheta}_{2},\dot{\vartheta}_{3},\dot{%
\vartheta}_{4,1},\dot{\vartheta}_{4,2},\dot{\vartheta}_{5,1},\dot{\vartheta}%
_{5,2},\dot{\vartheta}_{6,1},\dot{\vartheta}_{6,2}\right) ^{T}=\left( 
\begin{array}{cc}
\mathbf{H}_{\left( 1,l\right) } & \mathbf{0}_{3,2} \\ 
\mathbf{0}_{6,1} & \mathbf{H}_{\left( 2,l\right) }%
\end{array}%
\right) \dot{\mathbf{q}}_{\left( l\right) }=\mathbf{H}_{\left( l\right) }%
\dot{\mathbf{q}}_{\left( l\right) }.
\end{equation*}
\end{example}

\subsection{Taskspace Velocity}

Consider the complete PKM with all limbs assembled. The platform of the PKM
has a DOF $\delta _{\mathrm{p}}\leq 6$, and $\delta _{\mathrm{p}}$
'components' of the platform motion are regarded as kinematic output of the
PKM. The corresponding components of the platform twist (usually represented
in the platform frame $\mathcal{F}_{\mathrm{p}}$) form the \emph{task space
velocity} $\mathbf{V}_{\mathrm{t}}$ vector, i.e. the velocity output. The
task space velocity is formally related to the platform twist via%
\begin{equation}
\mathbf{V}_{\mathrm{p}}=\mathbf{P}_{\mathrm{p}}\mathbf{V}_{\mathrm{t}}
\label{VpVt}
\end{equation}%
where $\mathbf{P}_{\mathrm{p}}$ is a unimodular $6\times \delta _{\mathrm{p}%
} $ velocity distribution matrix, which assigns the $\delta _{\mathrm{p}}$
components of the task space velocity to the components of the platform
twist. Assuming apropriately attached platform (end-effector) frame $%
\mathcal{F}_{\mathrm{p}}$, typical choices are%
\begin{equation}
\mathbf{P}_{\mathrm{p}}^{\mathrm{trans}}=\left( 
\begin{array}{c}
\mathbf{0}_{3,3} \\ 
\mathbf{I}_{3}%
\end{array}%
\right) ,\ \mathbf{P}_{\mathrm{p}}^{\mathrm{rot}}=\left( 
\begin{array}{c}
\mathbf{I}_{3,3} \\ 
\mathbf{o}_{3}%
\end{array}%
\right) ,\ \mathbf{P}_{\mathrm{p}}^{\mathrm{planar}}=\left( 
\begin{array}{ccc}
0 & 0 & 0 \\ 
0 & 0 & 0 \\ 
0 & 0 & 1 \\ 
1 & 0 & 0 \\ 
0 & 1 & 0 \\ 
0 & 0 & 0%
\end{array}%
\right) ,\ \mathbf{P}_{\mathrm{p}}^{\mathrm{SCARA}}=\left( 
\begin{array}{ccc}
0 & 0 & 0 \\ 
0 & 0 & 0 \\ 
0 & 0 & 1 \\ 
1 & 0 & 0 \\ 
0 & 1 & 0 \\ 
0 & 0 & 1%
\end{array}%
\right) .  \label{Pp}
\end{equation}%
Here, $\mathbf{P}_{\mathrm{p}}^{\mathrm{trans}},\mathbf{P}_{\mathrm{p}}^{%
\mathrm{rot}}$, and $\mathbf{P}_{\mathrm{p}}^{\mathrm{planar}}$ account for
spatial translations and rotations, and for planar motions ($\delta _{%
\mathrm{p}}=3$), while $\mathbf{P}_{\mathrm{p}}^{\mathrm{SCARA}}$ is used
for Sch\"{o}nflies/SCARA motion ($\delta _{\mathrm{p}}=4$). The
corresponding task space velocities are $\mathbf{V}_{\mathrm{t}}^{\mathrm{%
trans}}=\left( \mathbf{v}\right) ^{T},\mathbf{V}_{\mathrm{t}}^{\mathrm{rot}%
}=\left( \bm{\omega}\right) ^{T}$, and $\mathbf{V}_{\mathrm{t}}^{\mathrm{%
planar}}=\left( \omega _{3},v_{1},v_{2}\right) ^{T},\mathbf{V}_{\mathrm{t}}^{%
\mathrm{SCARA}}=\left( \omega _{3},\mathbf{v}\right) ^{T}$.

Now consider the separated limb $l$, including the platform. The twist of
the platform is determined with (\ref{Lpl}) in terms of the independent
joint velocities as $\mathbf{V}_{\mathrm{p}\left( l\right) }=\mathbf{L}_{%
\mathrm{p}\left( l\right) }\dot{\mathbf{q}}_{\left( l\right) }$. The
instantaneous mobility of the platform is $\mathrm{rank}\,\mathbf{L}_{%
\mathrm{p}\left( l\right) }(%
%TCIMACRO{\TeXButton{eta}{\mathbold{\eta}}}%
%BeginExpansion
\mathbold{\eta}%
%EndExpansion
_{\left( l\right) })$. The rank in general depends on the configuration. It
may change in singular configurations of the limb. Moreover, $\mathbf{L}_{%
\mathrm{p}\left( l\right) }$ may exhibit a permanent drop of rank when the
PKM is assembled. The following assumptions is made throughout the paper.
Denote with $r_{l}\leq \delta _{l}$ the maximal rank of $\mathbf{L}_{\mathrm{%
p}\left( l\right) }$, i.e. its rank at a generic configuration $%
%TCIMACRO{\TeXButton{eta}{\mathbold{\eta}}}%
%BeginExpansion
\mathbold{\eta}%
%EndExpansion
_{\left( l\right) }$.

\begin{assumption}
It is assumed that in a regular configuration $%
%TCIMACRO{\TeXButton{eta}{\mathbold{\eta}}}%
%BeginExpansion
\mathbold{\eta}%
%EndExpansion
\in V$ of the PKM, the forward kinematics Jacobian $\mathbf{L}_{\mathrm{p}%
\left( l\right) }$ has maximal rank $r_{l}$.
\end{assumption}

\begin{remark}
This assumption does not exclude overconstrained mechanisms per se, but
rather allows for exceptionally overconstrained mechanisms, of which the
Delta is a good example. The platform motion of the latter is generated by
the intersection of the motion spaces of the platform in the separated
limbs. The compound Jacobian of each limb has maximal rank 4.
\end{remark}

\subsection{Geometric Forward Kinematics of Limb Mechanism%
%TCIMACRO{\TeXButton{secGeomFKLimb}{\label{secGeomFKLimb}}}%
%BeginExpansion
\label{secGeomFKLimb}%
%EndExpansion
}

Corresponding to the velocity relation (\ref{etaq}), there is a solution of
the geometric constraints of the form $%
%TCIMACRO{\TeXButton{red}{}}%
%BeginExpansion
%
%EndExpansion
%TCIMACRO{\TeXButton{vartheta}{\mathbold{\vartheta}}}%
%BeginExpansion
\mathbold{\vartheta}%
%EndExpansion
%TCIMACRO{\TeXButton{black}{\color{black}}}%
%BeginExpansion
\color{black}%
%EndExpansion
_{\left( l\right) }=\psi _{k\left( l\right) }(\mathbf{q}_{\left( l\right) })$%
, with a mapping $\psi _{k\left( l\right) }:{\mathbb{V}}^{\delta
_{l}}\rightarrow {\mathbb{V}}^{n_{l}}$. The geometric forward kinematics
problem of the limb is then solved by inserting this into (\ref{Cp}) to
provide $\mathbf{C}_{\mathrm{p}}(\mathbf{q}_{\left( l\right) })$,
respectively $\mathbf{x}(\mathbf{q}_{\left( l\right) })$. If a closed form
solution is available, it provides a solution of the velocity constraints,
instead of (\ref{GeomConsLoopCutJoint}). However, such an explicit solutions
can be determined in closed form only in special cases, such as the 3%
\underline{R}R[2RR]R Delta, for instance, and this is usually rather
involved. In the general situation, the solution of the velocity constraints
(\ref{etaq}) gives rise to a numerical solution of the geometric
constraints. A simple and efficient way is to use the Newton step%
\begin{equation}
\Delta 
%TCIMACRO{\TeXButton{red}{}}%
%BeginExpansion
%
%EndExpansion
%TCIMACRO{\TeXButton{vartheta}{\mathbold{\vartheta}}}%
%BeginExpansion
\mathbold{\vartheta}%
%EndExpansion
%TCIMACRO{\TeXButton{black}{\color{black}}}%
%BeginExpansion
\color{black}%
%EndExpansion
_{\left( l\right) }=\mathbf{H}_{\left( l\right) }(%
%TCIMACRO{\TeXButton{red}{}}%
%BeginExpansion
%
%EndExpansion
%TCIMACRO{\TeXButton{vartheta}{\mathbold{\vartheta}}}%
%BeginExpansion
\mathbold{\vartheta}%
%EndExpansion
%TCIMACRO{\TeXButton{black}{\color{black}}}%
%BeginExpansion
\color{black}%
%EndExpansion
_{\left( l\right) })\Delta \mathbf{q}_{\left( l\right) }.
\label{GeomForwardLimb}
\end{equation}%
A solution is found as $%
%TCIMACRO{\TeXButton{red}{}}%
%BeginExpansion
%
%EndExpansion
%TCIMACRO{\TeXButton{vartheta}{\mathbold{\vartheta}}}%
%BeginExpansion
\mathbold{\vartheta}%
%EndExpansion
%TCIMACRO{\TeXButton{black}{\color{black}}}%
%BeginExpansion
\color{black}%
%EndExpansion
_{\left( l\right) }:=%
%TCIMACRO{\TeXButton{red}{}}%
%BeginExpansion
%
%EndExpansion
%TCIMACRO{\TeXButton{vartheta}{\mathbold{\vartheta}}}%
%BeginExpansion
\mathbold{\vartheta}%
%EndExpansion
%TCIMACRO{\TeXButton{black}{\color{black}}}%
%BeginExpansion
\color{black}%
%EndExpansion
_{\left( l\right) }+\Delta 
%TCIMACRO{\TeXButton{red}{}}%
%BeginExpansion
%
%EndExpansion
%TCIMACRO{\TeXButton{vartheta}{\mathbold{\vartheta}}}%
%BeginExpansion
\mathbold{\vartheta}%
%EndExpansion
%TCIMACRO{\TeXButton{black}{\color{black}}}%
%BeginExpansion
\color{black}%
%EndExpansion
_{\left( l\right) }$ after only a few iteration steps (except near
singularities, where $\mathbf{G}_{\mathbf{y}\left( l\right) }$ or $\mathbf{G}%
_{\mathbf{q}\left( l\right) }$ in (\ref{GyGq}) become ill conditioned).
Alternatively, the velocity relation (\ref{etaq}) can be numerically
integrated for given $\mathbf{q}_{\left( l\right) }\left( t\right) $.

\section{Inverse Kinematics of PKM Mechanism%
%TCIMACRO{\TeXButton{secInvKinMech}{\label{secInvKinMech}}}%
%BeginExpansion
\label{secInvKinMech}%
%EndExpansion
}

The (standard) velocity inverse kinematics problem of the PKM is to
determine the actuator velocity for a given task space velocity. The
kinematics and dynamics modeling necessitates to determine the motion of all
bodies of the PKM, however. This amounts to solving the velocity \emph{%
inverse kinematics problem of the mechanism}, which consists in finding the
independent velocities $\dot{\mathbf{q}}_{\left( l\right) }$ of all limbs in
terms of the task space velocity. The latter allow to compute the motion of
all bodies and joints via the solutions (\ref{Hlambdal2}) for the individual
FCs.

The platform DOF $\delta _{\mathrm{p}}$ of the PKM and the maximal rank $%
r_{l}$ of limb Jacobian need not be equal. The subsequent formulation for
the inverse kinematics must distinguish these two cases.

\begin{definition}
If the platform DOF of the PKM is the same as the DOF of the platform when
it is connected to the separated limb only, i.e. $\delta _{\mathrm{p}}=r_{l}$%
, then the PKM is called \emph{equimobile}.
\end{definition}

\begin{definition}
If $r_{l}<\delta _{l}$, then limb $l$ is \emph{kinematically redundant}. A
PKM is kinematically redundant if it contains a kinematically redundant limb.
\end{definition}

The limb of an equimobile PKM admits the same platform mobility when it is
separated and when the PKM is assembled, whereas a limb of a non-equimobile
PKM is subjected to additional constraints when the limbs are assembled to
the PKM. The compound platform Jacobian of a kinematically non-redundant
limb is a regular $\delta _{l}\times \delta _{l}$-matrix. In the following,
only non-redundant PKM are considered. The formulation is straightforwardly
extended to kinematically redundant PKM \cite{MuellerAMR2020}.

\subsection{Velocity and Acceleration Inverse Kinematics of Equimobile PKM}

A \emph{task space Jacobian} $\mathbf{L}_{\mathrm{t}\left( l\right) }$ of
limb $l$ is constructed by selecting the relevant $\delta _{\mathrm{p}%
}=r_{l} $ rows of the forward kinematics Jacobian $\mathbf{L}_{\mathrm{p}%
\left( l\right) }$ in (\ref{Lpl}). The task space velocity is then
determined as%
\begin{equation}
\mathbf{V}_{\mathrm{t}}=\mathbf{L}_{\mathrm{t}\left( l\right) }\dot{\mathbf{q%
}}_{\left( l\right) },\ l=1,\ldots ,L.  \label{Vt}
\end{equation}%
When the PKM is non-redundant, $\mathbf{L}_{\mathrm{t}\left( l\right) }$ is
a $\delta _{\mathrm{p}}\times \delta _{\mathrm{p}}$-matrix, and (\ref{Vt})
can be solved to obtain the velocity inverse kinematics solution for limb $i$
as%
\begin{equation}
\dot{\mathbf{q}}_{\left( l\right) }=\mathbf{F}_{\left( l\right) }\mathbf{V}_{%
\mathrm{t}},\ \ \ \mathrm{with\ \ }\mathbf{F}_{\left( l\right) }:=\mathbf{L}%
_{\mathrm{t}\left( l\right) }^{-1}.  \label{Ftheta1}
\end{equation}%
The complete inverse kinematics for limb $l$ is then obtained with (\ref%
{etaq}) as%
\begin{equation}
%TCIMACRO{\TeXButton{red}{}}%
%BeginExpansion
%
%EndExpansion
\dot{%
%TCIMACRO{\TeXButton{vartheta}{\mathbold{\vartheta}}}%
%BeginExpansion
\mathbold{\vartheta}%
%EndExpansion
}%
%TCIMACRO{\TeXButton{black}{\color{black}}}%
%BeginExpansion
\color{black}%
%EndExpansion
_{\left( l\right) }=\mathbf{H}_{\left( l\right) }\mathbf{F}_{\left( l\right)
}\mathbf{V}_{\mathrm{t}}  \label{HFl}
\end{equation}%
where $\mathbf{H}_{\left( l\right) }\mathbf{F}_{\left( l\right) }$ serves as
inverse kinematics Jacobian of the limb mechanism. The latter can be
replaced by a closed form expression if a closed form solution of the
inverse kinematics is known.

Time derivative of (\ref{HFl}) yields a solution to the acceleration inverse
kinematics problem%
\begin{eqnarray}
%TCIMACRO{\TeXButton{red}{}}%
%BeginExpansion
%
%EndExpansion
\ddot{%
%TCIMACRO{\TeXButton{vartheta}{\mathbold{\vartheta}}}%
%BeginExpansion
\mathbold{\vartheta}%
%EndExpansion
}%
%TCIMACRO{\TeXButton{black}{\color{black}}}%
%BeginExpansion
\color{black}%
%EndExpansion
_{\left( l\right) } &=&\mathbf{H}_{\left( l\right) }\mathbf{F}_{\left(
l\right) }\dot{\mathbf{V}}_{\mathrm{t}}+(\dot{\mathbf{H}}_{\left( l\right) }-%
\mathbf{H}_{\left( l\right) }\mathbf{F}_{\left( l\right) }\dot{\mathbf{L}}_{%
\mathrm{t}\left( l\right) })\mathbf{F}_{\left( l\right) }\mathbf{V}_{\mathrm{%
t}}  \label{eta2d} \\
&=&\mathbf{H}_{\left( l\right) }\mathbf{F}_{\left( l\right) }\dot{\mathbf{V}}%
_{\mathrm{t}}+(\dot{\mathbf{H}}_{\left( l\right) }-\mathbf{H}_{\left(
l\right) }\mathbf{F}_{\left( l\right) }\dot{\mathbf{L}}_{\mathrm{t}\left(
l\right) })\dot{\mathbf{q}}_{\left( l\right) }  \notag
\end{eqnarray}%
with $\dot{\mathbf{H}}_{\left( l\right) }$ in (\ref{Hldot}), and $\dot{%
\mathbf{L}}_{\mathrm{t}\left( l\right) }$ is found with (\ref{Lpl}) as%
\begin{equation}
\dot{\mathbf{L}}_{\mathrm{p}\left( l\right) }(%
%TCIMACRO{\TeXButton{red}{}}%
%BeginExpansion
%
%EndExpansion
%TCIMACRO{\TeXButton{vartheta}{\mathbold{\vartheta}}}%
%BeginExpansion
\mathbold{\vartheta}%
%EndExpansion
%TCIMACRO{\TeXButton{black}{\color{black}}}%
%BeginExpansion
\color{black}%
%EndExpansion
_{\left( l\right) },%
%TCIMACRO{\TeXButton{red}{}}%
%BeginExpansion
%
%EndExpansion
\dot{%
%TCIMACRO{\TeXButton{vartheta}{\mathbold{\vartheta}}}%
%BeginExpansion
\mathbold{\vartheta}%
%EndExpansion
}%
%TCIMACRO{\TeXButton{black}{\color{black}}}%
%BeginExpansion
\color{black}%
%EndExpansion
_{\left( l\right) })=\dot{\mathbf{J}}_{\mathrm{p}\left( l\right) }\mathbf{H}%
_{\left( l\right) }+\mathbf{J}_{\mathrm{p}\left( l\right) }\dot{\mathbf{H}}%
_{\left( l\right) }.
\end{equation}

\subsection{Velocity and Acceleration Inverse Kinematics of Non-Equimobile
PKM}

The rank $r_{l}$ of the compound forward kinematics Jacobian $\mathbf{L}_{%
\mathrm{p}\left( l\right) }$ of limb $l$ of a non-equimobile PKM exceeds the
platform DOF $\delta _{\mathrm{p}}$. The task space Jacobian $\mathbf{L}_{%
\mathrm{t}\left( l\right) }$ of limb $l$ is constructed by selecting $r_{l}$
rows of $\mathbf{L}_{\mathrm{p}\left( l\right) }$. Then, only $\delta _{%
\mathrm{p}}$ components of $\mathbf{L}_{\mathrm{t}\left( l\right) }\dot{%
\mathbf{q}}_{\left( l\right) }$ correspond to the components of $\mathbf{V}_{%
\mathrm{t}}$, while the remaining $r_{l}-\delta _{\mathrm{p}}$ equations
represent constraints on the platform twist. The latter correspond to
constraints imposed on the motion of limb $l$. This is formalized with help
of a $r_{l}\times \delta _{\mathrm{p}}$ velocity distribution matrix $%
\mathbf{D}_{\mathrm{t}\left( l\right) }$, which assigns the components of
the task space velocity $\mathbf{V}_{\mathrm{t}}$ to the relevant rows of
the task space Jacobian of limb $l$. The equation%
\begin{equation}
\mathbf{D}_{\mathrm{t}\left( l\right) }\mathbf{V}_{\mathrm{t}}=\mathbf{L}_{%
\mathrm{t}\left( l\right) }\dot{\mathbf{q}}_{\left( l\right) },\ l=1,\ldots
,L  \label{VpD}
\end{equation}%
then summarizes the forward kinematics of the limb as well as the imposed
constraints.

The task space Jacobian $\mathbf{L}_{\mathrm{t}\left( l\right) }$ of a
kinematically non-redundant limb (i.e. $r_{l}=\delta _{l}$) is an invertible 
$r_{l}\times r_{l}$ matrix, and the solution of the velocity inverse
kinematics is%
\begin{equation}
\dot{\mathbf{q}}_{\left( l\right) }=\mathbf{F}_{l}\mathbf{V}_{\mathrm{t}},\
\ \ \mathrm{with\ \ }\mathbf{F}_{\left( l\right) }=\mathbf{L}_{\mathrm{t}%
\left( l\right) }^{-1}\mathbf{D}_{\mathrm{t}\left( l\right) }.
\label{Ftheta2}
\end{equation}%
The overall solution for the velocity inverse kinematics of the limb
mechanism is as in (\ref{HFl}).

\begin{example}[{3\protect\underline{R}R[2RR]R Delta --cont.}]
The platform of the 3-DOF 3\underline{R}R[2RR]R Delta can only translate,
and its DOF is $\delta _{\mathrm{p}}=3$. The task space velocity vector $%
\mathbf{V}_{\mathrm{t}}=\left( \mathbf{v}_{\mathrm{p}}\right) \in {\mathbb{R}%
}^{3}$ consists of the three components of the EE velocity $\mathbf{v}_{%
\mathrm{p}}$. The forward kinematics Jacobian of a limb has rank $r_{l}=4$,
which is equal to the DOF $\delta _{l}$ of the separated limb, and the PKM
is not equimobile. The platform of a separated limb can perform Sch\"{o}%
nflies motion, i.e. spatial translations plus an independent rotation about
an axis parallel to the axis of joint 1 (which is parallel to axes of joint
2 and 6). Expressed in the platform frame $\mathcal{F}_{\mathrm{p}}$ shown
in fig. \ref{figRR2RRRDelta}b), the 2-component of the angular velocity is
non-zero, and thus rows 2,4,5, and 6 are non-zero and are used to construct
the $4\times 4$ task space Jacobian $\mathbf{L}_{\mathrm{t}\left( l\right) }$%
. The selection matrix extracting the translation part of the platform twist
is (the zero row enforces the angular velocity be zero)%
\begin{equation}
\mathbf{D}_{\mathrm{t}\left( l\right) }=\left( 
\begin{array}{c}
\mathbf{0}_{1,3} \\ 
\mathbf{I}_{3}%
\end{array}%
\right) ,\ l=1,2,3.  \label{DtDelta}
\end{equation}
\end{example}

\begin{example}[IRSBot-2 --cont.]
%TCIMACRO{\TeXButton{IRSBot2-IK}{\label{IRSBot2-IK}}}%
%BeginExpansion
\label{IRSBot2-IK}%
%EndExpansion
The DOF of a separated limb is $\delta _{l}=3$, and its motion is
parameterized with $\mathbf{q}_{\left( l\right) }=\left( \vartheta
_{1},\vartheta _{5,1},\vartheta _{5,2}\right) ^{T}$. The platform motion is
due to the translation of the parallelogram loop in the 1-3-plane of the
platform frame $\mathcal{F}_{\mathrm{p}}$ shown in fig. \ref{figIRSBot}b)
combined with the rotation about the normal to the plane defined by the 4
U-joints and the translation along this normal. The 3-component of the
angular velocity vector $^{\mathrm{p}}\bm{\omega}$ is used to represent the
platform rotation. The task space Jacobian of the limb is accordingly
constructed from columns 3,4 and 6. When the limbs are assembled, the
platform of the 2-DOF IRSBot-2 can only perform planar translations ($\delta
_{\mathrm{p}}=2$) in the 1-3-plane of the platform frame. The task space
velocity vector is $\mathbf{V}_{\mathrm{t}}=\left( v_{1},v_{3}\right) ^{T}$.
The selection matrix assigning the two translation components is (the zero
row enforces the angular velocity be zero)%
\begin{equation}
\mathbf{D}_{\mathrm{t}\left( l\right) }=\left( 
\begin{array}{c}
\mathbf{0}_{1,2} \\ 
\mathbf{I}_{2}%
\end{array}%
\right) ,\ l=1,2.
\end{equation}
\end{example}

\subsection{Velocity Inverse Kinematics of general PKM}

In the preceding example \ref{IRSBot2-IK}, it must be noted that the
selection of $\omega _{3}$ fails to represent the angular motion when the 4U
loop is aligned vertical. This may not be relevant for practical
applications, but is is a good example for the fact that, when the forward
kinematics Jacobian (\ref{Lkl}) represented in platform frame does not
comprise exactly $r_{l}$ non-zero components, the selection of $r_{l}$ rows
may introduce singularities. In such cases the pseudoinverse solution can
always be used without preselection of a taskspace Jacobian. The unique
solution of $\mathbf{V}_{\mathrm{p}}=\mathbf{L}_{\mathrm{p}\left( l\right) }%
\dot{\mathbf{q}}_{\left( l\right) }$ is the inverse kinematics solution for
limb $l$ in terms of the left pseudoinverse $\mathbf{L}^{+}=\left( \mathbf{L}%
^{T}\mathbf{L}\right) ^{-1}\mathbf{L}$ 
\begin{equation}
\dot{\mathbf{q}}_{\left( l\right) }=\mathbf{L}_{\mathrm{p}\left( l\right)
}^{+}\mathbf{V}_{\mathrm{p}}  \label{Ftheta3}
\end{equation}%
which is then used in (\ref{HFl}).

\subsection{Geometric Inverse Kinematics of the Limb Mechanism}

The geometric inverse kinematics problem of the limb mechanism is to
determine the joint variables $%
%TCIMACRO{\TeXButton{red}{}}%
%BeginExpansion
%
%EndExpansion
%TCIMACRO{\TeXButton{vartheta}{\mathbold{\vartheta}}}%
%BeginExpansion
\mathbold{\vartheta}%
%EndExpansion
%TCIMACRO{\TeXButton{black}{\color{black}}}%
%BeginExpansion
\color{black}%
%EndExpansion
_{\left( l\right) }$ for given taskspace coordinates $\mathbf{x}$, i.e.
evaluation of the inverse kinematics mapping 
\begin{equation}
\psi _{\mathrm{IK}\left( l\right) }:{\mathbb{V}}^{\delta _{\mathrm{p}%
}}\rightarrow {\mathbb{V}}^{n_{l}},\ \ 
%TCIMACRO{\TeXButton{red}{}}%
%BeginExpansion
%
%EndExpansion
%TCIMACRO{\TeXButton{vartheta}{\mathbold{\vartheta}}}%
%BeginExpansion
\mathbold{\vartheta}%
%EndExpansion
%TCIMACRO{\TeXButton{black}{\color{black}}}%
%BeginExpansion
\color{black}%
%EndExpansion
_{\left( l\right) }=\psi _{\mathrm{IK}\left( l\right) }\left( \mathbf{x}%
\right) .  \label{GeomInvLimb}
\end{equation}%
For certain PKM, the inverse kinematics map $\psi _{\mathrm{IK}}$ can be
derived in closed form, as for the 3\underline{R}R[2RR]R Delta and the
IRSBot-2, for instance. In the general case, when this cannot (or is too
complicated to) be expressed in closed form, a solution can be obtained
numerically. The Jacobian of $\psi _{\mathrm{IK}\left( l\right) }$ is given
by $\mathbf{F}_{\left( l\right) }\mathbf{H}_{\left( l\right) }$ in (\ref{HFl}%
). The inverse kinematics map (\ref{GeomForwardLimb}) can be evaluated by
means of the simple iteration scheme,%
\begin{equation}
\Delta 
%TCIMACRO{\TeXButton{red}{}}%
%BeginExpansion
%
%EndExpansion
%TCIMACRO{\TeXButton{vartheta}{\mathbold{\vartheta}}}%
%BeginExpansion
\mathbold{\vartheta}%
%EndExpansion
%TCIMACRO{\TeXButton{black}{\color{black}}}%
%BeginExpansion
\color{black}%
%EndExpansion
_{\left( l\right) }=\mathbf{H}_{\left( l\right) }(%
%TCIMACRO{\TeXButton{red}{}}%
%BeginExpansion
%
%EndExpansion
%TCIMACRO{\TeXButton{vartheta}{\mathbold{\vartheta}}}%
%BeginExpansion
\mathbold{\vartheta}%
%EndExpansion
%TCIMACRO{\TeXButton{black}{\color{black}}}%
%BeginExpansion
\color{black}%
%EndExpansion
_{\left( l\right) })\mathbf{F}_{\left( l\right) }(%
%TCIMACRO{\TeXButton{red}{}}%
%BeginExpansion
%
%EndExpansion
%TCIMACRO{\TeXButton{vartheta}{\mathbold{\vartheta}}}%
%BeginExpansion
\mathbold{\vartheta}%
%EndExpansion
%TCIMACRO{\TeXButton{black}{\color{black}}}%
%BeginExpansion
\color{black}%
%EndExpansion
_{\left( l\right) })\Delta \mathbf{x}  \label{qetaIteration}
\end{equation}%
which yields the update $%
%TCIMACRO{\TeXButton{red}{}}%
%BeginExpansion
%
%EndExpansion
%TCIMACRO{\TeXButton{vartheta}{\mathbold{\vartheta}}}%
%BeginExpansion
\mathbold{\vartheta}%
%EndExpansion
%TCIMACRO{\TeXButton{black}{\color{black}}}%
%BeginExpansion
\color{black}%
%EndExpansion
_{\left( l\right) }:=%
%TCIMACRO{\TeXButton{red}{}}%
%BeginExpansion
%
%EndExpansion
%TCIMACRO{\TeXButton{vartheta}{\mathbold{\vartheta}}}%
%BeginExpansion
\mathbold{\vartheta}%
%EndExpansion
%TCIMACRO{\TeXButton{black}{\color{black}}}%
%BeginExpansion
\color{black}%
%EndExpansion
_{\left( l\right) }+\Delta 
%TCIMACRO{\TeXButton{red}{}}%
%BeginExpansion
%
%EndExpansion
%TCIMACRO{\TeXButton{vartheta}{\mathbold{\vartheta}}}%
%BeginExpansion
\mathbold{\vartheta}%
%EndExpansion
%TCIMACRO{\TeXButton{black}{\color{black}}}%
%BeginExpansion
\color{black}%
%EndExpansion
_{\left( l\right) }$. This iteration step is repeated until a solution with
the desired precision is obtained. When using (\ref{qetaIteration}), the
taskspace coordinates $\mathbf{x}$ are the canonical coordinates according
to the representation of $\mathbf{V}_{\mathrm{t}}$ in $\mathcal{F}_{\mathrm{p%
}}$.

\section{Inverse Kinematics of PKM%
%TCIMACRO{\TeXButton{secInvKin}{\label{secInvKin}}}%
%BeginExpansion
\label{secInvKin}%
%EndExpansion
}

The velocity inverse kinematics problem is to determine the velocity of the
actuated joints for given task space velocity. Assuming a fully actuated
PKM, the number of actuator variables $n_{\mathrm{act}}$ is equal to or
greater then the DOF $\delta $ of the PKM. A subset $%
%TCIMACRO{\TeXButton{vartheta}{\mathbold{\vartheta}}}%
%BeginExpansion
\mathbold{\vartheta}%
%EndExpansion
_{\left( l\right) \mathrm{act}}$ of $n_{\mathrm{act}\left( l\right) }$
variables of the independent coordinates $\mathbf{q}_{\left( l\right)
},l=1,\ldots ,L$ of the limbs corresponds to the actuated joints. The
tree-topology system can always be introduced so that the variables of
actuated joints are contained in $\mathbf{q}_{\left( l\right) }$. Denote the
overall vector of actuated joint coordinates with $%
%TCIMACRO{\TeXButton{vartheta}{\mathbold{\vartheta}}}%
%BeginExpansion
\mathbold{\vartheta}%
%EndExpansion
_{\mathrm{act}}\in {\mathbb{V}}^{n_{\mathrm{act}}}$. If $n_{\mathrm{act}%
}>\delta $, the PKM is called \emph{redundandantly actuated} \cite%
{MuellerRobotica2013}. For a non-redundantly actuated PKM, the actuator
coordinates represent generalized coordinates, which are usually denoted
with $\mathbf{q}:=%
%TCIMACRO{\TeXButton{vartheta}{\mathbold{\vartheta}}}%
%BeginExpansion
\mathbold{\vartheta}%
%EndExpansion
_{\mathrm{act}}\in {\mathbb{V}}^{\delta }$.

The actuator velocities of limb $l$ are readily obtained as%
\begin{equation}
\dot{%
%TCIMACRO{\TeXButton{vartheta}{\mathbold{\vartheta}}}%
%BeginExpansion
\mathbold{\vartheta}%
%EndExpansion
}_{\left( l\right) \mathrm{act}}=\mathbf{J}_{\mathrm{IK}\left( l\right) }(%
%TCIMACRO{\TeXButton{red}{}}%
%BeginExpansion
%
%EndExpansion
%TCIMACRO{\TeXButton{vartheta}{\mathbold{\vartheta}}}%
%BeginExpansion
\mathbold{\vartheta}%
%EndExpansion
%TCIMACRO{\TeXButton{black}{\color{black}}}%
%BeginExpansion
\color{black}%
%EndExpansion
_{\left( l\right) })\mathbf{V}_{\mathrm{t}}  \label{IKLimb}
\end{equation}%
where $\mathbf{J}_{\mathrm{IK}\left( l\right) }$ consists of the $n_{\mathrm{%
act}\left( l\right) }$ rows of matrix $\mathbf{F}_{\left( l\right) }$ in (%
\ref{Ftheta1}), (\ref{Ftheta2}) or (\ref{Ftheta3}). The velocity inverse
kinematics solution of the PKM is then $\dot{%
%TCIMACRO{\TeXButton{vartheta}{\mathbold{\vartheta}}}%
%BeginExpansion
\mathbold{\vartheta}%
%EndExpansion
}_{\mathrm{act}}=\mathbf{J}_{\mathrm{IK}}\mathbf{V}_{\mathrm{t}}$, with the $%
n_{\mathrm{act}}\times \delta _{\mathrm{p}}$ \emph{inverse kinematics
Jacobian of the manipulator} 
\begin{equation}
\mathbf{J}_{\mathrm{IK}}(%
%TCIMACRO{\TeXButton{red}{}}%
%BeginExpansion
%
%EndExpansion
%TCIMACRO{\TeXButton{vartheta}{\mathbold{\vartheta}}}%
%BeginExpansion
\mathbold{\vartheta}%
%EndExpansion
%TCIMACRO{\TeXButton{black}{\color{black}}}%
%BeginExpansion
\color{black}%
%EndExpansion
):=\left( 
\begin{array}{c}
\mathbf{J}_{\mathrm{IK}\left( 1\right) }(%
%TCIMACRO{\TeXButton{red}{}}%
%BeginExpansion
%
%EndExpansion
%TCIMACRO{\TeXButton{vartheta}{\mathbold{\vartheta}}}%
%BeginExpansion
\mathbold{\vartheta}%
%EndExpansion
%TCIMACRO{\TeXButton{black}{\color{black}}}%
%BeginExpansion
\color{black}%
%EndExpansion
_{\left( 1\right) }) \\ 
\vdots \\ 
\mathbf{J}_{\mathrm{IK}\left( L\right) }(%
%TCIMACRO{\TeXButton{red}{}}%
%BeginExpansion
%
%EndExpansion
%TCIMACRO{\TeXButton{vartheta}{\mathbold{\vartheta}}}%
%BeginExpansion
\mathbold{\vartheta}%
%EndExpansion
%TCIMACRO{\TeXButton{black}{\color{black}}}%
%BeginExpansion
\color{black}%
%EndExpansion
_{\left( L\right) })%
\end{array}%
\right) .  \label{IK}
\end{equation}%
The geometric inverse kinematics problem of the PKM consists in finding the
actuator coordinates $%
%TCIMACRO{\TeXButton{vartheta}{\mathbold{\vartheta}}}%
%BeginExpansion
\mathbold{\vartheta}%
%EndExpansion
_{\left( l\right) \mathrm{act}}$ for given taskspace coordinates $\mathbf{x}$%
, or platform pose $\mathbf{C}_{\mathrm{p}}$. This boils down to determine
the closed form expression of the inverse kinematics map $f_{\mathrm{IK}}:{%
\mathbb{V}}^{\delta _{\mathrm{p}}}\rightarrow {\mathbb{V}}^{n_{\mathrm{act}%
}} $, which satisfies $%
%TCIMACRO{\TeXButton{vartheta}{\mathbold{\vartheta}}}%
%BeginExpansion
\mathbold{\vartheta}%
%EndExpansion
_{\mathrm{act}}=f_{\mathrm{IK}}(\mathbf{x})$. Such explicit expressions are
available if the inverse kinematics of the limb mechanism can be solved as $%
\mathbf{q}_{\left( l\right) }=\mathbf{q}_{\left( l\right) }(\mathbf{x})$,
e.g. for the 3\underline{R}R[2RR]R Delta. If no closed form relations are
available, the numerical solution is already known with (\ref{qetaIteration}%
).

\begin{remark}
The DOF of the PKM and/or the assignment of actuators may be different in
different motion modes. If a PKM is operated with different motion modes, it
must be switched between the corresponding kinematic models.
\end{remark}

\section{Dynamic Equations of Motion (EOM)%
%TCIMACRO{\TeXButton{secEOM}{\label{secEOM}}}%
%BeginExpansion
\label{secEOM}%
%EndExpansion
}

In this section a task space formulation of the EOM for a PKM with complex
limbs is derived, i.e. EOM in terms of the task space velocity $\mathbf{V}_{%
\mathrm{t}}$ and acceleration. To this end, the platform is removed so to
obtain a tree-topology system with separated complex limbs not containing
the platform. For these limbs, the dynamic EOM are formulated. To this end,
the EOM of an associated tree-topology system is formulated and the loop
closure constraints are subsequently enforced employing the inverse
kinematics solution of the mechanism. These EOM along with the EOM of the
platform body give rise to the overall system of EOM for the PKM. This
approach gives rise to a systematic method for deriving EOM of PKM with
complex limbs that is easily implemented and is computationally efficient.
It facilitates flexible use of the dynamic formulation of the EOM of
tree-topology systems that is deemed most appropriate. It further admits
using recursive $O\left( n\right) $ evaluation and parallel/distributed
computation.

\subsection{Kinematics of a complex limb without platform}

\begin{figure}[b]
\centerline{
a)~\includegraphics[height=4.3cm]{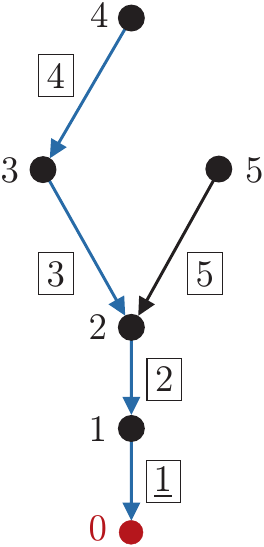}\hfil b)~\includegraphics[height=4.3cm]{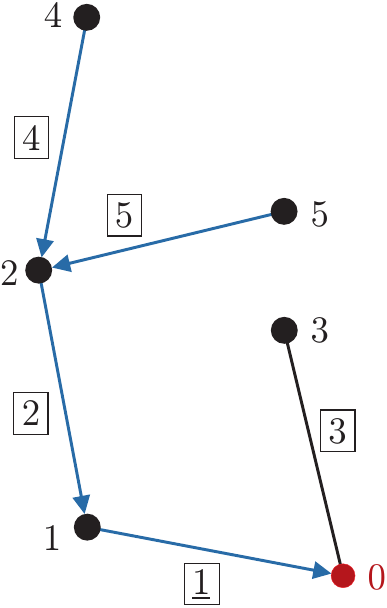}}
\caption{Directed graph of the tree-topology system for a limb of a) the 3%
\protect\underline{R}R[2RR]R Delta and b) the IRSBot-2, obtained after
removing the platform.}
\label{figTreeNoPlatform}
\end{figure}

A tree-topology system is obtained from the tree-topology according to $\vec{%
G}$ by eliminating the platform and all joints connecting it to the limbs.
This is shown in fig. \ref{figTreeNoPlatform} for the 3\underline{R}R[2RR]R
Delta and the IRSBot-2. Each limb gives rise to a tree-topology system where
the last joint connecting the limb to the platform removed. The remaining $%
\bar{n}_{l}<n_{l}$ joint variables, summarized in $%
%TCIMACRO{\TeXButton{red}{}}%
%BeginExpansion
%
%EndExpansion
\bar{%
%TCIMACRO{\TeXButton{vartheta}{\mathbold{\vartheta}}}%
%BeginExpansion
\mathbold{\vartheta}%
%EndExpansion
}%
%TCIMACRO{\TeXButton{black}{\color{black}}}%
%BeginExpansion
\color{black}%
%EndExpansion
_{\left( l\right) }\in {\mathbb{V}}^{\bar{n}_{l}}$, determine the
configuration of this tree-topology system. The vector of the remaining
independent joint variables is denoted with $\bar{\mathbf{q}}$ (possibly the
same as $\mathbf{q}$). The joint velocities $%
%TCIMACRO{\TeXButton{red}{}}%
%BeginExpansion
%
%EndExpansion
\dot{%
%TCIMACRO{\TeXButton{vartheta}{\mathbold{\vartheta}}}%
%BeginExpansion
\mathbold{\vartheta}%
%EndExpansion
}%
%TCIMACRO{\TeXButton{black}{\color{black}}}%
%BeginExpansion
\color{black}%
%EndExpansion
_{\left( l\right) }$ are determined in terms of $\dot{\mathbf{q}}_{\left(
l\right) }$ by the solution (\ref{etaq}) of the loop constraints. Denoting
with $\bar{\mathbf{H}}_{\left( l\right) }$ the submatrix of $\mathbf{H}%
_{\left( l\right) }$ with the rows corresponding to the remaining $\bar{n}%
_{l}$ joint variables, then%
\begin{equation}
%TCIMACRO{\TeXButton{red}{}}%
%BeginExpansion
%
%EndExpansion
\dot{\bar{%
%TCIMACRO{\TeXButton{vartheta}{\mathbold{\vartheta}}}%
%BeginExpansion
\mathbold{\vartheta}%
%EndExpansion
}}%
%TCIMACRO{\TeXButton{black}{\color{black}}}%
%BeginExpansion
\color{black}%
%EndExpansion
_{\left( l\right) }=\bar{\mathbf{H}}_{\left( l\right) }\dot{\bar{\mathbf{q}}}%
_{\left( l\right) }.  \label{etaq2}
\end{equation}%

\subsection{EOM of the tree-topology system of a separated limb%
%TCIMACRO{\TeXButton{secEOMLimb}{\label{secEOMLimb}}}%
%BeginExpansion
\label{secEOMLimb}%
%EndExpansion
}

The dynamic EOM of the tree-topology system of limb $l$ without platform can
be written in terms of the coordinates $%
%TCIMACRO{\TeXButton{red}{}}%
%BeginExpansion
%
%EndExpansion
\bar{%
%TCIMACRO{\TeXButton{vartheta}{\mathbold{\vartheta}}}%
%BeginExpansion
\mathbold{\vartheta}%
%EndExpansion
}%
%TCIMACRO{\TeXButton{black}{\color{black}}}%
%BeginExpansion
\color{black}%
%EndExpansion
_{\left( l\right) }$ as%
\begin{equation}
\bar{\mathbf{M}}_{\left( l\right) }%
%TCIMACRO{\TeXButton{red}{}}%
%BeginExpansion
%
%EndExpansion
\ddot{\bar{%
%TCIMACRO{\TeXButton{vartheta}{\mathbold{\vartheta}}}%
%BeginExpansion
\mathbold{\vartheta}%
%EndExpansion
}}%
%TCIMACRO{\TeXButton{black}{\color{black}}}%
%BeginExpansion
\color{black}%
%EndExpansion
_{\left( l\right) }+\bar{\mathbf{C}}_{\left( l\right) }%
%TCIMACRO{\TeXButton{red}{}}%
%BeginExpansion
%
%EndExpansion
\dot{\bar{%
%TCIMACRO{\TeXButton{vartheta}{\mathbold{\vartheta}}}%
%BeginExpansion
\mathbold{\vartheta}%
%EndExpansion
}}%
%TCIMACRO{\TeXButton{black}{\color{black}}}%
%BeginExpansion
\color{black}%
%EndExpansion
_{\left( l\right) }+\bar{\mathbf{Q}}_{\left( l\right) }^{\mathrm{grav}}+\bar{%
\mathbf{Q}}_{\left( l\right) }=\bar{\mathbf{Q}}_{\left( l\right) }^{\mathrm{%
act}}  \label{EOMLimb}
\end{equation}%
where $\bar{\mathbf{M}}_{\left( l\right) }(%
%TCIMACRO{\TeXButton{red}{}}%
%BeginExpansion
%
%EndExpansion
\bar{%
%TCIMACRO{\TeXButton{vartheta}{\mathbold{\vartheta}}}%
%BeginExpansion
\mathbold{\vartheta}%
%EndExpansion
}%
%TCIMACRO{\TeXButton{black}{\color{black}}}%
%BeginExpansion
\color{black}%
%EndExpansion
_{\left( l\right) })$ is the generalized mass matrix, $\bar{\mathbf{C}}%
_{\left( l\right) }(%
%TCIMACRO{\TeXButton{red}{}}%
%BeginExpansion
%
%EndExpansion
\dot{\bar{%
%TCIMACRO{\TeXButton{vartheta}{\mathbold{\vartheta}}}%
%BeginExpansion
\mathbold{\vartheta}%
%EndExpansion
}}%
%TCIMACRO{\TeXButton{black}{\color{black}}}%
%BeginExpansion
\color{black}%
%EndExpansion
_{\left( l\right) },%
%TCIMACRO{\TeXButton{red}{}}%
%BeginExpansion
%
%EndExpansion
\dot{%
%TCIMACRO{\TeXButton{vartheta}{\mathbold{\vartheta}}}%
%BeginExpansion
\mathbold{\vartheta}%
%EndExpansion
}%
%TCIMACRO{\TeXButton{black}{\color{black}}}%
%BeginExpansion
\color{black}%
%EndExpansion
_{\left( l\right) })$ is the generalized Coriolis/centrifugal matrix, and $%
\bar{\mathbf{Q}}_{\left( l\right) }^{\mathrm{grav}}(%
%TCIMACRO{\TeXButton{red}{}}%
%BeginExpansion
%
%EndExpansion
\bar{%
%TCIMACRO{\TeXButton{vartheta}{\mathbold{\vartheta}}}%
%BeginExpansion
\mathbold{\vartheta}%
%EndExpansion
}%
%TCIMACRO{\TeXButton{black}{\color{black}}}%
%BeginExpansion
\color{black}%
%EndExpansion
_{\left( l\right) })$ are the generalized forces due to gravity. The vector
of generalized forced $\bar{\mathbf{Q}}_{\left( l\right) }(%
%TCIMACRO{\TeXButton{red}{}}%
%BeginExpansion
%
%EndExpansion
\dot{\bar{%
%TCIMACRO{\TeXButton{vartheta}{\mathbold{\vartheta}}}%
%BeginExpansion
\mathbold{\vartheta}%
%EndExpansion
}}%
%TCIMACRO{\TeXButton{black}{\color{black}}}%
%BeginExpansion
\color{black}%
%EndExpansion
_{\left( l\right) },%
%TCIMACRO{\TeXButton{red}{}}%
%BeginExpansion
%
%EndExpansion
\bar{%
%TCIMACRO{\TeXButton{vartheta}{\mathbold{\vartheta}}}%
%BeginExpansion
\mathbold{\vartheta}%
%EndExpansion
}%
%TCIMACRO{\TeXButton{black}{\color{black}}}%
%BeginExpansion
\color{black}%
%EndExpansion
_{\left( l\right) })$ accounts for joint friction, elastic forces, and all
other effects, and the non-zero entries in $\bar{\mathbf{Q}}_{\left(
l\right) }^{\mathrm{act}}\left( t\right) $ are the drive forces/torques
collocated to the $n_{\mathrm{act}}$ variables of actuated joints. In
appendix \ref{AppendixEOMTree}, the Lie group formulation from \cite%
{MUBOScrew1,MUBOScrew2} is summarized, which allows deriving (\ref{EOMLimb})
without resorting to specific parameterization of joint geometries.

There are various ways to derive the EOM of multibody systems, such as the
limb of a PKM, in terms of relative coordinates (joint variables). So-called
matrix methods \cite{UickerRavaniSheth2013}, in particular, provide
systematic formulations that can be easily implemented and are
computationally efficient at the same time. This was formalized by the
spatial operator algebra approach (SOA) \cite%
{Rodriguez1991,Jain1991,JainBook}, which gave rise to efficient $O\left(
n\right) $-algorithms \cite{Featherstone2008}. Taking into account the
geometric nature of the EOM, Lie group methods for MBS dynamics have been
proposed in \cite{ParkBobrowPloen1995}, which represent a geometric version
of the SOA formulation. Lie group formulations for general MBS with
tree-topology were reported in \cite{ParkKim2000,MUBOScrew1,MUBOScrew2},
where the closed form EOM are determined on system level in terms of simple
matrix operations. Also recursive $O\left( n\right) $-algorithms were
reported \cite{ParkBobrowPloen1995,ICRA2017,RAL2020}. They are already an
established approach for robotic systems \cite{ModernRobotics}. Lie group
methods provide powerful tools which not only allow computationally
efficient evaluation of the EOM but also enable efficient formulations of
the linearized equations (for stability analysis), parameter sensitivity, as
well as higher-order derivatives, which are crucial for model-based control
and optimization \cite{ICRA2017,RAL2020}. In this context, the conceptual
similarity of the {natural orthogonal complement (NOC) approach \cite%
{AngelesLee1988,AngelesBook} and the Lie group method }should be mentioned.

In order to account for general formulations, and to allow for using
recursive algorithms to evaluate the equations (\ref{EOMLimb}), the left
hand side of the ODE system is written as%
\begin{equation}
\varphi _{\left( l\right) }(%
%TCIMACRO{\TeXButton{red}{}}%
%BeginExpansion
%
%EndExpansion
\bar{%
%TCIMACRO{\TeXButton{vartheta}{\mathbold{\vartheta}}}%
%BeginExpansion
\mathbold{\vartheta}%
%EndExpansion
}%
%TCIMACRO{\TeXButton{black}{\color{black}}}%
%BeginExpansion
\color{black}%
%EndExpansion
_{\left( l\right) },%
%TCIMACRO{\TeXButton{red}{}}%
%BeginExpansion
%
%EndExpansion
\dot{\bar{%
%TCIMACRO{\TeXButton{vartheta}{\mathbold{\vartheta}}}%
%BeginExpansion
\mathbold{\vartheta}%
%EndExpansion
}}%
%TCIMACRO{\TeXButton{black}{\color{black}}}%
%BeginExpansion
\color{black}%
%EndExpansion
_{\left( l\right) },%
%TCIMACRO{\TeXButton{red}{}}%
%BeginExpansion
%
%EndExpansion
\ddot{\bar{%
%TCIMACRO{\TeXButton{vartheta}{\mathbold{\vartheta}}}%
%BeginExpansion
\mathbold{\vartheta}%
%EndExpansion
}}%
%TCIMACRO{\TeXButton{black}{\color{black}}}%
%BeginExpansion
\color{black}%
%EndExpansion
_{\left( l\right) }):=\bar{\mathbf{M}}_{\left( l\right) }%
%TCIMACRO{\TeXButton{red}{}}%
%BeginExpansion
%
%EndExpansion
\ddot{\bar{%
%TCIMACRO{\TeXButton{vartheta}{\mathbold{\vartheta}}}%
%BeginExpansion
\mathbold{\vartheta}%
%EndExpansion
}}%
%TCIMACRO{\TeXButton{black}{\color{black}}}%
%BeginExpansion
\color{black}%
%EndExpansion
_{\left( l\right) }+\bar{\mathbf{C}}_{\left( l\right) }%
%TCIMACRO{\TeXButton{red}{}}%
%BeginExpansion
%
%EndExpansion
\dot{\bar{%
%TCIMACRO{\TeXButton{vartheta}{\mathbold{\vartheta}}}%
%BeginExpansion
\mathbold{\vartheta}%
%EndExpansion
}}%
%TCIMACRO{\TeXButton{black}{\color{black}}}%
%BeginExpansion
\color{black}%
%EndExpansion
_{\left( l\right) }+\bar{\mathbf{Q}}_{\left( l\right) }^{\mathrm{grav}}+\bar{%
\mathbf{Q}}_{\left( l\right) }.  \label{phil}
\end{equation}%
This admits flexible use of alternative formulations or algorithms to
evaluate the dynamics of the $L$ limbs. The EOM of the limbs are
occasionally simplified, and then the term $\varphi _{\left( l\right) }(%
%TCIMACRO{\TeXButton{red}{}}%
%BeginExpansion
%
%EndExpansion
\bar{%
%TCIMACRO{\TeXButton{vartheta}{\mathbold{\vartheta}}}%
%BeginExpansion
\mathbold{\vartheta}%
%EndExpansion
}%
%TCIMACRO{\TeXButton{black}{\color{black}}}%
%BeginExpansion
\color{black}%
%EndExpansion
_{\left( l\right) },%
%TCIMACRO{\TeXButton{red}{}}%
%BeginExpansion
%
%EndExpansion
\dot{\bar{%
%TCIMACRO{\TeXButton{vartheta}{\mathbold{\vartheta}}}%
%BeginExpansion
\mathbold{\vartheta}%
%EndExpansion
}}%
%TCIMACRO{\TeXButton{black}{\color{black}}}%
%BeginExpansion
\color{black}%
%EndExpansion
_{\left( l\right) },%
%TCIMACRO{\TeXButton{red}{}}%
%BeginExpansion
%
%EndExpansion
\ddot{\bar{%
%TCIMACRO{\TeXButton{vartheta}{\mathbold{\vartheta}}}%
%BeginExpansion
\mathbold{\vartheta}%
%EndExpansion
}}%
%TCIMACRO{\TeXButton{black}{\color{black}}}%
%BeginExpansion
\color{black}%
%EndExpansion
_{\left( l\right) })$ is substituted accordingly. In \cite%
{Pierrot1990,ChemoriPierrot2018}, for instance, the lower arm segments
(forming the parallelogram) of Delta-like PKM were split into two halves,
and their mass was added to the upper arm and the platform, respectively.

\subsection{EOM of a complex limb}

The coordinates $%
%TCIMACRO{\TeXButton{red}{}}%
%BeginExpansion
%
%EndExpansion
\bar{%
%TCIMACRO{\TeXButton{vartheta}{\mathbold{\vartheta}}}%
%BeginExpansion
\mathbold{\vartheta}%
%EndExpansion
}%
%TCIMACRO{\TeXButton{black}{\color{black}}}%
%BeginExpansion
\color{black}%
%EndExpansion
_{\left( l\right) }$ are subjected to the loop closure constraints due to
the $\gamma _{l}$ FCs of limb $l$. The relation (\ref{etaq2}) provides an
explicit solution of the velocity constraints in terms of $\delta _{l}$
independent velocities $\dot{\bar{\mathbf{q}}}_{\left( l\right) }$. The
coordinates $\bar{\mathbf{q}}_{\left( l\right) }$ serve as generalized
coordinates of limb $l$ (without platform). Jourdain's principle of virtual
power applied to {(\ref{EOMLimb}) along with the variations }$\delta 
%TCIMACRO{\TeXButton{red}{}}%
%BeginExpansion
%
%EndExpansion
\dot{\bar{%
%TCIMACRO{\TeXButton{vartheta}{\mathbold{\vartheta}}}%
%BeginExpansion
\mathbold{\vartheta}%
%EndExpansion
}}%
%TCIMACRO{\TeXButton{black}{\color{black}}}%
%BeginExpansion
\color{black}%
%EndExpansion
_{\left( l\right) }=\bar{\mathbf{H}}_{\left( l\right) }\delta \dot{\bar{%
\mathbf{q}}}_{\left( l\right) }${\ }yields%
\begin{equation}
\bar{\bar{\mathbf{M}}}_{\left( l\right) }\ddot{\bar{\mathbf{q}}}_{\left(
l\right) }+\bar{\bar{\mathbf{C}}}_{\left( l\right) }\dot{\bar{\mathbf{q}}}%
_{\left( l\right) }+\bar{\bar{\mathbf{Q}}}_{\left( l\right) }^{\mathrm{grav}%
}+\bar{\bar{\mathbf{Q}}}_{\left( l\right) }=\bar{\bar{\mathbf{Q}}}_{\left(
l\right) }^{\mathrm{act}}  \label{EOMLimb2}
\end{equation}%
where%
\begin{align}
\bar{\bar{\mathbf{M}}}_{\left( l\right) }(%
%TCIMACRO{\TeXButton{red}{}}%
%BeginExpansion
%
%EndExpansion
\bar{%
%TCIMACRO{\TeXButton{vartheta}{\mathbold{\vartheta}}}%
%BeginExpansion
\mathbold{\vartheta}%
%EndExpansion
}%
%TCIMACRO{\TeXButton{black}{\color{black}}}%
%BeginExpansion
\color{black}%
%EndExpansion
_{\left( l\right) })& :=\bar{\mathbf{H}}_{\left( l\right) }^{T}\bar{\mathbf{M%
}}_{\left( l\right) }\bar{\mathbf{H}}_{\left( l\right) }  \notag \\
\bar{\bar{\mathbf{C}}}_{\left( l\right) }(%
%TCIMACRO{\TeXButton{red}{}}%
%BeginExpansion
%
%EndExpansion
\bar{%
%TCIMACRO{\TeXButton{vartheta}{\mathbold{\vartheta}}}%
%BeginExpansion
\mathbold{\vartheta}%
%EndExpansion
}%
%TCIMACRO{\TeXButton{black}{\color{black}}}%
%BeginExpansion
\color{black}%
%EndExpansion
_{\left( l\right) },%
%TCIMACRO{\TeXButton{red}{}}%
%BeginExpansion
%
%EndExpansion
\dot{\bar{%
%TCIMACRO{\TeXButton{vartheta}{\mathbold{\vartheta}}}%
%BeginExpansion
\mathbold{\vartheta}%
%EndExpansion
}}%
%TCIMACRO{\TeXButton{black}{\color{black}}}%
%BeginExpansion
\color{black}%
%EndExpansion
_{\left( l\right) })& :=\bar{\mathbf{H}}_{\left( l\right) }^{T}(\bar{\mathbf{%
M}}_{\left( l\right) }\dot{\bar{\mathbf{H}}}_{\left( l\right) }+\bar{\mathbf{%
C}}_{\left( l\right) }\bar{\mathbf{H}}_{\left( l\right) })  \label{MCQLimb}
\\
\bar{\bar{\mathbf{Q}}}_{\left( l\right) }(%
%TCIMACRO{\TeXButton{red}{}}%
%BeginExpansion
%
%EndExpansion
\bar{%
%TCIMACRO{\TeXButton{vartheta}{\mathbold{\vartheta}}}%
%BeginExpansion
\mathbold{\vartheta}%
%EndExpansion
}_{\left( l\right) }%
%TCIMACRO{\TeXButton{black}{\color{black}}}%
%BeginExpansion
\color{black}%
%EndExpansion
)& :=\bar{\mathbf{H}}_{\left( l\right) }^{T}\bar{\mathbf{Q}}_{\left(
l\right) },\ \bar{\bar{\mathbf{Q}}}_{\left( l\right) }^{\mathrm{act}}(%
%TCIMACRO{\TeXButton{red}{}}%
%BeginExpansion
%
%EndExpansion
\bar{%
%TCIMACRO{\TeXButton{vartheta}{\mathbold{\vartheta}}}%
%BeginExpansion
\mathbold{\vartheta}%
%EndExpansion
}_{\left( l\right) }%
%TCIMACRO{\TeXButton{black}{\color{black}}}%
%BeginExpansion
\color{black}%
%EndExpansion
):=\bar{\mathbf{H}}_{\left( l\right) }^{T}\bar{\mathbf{Q}}_{\left( l\right)
}^{\mathrm{act}},\ \bar{\bar{\mathbf{Q}}}_{\left( l\right) }^{\mathrm{grav}}(%
%TCIMACRO{\TeXButton{red}{}}%
%BeginExpansion
%
%EndExpansion
\bar{%
%TCIMACRO{\TeXButton{vartheta}{\mathbold{\vartheta}}}%
%BeginExpansion
\mathbold{\vartheta}%
%EndExpansion
}%
%TCIMACRO{\TeXButton{black}{\color{black}}}%
%BeginExpansion
\color{black}%
%EndExpansion
):=\bar{\mathbf{H}}_{\left( l\right) }^{T}\bar{\mathbf{Q}}_{\left( l\right)
}^{\mathrm{grav}}.  \notag
\end{align}%
The matrices (\ref{MCQLimb}) depend on the coordinates $%
%TCIMACRO{\TeXButton{red}{}}%
%BeginExpansion
%
%EndExpansion
\bar{%
%TCIMACRO{\TeXButton{vartheta}{\mathbold{\vartheta}}}%
%BeginExpansion
\mathbold{\vartheta}%
%EndExpansion
}%
%TCIMACRO{\TeXButton{black}{\color{black}}}%
%BeginExpansion
\color{black}%
%EndExpansion
_{\left( l\right) }$ and velocity of the tree-system. The velocity $%
%TCIMACRO{\TeXButton{red}{}}%
%BeginExpansion
%
%EndExpansion
\dot{\bar{%
%TCIMACRO{\TeXButton{vartheta}{\mathbold{\vartheta}}}%
%BeginExpansion
\mathbold{\vartheta}%
%EndExpansion
}}%
%TCIMACRO{\TeXButton{black}{\color{black}}}%
%BeginExpansion
\color{black}%
%EndExpansion
_{\left( l\right) }$ can be replace with $\dot{\bar{\mathbf{q}}}_{\left(
l\right) }$ using (\ref{etaq}), and $%
%TCIMACRO{\TeXButton{red}{}}%
%BeginExpansion
%
%EndExpansion
\bar{%
%TCIMACRO{\TeXButton{vartheta}{\mathbold{\vartheta}}}%
%BeginExpansion
\mathbold{\vartheta}%
%EndExpansion
}%
%TCIMACRO{\TeXButton{black}{\color{black}}}%
%BeginExpansion
\color{black}%
%EndExpansion
_{\left( l\right) }$ can be replaced with a (closed form or numerical)
solution on terms of $\bar{\mathbf{q}}_{\left( l\right) }$. The equations (%
\ref{MCQLimb}) are also referred to as the Woronets equations \cite%
{MuellerJNLS2021} as a similar form first appeared in \cite%
{Voronets1901,Woronetz1910}.

\subsection{EOM of the Platform}

The platform twist $\mathbf{V}_{\mathrm{p}}=\left( 
%TCIMACRO{\TeXButton{w}{\bm{\omega}}}%
%BeginExpansion
\bm{\omega}%
%EndExpansion
_{\mathrm{p}},\mathbf{v}_{\mathrm{p}}\right) ^{T}$ in body-fixed
representation consists of the linear velocity $\mathbf{v}_{\mathrm{p}}$ and
angular velocity $%
%TCIMACRO{\TeXButton{w}{\bm{\omega}}}%
%BeginExpansion
\bm{\omega}%
%EndExpansion
_{\mathrm{p}}$ of the platform frame $\mathcal{F}_{\mathrm{p}}$, relative to
the world frame $\mathcal{F}_{0}$. The dynamics of the platform body is
governed by the Newton-Euler equations, which can be summarized as (see
appendix \ref{AppendixEOMTree})%
\begin{equation}
\mathbf{M}_{\mathrm{p}}\dot{\mathbf{V}}_{\mathrm{p}}+\mathbf{G}_{\mathrm{p}}%
\mathbf{M}_{\mathrm{p}}\mathbf{V}_{\mathrm{p}}+\mathbf{W}_{\mathrm{p}}^{%
\mathrm{grav}}=\mathbf{W}_{\mathrm{p}}^{\mathrm{EE}}  \label{EOMPlat}
\end{equation}%
where the constant $6\times 6$ inertia matrix $\mathbf{M}_{\mathrm{p}}$ and
matrix $\mathbf{G}_{\mathrm{p}}$ is, respectively,%
\begin{equation}
\mathbf{M}_{\mathrm{p}}=\left( 
\begin{array}{cc}
%TCIMACRO{\TeXButton{Theta}{\bm{\Theta}} }%
%BeginExpansion
\bm{\Theta}
%EndExpansion
& m\widetilde{\mathbf{d}} \\ 
-m\widetilde{\mathbf{d}}\ \ \  & m\mathbf{I}%
\end{array}%
\right) ,\ \mathbf{G}_{\mathrm{p}}\left( \mathbf{V}_{\mathrm{p}}\right)
=\left( 
\begin{array}{cc}
\widetilde{%
%TCIMACRO{\TeXButton{w}{\bm{\omega}}}%
%BeginExpansion
\bm{\omega}%
%EndExpansion
}_{\mathrm{p}} & \widetilde{\mathbf{v}}_{\mathrm{p}} \\ 
\mathbf{0} & \widetilde{%
%TCIMACRO{\TeXButton{w}{\bm{\omega}}}%
%BeginExpansion
\bm{\omega}%
%EndExpansion
}_{\mathrm{p}}%
\end{array}%
\right)  \label{MGPlatform}
\end{equation}%
with the body-fixed inertia tensor $%
%TCIMACRO{\TeXButton{Theta}{\bm{\Theta}}}%
%BeginExpansion
\bm{\Theta}%
%EndExpansion
$ w.r.t. $\mathcal{F}_{\mathrm{p}}$, and $\mathbf{d}$ is the position vector
of the COM represented in $\mathcal{F}_{\mathrm{p}}$. Acting at the platform
are the wrench $\mathbf{W}_{\mathrm{p}}^{\mathrm{grav}}\left( \mathbf{x}%
\right) $ due to gravity and the EE-wrench $\mathbf{W}_{\mathrm{p}}^{\mathrm{%
EE}}\left( t\right) $ (due to interaction of the PKM), where a wrench $%
\mathbf{W}_{\mathrm{p}}=\left( \mathbf{m}_{\mathrm{p}}^{T},\mathbf{f}_{%
\mathrm{p}}^{T}\right) $, represented in $\mathcal{F}_{\mathrm{p}}$,
consists of a torque $\mathbf{m}_{\mathrm{p}}$ and a force $\mathbf{f}_{%
\mathrm{p}}$. The gyroscopic matrix is related to the matrix of the
coadjoint action on $se\left( 3\right) $ by $\mathbf{G}_{\mathrm{p}}\left( 
\mathbf{V}_{\mathrm{p}}\right) =-\mathbf{ad}_{\mathbf{V}_{\mathrm{p}}}^{T}$.
Moreover, the left-hand side of (\ref{EOMPlat}) are the Euler-Poincar\'{e}
equations of the rigid body on $SE\left( 3\right) $ \cite%
{MarsdenBook,HolmBook2,MuellerJNLS2021}. The equation (\ref{EOMPlat}) holds
true for an arbitrary body-fixed reference frame.

The gravity wrench is 
\begin{equation}
\mathbf{W}_{\mathrm{p}}^{\mathrm{grav}}=-\mathbf{M}_{\mathrm{p}}\mathbf{Ad}_{%
\mathbf{C}_{\mathrm{p}}}^{-1}\left( 
\begin{array}{c}
\mathbf{0} \\ 
{^{0}\mathbf{g}}%
\end{array}%
\right)
\end{equation}%
where ${^{0}\mathbf{g}}$ is the vector of gravitational acceleration
expressed in the inertial frame.

\subsection{Task Space Formulation of the EOM for PKM with Complex Limbs%
%TCIMACRO{\TeXButton{secEOMTaskSpace}{\label{secEOMTaskSpace}}}%
%BeginExpansion
\label{secEOMTaskSpace}%
%EndExpansion
}

The dynamics of the disconnected platform and of the $L$ separated limbs are
governed by the EOM (\ref{EOMPlat}) and (\ref{EOMLimb2}), respectively. When
assembled their motion is constrained. The task space velocity determines
the platform twists via (\ref{VpVt}) and the velocity of the limbs via (\ref%
{Ftheta1}), (\ref{Ftheta2}) or (\ref{Ftheta3}). Denote with $\bar{\mathbf{F}}%
_{\left( l\right) }$ the submatrix of $\mathbf{F}_{\left( l\right) }$ with
rows corresponding to the generalized coordinates $\bar{\mathbf{q}}_{\left(
l\right) }$, so that $\dot{\bar{\mathbf{q}}}_{\left( l\right) }=\bar{\mathbf{%
F}}_{\left( l\right) }\mathbf{V}_{\mathrm{t}}$ and hence $%
%TCIMACRO{\TeXButton{red}{}}%
%BeginExpansion
%
%EndExpansion
\dot{\bar{%
%TCIMACRO{\TeXButton{vartheta}{\mathbold{\vartheta}}}%
%BeginExpansion
\mathbold{\vartheta}%
%EndExpansion
}}%
%TCIMACRO{\TeXButton{black}{\color{black}}}%
%BeginExpansion
\color{black}%
%EndExpansion
_{\left( l\right) }=\bar{\mathbf{H}}_{\left( l\right) }\bar{\mathbf{F}}%
_{\left( l\right) }\mathbf{V}_{\mathrm{t}}$.

The principle of virtual power finally yields the dynamic EOM in task space
velocity coordinates%
\begin{equation}
\mathbf{M}_{\mathrm{t}}\dot{\mathbf{V}}_{\mathrm{t}}+\mathbf{C}_{\mathrm{t}}%
\mathbf{V}_{\mathrm{t}}+\mathbf{W}_{\mathrm{t}}^{\mathrm{grav}}+\mathbf{W}_{%
\mathrm{t}}=\mathbf{W}_{\mathrm{t}}^{\mathrm{EE}}+\mathbf{J}_{\mathrm{IK}%
}^{T}\mathbf{u}\left( t\right)  \label{EOMTask}
\end{equation}%
with the $\delta \times \delta $ generalized mass matrix and Coriolis matrix%
\begin{align}
\mathbf{M}_{\mathrm{t}}(%
%TCIMACRO{\TeXButton{red}{}}%
%BeginExpansion
%
%EndExpansion
%TCIMACRO{\TeXButton{vartheta}{\mathbold{\vartheta}}}%
%BeginExpansion
\mathbold{\vartheta}%
%EndExpansion
%TCIMACRO{\TeXButton{black}{\color{black}}}%
%BeginExpansion
\color{black}%
%EndExpansion
):=& \sum_{l=1}^{L}\bar{\mathbf{F}}_{\left( l\right) }^{T}\bar{\bar{\mathbf{M%
}}}_{\left( l\right) }\bar{\mathbf{F}}_{\left( l\right) }+\mathbf{P}_{%
\mathrm{p}}^{T}\mathbf{M}_{\mathrm{p}}\mathbf{P}_{\mathrm{p}}  \label{Mbar}
\\
=& \sum_{l=1}^{L}\bar{\mathbf{F}}_{\left( l\right) }^{T}\bar{\mathbf{H}}%
_{\left( l\right) }^{T}\bar{\mathbf{M}}_{\left( l\right) }\bar{\mathbf{H}}%
_{\left( l\right) }\bar{\mathbf{F}}_{\left( l\right) }+\mathbf{P}_{\mathrm{p}%
}^{T}\mathbf{M}_{\mathrm{p}}\mathbf{P}_{\mathrm{p}}  \notag \\
\mathbf{C}_{\mathrm{t}}(%
%TCIMACRO{\TeXButton{red}{}}%
%BeginExpansion
%
%EndExpansion
%TCIMACRO{\TeXButton{vartheta}{\mathbold{\vartheta}}}%
%BeginExpansion
\mathbold{\vartheta}%
%EndExpansion
%TCIMACRO{\TeXButton{black}{\color{black}}}%
%BeginExpansion
\color{black}%
%EndExpansion
,%
%TCIMACRO{\TeXButton{red}{}}%
%BeginExpansion
%
%EndExpansion
\dot{%
%TCIMACRO{\TeXButton{vartheta}{\mathbold{\vartheta}}}%
%BeginExpansion
\mathbold{\vartheta}%
%EndExpansion
}%
%TCIMACRO{\TeXButton{black}{\color{black}}}%
%BeginExpansion
\color{black}%
%EndExpansion
):=& \sum_{l=1}^{L}\bar{\mathbf{F}}_{\left( l\right) }^{T}(\bar{\bar{\mathbf{%
C}}}_{\left( l\right) }\bar{\mathbf{F}}_{\left( l\right) }+\bar{\bar{\mathbf{%
M}}}_{\left( l\right) }\dot{\bar{\mathbf{F}}}_{\left( l\right) })+\mathbf{P}%
_{\mathrm{p}}^{T}\mathbf{G}_{\mathrm{p}}\mathbf{M}_{\mathrm{p}}\mathbf{P}_{%
\mathrm{p}}  \label{Cbar} \\
=& \sum_{l=1}^{L}\bar{\mathbf{F}}_{\left( l\right) }^{T}\bar{\mathbf{H}}%
_{\left( l\right) }^{T}%
%TCIMACRO{\TeXButton{big}{\big}}%
%BeginExpansion
\big%
%EndExpansion
(\bar{\mathbf{C}}_{\left( l\right) }\bar{\mathbf{H}}_{\left( l\right) }\bar{%
\mathbf{F}}_{\left( l\right) }+\bar{\mathbf{M}}_{\left( l\right) }%
%TCIMACRO{\TeXButton{big}{\big}}%
%BeginExpansion
\big%
%EndExpansion
(\dot{\bar{\mathbf{H}}}_{\left( l\right) }\bar{\mathbf{F}}_{\left( l\right)
}+\bar{\mathbf{H}}_{\left( l\right) }\dot{\bar{\mathbf{F}}}_{\left( l\right)
}%
%TCIMACRO{\TeXButton{big}{\big}}%
%BeginExpansion
\big%
%EndExpansion
)%
%TCIMACRO{\TeXButton{big}{\big}}%
%BeginExpansion
\big%
%EndExpansion
)+\mathbf{P}_{\mathrm{p}}^{T}\mathbf{G}_{\mathrm{p}}\mathbf{M}_{\mathrm{p}}%
\mathbf{P}_{\mathrm{p}},  \notag
\end{align}%
the inverse kinematics Jacobian $\mathbf{J}_{\mathrm{IK}}(%
%TCIMACRO{\TeXButton{red}{}}%
%BeginExpansion
%
%EndExpansion
%TCIMACRO{\TeXButton{vartheta}{\mathbold{\vartheta}}}%
%BeginExpansion
\mathbold{\vartheta}%
%EndExpansion
%TCIMACRO{\TeXButton{black}{\color{black}}}%
%BeginExpansion
\color{black}%
%EndExpansion
)$ in (\ref{IK}), and the vector $\mathbf{u}\in {\mathbb{R}}^{N_{\mathrm{act}%
}}$ of $N_{\mathrm{act}}\geq \delta $ actuator forces/torques. The vector of
generalized forces%
\begin{align}
\ \mathbf{W}_{\mathrm{t}}^{\mathrm{EE}}\left( t\right) & :=\mathbf{P}_{%
\mathrm{p}}^{T}\mathbf{W}_{\mathrm{p}}^{\mathrm{EE}}\left( t\right)
\label{WEE} \\
\mathbf{W}_{\mathrm{t}}(%
%TCIMACRO{\TeXButton{red}{}}%
%BeginExpansion
%
%EndExpansion
%TCIMACRO{\TeXButton{vartheta}{\mathbold{\vartheta}}}%
%BeginExpansion
\mathbold{\vartheta}%
%EndExpansion
%TCIMACRO{\TeXButton{black}{\color{black}}}%
%BeginExpansion
\color{black}%
%EndExpansion
,%
%TCIMACRO{\TeXButton{red}{}}%
%BeginExpansion
%
%EndExpansion
\dot{%
%TCIMACRO{\TeXButton{vartheta}{\mathbold{\vartheta}}}%
%BeginExpansion
\mathbold{\vartheta}%
%EndExpansion
}%
%TCIMACRO{\TeXButton{black}{\color{black}}}%
%BeginExpansion
\color{black}%
%EndExpansion
,t)& :=\sum_{l=1}^{L}\bar{\mathbf{F}}_{\left( l\right) }^{T}\bar{\bar{%
\mathbf{Q}}}_{\left( l\right) }=\sum_{l=1}^{L}\bar{\mathbf{F}}_{\left(
l\right) }^{T}\bar{\mathbf{H}}_{\left( l\right) }^{T}\bar{\mathbf{Q}}%
_{\left( l\right) }  \label{W} \\
\mathbf{W}_{\mathrm{t}}^{\mathrm{grav}}(%
%TCIMACRO{\TeXButton{red}{}}%
%BeginExpansion
%
%EndExpansion
%TCIMACRO{\TeXButton{vartheta}{\mathbold{\vartheta}}}%
%BeginExpansion
\mathbold{\vartheta}%
%EndExpansion
%TCIMACRO{\TeXButton{black}{\color{black}}}%
%BeginExpansion
\color{black}%
%EndExpansion
,\mathbf{x})& :=\sum_{l=1}^{L}\bar{\mathbf{F}}_{\left( l\right) }^{T}\bar{%
\bar{\mathbf{Q}}}_{\left( l\right) }^{\mathrm{grav}}+\mathbf{P}_{\mathrm{p}%
}^{T}\mathbf{W}_{\mathrm{p}}^{\mathrm{grav}}=\sum_{l=1}^{L}\bar{\mathbf{F}}%
_{\left( l\right) }^{T}\bar{\mathbf{H}}_{\left( l\right) }^{T}\bar{\mathbf{Q}%
}_{\left( l\right) }^{\mathrm{grav}}+\mathbf{P}_{\mathrm{p}}^{T}\mathbf{W}_{%
\mathrm{p}}^{\mathrm{grav}}
\end{align}%
accounts for EE-loads and all other loads. The gyroscopic matrix in (\ref%
{MGPlatform}) is evaluated with the task space velocity: $\mathbf{G}_{%
\mathrm{p}}=\mathbf{G}_{\mathrm{p}}\left( \mathbf{P}_{\mathrm{p}}\mathbf{V}_{%
\mathrm{t}}\right) $. The dependency of the terms in (\ref{EOMTask}) on $%
%TCIMACRO{\TeXButton{red}{}}%
%BeginExpansion
%
%EndExpansion
%TCIMACRO{\TeXButton{vartheta}{\mathbold{\vartheta}}}%
%BeginExpansion
\mathbold{\vartheta}%
%EndExpansion
%TCIMACRO{\TeXButton{black}{\color{black}}}%
%BeginExpansion
\color{black}%
%EndExpansion
$ is kept as all expressions in (\ref{MCQLimb}) depend on $%
%TCIMACRO{\TeXButton{red}{}}%
%BeginExpansion
%
%EndExpansion
\bar{%
%TCIMACRO{\TeXButton{vartheta}{\mathbold{\vartheta}}}%
%BeginExpansion
\mathbold{\vartheta}%
%EndExpansion
}%
%TCIMACRO{\TeXButton{black}{\color{black}}}%
%BeginExpansion
\color{black}%
%EndExpansion
$, and further $\bar{\mathbf{F}}_{\left( l\right) }$ depends on $%
%TCIMACRO{\TeXButton{red}{}}%
%BeginExpansion
%
%EndExpansion
%TCIMACRO{\TeXButton{vartheta}{\mathbold{\vartheta}}}%
%BeginExpansion
\mathbold{\vartheta}%
%EndExpansion
%TCIMACRO{\TeXButton{black}{\color{black}}}%
%BeginExpansion
\color{black}%
%EndExpansion
_{\left( l\right) }$ (the complete set of tree-joint variables including the
joint connecting limb $l$ to the platform).

The above equations in terms of $%
%TCIMACRO{\TeXButton{red}{}}%
%BeginExpansion
%
%EndExpansion
%TCIMACRO{\TeXButton{vartheta}{\mathbold{\vartheta}}}%
%BeginExpansion
\mathbold{\vartheta}%
%EndExpansion
%TCIMACRO{\TeXButton{black}{\color{black}}}%
%BeginExpansion
\color{black}%
%EndExpansion
$ are derived making use of the relations (\ref{etaq2}) and (\ref{HFl}). If
the inverse kinematics map $\psi _{\mathrm{IK}}$ of the mechanism can be
expressed in closed form, the dynamic equations (\ref{EOMTask}) can be
derived solely in terms of the task space coordinates $\mathbf{x}$ and
velocity $\mathbf{V}_{\mathrm{t}}$. Then, $\bar{\mathbf{H}}_{\left( l\right)
}\bar{\mathbf{F}}_{\left( l\right) }$ is replaced by the (inverse
kinematics) Jacobian of $\psi _{\mathrm{IK}}$. Yet, this usually leads to
very complex expressions.

The EOM (\ref{EOMTask}) can be expressed in terms of the general form of the
EOM (\ref{phil}), and introducing $\varphi _{\mathrm{p}}\left( \mathbf{x},%
\mathbf{V}_{\mathrm{t}},\dot{\mathbf{V}}_{\mathrm{t}}\right) :=\mathbf{M}_{%
\mathrm{p}}\mathbf{P}_{\mathrm{p}}\dot{\mathbf{V}}_{\mathrm{t}}+\mathbf{G}_{%
\mathrm{p}}\mathbf{M}_{\mathrm{p}}\mathbf{P}_{\mathrm{p}}\mathbf{V}_{\mathrm{%
t}}+\mathbf{W}_{\mathrm{p}}^{\mathrm{grav}}$, as 
\begin{equation}
\sum_{l=1}^{L}\bar{\mathbf{H}}_{\left( l\right) }^{T}\bar{\mathbf{F}}%
_{\left( l\right) }^{T}\varphi _{\left( l\right) }(%
%TCIMACRO{\TeXButton{red}{}}%
%BeginExpansion
%
%EndExpansion
\bar{%
%TCIMACRO{\TeXButton{vartheta}{\mathbold{\vartheta}}}%
%BeginExpansion
\mathbold{\vartheta}%
%EndExpansion
}%
%TCIMACRO{\TeXButton{black}{\color{black}}}%
%BeginExpansion
\color{black}%
%EndExpansion
_{\left( l\right) },%
%TCIMACRO{\TeXButton{red}{}}%
%BeginExpansion
%
%EndExpansion
\dot{\bar{%
%TCIMACRO{\TeXButton{vartheta}{\mathbold{\vartheta}}}%
%BeginExpansion
\mathbold{\vartheta}%
%EndExpansion
}}%
%TCIMACRO{\TeXButton{black}{\color{black}}}%
%BeginExpansion
\color{black}%
%EndExpansion
_{\left( l\right) },%
%TCIMACRO{\TeXButton{red}{}}%
%BeginExpansion
%
%EndExpansion
\ddot{\bar{%
%TCIMACRO{\TeXButton{vartheta}{\mathbold{\vartheta}}}%
%BeginExpansion
\mathbold{\vartheta}%
%EndExpansion
}}%
%TCIMACRO{\TeXButton{black}{\color{black}}}%
%BeginExpansion
\color{black}%
%EndExpansion
_{\left( l\right) })+\mathbf{P}_{\mathrm{p}}^{T}\varphi _{\mathrm{p}}\left( 
\mathbf{x},\mathbf{V}_{\mathrm{t}},\dot{\mathbf{V}}_{\mathrm{t}}\right) =%
\mathbf{W}_{\mathrm{t}}^{\mathrm{EE}}+\mathbf{J}_{\mathrm{IK}}^{T}\mathbf{u}
\end{equation}%
with the kinematic relations (\ref{HFl}) and (\ref{eta2d}).

\subsection{Formulation of the EOM for non-redundant PKM in terms of
actuator coordinates%
%TCIMACRO{\TeXButton{secEOMActCoord}{\label{secEOMActCoord}}}%
%BeginExpansion
\label{secEOMActCoord}%
%EndExpansion
}

The solution of the velocity forward kinematics problem of a non-redundant
PKM (not kinematically redundant nor redundantly actuated) is obtain from (%
\ref{IKLimb}) as%
\begin{equation}
\mathbf{V}_{\mathrm{t}}=\mathbf{J}_{\mathrm{FK}}\dot{%
%TCIMACRO{\TeXButton{vartheta}{\mathbold{\vartheta}}}%
%BeginExpansion
\mathbold{\vartheta}%
%EndExpansion
}_{\mathrm{act}}
\end{equation}%
where $\mathbf{J}_{\mathrm{FK}}:=\mathbf{J}_{\mathrm{IK}}^{-1}$ is the
forward kinematics Jacobian. Combined with (\ref{HFl}), this yields a
solution of the velocity forward kinematics problem of the mechanism as%
\begin{equation}
%TCIMACRO{\TeXButton{red}{}}%
%BeginExpansion
%
%EndExpansion
\dot{%
%TCIMACRO{\TeXButton{vartheta}{\mathbold{\vartheta}}}%
%BeginExpansion
\mathbold{\vartheta}%
%EndExpansion
}%
%TCIMACRO{\TeXButton{black}{\color{black}}}%
%BeginExpansion
\color{black}%
%EndExpansion
_{\left( l\right) }=\mathbf{H}_{\left( l\right) }\mathbf{F}_{\left( l\right)
}\mathbf{J}_{\mathrm{FK}}\dot{%
%TCIMACRO{\TeXButton{vartheta}{\mathbold{\vartheta}}}%
%BeginExpansion
\mathbold{\vartheta}%
%EndExpansion
}_{\mathrm{act}}.  \label{HFJl}
\end{equation}%
With this relation, the principle of virtual power applied to (\ref{EOMTask}%
) yields the dynamic equation in actuator coordinates%
\begin{equation}
\mathbf{M}_{\mathrm{a}}\ddot{%
%TCIMACRO{\TeXButton{vartheta}{\mathbold{\vartheta}}}%
%BeginExpansion
\mathbold{\vartheta}%
%EndExpansion
}_{\mathrm{a}}+\mathbf{C}_{\mathrm{a}}\dot{%
%TCIMACRO{\TeXButton{vartheta}{\mathbold{\vartheta}}}%
%BeginExpansion
\mathbold{\vartheta}%
%EndExpansion
}_{\mathrm{a}}+\mathbf{Q}_{\mathrm{a}}^{\mathrm{grav}}+\mathbf{Q}_{\mathrm{a}%
}=\mathbf{Q}_{\mathrm{a}}^{\mathrm{EE}}+\mathbf{u}  \label{EOMact}
\end{equation}%
with the $\delta \times \delta $ generalized mass and Coriolis matrix%
\begin{align}
\mathbf{M}_{\mathrm{a}}(%
%TCIMACRO{\TeXButton{eta}{\mathbold{\eta}}}%
%BeginExpansion
\mathbold{\eta}%
%EndExpansion
)& :=\mathbf{J}_{\mathrm{FK}}^{T}\mathbf{M}_{\mathrm{t}}\mathbf{J}_{\mathrm{%
FK}}  \label{Ma} \\
\mathbf{C}_{\mathrm{a}}(%
%TCIMACRO{\TeXButton{red}{}}%
%BeginExpansion
%
%EndExpansion
\bar{%
%TCIMACRO{\TeXButton{vartheta}{\mathbold{\vartheta}}}%
%BeginExpansion
\mathbold{\vartheta}%
%EndExpansion
}%
%TCIMACRO{\TeXButton{black}{\color{black}}}%
%BeginExpansion
\color{black}%
%EndExpansion
,%
%TCIMACRO{\TeXButton{red}{}}%
%BeginExpansion
%
%EndExpansion
\dot{%
%TCIMACRO{\TeXButton{vartheta}{\mathbold{\vartheta}}}%
%BeginExpansion
\mathbold{\vartheta}%
%EndExpansion
}%
%TCIMACRO{\TeXButton{black}{\color{black}}}%
%BeginExpansion
\color{black}%
%EndExpansion
)& :=\mathbf{J}_{\mathrm{FK}}^{T}(\mathbf{C}_{\mathrm{t}}\mathbf{J}_{\mathrm{%
FK}}+\mathbf{M}_{\mathrm{t}}\dot{\mathbf{J}}_{\mathrm{FK}})=\mathbf{J}_{%
\mathrm{FK}}^{T}(\mathbf{C}_{\mathrm{t}}-\mathbf{M}_{\mathrm{t}}\mathbf{J}_{%
\mathrm{FK}}\dot{\mathbf{J}}_{\mathrm{IK}})\mathbf{J}_{\mathrm{FK}}.
\end{align}%
The vector $\mathbf{u}$ of actuator forces/torques appears explicitly, while
the generalized EE forces and the vector of all remaining forces are%
\begin{align}
\mathbf{Q}_{\mathrm{a}}(%
%TCIMACRO{\TeXButton{red}{}}%
%BeginExpansion
%
%EndExpansion
%TCIMACRO{\TeXButton{vartheta}{\mathbold{\vartheta}}}%
%BeginExpansion
\mathbold{\vartheta}%
%EndExpansion
%TCIMACRO{\TeXButton{black}{\color{black}}}%
%BeginExpansion
\color{black}%
%EndExpansion
,%
%TCIMACRO{\TeXButton{red}{}}%
%BeginExpansion
%
%EndExpansion
\dot{%
%TCIMACRO{\TeXButton{vartheta}{\mathbold{\vartheta}}}%
%BeginExpansion
\mathbold{\vartheta}%
%EndExpansion
}%
%TCIMACRO{\TeXButton{black}{\color{black}}}%
%BeginExpansion
\color{black}%
%EndExpansion
,t)& :=\mathbf{J}_{\mathrm{FK}}^{T}\mathbf{W}_{\mathrm{t}},\ \mathbf{Q}_{%
\mathrm{a}}^{\mathrm{grav}}(%
%TCIMACRO{\TeXButton{red}{}}%
%BeginExpansion
%
%EndExpansion
%TCIMACRO{\TeXButton{vartheta}{\mathbold{\vartheta}}}%
%BeginExpansion
\mathbold{\vartheta}%
%EndExpansion
%TCIMACRO{\TeXButton{black}{\color{black}}}%
%BeginExpansion
\color{black}%
%EndExpansion
):=\mathbf{J}_{\mathrm{FK}}^{T}\mathbf{W}_{\mathrm{t}}^{\mathrm{grav}}
\label{Qa1} \\
\mathbf{Q}_{\mathrm{a}}^{\mathrm{EE}}(%
%TCIMACRO{\TeXButton{red}{}}%
%BeginExpansion
%
%EndExpansion
%TCIMACRO{\TeXButton{vartheta}{\mathbold{\vartheta}}}%
%BeginExpansion
\mathbold{\vartheta}%
%EndExpansion
%TCIMACRO{\TeXButton{black}{\color{black}}}%
%BeginExpansion
\color{black}%
%EndExpansion
,t)& :=\mathbf{J}_{\mathrm{FK}}^{T}\mathbf{W}_{\mathrm{t}}^{\mathrm{EE}}.
\label{Qa2}
\end{align}%
By construction, (\ref{EOMact}) depends on the joint coordinates $%
%TCIMACRO{\TeXButton{red}{}}%
%BeginExpansion
%
%EndExpansion
%TCIMACRO{\TeXButton{vartheta}{\mathbold{\vartheta}}}%
%BeginExpansion
\mathbold{\vartheta}%
%EndExpansion
%TCIMACRO{\TeXButton{black}{\color{black}}}%
%BeginExpansion
\color{black}%
%EndExpansion
$. It can be transformed to actuator coordinated by solving the geometric
forward kinematics problem of the mechanism, which is to find the joint
coordinates $%
%TCIMACRO{\TeXButton{red}{}}%
%BeginExpansion
%
%EndExpansion
%TCIMACRO{\TeXButton{vartheta}{\mathbold{\vartheta}}}%
%BeginExpansion
\mathbold{\vartheta}%
%EndExpansion
%TCIMACRO{\TeXButton{black}{\color{black}}}%
%BeginExpansion
\color{black}%
%EndExpansion
\left( t\right) $ for given actuator coordinates $%
%TCIMACRO{\TeXButton{vartheta}{\mathbold{\vartheta}}}%
%BeginExpansion
\mathbold{\vartheta}%
%EndExpansion
_{\mathrm{act}}\left( t\right) $. The solution is given by the forward
kinematics map $\psi _{\mathrm{FK}}:{\mathbb{V}}^{n_{\mathrm{act}%
}}\rightarrow {\mathbb{V}}^{%
%TCIMACRO{\TeXButton{red}{}}%
%BeginExpansion
%
%EndExpansion
n%
%TCIMACRO{\TeXButton{black}{\color{black}}}%
%BeginExpansion
\color{black}%
%EndExpansion
}$, so that $%
%TCIMACRO{\TeXButton{red}{}}%
%BeginExpansion
%
%EndExpansion
%TCIMACRO{\TeXButton{vartheta}{\mathbold{\vartheta}}}%
%BeginExpansion
\mathbold{\vartheta}%
%EndExpansion
%TCIMACRO{\TeXButton{black}{\color{black}}}%
%BeginExpansion
\color{black}%
%EndExpansion
=\psi _{\mathrm{FK}}(%
%TCIMACRO{\TeXButton{vartheta}{\mathbold{\vartheta}}}%
%BeginExpansion
\mathbold{\vartheta}%
%EndExpansion
_{\mathrm{act}})$. For PKM in general, $\psi _{\mathrm{FK}}$ cannot be
expressed in closed form. Again, in special cases, such as the 3\underline{R}%
R[2RR]R Delta, a closed form expression is available (and the non-uniqueness
problem can be tackled). In more complicated situations, such as the active
ankle PKM module reported in \cite{Stoeffler2018}, explicit solutions can be
obtained by means of an algebraic description of the constraints, and their
solution using algorithms from computational algebraic geometry. Then, $%
%TCIMACRO{\TeXButton{red}{}}%
%BeginExpansion
%
%EndExpansion
%TCIMACRO{\TeXButton{vartheta}{\mathbold{\vartheta}}}%
%BeginExpansion
\mathbold{\vartheta}%
%EndExpansion
%TCIMACRO{\TeXButton{black}{\color{black}}}%
%BeginExpansion
\color{black}%
%EndExpansion
$ and its time derivatives can be substituted by the actuator coordinates $%
%TCIMACRO{\TeXButton{vartheta}{\mathbold{\vartheta}}}%
%BeginExpansion
\mathbold{\vartheta}%
%EndExpansion
_{\mathrm{act}}$ and their derivatives. The term $\bar{\mathbf{H}}_{\left(
l\right) }\bar{\mathbf{F}}_{\left( l\right) }\mathbf{J}_{\mathrm{FK}}$ is
then replaced by the Jacobian of $\psi _{\mathrm{FK}}$. For a
non-redundantly actuated PKM, the actuator variables serve as generalized
coordinates: $\mathbf{q}\left( t\right) :=%
%TCIMACRO{\TeXButton{vartheta}{\mathbold{\vartheta}}}%
%BeginExpansion
\mathbold{\vartheta}%
%EndExpansion
_{\mathrm{act}}$.

\section{Modular Modeling%
%TCIMACRO{\TeXButton{secModularModeling}{\label{secModularModeling}}}%
%BeginExpansion
\label{secModularModeling}%
%EndExpansion
}

Almost all PKM are built from structurally identical limbs. Consequently,
the PKM can be regarded as being assembled from $L$ instances of a \emph{%
representative limb (RL)}, which implies that the topological graph $\Gamma $
is the union of $L$ congruent subgraphs $\Gamma _{l}$. This allows reusing
the kinematics and dynamics equations of a single RL for deriving the
overall PKM model. To this end, $L$ instances of the RL are mounted at the
base and platform, respectively.

To account for the different locations of the limbs, the IFR $\mathcal{F}%
_{0} $ and the platform frame $\mathcal{F}_{\mathrm{p}}$ in the model for
the RL are replaced by an arbitrarily located \emph{construction frame} at
the base, denoted $\mathcal{F}_{0}^{\prime }$, and at the platform, denoted $%
\mathcal{F}_{\mathrm{p}}^{\prime }$, respectively. This is shown in Fig. \ref%
{figReferenceLimbDelta} for the RL of the 3\underline{R}R[2RR]R Delta. In
order to locate the $L$ instances of the RL within the PKM model, a \emph{%
mount frame} is defined at the ground and the platform, denoted with $%
\mathcal{F}_{0\left( l\right) }$ and $\mathcal{F}_{\mathrm{p}\left( l\right)
}$, respectively. Fig. \ref{figMountDelta} shows this for the Delta example.
The $l$th instance of the RL is inserted between these two frames. That is,
the overall PKM is assembled by identifying the mount frame $\mathcal{F}%
_{0\left( l\right) }$ for limb $l$ at the base with the construction frame $%
\mathcal{F}_{0}^{\prime }$, and the mount frame at the platform $\mathcal{F}%
_{\mathrm{p}\left( l\right) }$ with the construction frame $\mathcal{F}_{%
\mathrm{p}}^{\prime }$ of the $l$th instance of the RL. 
\begin{figure}[h]
\centerline{
\includegraphics[height=6cm]{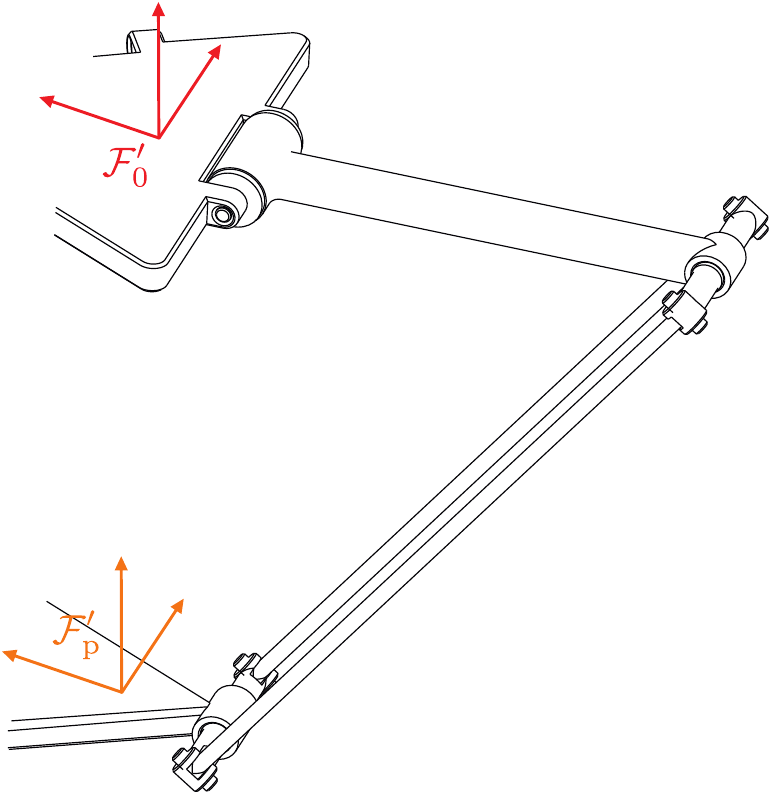}}
\caption{Representative limb of the 3\protect\underline{R}R[2RR]R Delta
robot. The construction frames $\mathcal{F}_{0}^{\prime }$ and $\mathcal{F}_{%
\mathrm{p}}^{\prime }$ are arbitrarily located. }
\label{figReferenceLimbDelta}
\end{figure}
\begin{figure}[h]
\centerline{
a)~\includegraphics[height=3.5cm]{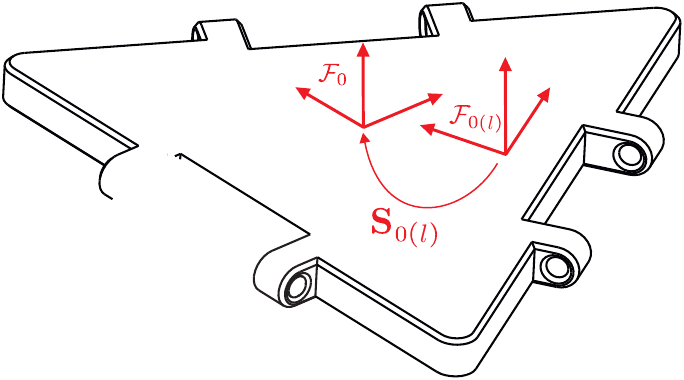}~~~b)~\includegraphics[height=3.5cm]{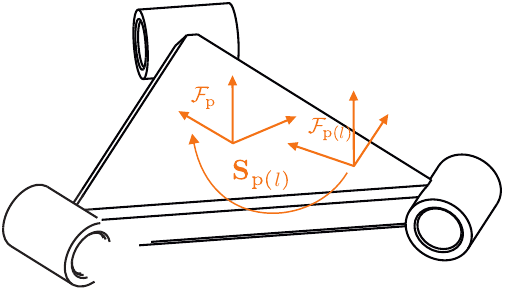}}
\caption{Mount frames $\mathcal{F}_{0\left( l\right) }$ and $\mathcal{F}_{%
\mathrm{p}\left( l\right) }$ at the ground a) and platform b) for the 3%
\protect\underline{R}R[2RR]R Delta. Instances of the representative limb are
inserted between these mount frames, which means that $\mathcal{F}%
_{0}^{\prime }$ (fig. \protect\ref{figReferenceLimbDelta}) of the $l$th
instance is identified with $\mathcal{F}_{0\left( l\right) }$ at the ground,
and $\mathcal{F}_{\mathrm{p}}^{\prime }$ is identified with $\mathcal{F}_{%
\mathrm{p}\left( l\right) }$ at the platform.}
\label{figMountDelta}
\end{figure}

The kinematic and dynamic model of the particular limb $l$ are derived from
the models of the RL as follows.

\paragraph{Kinematics:}

A change of IFR affects the body configurations (\ref{Ckl}), while a change
of platform frame affects the forward kinematics and task space Jacobian of
the limb in (\ref{Lpl}) and (\ref{Vt}), respectively, and hence the inverse
kinematics Jacobian $\mathbf{F}_{\left( l\right) }$ in (\ref{Ftheta1}). The
geometric Jacobians of bodies $i=1,\ldots ,n_{l}$ in (\ref{Jk}) are
unaffected, as they are represented in the body-fixed frames, which are
identically located at all limbs, and the same applies to the compound
Jacobians (\ref{Lkl}).

Denote with $\mathbf{C}_{i}^{\prime }\in SE\left( 3\right) $ the
configuration of body $i$ of the RL relative to the construction frame $%
\mathcal{F}_{0}^{\prime }$. Using (\ref{Ckl}), this is given by $\mathbf{C}%
_{i}^{\prime }(%
%TCIMACRO{\TeXButton{red}{}}%
%BeginExpansion
%
%EndExpansion
%TCIMACRO{\TeXButton{vartheta}{\mathbold{\vartheta}}}%
%BeginExpansion
\mathbold{\vartheta}%
%EndExpansion
%TCIMACRO{\TeXButton{black}{\color{black}}}%
%BeginExpansion
\color{black}%
%EndExpansion
^{\prime })=f_{i}^{\prime }(%
%TCIMACRO{\TeXButton{red}{}}%
%BeginExpansion
%
%EndExpansion
%TCIMACRO{\TeXButton{vartheta}{\mathbold{\vartheta}}}%
%BeginExpansion
\mathbold{\vartheta}%
%EndExpansion
%TCIMACRO{\TeXButton{black}{\color{black}}}%
%BeginExpansion
\color{black}%
%EndExpansion
^{\prime })\mathbf{A}_{i}^{\prime }$ with%
\begin{equation}
f_{i}^{\prime }(%
%TCIMACRO{\TeXButton{red}{}}%
%BeginExpansion
%
%EndExpansion
%TCIMACRO{\TeXButton{vartheta}{\mathbold{\vartheta}}}%
%BeginExpansion
\mathbold{\vartheta}%
%EndExpansion
%TCIMACRO{\TeXButton{black}{\color{black}}}%
%BeginExpansion
\color{black}%
%EndExpansion
^{\prime })=\exp \left( \vartheta _{\underline{i}}^{\prime }\mathbf{Y}_{%
\underline{i}}^{\prime }\right) \cdot \ldots \cdot \exp \left( \vartheta
_{i-1}^{\prime }\mathbf{Y}_{i-1}^{\prime }\right) \exp \left( \vartheta
_{i}^{\prime }\mathbf{Y}_{i}^{\prime }\right)  \label{fprime}
\end{equation}%
where $\mathbf{A}_{i}^{\prime }$ is the zero reference configuration
relative to $\mathcal{F}_{0}^{\prime }$, $\mathbf{Y}_{\underline{i}}^{\prime
}$ are the screw coordinate vectors represented in the construction frame $%
\mathcal{F}_{0}^{\prime }$, and $%
%TCIMACRO{\TeXButton{eta}{\mathbold{\eta}}}%
%BeginExpansion
\mathbold{\eta}%
%EndExpansion
^{\prime }$ is the joint coordinate vector of the RL.

The configuration of limb $l$ is obtained by transforming (\ref{fprime}) to $%
\mathcal{F}_{0}$. The transformation from mount frame $\mathcal{F}_{0\left(
l\right) }$ (which is identical to $\mathcal{F}_{0}^{\prime }$ when the limb
is mounted) to the global IFR $\mathcal{F}_{0}$ is denoted with $\mathbf{S}%
_{0\left( l\right) }\in SE\left( 3\right) $. Thus, the configuration of body 
$i$ of limb $l$ relative to the IFR $\mathcal{F}_{0}$ is 
\begin{equation}
\mathbf{C}_{i\left( l\right) }=\mathbf{S}_{0\left( l\right) }\mathbf{C}%
_{i\left( l\right) }^{\prime }=\mathbf{S}_{0\left( l\right) }f_{i}^{\prime }(%
%TCIMACRO{\TeXButton{red}{}}%
%BeginExpansion
%
%EndExpansion
%TCIMACRO{\TeXButton{vartheta}{\mathbold{\vartheta}}}%
%BeginExpansion
\mathbold{\vartheta}%
%EndExpansion
%TCIMACRO{\TeXButton{black}{\color{black}}}%
%BeginExpansion
\color{black}%
%EndExpansion
_{\left( l\right) })\mathbf{A}_{i}^{\prime }=f_{i\left( l\right) }(%
%TCIMACRO{\TeXButton{red}{}}%
%BeginExpansion
%
%EndExpansion
%TCIMACRO{\TeXButton{vartheta}{\mathbold{\vartheta}}}%
%BeginExpansion
\mathbold{\vartheta}%
%EndExpansion
%TCIMACRO{\TeXButton{black}{\color{black}}}%
%BeginExpansion
\color{black}%
%EndExpansion
_{\left( l\right) })\mathbf{A}_{i\left( l\right) }.  \label{CiS}
\end{equation}%
The last term is of form (\ref{fk}) where the joint screw coordinates and
the reference configuration are%
\begin{equation}
\mathbf{Y}_{i\left( l\right) }=\mathbf{Ad}_{\mathbf{S}_{0\left( l\right) }}%
\mathbf{Y}_{i}^{\prime },\ \ \mathbf{A}_{i\left( l\right) }=\mathbf{S}%
_{0\left( l\right) }\mathbf{A}_{i}^{\prime }.  \label{S0}
\end{equation}

Denote with $\mathbf{S}_{\mathrm{p}\left( l\right) }\in SE\left( 3\right) $
the transformation from mount frame $\mathcal{F}_{\mathrm{p}\left( l\right)
} $ to the platform frame $\mathcal{F}_{\mathrm{p}}$ of the PKM. The
reference configuration of $\mathcal{F}_{\mathrm{p}}^{\prime }$ relative to $%
\mathcal{F}_{0}^{\prime }$ of the RL is $\mathbf{A}_{\mathrm{p}}^{\prime }$.
When the limb is mounted, $\mathcal{F}_{\mathrm{p}}^{\prime }$ is identical
to $\mathcal{F}_{\mathrm{p}\left( l\right) }$, and $\mathcal{F}_{0}^{\prime }
$ is identical to $\mathcal{F}_{0\left( l\right) }$. The platform pose of
the PKM is then, with (\ref{Cp}), determined by limb $l$ as $\mathbf{C}_{%
\mathrm{p}}=\mathbf{S}_{0\left( l\right) }\mathbf{C}_{\mathrm{p}}^{\prime }%
\mathbf{S}_{\mathrm{p}\left( l\right) }^{-1}=f_{\mathrm{p}\left( l\right) }(%
%TCIMACRO{\TeXButton{red}{}}%
%BeginExpansion
%
%EndExpansion
%TCIMACRO{\TeXButton{vartheta}{\mathbold{\vartheta}}}%
%BeginExpansion
\mathbold{\vartheta}%
%EndExpansion
%TCIMACRO{\TeXButton{black}{\color{black}}}%
%BeginExpansion
\color{black}%
%EndExpansion
_{\left( l\right) })\mathbf{A}_{\mathrm{p}\left( l\right) }$, where $\mathbf{%
A}_{\mathrm{p}}=\mathbf{S}_{0\left( l\right) }\mathbf{A}_{\mathrm{p}%
}^{\prime }\mathbf{S}_{\mathrm{p}\left( l\right) }^{-1}$.

The twist $\mathbf{V}_{\mathrm{p}}^{\prime }$ of frame $\mathcal{F}_{\mathrm{%
p}}^{\prime }$ of the RL is determined by the forward kinematics Jacobian $%
\mathbf{J}_{\mathrm{p}}^{\prime }$ of the RL, and is related to the platform
twist as $\mathbf{V}_{\mathrm{p}\left( l\right) }=\mathbf{Ad}_{\mathbf{S}_{%
\mathrm{p}\left( l\right) }}\mathbf{V}_{\mathrm{p}}^{\prime }$. The forward
kinematics Jacobian of the tree-topology system of limb $l$ is thus%
\begin{equation}
\mathbf{J}_{\mathrm{p}\left( l\right) }=\mathbf{Ad}_{\mathbf{S}_{\mathrm{p}%
\left( l\right) }}\mathbf{J}_{\mathrm{p}}^{\prime }  \label{Jpl3}
\end{equation}%
and hence the compound forward kinematics Jacobian in (\ref{Lpl}) is%
\begin{equation}
\mathbf{L}_{\mathrm{p}\left( l\right) }=\mathbf{Ad}_{\mathbf{S}_{\mathrm{p}%
\left( l\right) }}\mathbf{L}_{\mathrm{p}}^{\prime }  \label{LpS}
\end{equation}%
which gives rise to the task space Jacobian (\ref{Vt}) and the inverse
kinematics Jacobian $\mathbf{F}_{\left( l\right) }$.

\paragraph{Dynamics:}

The generalized mass and centrifugal/Coriolis matrix in the EOM (\ref%
{EOMLimb}) of the tree-topology system are invariant w.r.t. to a change of
IFR (i.e. they are left-invariant under $SE\left( 3\right) $ actions). They
can be derived and implemented for the RL using (\ref{Ml}) and (\ref%
{Coriolisl}). The only term in the EOM (\ref{EOMLimb}) that depends on the
IFR is the vector of generalized gravity forces, which is determined by the
relation (\ref{Qgravl}) in appendix \ref{AppendixEOMTree}. Let $\mathsf{J}%
^{\prime }$ be the system Jacobian of the RL defined in (\ref{JbSys}).
Introducing the transformation $\mathbf{S}_{0\left( l\right) }$ from $%
\mathcal{F}_{0\left( l\right) }$ to $\mathcal{F}_{0}$ yields the generalized
gravity forces of limb $l$ 
\begin{equation}
\mathbf{Q}_{\left( l\right) }^{\mathrm{grav}}=-\mathsf{J}^{\prime T}\left( 
\begin{array}{c}
\mathbf{M}_{1}\mathbf{Ad}_{\mathbf{C}_{1}^{\prime }}^{-1} \\ 
\mathbf{M}_{2}\mathbf{Ad}_{\mathbf{C}_{2}^{\prime }}^{-1} \\ 
\vdots \\ 
\mathbf{M}_{n_{l}}\mathbf{Ad}_{\mathbf{C}_{n_{l}}^{\prime }}^{-1}%
\end{array}%
\right) _{%
%TCIMACRO{\TeXButton{-1ex}{\hspace{-1ex}}}%
%BeginExpansion
\hspace{-1ex}%
%EndExpansion
\left( l\right) }\mathbf{Ad}_{\mathbf{S}_{0\left( l\right) }}^{-1}\left( 
\begin{array}{c}
\mathbf{0} \\ 
^{0}\mathbf{g}%
\end{array}%
\right) =-\mathsf{J}^{\prime T}\left( 
\begin{array}{c}
\mathbf{M}_{1}\mathbf{Ad}_{\mathbf{C}_{1}^{\prime }}^{-1} \\ 
\mathbf{M}_{2}\mathbf{Ad}_{\mathbf{C}_{2}^{\prime }}^{-1} \\ 
\vdots \\ 
\mathbf{M}_{n_{l}}\mathbf{Ad}_{\mathbf{C}_{n_{l}}^{\prime }}^{-1}%
\end{array}%
\right) _{%
%TCIMACRO{\TeXButton{-1ex}{\hspace{-1ex}}}%
%BeginExpansion
\hspace{-1ex}%
%EndExpansion
\left( l\right) }\left( 
\begin{array}{c}
\mathbf{0} \\ 
\mathbf{R}_{0\left( l\right) }^{T}{^{0}}\mathbf{g}%
\end{array}%
\right) .
\end{equation}%
Obviously, the generalized gravity forces are obtained from the model for
the RL by using the gravity vector $\mathbf{R}_{0\left( l\right) }^{T}{^{0}}%
\mathbf{g}$ (instead of ${^{0}}\mathbf{g}$) where $\mathbf{R}_{0\left(
l\right) }$ is the rotation matrix in $\mathbf{S}_{0\left( l\right) }$.

\section{Applications of the Dynamic Equations%
%TCIMACRO{\TeXButton{secApplications}{\label{secApplications}}}%
%BeginExpansion
\label{secApplications}%
%EndExpansion
}

The dynamic model can be employed to various means, in particular for
solving the forward and inverse dynamics problem. By construction, (\ref%
{EOMTask}) depends on $%
%TCIMACRO{\TeXButton{red}{}}%
%BeginExpansion
%
%EndExpansion
%TCIMACRO{\TeXButton{vartheta}{\mathbold{\vartheta}}}%
%BeginExpansion
\mathbold{\vartheta}%
%EndExpansion
%TCIMACRO{\TeXButton{black}{\color{black}}}%
%BeginExpansion
\color{black}%
%EndExpansion
,%
%TCIMACRO{\TeXButton{red}{}}%
%BeginExpansion
%
%EndExpansion
\dot{%
%TCIMACRO{\TeXButton{vartheta}{\mathbold{\vartheta}}}%
%BeginExpansion
\mathbold{\vartheta}%
%EndExpansion
}%
%TCIMACRO{\TeXButton{black}{\color{black}}}%
%BeginExpansion
\color{black}%
%EndExpansion
$, as well as on $\mathbf{V}_{\mathrm{t}},\dot{\mathbf{V}}_{\mathrm{t}}$,
which must be taken into account when applying the model. In the following
this is discussed in detail.

\subsection{Forward Dynamics --Time Integration of the EOM%
%TCIMACRO{\TeXButton{secAppForwDyn}{\label{secAppForwDyn}}}%
%BeginExpansion
\label{secAppForwDyn}%
%EndExpansion
}

\subsubsection{EOM in terms of taskspace coordinates and velocity}

If the inverse kinematics map $\psi _{\mathrm{IK}}$ of the mechanism in (\ref%
{GeomInvLimb}) is available in closed form, $%
%TCIMACRO{\TeXButton{red}{}}%
%BeginExpansion
%
%EndExpansion
%TCIMACRO{\TeXButton{vartheta}{\mathbold{\vartheta}}}%
%BeginExpansion
\mathbold{\vartheta}%
%EndExpansion
%TCIMACRO{\TeXButton{black}{\color{black}}}%
%BeginExpansion
\color{black}%
%EndExpansion
$ can be replaced with $\mathbf{x}$. The joint velocity $%
%TCIMACRO{\TeXButton{red}{}}%
%BeginExpansion
%
%EndExpansion
\dot{%
%TCIMACRO{\TeXButton{vartheta}{\mathbold{\vartheta}}}%
%BeginExpansion
\mathbold{\vartheta}%
%EndExpansion
}%
%TCIMACRO{\TeXButton{black}{\color{black}}}%
%BeginExpansion
\color{black}%
%EndExpansion
$ can be replaced with $\mathbf{V}_{\mathrm{t}}$ using one of the relations (%
\ref{Ftheta1}), (\ref{Ftheta2}) or (\ref{Ftheta3}), or using the closed form
relation for the particular PKM. Complementing (\ref{EOMTask}) with the
kinematic equations (\ref{Vpx}) leads to the EOM in terms of taskspace
coordinates and velocity%
\begin{eqnarray}
\mathbf{M}_{\mathrm{t}}\left( \mathbf{x}\right) \dot{\mathbf{V}}_{\mathrm{t}%
}+\mathbf{C}_{\mathrm{t}}\left( \mathbf{x,V}_{\mathrm{t}}\right) \mathbf{V}_{%
\mathrm{t}}+\mathbf{W}_{\mathrm{t}}^{\mathrm{grav}}\left( \mathbf{x}\right) +%
\mathbf{W}_{\mathrm{t}}\left( \mathbf{x},\mathbf{V}_{\mathrm{t}},t\right) &=&%
\mathbf{W}_{\mathrm{t}}^{\mathrm{EE}}\left( t\right) +\mathbf{J}_{\mathrm{IK}%
}^{T}\left( \mathbf{x}\right) \mathbf{u}\left( t\right)  \label{EOM11} \\
\mathbf{V}_{\mathrm{t}} &=&\mathbf{H}_{\mathrm{t}}\left( \mathbf{x}\right) 
\dot{\mathbf{x}}.  \label{EOM12}
\end{eqnarray}%
This is a system of $2\delta _{\mathrm{p}}$ first-order ODEs in terms of the
taskspace coordinates $\mathbf{x}$ and velocities $\mathbf{V}_{\mathrm{t}}$,
which is linear in $\dot{\mathbf{x}}$ and $\dot{\mathbf{V}}_{\mathrm{t}}$.
It can be reformulated as the explicit ODE system%
\begin{eqnarray}
\dot{\mathbf{V}}_{\mathrm{t}} &=&\mathbf{M}_{\mathrm{t}}^{-1}\left( \mathbf{x%
}\right) \left( \mathbf{W}_{\mathrm{t}}^{\mathrm{EE}}\left( t\right) +%
\mathbf{J}_{\mathrm{IK}}^{T}\left( \mathbf{x}\right) \mathbf{u}\left(
t\right) -\mathbf{C}_{\mathrm{t}}\left( \mathbf{x,V}_{\mathrm{t}}\right) 
\mathbf{V}_{\mathrm{t}}-\mathbf{W}_{\mathrm{t}}^{\mathrm{grav}}\left( 
\mathbf{x}\right) -\mathbf{W}_{\mathrm{t}}\left( \mathbf{x},\mathbf{V}_{%
\mathrm{t}},t\right) \right)  \label{EOM13} \\
\dot{\mathbf{x}} &=&\mathbf{H}_{\mathrm{t}}^{-1}\left( \mathbf{x}\right) 
\mathbf{V}_{\mathrm{t}}.  \label{EOM14}
\end{eqnarray}%
The parameterization of the platform motion with coordinates $\mathbf{x}$
may introduce parameterization singularities, which happens when a
three-parametric description of the platform orientation used. In such
singularities, the inverse in (\ref{EOM14}) does not exist. The kinematic
equations (\ref{EOM12}) could be inserted into (\ref{EOM11}) to yield a
system of $\delta _{\mathrm{p}}=\delta $ second-order ODEs in $\mathbf{x}%
\left( t\right) $. This, however, usually leads to very complicated
expressions, and it is advisable to use the platform twist as velocity
variables in (\ref{EOM11}). The singularity problem remains.

\subsubsection{EOM in terms of joint coordinates and taskspace velocity}

In general there is no closed form expression of the inverse kinematics map $%
\psi _{\mathrm{IK}}$. Moreover, inserting it into (\ref{EOMTask}) often
yields equations (\ref{EOM11}) with very complicated terms. Instead, the
dynamic equations (\ref{EOMTask}) are complemented with the kinematic
equations (\ref{HFl}). This yields the EOM of the PKM with complex limbs%
\begin{eqnarray}
\mathbf{M}_{\mathrm{t}}(%
%TCIMACRO{\TeXButton{red}{}}%
%BeginExpansion
%
%EndExpansion
%TCIMACRO{\TeXButton{vartheta}{\mathbold{\vartheta}}}%
%BeginExpansion
\mathbold{\vartheta}%
%EndExpansion
%TCIMACRO{\TeXButton{black}{\color{black}}}%
%BeginExpansion
\color{black}%
%EndExpansion
)\dot{\mathbf{V}}_{\mathrm{t}}+\mathbf{C}_{\mathrm{t}}(%
%TCIMACRO{\TeXButton{red}{}}%
%BeginExpansion
%
%EndExpansion
%TCIMACRO{\TeXButton{vartheta}{\mathbold{\vartheta}}}%
%BeginExpansion
\mathbold{\vartheta}%
%EndExpansion
%TCIMACRO{\TeXButton{black}{\color{black}}}%
%BeginExpansion
\color{black}%
%EndExpansion
,%
%TCIMACRO{\TeXButton{red}{}}%
%BeginExpansion
%
%EndExpansion
\dot{%
%TCIMACRO{\TeXButton{vartheta}{\mathbold{\vartheta}}}%
%BeginExpansion
\mathbold{\vartheta}%
%EndExpansion
}%
%TCIMACRO{\TeXButton{black}{\color{black}}}%
%BeginExpansion
\color{black}%
%EndExpansion
)\mathbf{V}_{\mathrm{t}}+\mathbf{W}_{\mathrm{t}}^{\mathrm{grav}}\left( 
%TCIMACRO{\TeXButton{red}{}}%
%BeginExpansion
%
%EndExpansion
%TCIMACRO{\TeXButton{vartheta}{\mathbold{\vartheta}}}%
%BeginExpansion
\mathbold{\vartheta}%
%EndExpansion
%TCIMACRO{\TeXButton{black}{\color{black}}}%
%BeginExpansion
\color{black}%
%EndExpansion
\right) +\mathbf{W}_{\mathrm{t}}(%
%TCIMACRO{\TeXButton{red}{}}%
%BeginExpansion
%
%EndExpansion
%TCIMACRO{\TeXButton{vartheta}{\mathbold{\vartheta}}}%
%BeginExpansion
\mathbold{\vartheta}%
%EndExpansion
%TCIMACRO{\TeXButton{black}{\color{black}}}%
%BeginExpansion
\color{black}%
%EndExpansion
,%
%TCIMACRO{\TeXButton{red}{}}%
%BeginExpansion
%
%EndExpansion
\dot{%
%TCIMACRO{\TeXButton{vartheta}{\mathbold{\vartheta}}}%
%BeginExpansion
\mathbold{\vartheta}%
%EndExpansion
}%
%TCIMACRO{\TeXButton{black}{\color{black}}}%
%BeginExpansion
\color{black}%
%EndExpansion
,t) &=&\mathbf{W}_{\mathrm{t}}^{\mathrm{EE}}\left( t\right) +\mathbf{J}_{%
\mathrm{IK}}^{T}(%
%TCIMACRO{\TeXButton{red}{}}%
%BeginExpansion
%
%EndExpansion
%TCIMACRO{\TeXButton{vartheta}{\mathbold{\vartheta}}}%
%BeginExpansion
\mathbold{\vartheta}%
%EndExpansion
%TCIMACRO{\TeXButton{black}{\color{black}}}%
%BeginExpansion
\color{black}%
%EndExpansion
)\mathbf{u}\left( t\right)  \label{EOM21} \\
%TCIMACRO{\TeXButton{red}{}}%
%BeginExpansion
%
%EndExpansion
\dot{%
%TCIMACRO{\TeXButton{vartheta}{\mathbold{\vartheta}}}%
%BeginExpansion
\mathbold{\vartheta}%
%EndExpansion
}%
%TCIMACRO{\TeXButton{black}{\color{black}}}%
%BeginExpansion
\color{black}%
%EndExpansion
_{\left( l\right) } &=&\mathbf{H}_{\left( l\right) }(%
%TCIMACRO{\TeXButton{red}{}}%
%BeginExpansion
%
%EndExpansion
%TCIMACRO{\TeXButton{vartheta}{\mathbold{\vartheta}}}%
%BeginExpansion
\mathbold{\vartheta}%
%EndExpansion
%TCIMACRO{\TeXButton{black}{\color{black}}}%
%BeginExpansion
\color{black}%
%EndExpansion
)\mathbf{F}_{\left( l\right) }(%
%TCIMACRO{\TeXButton{red}{}}%
%BeginExpansion
%
%EndExpansion
%TCIMACRO{\TeXButton{vartheta}{\mathbold{\vartheta}}}%
%BeginExpansion
\mathbold{\vartheta}%
%EndExpansion
%TCIMACRO{\TeXButton{black}{\color{black}}}%
%BeginExpansion
\color{black}%
%EndExpansion
)\mathbf{J}_{\mathrm{FK}}(%
%TCIMACRO{\TeXButton{red}{}}%
%BeginExpansion
%
%EndExpansion
%TCIMACRO{\TeXButton{vartheta}{\mathbold{\vartheta}}}%
%BeginExpansion
\mathbold{\vartheta}%
%EndExpansion
%TCIMACRO{\TeXButton{black}{\color{black}}}%
%BeginExpansion
\color{black}%
%EndExpansion
)\mathbf{V}_{\mathrm{t}}  \label{EOM22}
\end{eqnarray}%
which is a system of $\delta _{\mathrm{p}}+n$ first-order ODEs in terms of
joint variables $%
%TCIMACRO{\TeXButton{red}{}}%
%BeginExpansion
%
%EndExpansion
%TCIMACRO{\TeXButton{vartheta}{\mathbold{\vartheta}}}%
%BeginExpansion
\mathbold{\vartheta}%
%EndExpansion
%TCIMACRO{\TeXButton{black}{\color{black}}}%
%BeginExpansion
\color{black}%
%EndExpansion
$ of the tree-topology system and the taskspace velocity $\mathbf{V}_{%
\mathrm{t}}$ that can be regarded as non-collocated state variables. It can
written as explicit ODE system%
\begin{eqnarray}
\dot{\mathbf{V}}_{\mathrm{t}} &=&\mathbf{M}_{\mathrm{t}}^{-1}(%
%TCIMACRO{\TeXButton{red}{}}%
%BeginExpansion
%
%EndExpansion
%TCIMACRO{\TeXButton{vartheta}{\mathbold{\vartheta}}}%
%BeginExpansion
\mathbold{\vartheta}%
%EndExpansion
%TCIMACRO{\TeXButton{black}{\color{black}}}%
%BeginExpansion
\color{black}%
%EndExpansion
)\left( \mathbf{W}_{\mathrm{t}}^{\mathrm{EE}}\left( t\right) +\mathbf{J}_{%
\mathrm{IK}}^{T}(%
%TCIMACRO{\TeXButton{red}{}}%
%BeginExpansion
%
%EndExpansion
%TCIMACRO{\TeXButton{vartheta}{\mathbold{\vartheta}}}%
%BeginExpansion
\mathbold{\vartheta}%
%EndExpansion
%TCIMACRO{\TeXButton{black}{\color{black}}}%
%BeginExpansion
\color{black}%
%EndExpansion
)\mathbf{u}\left( t\right) -\mathbf{C}_{\mathrm{t}}(%
%TCIMACRO{\TeXButton{red}{}}%
%BeginExpansion
%
%EndExpansion
%TCIMACRO{\TeXButton{vartheta}{\mathbold{\vartheta}}}%
%BeginExpansion
\mathbold{\vartheta}%
%EndExpansion
%TCIMACRO{\TeXButton{black}{\color{black}}}%
%BeginExpansion
\color{black}%
%EndExpansion
,\mathbf{V}_{\mathrm{t}})\mathbf{V}_{\mathrm{t}}-\mathbf{W}_{\mathrm{t}}^{%
\mathrm{grav}}\left( 
%TCIMACRO{\TeXButton{red}{}}%
%BeginExpansion
%
%EndExpansion
%TCIMACRO{\TeXButton{vartheta}{\mathbold{\vartheta}}}%
%BeginExpansion
\mathbold{\vartheta}%
%EndExpansion
%TCIMACRO{\TeXButton{black}{\color{black}}}%
%BeginExpansion
\color{black}%
%EndExpansion
\right) -\mathbf{W}_{\mathrm{t}}(%
%TCIMACRO{\TeXButton{red}{}}%
%BeginExpansion
%
%EndExpansion
%TCIMACRO{\TeXButton{vartheta}{\mathbold{\vartheta}}}%
%BeginExpansion
\mathbold{\vartheta}%
%EndExpansion
%TCIMACRO{\TeXButton{black}{\color{black}}}%
%BeginExpansion
\color{black}%
%EndExpansion
,\mathbf{V}_{\mathrm{t}},t)\right)  \label{EOM23} \\
%TCIMACRO{\TeXButton{red}{}}%
%BeginExpansion
%
%EndExpansion
\dot{%
%TCIMACRO{\TeXButton{vartheta}{\mathbold{\vartheta}}}%
%BeginExpansion
\mathbold{\vartheta}%
%EndExpansion
}%
%TCIMACRO{\TeXButton{black}{\color{black}}}%
%BeginExpansion
\color{black}%
%EndExpansion
_{\left( l\right) } &=&\mathbf{H}_{\left( l\right) }(%
%TCIMACRO{\TeXButton{red}{}}%
%BeginExpansion
%
%EndExpansion
%TCIMACRO{\TeXButton{vartheta}{\mathbold{\vartheta}}}%
%BeginExpansion
\mathbold{\vartheta}%
%EndExpansion
%TCIMACRO{\TeXButton{black}{\color{black}}}%
%BeginExpansion
\color{black}%
%EndExpansion
)\mathbf{F}_{\left( l\right) }(%
%TCIMACRO{\TeXButton{red}{}}%
%BeginExpansion
%
%EndExpansion
%TCIMACRO{\TeXButton{vartheta}{\mathbold{\vartheta}}}%
%BeginExpansion
\mathbold{\vartheta}%
%EndExpansion
%TCIMACRO{\TeXButton{black}{\color{black}}}%
%BeginExpansion
\color{black}%
%EndExpansion
)\mathbf{V}_{\mathrm{t}}  \label{EOM24}
\end{eqnarray}%
where $\mathbf{C}_{\mathrm{t}}(%
%TCIMACRO{\TeXButton{red}{}}%
%BeginExpansion
%
%EndExpansion
%TCIMACRO{\TeXButton{vartheta}{\mathbold{\vartheta}}}%
%BeginExpansion
\mathbold{\vartheta}%
%EndExpansion
%TCIMACRO{\TeXButton{black}{\color{black}}}%
%BeginExpansion
\color{black}%
%EndExpansion
,%
%TCIMACRO{\TeXButton{red}{}}%
%BeginExpansion
%
%EndExpansion
\dot{%
%TCIMACRO{\TeXButton{vartheta}{\mathbold{\vartheta}}}%
%BeginExpansion
\mathbold{\vartheta}%
%EndExpansion
}%
%TCIMACRO{\TeXButton{black}{\color{black}}}%
%BeginExpansion
\color{black}%
%EndExpansion
)$ and $\mathbf{W}_{\mathrm{t}}(%
%TCIMACRO{\TeXButton{red}{}}%
%BeginExpansion
%
%EndExpansion
%TCIMACRO{\TeXButton{vartheta}{\mathbold{\vartheta}}}%
%BeginExpansion
\mathbold{\vartheta}%
%EndExpansion
%TCIMACRO{\TeXButton{black}{\color{black}}}%
%BeginExpansion
\color{black}%
%EndExpansion
,%
%TCIMACRO{\TeXButton{red}{}}%
%BeginExpansion
%
%EndExpansion
\dot{%
%TCIMACRO{\TeXButton{vartheta}{\mathbold{\vartheta}}}%
%BeginExpansion
\mathbold{\vartheta}%
%EndExpansion
}%
%TCIMACRO{\TeXButton{black}{\color{black}}}%
%BeginExpansion
\color{black}%
%EndExpansion
,t)$ in (\ref{EOM23}) are evaluated with (\ref{EOM24}) for given state
vector $(\mathbf{V}_{\mathrm{t}},%
%TCIMACRO{\TeXButton{red}{}}%
%BeginExpansion
%
%EndExpansion
%TCIMACRO{\TeXButton{vartheta}{\mathbold{\vartheta}}}%
%BeginExpansion
\mathbold{\vartheta}%
%EndExpansion
%TCIMACRO{\TeXButton{black}{\color{black}}}%
%BeginExpansion
\color{black}%
%EndExpansion
)$.

The joint velocities $%
%TCIMACRO{\TeXButton{red}{}}%
%BeginExpansion
%
%EndExpansion
\dot{%
%TCIMACRO{\TeXButton{vartheta}{\mathbold{\vartheta}}}%
%BeginExpansion
\mathbold{\vartheta}%
%EndExpansion
}%
%TCIMACRO{\TeXButton{black}{\color{black}}}%
%BeginExpansion
\color{black}%
%EndExpansion
$ can be replaced with $\mathbf{V}_{\mathrm{t}}$ using (\ref{Ftheta1}), (\ref%
{Ftheta2}) or (\ref{Ftheta3}). Moreover, if the inverse kinematics of the
mechanism can be solved explicitly $%
%TCIMACRO{\TeXButton{red}{}}%
%BeginExpansion
%
%EndExpansion
%TCIMACRO{\TeXButton{vartheta}{\mathbold{\vartheta}}}%
%BeginExpansion
\mathbold{\vartheta}%
%EndExpansion
%TCIMACRO{\TeXButton{black}{\color{black}}}%
%BeginExpansion
\color{black}%
%EndExpansion
$ can be replaced by $\mathbf{x}$, so that all terms in (\ref{EOMTask})
depend on the state $(\mathbf{x},\mathbf{V}_{\mathrm{t}})$ of the platform.
This often leads to very complex expressions, and it may be computationally
more efficient to separately evaluate the inverse kinematics and to insert
the result into the dynamic EOM (see sec. \ref{secImplement}).

\subsubsection{EOM in terms of actuator coordinates}

The dynamic equations (\ref{EOMact}) are complemented with the kinematic
relation (\ref{HFJl}) 
\begin{eqnarray}
\mathbf{M}_{\mathrm{a}}(%
%TCIMACRO{\TeXButton{red}{}}%
%BeginExpansion
%
%EndExpansion
%TCIMACRO{\TeXButton{vartheta}{\mathbold{\vartheta}}}%
%BeginExpansion
\mathbold{\vartheta}%
%EndExpansion
%TCIMACRO{\TeXButton{black}{\color{black}}}%
%BeginExpansion
\color{black}%
%EndExpansion
)\ddot{%
%TCIMACRO{\TeXButton{vartheta}{\mathbold{\vartheta}}}%
%BeginExpansion
\mathbold{\vartheta}%
%EndExpansion
}_{\mathrm{act}}+\mathbf{C}_{\mathrm{a}}(%
%TCIMACRO{\TeXButton{red}{}}%
%BeginExpansion
%
%EndExpansion
%TCIMACRO{\TeXButton{vartheta}{\mathbold{\vartheta}}}%
%BeginExpansion
\mathbold{\vartheta}%
%EndExpansion
%TCIMACRO{\TeXButton{black}{\color{black}}}%
%BeginExpansion
\color{black}%
%EndExpansion
,%
%TCIMACRO{\TeXButton{red}{}}%
%BeginExpansion
%
%EndExpansion
\dot{%
%TCIMACRO{\TeXButton{vartheta}{\mathbold{\vartheta}}}%
%BeginExpansion
\mathbold{\vartheta}%
%EndExpansion
}%
%TCIMACRO{\TeXButton{black}{\color{black}}}%
%BeginExpansion
\color{black}%
%EndExpansion
)\dot{%
%TCIMACRO{\TeXButton{vartheta}{\mathbold{\vartheta}}}%
%BeginExpansion
\mathbold{\vartheta}%
%EndExpansion
}_{\mathrm{act}}+\mathbf{Q}_{\mathrm{a}}^{\mathrm{grav}}(%
%TCIMACRO{\TeXButton{red}{}}%
%BeginExpansion
%
%EndExpansion
\dot{%
%TCIMACRO{\TeXButton{vartheta}{\mathbold{\vartheta}}}%
%BeginExpansion
\mathbold{\vartheta}%
%EndExpansion
}%
%TCIMACRO{\TeXButton{black}{\color{black}}}%
%BeginExpansion
\color{black}%
%EndExpansion
)+\mathbf{Q}_{\mathrm{a}}(%
%TCIMACRO{\TeXButton{red}{}}%
%BeginExpansion
%
%EndExpansion
%TCIMACRO{\TeXButton{vartheta}{\mathbold{\vartheta}}}%
%BeginExpansion
\mathbold{\vartheta}%
%EndExpansion
%TCIMACRO{\TeXButton{black}{\color{black}}}%
%BeginExpansion
\color{black}%
%EndExpansion
,%
%TCIMACRO{\TeXButton{red}{}}%
%BeginExpansion
%
%EndExpansion
\dot{%
%TCIMACRO{\TeXButton{vartheta}{\mathbold{\vartheta}}}%
%BeginExpansion
\mathbold{\vartheta}%
%EndExpansion
}%
%TCIMACRO{\TeXButton{black}{\color{black}}}%
%BeginExpansion
\color{black}%
%EndExpansion
,t) &=&\mathbf{Q}_{\mathrm{a}}^{\mathrm{EE}}(%
%TCIMACRO{\TeXButton{red}{}}%
%BeginExpansion
%
%EndExpansion
%TCIMACRO{\TeXButton{vartheta}{\mathbold{\vartheta}}}%
%BeginExpansion
\mathbold{\vartheta}%
%EndExpansion
%TCIMACRO{\TeXButton{black}{\color{black}}}%
%BeginExpansion
\color{black}%
%EndExpansion
)+\mathbf{u}  \label{ODE1} \\
%TCIMACRO{\TeXButton{red}{}}%
%BeginExpansion
%
%EndExpansion
\dot{%
%TCIMACRO{\TeXButton{vartheta}{\mathbold{\vartheta}}}%
%BeginExpansion
\mathbold{\vartheta}%
%EndExpansion
}%
%TCIMACRO{\TeXButton{black}{\color{black}}}%
%BeginExpansion
\color{black}%
%EndExpansion
_{\left( l\right) } &=&\mathbf{H}_{\left( l\right) }(%
%TCIMACRO{\TeXButton{red}{}}%
%BeginExpansion
%
%EndExpansion
%TCIMACRO{\TeXButton{vartheta}{\mathbold{\vartheta}}}%
%BeginExpansion
\mathbold{\vartheta}%
%EndExpansion
%TCIMACRO{\TeXButton{black}{\color{black}}}%
%BeginExpansion
\color{black}%
%EndExpansion
)\mathbf{F}_{\left( l\right) }(%
%TCIMACRO{\TeXButton{red}{}}%
%BeginExpansion
%
%EndExpansion
%TCIMACRO{\TeXButton{vartheta}{\mathbold{\vartheta}}}%
%BeginExpansion
\mathbold{\vartheta}%
%EndExpansion
%TCIMACRO{\TeXButton{black}{\color{black}}}%
%BeginExpansion
\color{black}%
%EndExpansion
)\dot{%
%TCIMACRO{\TeXButton{vartheta}{\mathbold{\vartheta}}}%
%BeginExpansion
\mathbold{\vartheta}%
%EndExpansion
}_{\mathrm{act}}.  \label{ODE2}
\end{eqnarray}%
This is a system of $\delta +n$ ODEs. The dynamic equations (\ref{ODE1})
form a second-order ODE system in $%
%TCIMACRO{\TeXButton{vartheta}{\mathbold{\vartheta}}}%
%BeginExpansion
\mathbold{\vartheta}%
%EndExpansion
_{\mathrm{act}}$, while the kinematic equations (\ref{ODE2}) form a
first-order system. Remember that $%
%TCIMACRO{\TeXButton{vartheta}{\mathbold{\vartheta}}}%
%BeginExpansion
\mathbold{\vartheta}%
%EndExpansion
_{\mathrm{act}}$ is a subset of $%
%TCIMACRO{\TeXButton{red}{}}%
%BeginExpansion
%
%EndExpansion
\dot{%
%TCIMACRO{\TeXButton{vartheta}{\mathbold{\vartheta}}}%
%BeginExpansion
\mathbold{\vartheta}%
%EndExpansion
}%
%TCIMACRO{\TeXButton{black}{\color{black}}}%
%BeginExpansion
\color{black}%
%EndExpansion
$.

\subsection{Inverse Dynamics Formulation in Taskspace%
%TCIMACRO{\TeXButton{secAppInvDyn}{\label{secAppInvDyn}}}%
%BeginExpansion
\label{secAppInvDyn}%
%EndExpansion
}

For model-based control, the inverse dynamics in task space coordinates is
most relevant since task space control schemes directly regulate the EE
motion and achieves better tracking performance. Therefore, in the
following, the inverse dynamics in task space is discussed. A joint space
formulation is obtained from (\ref{EOMact}) if desired.

\paragraph{Inverse dynamics solution:}

The inverse dynamics amounts to determine the actuator forces/torques $%
\mathbf{u}\left( t\right) $ for a given motion of the PKM. In the following,
the dynamics model (\ref{EOMTask}) is used. The PKM motion is then
represented by the joint trajectory $%
%TCIMACRO{\TeXButton{eta}{\mathbold{\eta}}}%
%BeginExpansion
\mathbold{\eta}%
%EndExpansion
\left( t\right) $. The inverse dynamics solution for a non-redundantly
actuated PKM ($\delta _{\mathrm{p}}=n_{\mathrm{act}}$) is%
\begin{eqnarray}
\mathbf{u}\left( t\right) &=&\mathbf{J}_{\mathrm{IK}}^{-T}(%
%TCIMACRO{\TeXButton{red}{}}%
%BeginExpansion
%
%EndExpansion
%TCIMACRO{\TeXButton{vartheta}{\mathbold{\vartheta}}}%
%BeginExpansion
\mathbold{\vartheta}%
%EndExpansion
%TCIMACRO{\TeXButton{black}{\color{black}}}%
%BeginExpansion
\color{black}%
%EndExpansion
)\left( \mathbf{M}_{\mathrm{t}}(%
%TCIMACRO{\TeXButton{red}{}}%
%BeginExpansion
%
%EndExpansion
%TCIMACRO{\TeXButton{vartheta}{\mathbold{\vartheta}}}%
%BeginExpansion
\mathbold{\vartheta}%
%EndExpansion
%TCIMACRO{\TeXButton{black}{\color{black}}}%
%BeginExpansion
\color{black}%
%EndExpansion
)\dot{\mathbf{V}}_{\mathrm{t}}+\mathbf{C}_{\mathrm{t}}(%
%TCIMACRO{\TeXButton{red}{}}%
%BeginExpansion
%
%EndExpansion
%TCIMACRO{\TeXButton{vartheta}{\mathbold{\vartheta}}}%
%BeginExpansion
\mathbold{\vartheta}%
%EndExpansion
%TCIMACRO{\TeXButton{black}{\color{black}}}%
%BeginExpansion
\color{black}%
%EndExpansion
,%
%TCIMACRO{\TeXButton{red}{}}%
%BeginExpansion
%
%EndExpansion
\dot{%
%TCIMACRO{\TeXButton{vartheta}{\mathbold{\vartheta}}}%
%BeginExpansion
\mathbold{\vartheta}%
%EndExpansion
}%
%TCIMACRO{\TeXButton{black}{\color{black}}}%
%BeginExpansion
\color{black}%
%EndExpansion
)\mathbf{V}_{\mathrm{t}}+\mathbf{W}_{\mathrm{t}}^{\mathrm{grav}}\left( 
%TCIMACRO{\TeXButton{red}{}}%
%BeginExpansion
%
%EndExpansion
%TCIMACRO{\TeXButton{vartheta}{\mathbold{\vartheta}}}%
%BeginExpansion
\mathbold{\vartheta}%
%EndExpansion
%TCIMACRO{\TeXButton{black}{\color{black}}}%
%BeginExpansion
\color{black}%
%EndExpansion
\right) +\mathbf{W}_{\mathrm{t}}(%
%TCIMACRO{\TeXButton{red}{}}%
%BeginExpansion
%
%EndExpansion
%TCIMACRO{\TeXButton{vartheta}{\mathbold{\vartheta}}}%
%BeginExpansion
\mathbold{\vartheta}%
%EndExpansion
%TCIMACRO{\TeXButton{black}{\color{black}}}%
%BeginExpansion
\color{black}%
%EndExpansion
,%
%TCIMACRO{\TeXButton{red}{}}%
%BeginExpansion
%
%EndExpansion
\dot{%
%TCIMACRO{\TeXButton{vartheta}{\mathbold{\vartheta}}}%
%BeginExpansion
\mathbold{\vartheta}%
%EndExpansion
}%
%TCIMACRO{\TeXButton{black}{\color{black}}}%
%BeginExpansion
\color{black}%
%EndExpansion
,t)-\mathbf{W}_{\mathrm{t}}^{\mathrm{EE}}\left( t\right) \right)
\label{InvDyn} \\
&=&\mathbf{J}_{\mathrm{IK}}^{-T}\left( \varphi _{\mathrm{t}}\left( 
%TCIMACRO{\TeXButton{red}{}}%
%BeginExpansion
%
%EndExpansion
%TCIMACRO{\TeXButton{vartheta}{\mathbold{\vartheta}}}%
%BeginExpansion
\mathbold{\vartheta}%
%EndExpansion
%TCIMACRO{\TeXButton{black}{\color{black}}}%
%BeginExpansion
\color{black}%
%EndExpansion
,%
%TCIMACRO{\TeXButton{red}{}}%
%BeginExpansion
%
%EndExpansion
\dot{%
%TCIMACRO{\TeXButton{vartheta}{\mathbold{\vartheta}}}%
%BeginExpansion
\mathbold{\vartheta}%
%EndExpansion
}%
%TCIMACRO{\TeXButton{black}{\color{black}}}%
%BeginExpansion
\color{black}%
%EndExpansion
,%
%TCIMACRO{\TeXButton{red}{}}%
%BeginExpansion
%
%EndExpansion
\ddot{%
%TCIMACRO{\TeXButton{vartheta}{\mathbold{\vartheta}}}%
%BeginExpansion
\mathbold{\vartheta}%
%EndExpansion
}%
%TCIMACRO{\TeXButton{black}{\color{black}}}%
%BeginExpansion
\color{black}%
%EndExpansion
,\mathbf{x},\mathbf{V}_{\mathrm{t}},\dot{\mathbf{V}}_{\mathrm{t}},t\right) -%
\mathbf{W}_{\mathrm{t}}^{\mathrm{EE}}\left( t\right) \right)  \label{InvDyn2}
\end{eqnarray}%
with 
\begin{equation}
\varphi _{\mathrm{t}}\left( 
%TCIMACRO{\TeXButton{red}{}}%
%BeginExpansion
%
%EndExpansion
%TCIMACRO{\TeXButton{vartheta}{\mathbold{\vartheta}}}%
%BeginExpansion
\mathbold{\vartheta}%
%EndExpansion
%TCIMACRO{\TeXButton{black}{\color{black}}}%
%BeginExpansion
\color{black}%
%EndExpansion
,%
%TCIMACRO{\TeXButton{red}{}}%
%BeginExpansion
%
%EndExpansion
\dot{%
%TCIMACRO{\TeXButton{vartheta}{\mathbold{\vartheta}}}%
%BeginExpansion
\mathbold{\vartheta}%
%EndExpansion
}%
%TCIMACRO{\TeXButton{black}{\color{black}}}%
%BeginExpansion
\color{black}%
%EndExpansion
,%
%TCIMACRO{\TeXButton{red}{}}%
%BeginExpansion
%
%EndExpansion
\ddot{%
%TCIMACRO{\TeXButton{vartheta}{\mathbold{\vartheta}}}%
%BeginExpansion
\mathbold{\vartheta}%
%EndExpansion
}%
%TCIMACRO{\TeXButton{black}{\color{black}}}%
%BeginExpansion
\color{black}%
%EndExpansion
,\mathbf{x},\mathbf{V}_{\mathrm{t}},\dot{\mathbf{V}}_{\mathrm{t}},t\right) :=%
\mathbf{M}_{\mathrm{t}}\dot{\mathbf{V}}_{\mathrm{t}}+\mathbf{C}_{\mathrm{t}}%
\mathbf{V}_{\mathrm{t}}+\mathbf{W}_{\mathrm{t}}^{\mathrm{grav}}+\mathbf{W}_{%
\mathrm{t}}.  \label{phit}
\end{equation}%
If the PKM is redundantly actuated ($\delta _{\mathrm{p}}<n_{\mathrm{act}}$%
), a (weighted) pseudoinverse of $\mathbf{J}_{\mathrm{IK}}^{T}$ is used \cite%
{TRORedPKM}.

The joint trajectory $%
%TCIMACRO{\TeXButton{red}{}}%
%BeginExpansion
%
%EndExpansion
%TCIMACRO{\TeXButton{vartheta}{\mathbold{\vartheta}}}%
%BeginExpansion
\mathbold{\vartheta}%
%EndExpansion
%TCIMACRO{\TeXButton{black}{\color{black}}}%
%BeginExpansion
\color{black}%
%EndExpansion
\left( t\right) $ is determined from the taskspace motion, i.e. from $%
\mathbf{x}\left( t\right) $ and $\mathbf{V}_{\mathrm{t}}$, by evaluating the
inverse kinematics map (\ref{GeomInvLimb}) and inverse kinematics Jacobian (%
\ref{HFl}) of the mechanism. If the joint motion can be expressed explicitly
in terms of task space coordinates $\mathbf{x}$ and velocities $\mathbf{V}_{%
\mathrm{t}}$, then the dynamics model (\ref{EOM11}) is used, and the PKM
motion is deduced from taskspace motion via the inverse kinematics map $f_{%
\mathrm{IK}}$ and Jacobian (\ref{IK}) of the PKM.

\paragraph{Parallel computation:%
%TCIMACRO{\TeXButton{secImplement}{\label{secImplement}}}%
%BeginExpansion
\label{secImplement}%
%EndExpansion
}

The inherent parallel kinematic structure of PKM can be exploited for
separate evaluation of the kinematics and dynamics of the complex limbs in
parallel. To this end, the taskspace formulation (\ref{EOMTask}) is
expressed without substituting the joint velocity and acceleration into (\ref%
{EOMLimb}) using (\ref{etaq2}), as in (\ref{EOMLimb2}), to obtain%
\begin{eqnarray}
\mathbf{J}_{\mathrm{IK}}^{T}\mathbf{u}\left( t\right) &=&\sum_{l=1}^{L}\bar{%
\mathbf{F}}_{\left( l\right) }^{T}\bar{\mathbf{H}}_{\left( l\right) }^{T}%
%TCIMACRO{\TeXButton{big}{\big}}%
%BeginExpansion
\big%
%EndExpansion
(\bar{\mathbf{M}}_{\left( l\right) }%
%TCIMACRO{\TeXButton{red}{}}%
%BeginExpansion
%
%EndExpansion
\ddot{\bar{%
%TCIMACRO{\TeXButton{vartheta}{\mathbold{\vartheta}}}%
%BeginExpansion
\mathbold{\vartheta}%
%EndExpansion
}}%
%TCIMACRO{\TeXButton{black}{\color{black}}}%
%BeginExpansion
\color{black}%
%EndExpansion
_{\left( l\right) }+\bar{\mathbf{C}}_{\left( l\right) }%
%TCIMACRO{\TeXButton{red}{}}%
%BeginExpansion
%
%EndExpansion
\dot{\bar{%
%TCIMACRO{\TeXButton{vartheta}{\mathbold{\vartheta}}}%
%BeginExpansion
\mathbold{\vartheta}%
%EndExpansion
}}%
%TCIMACRO{\TeXButton{black}{\color{black}}}%
%BeginExpansion
\color{black}%
%EndExpansion
_{\left( l\right) }+\bar{\mathbf{Q}}_{\left( l\right) }%
%TCIMACRO{\TeXButton{big}{\big}}%
%BeginExpansion
\big%
%EndExpansion
)+\mathbf{W}_{\mathrm{t}}^{\mathrm{grav}}+\mathbf{P}_{\mathrm{p}}^{T}\left( 
\mathbf{M}_{\mathrm{p}}\mathbf{P}_{\mathrm{p}}\dot{\mathbf{V}}_{\mathrm{t}}+%
\mathbf{G}_{\mathrm{p}}\mathbf{M}_{\mathrm{p}}\mathbf{P}_{\mathrm{p}}\mathbf{%
V}_{\mathrm{t}}+\mathbf{W}_{\mathrm{p}}^{\mathrm{grav}}\right) -\mathbf{W}_{%
\mathrm{t}}^{\mathrm{EE}}  \label{EOMSum} \\
&=&\sum_{l=1}^{L}\bar{\mathbf{F}}_{\left( l\right) }^{T}\bar{\mathbf{H}}%
_{\left( l\right) }^{T}\varphi _{\left( l\right) }(%
%TCIMACRO{\TeXButton{red}{}}%
%BeginExpansion
%
%EndExpansion
\bar{%
%TCIMACRO{\TeXButton{vartheta}{\mathbold{\vartheta}}}%
%BeginExpansion
\mathbold{\vartheta}%
%EndExpansion
}%
%TCIMACRO{\TeXButton{black}{\color{black}}}%
%BeginExpansion
\color{black}%
%EndExpansion
_{\left( l\right) },%
%TCIMACRO{\TeXButton{red}{}}%
%BeginExpansion
%
%EndExpansion
\dot{\bar{%
%TCIMACRO{\TeXButton{vartheta}{\mathbold{\vartheta}}}%
%BeginExpansion
\mathbold{\vartheta}%
%EndExpansion
}}%
%TCIMACRO{\TeXButton{black}{\color{black}}}%
%BeginExpansion
\color{black}%
%EndExpansion
_{\left( l\right) },%
%TCIMACRO{\TeXButton{red}{}}%
%BeginExpansion
%
%EndExpansion
\ddot{\bar{%
%TCIMACRO{\TeXButton{vartheta}{\mathbold{\vartheta}}}%
%BeginExpansion
\mathbold{\vartheta}%
%EndExpansion
}}%
%TCIMACRO{\TeXButton{black}{\color{black}}}%
%BeginExpansion
\color{black}%
%EndExpansion
_{\left( l\right) })+\mathbf{P}_{\mathrm{p}}^{T}\varphi _{\mathrm{p}}\left( 
\mathbf{x},\mathbf{V}_{\mathrm{t}},\dot{\mathbf{V}}_{\mathrm{t}}\right) -%
\mathbf{W}_{\mathrm{t}}^{\mathrm{EE}}  \notag
\end{eqnarray}%
denoting $\varphi _{\mathrm{p}}\left( \mathbf{x},\mathbf{V}_{\mathrm{t}},%
\dot{\mathbf{V}}_{\mathrm{t}}\right) :=\mathbf{M}_{\mathrm{p}}\mathbf{P}_{%
\mathrm{p}}\dot{\mathbf{V}}_{\mathrm{t}}+\mathbf{G}_{\mathrm{p}}\mathbf{M}_{%
\mathrm{p}}\mathbf{P}_{\mathrm{p}}\mathbf{V}_{\mathrm{t}}+\mathbf{W}_{%
\mathrm{p}}^{\mathrm{grav}}$. The crucial observation is that the term $%
\varphi _{\left( l\right) }$, accounting for the dynamics of limb $l$, as
well as $\bar{\mathbf{F}}_{\left( l\right) }$ and $\bar{\mathbf{H}}_{\left(
l\right) }$ solely depend on the joint variables $%
%TCIMACRO{\TeXButton{red}{}}%
%BeginExpansion
%
%EndExpansion
\bar{%
%TCIMACRO{\TeXButton{vartheta}{\mathbold{\vartheta}}}%
%BeginExpansion
\mathbold{\vartheta}%
%EndExpansion
}%
%TCIMACRO{\TeXButton{black}{\color{black}}}%
%BeginExpansion
\color{black}%
%EndExpansion
_{\left( l\right) }$ and its time derivatives, and can thus be evaluated
independently. Also the remaining term $\varphi _{\mathrm{p}}$ (NE equations
of platform) depends on $\mathbf{x}$ and on $\mathbf{V}_{\mathrm{t}}$ and
its derivative only. Consequently, instead of evaluating the monolithic
system (\ref{EOMTask}), the individual terms in (\ref{EOMSum}) can be
evaluated in parallel by means of distributed computing. The same holds true
for the inverse kinematics of the limbs. 
\begin{figure}[b]
\caption{Computational scheme for parallel evaluation of inverse kinematics
and dynamics. Input: taskspace coordinates $\mathbf{x}$, velocity $\mathbf{V}%
_{\mathrm{t}}$, and acceleration $\dot{\mathbf{V}}_{\mathrm{t}}$, and
EE-loads $\mathbf{W}_{\mathrm{t}}^{\mathrm{EE}}$. Output: joint variables $%
%TCIMACRO{\TeXButton{red}{}}%
%BeginExpansion
%
%EndExpansion
%TCIMACRO{\TeXButton{vartheta}{\mathbold{\vartheta}}}%
%BeginExpansion
\mathbold{\vartheta}%
%EndExpansion
%TCIMACRO{\TeXButton{black}{\color{black}}}%
%BeginExpansion
\color{black}%
%EndExpansion
$, velocities $%
%TCIMACRO{\TeXButton{red}{}}%
%BeginExpansion
%
%EndExpansion
\dot{%
%TCIMACRO{\TeXButton{vartheta}{\mathbold{\vartheta}}}%
%BeginExpansion
\mathbold{\vartheta}%
%EndExpansion
}%
%TCIMACRO{\TeXButton{black}{\color{black}}}%
%BeginExpansion
\color{black}%
%EndExpansion
$, acceleration $%
%TCIMACRO{\TeXButton{red}{}}%
%BeginExpansion
%
%EndExpansion
\ddot{%
%TCIMACRO{\TeXButton{vartheta}{\mathbold{\vartheta}}}%
%BeginExpansion
\mathbold{\vartheta}%
%EndExpansion
}%
%TCIMACRO{\TeXButton{black}{\color{black}}}%
%BeginExpansion
\color{black}%
%EndExpansion
$, and actution forces/torques $\mathbf{u}$. }
\label{figScheme}%
\centerline{
\includegraphics[height=13cm]{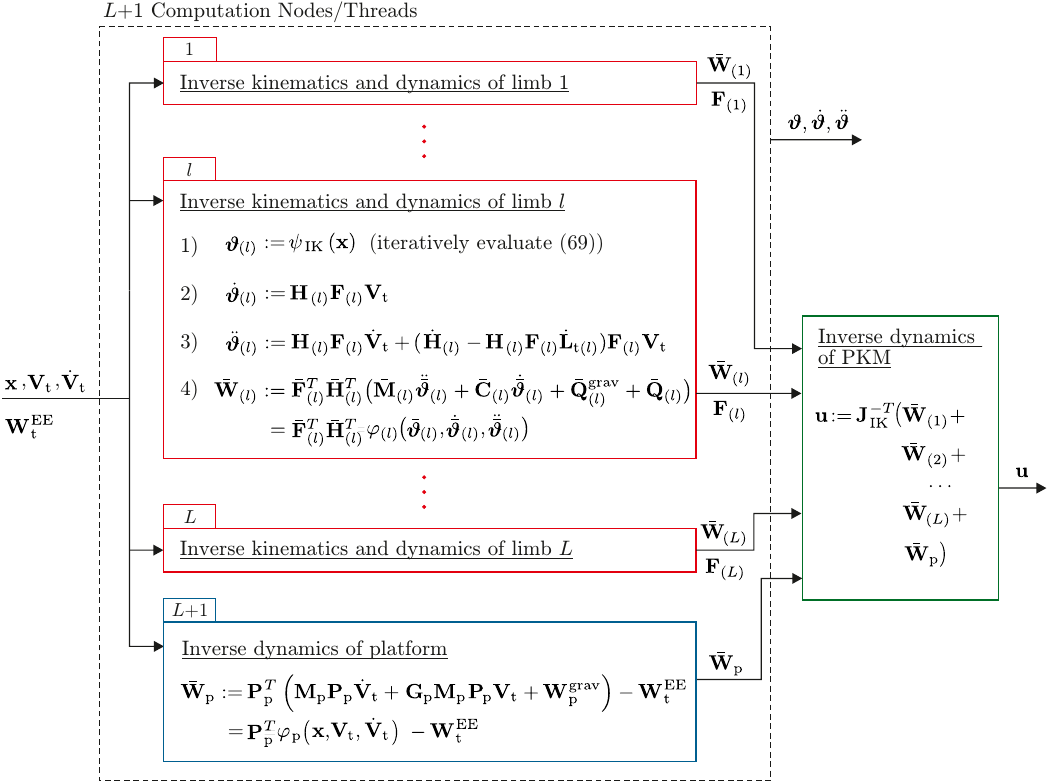}}
\end{figure}

A parallel/distributed evaluation scheme is summarized in fig. \ref%
{figScheme}. There are $L+1$ parallel computation threats/nodes. The first $%
L $ computation nodes are allocated for computing the inverse kinematics and
dynamics of the separated complex limbs. In the $l$th threat, the geometric
inverse kinematics (\ref{GeomInvLimb}), and the velocity and acceleration
inverse kinematics, (\ref{HFl}) and (\ref{eta2d}), of limb $l$ are solved in
subsequent steps 1)--3). If the inverse kinematics map $\psi _{\mathrm{IK}%
\left( l\right) }$ (\ref{GeomInvLimb}) is available in closed form, these
steps can be simplified, and $\bar{\mathbf{H}}_{\left( l\right) }\bar{%
\mathbf{F}}_{\left( l\right) }$ are replaced by the corresponding Jacobians.
The dynamics equations of the limbs are evaluated in step 4). The $L+1$-st
computation node evaluates the dynamic equations (\ref{EOMPlat}) of the
platform. Results of these $L+1$ computation runs are the auxiliary wrenches 
$\bar{\mathbf{W}}_{\left( l\right) }$ and $\bar{\mathbf{W}}_{\mathrm{p}}$,
but also the inverse kinematics Jacobians $\mathbf{F}_{\left( l\right) }$ of
the limbs, which deliver the rows of the inverse kinematics Jacobian $%
\mathbf{J}_{\mathrm{IK}}$ of the PKM. The latter is inverted in the final
computation step to compute the actuations $\mathbf{u}$. The overall inverse
kinematics output are $%
%TCIMACRO{\TeXButton{red}{}}%
%BeginExpansion
%
%EndExpansion
%TCIMACRO{\TeXButton{vartheta}{\mathbold{\vartheta}}}%
%BeginExpansion
\mathbold{\vartheta}%
%EndExpansion
%TCIMACRO{\TeXButton{red}{}}%
%BeginExpansion
%
%EndExpansion
,\dot{%
%TCIMACRO{\TeXButton{vartheta}{\mathbold{\vartheta}}}%
%BeginExpansion
\mathbold{\vartheta}%
%EndExpansion
}%
%TCIMACRO{\TeXButton{black}{\color{black}}}%
%BeginExpansion
\color{black}%
%EndExpansion
$, and $%
%TCIMACRO{\TeXButton{red}{}}%
%BeginExpansion
%
%EndExpansion
\ddot{%
%TCIMACRO{\TeXButton{vartheta}{\mathbold{\vartheta}}}%
%BeginExpansion
\mathbold{\vartheta}%
%EndExpansion
}%
%TCIMACRO{\TeXButton{black}{\color{black}}}%
%BeginExpansion
\color{black}%
%EndExpansion
$.

The equations (\ref{EOMLimb}) of the tree-topology system without platform
(in step 4 of the $L$ evaluation blocks) can be evaluated with an $O\left(
n\right) $ inverse dynamics algorithm, thus replacing evaluation of $\varphi
_{\left( l\right) }$. There are various such algorithms using classical 3D
vector formulations and DH-parameterization \cite%
{LuhWalkerPaul1980,Featherstone2008}, but also such using Lie group
formulations \cite%
{ParkBobrowPloen1995,ModernRobotics,MUBOScrew2,ICRA2017,RAL2020}.

The critical aspect deciding about the efficacy of a parallel implementation
is the data exchange between the $N_{l}+1$ compute notes. Moreover, a
massive parallel computing hardware with minimal communication overhead and
latency is crucial.%
%TCIMACRO{\TeXButton{newpage}{\newpage}}%
%BeginExpansion
\newpage%
%EndExpansion

\section{Example: Inverse Dynamics of a 3\protect\underline{R}R[2RR]R Delta%
%TCIMACRO{\TeXButton{AppendixDelta}{\label{AppendixDelta}}}%
%BeginExpansion
\label{AppendixDelta}%
%EndExpansion
}

In this section, the complete kinematic and dynamic model of a 3\underline{R}%
R[2RR]R Delta robot is derived. To this end, the Lie group formulation
summarized in appendix \ref{AppendixEOMTree} is used. This formulation and
the example model have been implemented in Mathematica, \color{red} which
available as e-Component to this paper and at \cite{MendeleyData}. %
\color{black}For numerical evaluation, the geometric and dynamic parameters
are used that shall roughly resemble a MOTOMAN-MPP3H Delta robot.

\subsection{Kinematics of representative limb%
%TCIMACRO{\TeXButton{secDeltaKin}{\label{secDeltaKin}}}%
%BeginExpansion
\label{secDeltaKin}%
%EndExpansion
}

The model of the 3\underline{R}R[2RR]R Delta robot possesses $N=3\times 6$
revolute joints. The components of the platform position vector serve as
task space coordinates, $\mathbf{x}:=\mathbf{r}_{\mathrm{p}}$. The joint
angles of the three actuated revolute joints at the base serve as actuator
coordinates, $%
%TCIMACRO{\TeXButton{vartheta}{\mathbold{\vartheta}}}%
%BeginExpansion
\mathbold{\vartheta}%
%EndExpansion
_{\mathrm{act}}:=\left( \vartheta _{1\left( 1\right) },\vartheta _{1\left(
2\right) },\vartheta _{1\left( 3\right) }\right) ^{T}$. For the Delta robot,
the solutions to the inverse and forward kinematics problem can be expressed
in closed form. This is not used here, but an iterative solution is pursued
in order to capture the general situation. Bodies and joints are numbered as
in Fig. \ref{figRR2RRRDelta}b), the PKM topology is represented by the graph
in Fig. \ref{figDeltaLimbGraph}a), and the tree-topology system is defined
as in Fig. \ref{figDeltaLimbGraph}b).

\paragraph{Joint screws and reference configurations}

\begin{figure}[h]
\centerline{
\includegraphics[height=7cm]{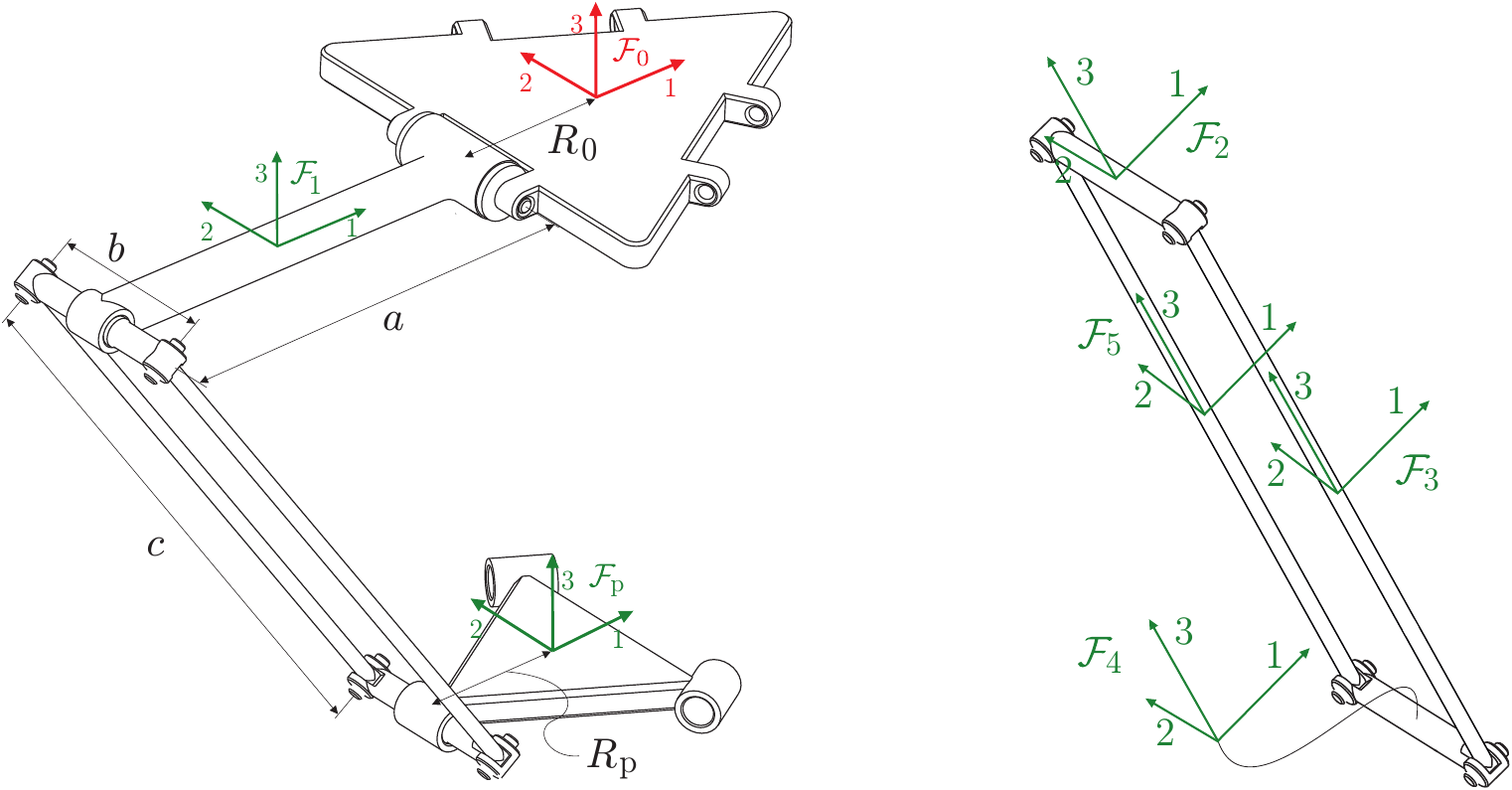}}
\caption{Geometric parameters, and location of construction frames and
body-frames of the representative limb of the 3\protect\underline{R}R[2RR]R
Delta robot. The construction frames $\mathcal{F}_{0}^{\prime }$ and $%
\mathcal{F}_{\mathrm{p}}^{\prime }$ are located at center of the base and
platform, respectively. In the reference configuration, the upper arm is
aligned horizontally. }
\label{figRefLimbGeom}
\end{figure}

Fig. \ref{figRefLimbGeom} shows the RL in the reference configuration, where
the upper arms of all limbs are aligned horizontally. The construction
frames $\mathcal{F}_{0}^{\prime }$ and $\mathcal{F}_{\mathrm{p}}^{\prime }$
are located at the center of the base and platform, respectively. The length
of the upper arm (body 1) is denoted with $a$, the length of each of the
rods (bodies 3 and 5) with $c$, and $b$ denotes the length of the links
(bodies 2 and 4) connecting the two rods. The R-joints at the base and
platform are located at a distance of $R_{0}$ and $R_{\mathrm{p}}$ from the
respective center, mutually aligned with 120$%
%TCIMACRO{\U{b0}}%
%BeginExpansion
{{}^\circ}%
%EndExpansion
$. 

In order to simplify the expressions, the abbreviations $d:=a+R_{0}-R_{%
\mathrm{p}}$ and $h:=\sqrt{c^{2}-d^{2}}$ (the height of the platform in
reference configuration) are introduced. The position vectors $\mathbf{y}%
_{i}^{\prime }$ to the joint axes and the unit vectors $\mathbf{e}%
_{i}^{\prime }$ along the joint axes expressed in $\mathcal{F}_{0}^{\prime }$
are%
\begin{eqnarray*}
\mathbf{y}_{1}^{\prime } &=&\left( -R_{0},0,0\right) ^{T},\mathbf{y}%
_{2}^{\prime }=\left( -R_{0}-a,0,0\right) ^{T},\mathbf{y}_{3}^{\prime
}=\left( -R_{0}-a,-b/2,0\right) ^{T}, \\
\mathbf{y}_{4}^{\prime } &=&\left( -R_{\mathrm{p}},-b/2,-h\right) ^{T},%
\mathbf{y}_{5}^{\prime }=\left( -R_{0}-a,b/2,0\right) ^{T},\mathbf{y}%
_{6}^{\prime }=\left( -R_{\mathrm{p}},0,-h\right) ^{T} \\
\mathbf{e}_{i}^{\prime } &=&\left( 0,-1,0\right) ^{T},i=1,2,6;\ \ \ \mathbf{e%
}_{i}^{\prime }=\left( h/c,0,d/c\right) ^{T},i=3,4
\end{eqnarray*}%
which, according to (\ref{XY}), give rise the the screw coordinate vectors
represented in $\mathcal{F}_{0}^{\prime }$ 
\begin{equation*}
{\mathbf{Y}}_{1}^{\prime }=\left( 
\begin{array}{c}
0 \\ 
-1 \\ 
0 \\ 
0 \\ 
0 \\ 
R_{0}%
\end{array}%
\right) ,{\mathbf{Y}}_{2}^{\prime }=\left( 
\begin{array}{c}
0 \\ 
-1 \\ 
0 \\ 
0 \\ 
0 \\ 
a+R_{0}%
\end{array}%
\right) ,\mathbf{Y}_{3}^{\prime }=\frac{1}{c}\left( 
\begin{array}{c}
h \\ 
0 \\ 
d \\ 
-bd/2 \\ 
d\left( a+R_{0}\right) \\ 
bh/2%
\end{array}%
\right) ,\mathbf{Y}_{4}^{\prime }=\frac{1}{c}\left( 
\begin{array}{c}
h \\ 
0 \\ 
d \\ 
-bd/2 \\ 
h^{2}+R_{\mathrm{p}}d \\ 
bh/2%
\end{array}%
\right) ,\mathbf{Y}_{5}^{\prime }=\frac{1}{c}\left( 
\begin{array}{c}
h \\ 
0 \\ 
d \\ 
bd/2 \\ 
d\left( a+R_{0}\right) \\ 
-bh/2%
\end{array}%
\right) ,\mathbf{Y}_{6}^{\prime }=\left( 
\begin{array}{c}
0 \\ 
-1 \\ 
0 \\ 
-h \\ 
0 \\ 
R_{\mathrm{p}}%
\end{array}%
\right) .
\end{equation*}%
The body-fixed frames are shown in fig. \ref{figRefLimbGeom}. The rotation
matrices $\mathbf{R}_{i}^{\prime }$ and position vectors $\mathbf{p}%
_{i}^{\prime }$, which determine the reference configurations $\mathbf{A}%
_{i}^{\prime }$ of the bodies, are%
\begin{equation*}
\mathbf{R}_{1}^{\prime }=\mathbf{R}_{6}^{\prime }=\mathbf{I},\ \ \mathbf{R}%
_{i}^{\prime }=\frac{1}{c}\left( 
\begin{array}{ccc}
h\ \  & 0 & -d \\ 
0\ \  & 1 & 0 \\ 
d\ \  & 0 & h%
\end{array}%
\right) ,\ i=2,\ldots ,5
\end{equation*}%
\begin{equation*}
\mathbf{p}_{1}^{\prime }=\left( 
\begin{array}{c}
-a/2-R_{0} \\ 
0 \\ 
0%
\end{array}%
\right) ,\mathbf{p}_{2}^{\prime }=\left( 
\begin{array}{c}
-a-R_{0} \\ 
0 \\ 
0%
\end{array}%
\right) ,\mathbf{p}_{3}^{\prime }=\left( 
\begin{array}{c}
-d/2-R_{\mathrm{p}} \\ 
-b/2 \\ 
-h/2%
\end{array}%
\right) ,\mathbf{p}_{4}^{\prime }=\left( 
\begin{array}{c}
-R_{\mathrm{p}} \\ 
0 \\ 
-h%
\end{array}%
\right) ,\mathbf{p}_{5}^{\prime }=\left( 
\begin{array}{c}
-d/2-R_{\mathrm{p}} \\ 
b/2 \\ 
-h/2%
\end{array}%
\right) ,\mathbf{p}_{6}^{\prime }=\left( 
\begin{array}{c}
0 \\ 
0 \\ 
-h%
\end{array}%
\right) .
\end{equation*}%
The joint screw coordinates in body-fixed representation are then found with
(\ref{XtoY}) as%
\begin{equation*}
{^{1}\mathbf{X}}_{1}^{\prime }=\left( 
\begin{array}{c}
0 \\ 
-1 \\ 
0 \\ 
0 \\ 
0 \\ 
-a/2%
\end{array}%
\right) ,\ {^{2}\mathbf{X}}_{2}^{\prime }=\left( 
\begin{array}{c}
0 \\ 
-1 \\ 
0 \\ 
0 \\ 
0 \\ 
0%
\end{array}%
\right) ,\ {^{3}\mathbf{X}}_{3}^{\prime }={^{5}\mathbf{X}}_{5}^{\prime
}=\left( 
\begin{array}{c}
(d^{2}+h^{2})/c^{2} \\ 
0 \\ 
0 \\ 
0 \\ 
2ad+h^{2}-d(d-2R_{0}+2R_{\mathrm{p}}) \\ 
0%
\end{array}%
\right) ,\ {^{4}\mathbf{X}}_{4}^{\prime }=\left( 
\begin{array}{c}
(d^{2}+h^{2})/c^{2} \\ 
0 \\ 
0 \\ 
0 \\ 
0 \\ 
b(d^{2}+h^{2})/(2c^{2})%
\end{array}%
\right) ,\ {^{6}\mathbf{X}}_{6}^{\prime }=\left( 
\begin{array}{c}
0 \\ 
-1 \\ 
0 \\ 
0 \\ 
0 \\ 
R_{\mathrm{p}}%
\end{array}%
\right)
\end{equation*}%
A separated limb without platform comprises $\bar{n}_{l}=5$ bodies. The
kinematics model is expressed in terms of the matrices $\mathsf{A}^{\prime }$
and $\mathsf{X}^{\prime }$ in (\ref{Ab}). According to the ordering and
predecessor relation induced by the graph in fig. \ref{figTreeNoPlatform}a),
these are%
\begin{equation}
\mathsf{A}^{\prime }=\left( 
\begin{array}{ccccc}
\mathbf{I} & \mathbf{0} & \mathbf{0} & \mathbf{0\ \ \ \ } & \mathbf{0} \\ 
\mathbf{Ad}_{\mathbf{C}_{2,1}^{\prime }} & \mathbf{I} & \mathbf{0} & \mathbf{%
0\ \ \ \ } & \mathbf{0} \\ 
\mathbf{Ad}_{\mathbf{C}_{3,1}^{\prime }} & \mathbf{Ad}_{\mathbf{C}%
_{3,2}^{\prime }} & \mathbf{I} & \mathbf{0\ \ \ \ } & \mathbf{0} \\ 
\mathbf{Ad}_{\mathbf{C}_{4,1}^{\prime }} & \mathbf{Ad}_{\mathbf{C}%
_{4,2}^{\prime }} & \mathbf{Ad}_{\mathbf{C}_{4,3}^{\prime }} & \mathbf{I\ \
\ \ } & \mathbf{0} \\ 
\mathbf{Ad}_{\mathbf{C}_{5,1}^{\prime }} & \mathbf{Ad}_{\mathbf{C}%
_{5,2}^{\prime }} & \mathbf{0} & \mathbf{0\ \ \ \ } & \mathbf{I}%
\end{array}%
\right) ,\ \ \mathsf{X}^{\prime }=\mathrm{diag~}\left( {^{1}\mathbf{X}}%
_{1}^{\prime },\ldots ,{^{5}}\mathbf{X}_{5}^{\prime }\right) .
\label{ADelta}
\end{equation}

\paragraph{Particular geometry parameter}

The geometric parameters are chosen so to roughly resemble a MOTOMAN-MPP3H
Delta robot. All moving parts are approximated by geometric primitives for
which the inertia parameters are determined. The principal dimensions are
set to%
\begin{equation*}
R_{0}=150\,\mathrm{mm},R_{\mathrm{p}}=70\,\mathrm{mm},a=250\,\mathrm{mm}%
,b=80\,\mathrm{mm},c=1000\,\mathrm{mm}\ .
\end{equation*}%
Then the body-fixed joint coordinate vectors are%
\begin{equation*}
{^{1}\mathbf{X}}_{1}^{\prime }=\left( 
\begin{array}{c}
0 \\ 
-1 \\ 
0 \\ 
0 \\ 
0 \\ 
-1/8%
\end{array}%
\right) ,\ {^{2}\mathbf{X}}_{2}^{\prime }=\left( 
\begin{array}{c}
0 \\ 
-1 \\ 
0 \\ 
0 \\ 
0 \\ 
0%
\end{array}%
\right) ,\ {^{3}\mathbf{X}}_{3}^{\prime }={^{5}\mathbf{X}}_{5}^{\prime
}=\left( 
\begin{array}{c}
1 \\ 
0 \\ 
0 \\ 
0 \\ 
1/2 \\ 
0%
\end{array}%
\right) ,\ {^{4}\mathbf{X}}_{4}^{\prime }=\left( 
\begin{array}{c}
1 \\ 
0 \\ 
0 \\ 
0 \\ 
0 \\ 
1/25%
\end{array}%
\right) ,\ {^{6}\mathbf{X}}_{6}^{\prime }=\left( 
\begin{array}{c}
0 \\ 
-1 \\ 
0 \\ 
0 \\ 
0 \\ 
7/100%
\end{array}%
\right)
\end{equation*}%
which are particularly simple since the joint axes are aligned with the
body-fixed reference frames and could be readily deduced from the model,
which is one advantage of the geometric Lie group formulation.

The platform configuration $\mathbf{C}_{\mathrm{p}}^{\prime }$ is expressed
in terms of the joint angles $\vartheta _{1}^{\prime },\vartheta
_{2}^{\prime },\vartheta _{3}^{\prime },\vartheta _{4}^{\prime },\vartheta
_{6}^{\prime }$ via the POE (\ref{Cp}). In the following, the closed form
solution $\vartheta _{3}^{\prime }=\vartheta _{5}^{\prime }=-\vartheta
_{4}^{\prime }$ of the loop constraints will be used (see example \ref%
{exDeltaLoopSol}), so that the vector of generalized coordinates is $\mathbf{%
q}_{\left( l\right) }=\left( \vartheta _{4}^{\prime },\vartheta _{1}^{\prime
},\vartheta _{2}^{\prime },\vartheta _{6}^{\prime }\right) ^{T}$ (see
example \ref{exDeltaq}). The platform pose is then%
\begin{equation*}
\mathbf{C}_{\mathrm{p}}^{\prime }(\mathbf{q}_{\left( l\right) }^{\prime
})=\left( 
\begin{array}{cccc}
{c}_{{1+2+6}} & 0 & -{s}_{{1+2+6}} & \frac{1}{200}\left( 33\left( {c_{{1+2+4}%
}}+{c_{{1+2-4}}}\right) +14{c}_{{1+2+6}}-50c_{1}+2\xi {s}_{{1+2}%
}c_{4}-30\right) \\ 
0 & 1 & 0 & -s_{4} \\ 
{s}_{{1+2+6}} & 0 & {c}_{{1+2+6}} & \frac{1}{200}\left( 14{s}_{{1+2+6}}+66{s}%
_{{1+2}}c_{4}-50s_{1}-\xi \left( {c_{{1+2+4}}}+{c_{{1+2-4}}}\right) \right)
\\ 
0 & 0 & 0 & 1%
\end{array}%
\right)
\end{equation*}%
with $s_{i}:=\sin \vartheta _{i}^{\prime },c_{i}:=\cos \vartheta
_{i}^{\prime },s_{i\pm j\pm k}:=\sin (\vartheta _{i}^{\prime }\pm \vartheta
_{j}^{\prime }\pm \vartheta _{k}^{\prime }),c_{i\pm j\pm k}:=\cos (\vartheta
_{i}^{\prime }\pm \vartheta _{j}^{\prime }\pm \vartheta _{k}^{\prime })$,
and $\xi :=\sqrt{8911}$.

The system Jacobian (\ref{JbSys2}) is readily found with the above joint
screw coordinates and the matrix (\ref{ADelta}) according to (\ref{JbSys}).
The last block row in (\ref{JbSys}) is the geometric Jacobian of the
platform $\mathbf{J}_{\mathrm{p}}^{\prime }:=\mathbf{J}_{n_{l}=6}^{\prime }$%
. Along with $\mathbf{H}_{\left( l\right) }^{\prime }$ in (\ref{HlDelta}),
this yields the compound forward kinematics Jacobian $\mathbf{L}_{\mathrm{p}%
\left( l\right) }^{\prime }$ in (\ref{Lpl}). Rows 2,4,5,6 are used to
construct $\mathbf{L}_{\mathrm{t}\left( l\right) }^{\prime }$ in (\ref{Vt}).
The Jacobian in the velocity inverse kinematics solution of the mechanism (%
\ref{Ftheta1}) is finally obtained with (\ref{Ftheta2}), using the selection
matrix in (\ref{DtDelta}), as%
\begin{equation}
\mathbf{F}_{\left( l\right) }^{\prime }(%
%TCIMACRO{\TeXButton{vartheta}{\mathbold{\vartheta}}}%
%BeginExpansion
\mathbold{\vartheta}%
%EndExpansion
_{\left( l\right) })=\frac{1}{w_{2}}\left( 
\begin{array}{ccc}
0 & -{w}_{2}\sec {\vartheta }_{{4}}^{\prime } & 0 \\ 
4{w}_{{6}} & -400\tan {\vartheta }_{{4}}^{\prime } & -4{u}_{{6}} \\ 
-4({w}_{{6}}-25\sec {\vartheta }_{{4}}^{\prime }\cos ({\vartheta }_{{2}%
}^{\prime }+{\vartheta }_{{6}}^{\prime })) & \tan {\vartheta }_{{4}}^{\prime
}(400-{u}_{{2}}\sec {\vartheta }_{{4}}^{\prime }) & 4({u}_{{6}}-25\sec {%
\vartheta }_{{4}}^{\prime }\sin ({\vartheta }_{{2}}^{\prime }+{\vartheta }_{{%
6}}^{\prime })) \\ 
-100\sec {\vartheta }_{4}^{\prime }\cos ({\vartheta }_{{2}}^{\prime }+{%
\vartheta }_{{6}}^{\prime }) & {u}_{{2}}\tan {\vartheta }_{{4}}^{\prime
}\sec {\vartheta }_{{4}}^{\prime } & 100\sec {\vartheta }_{{4}}^{\prime
}\sin ({\vartheta }_{{2}}^{\prime }+{\vartheta }_{{6}}^{\prime })%
\end{array}%
\right)  \label{FlMotoman}
\end{equation}%
where%
\begin{align*}
u_{2}& :=\xi \sin {\vartheta }_{{2}}^{\prime }+33\cos {\vartheta }_{{2}%
}^{\prime },\ \ {w}_{{2}}:=\xi \cos {\vartheta }_{{2}}^{\prime }-33\sin {%
\vartheta }_{{2}}^{\prime } \\
u_{6}& :=33\sin {\vartheta }_{{6}}^{\prime }+\xi \cos {\vartheta }_{{6}%
}^{\prime },\ \ w_{6}:=33\cos {\vartheta }_{{6}}^{\prime }-\xi \sin {%
\vartheta }_{{6}}^{\prime }
\end{align*}%
The compound inverse kinematics Jacobian in (\ref{HFl}) is obtained with $%
\mathbf{H}_{\left( l\right) }^{\prime }$ in (\ref{HlDelta}).

\subsection{Kinematics of the $L=3$ limbs}

The construction and mount frames of the limbs are all located the center of
the base and platform, respectively. They are only rotated by $\pm 2/3\pi $.
The respective transformations are%
\begin{equation}
\mathbf{S}_{0\left( 1\right) }=\mathbf{S}_{\mathrm{p}\left( 1\right) }=%
\mathbf{I,S}_{0\left( 2\right) }=\mathbf{S}_{\mathrm{p}\left( 2\right)
}=\left( 
\begin{array}{cccc}
-1/2 & -\sqrt{3}/2 & 0 & 0 \\ 
\sqrt{3}/2 & -1/2 & 0 & 0 \\ 
0 & 0 & 1 & 0 \\ 
0 & 0 & 0 & 1%
\end{array}%
\right) ,\mathbf{S}_{0\left( 3\right) }=\mathbf{S}_{\mathrm{p}\left(
3\right) }=\left( 
\begin{array}{cccc}
-1/2 & -\sqrt{3}/2 & 0 & 0 \\ 
\sqrt{3}/2 & -1/2 & 0 & 0 \\ 
0 & 0 & 1 & 0 \\ 
0 & 0 & 0 & 1%
\end{array}%
\right) .
\end{equation}%
The configuration of body $i$ is thus determined by (\ref{CiS}), and the
forward and inverse Jacobian by (\ref{Jpl3}) and (\ref{LpS}), respectively.
Therewith, according to (\ref{IK}), the inverse kinematics Jacobian $\mathbf{%
J}_{\mathrm{IK}}$ is constructed from the second row of $\mathbf{F}_{\left(
l\right) }^{\prime },l=1,2,3$ in (\ref{FlMotoman}).

\subsection{Dynamic parameters}

The upper arm is represented by a solid cylinder with length $a=250$%
\thinspace mm and 30\thinspace mm diameter. The bars forming the
parallelogram are modeled as cylindrical rods with length $c=1$\thinspace m
and diameter of 10\thinspace mm. The link connecting the rods is a cylinder
with length $b=80$\thinspace mm and diameter 20\thinspace mm. The T-axis
flange (which allows mounting a 4th rotary axis at the platform) is regarded
as the dominant element contributing to the inertia of the moving platform.
It is modeled as a solid cylinder with 90\thinspace mm diameter and
100\thinspace mm height. As the platform cannot rotate, with $\mathbf{P}_{%
\mathrm{p}}:=\mathbf{P}_{\mathrm{p}}^{\mathrm{trans}}$ in (\ref{Pp}), the
contribution of the platform inertia to the mass matrix in (\ref{Mbar})
reduced to $\mathbf{P}_{\mathrm{p}}^{T}\mathbf{M}_{\mathrm{p}}\mathbf{P}_{%
\mathrm{p}}=m_{\mathrm{p}}\mathbf{I}$, and the Coriolis/centrifugal term in (%
\ref{Cbar}) vanishes since $\mathbf{P}_{\mathrm{p}}^{T}\mathbf{G}_{\mathrm{p}%
}\mathbf{M}_{\mathrm{p}}\mathbf{P}_{\mathrm{p}}=\mathbf{0}$. This is clear,
as the NE-equations (\ref{EOMPlat}) reduce to the balance of linear
momentum. Thus only the mass of the platform must be determined.

All links are assumed to be made of aluminum. The inertia parameter are
indeed not those of a real Delta robot, but serve the purpose of a numerical
example.

\subsection{Implementation and Results}

\paragraph{Code Generation and Numerical Results}

The dynamic EOM (\ref{EOMTask}) were generated symbolically in closed form
using a Mathematica implementation of the Lie group formulation. The
kinematic relations (\ref{HFl}) and (\ref{eta2d}) were incorporated to
obtain (\ref{EOMTask}) solely in terms of $%
%TCIMACRO{\TeXButton{red}{}}%
%BeginExpansion
%
%EndExpansion
%TCIMACRO{\TeXButton{vartheta}{\mathbold{\vartheta}}}%
%BeginExpansion
\mathbold{\vartheta}%
%EndExpansion
%TCIMACRO{\TeXButton{red}{}}%
%BeginExpansion
%
%EndExpansion
,\mathbf{V}_{\mathrm{t}},\dot{\mathbf{V}}_{\mathrm{t}}$. Further, the
transposed inverse $\mathbf{J}_{\mathrm{IK}}^{-T}$ of the inverse kinematics
Jacobian was generated symbolically. All equations were exported in C
language, and implemented as C-mex function in Matlab compiled with the
standard MinGW64 compiler.

\begin{figure}[h]
%\vspace{-3ex} 
\centerline{
\includegraphics[height=8cm]{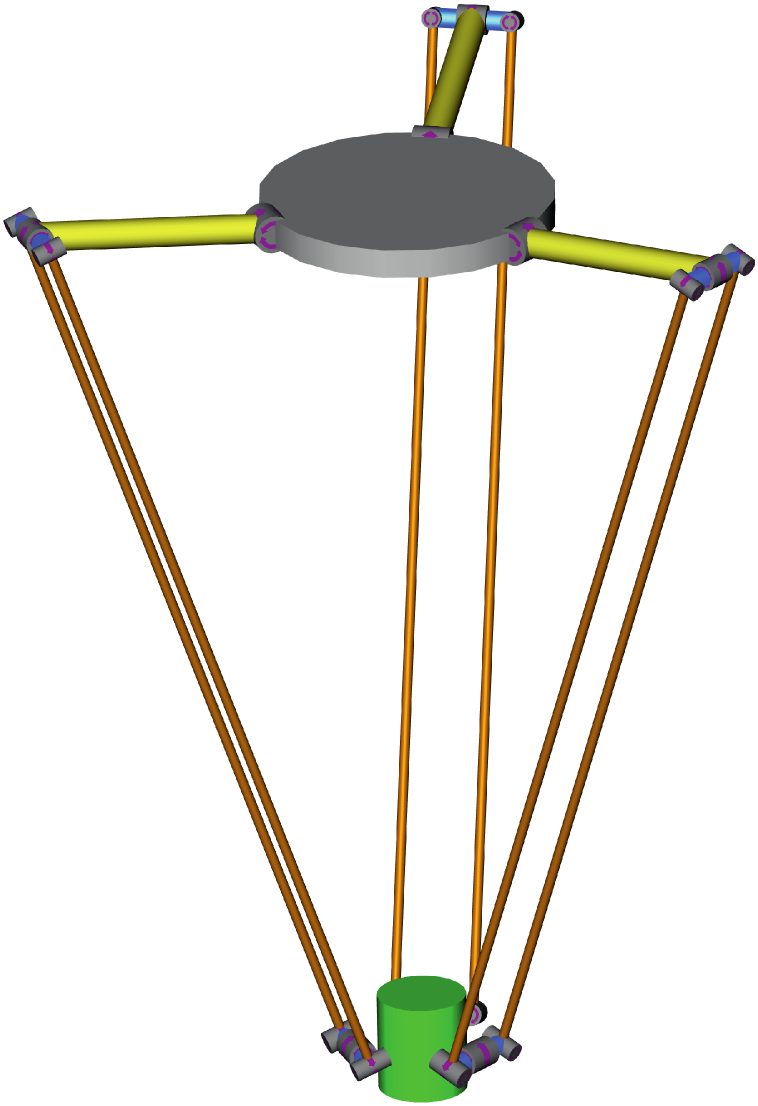}} %\vspace{-4ex}
\caption{Alaska model resembling the MOTOMAN-MPP3H\ Delta robot.}
\label{figAlaska}
\end{figure}

\begin{figure}[h]
\centerline{
\includegraphics[height=6.5cm]{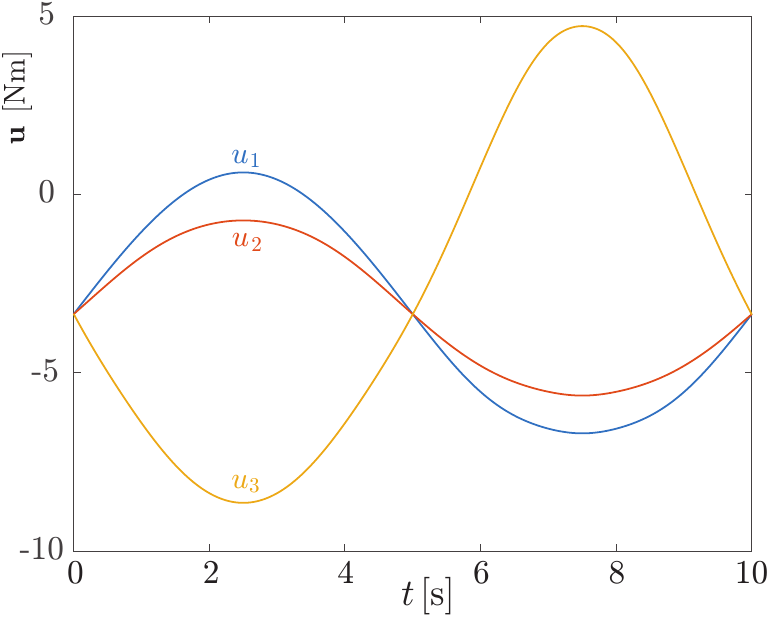}~~~~\includegraphics[height=6.5cm]{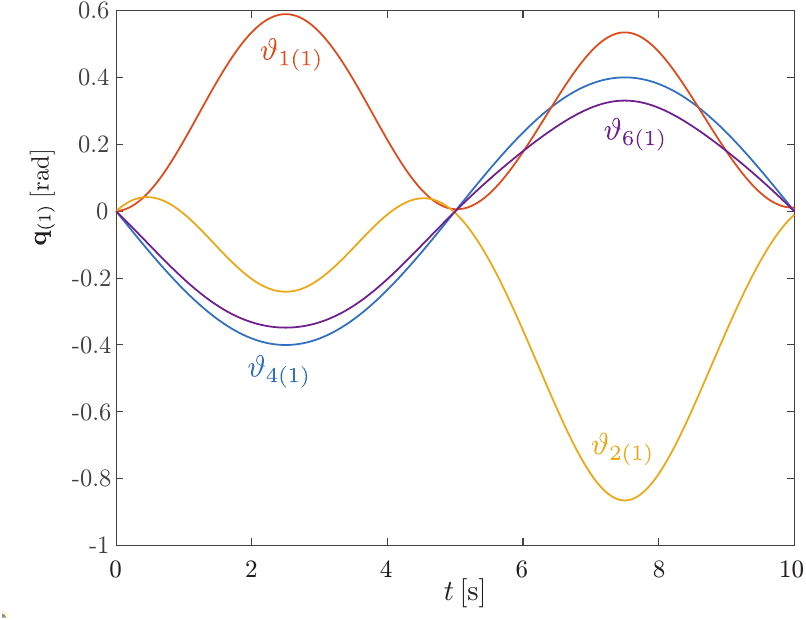}}
\centerline{
\includegraphics[height=6.5cm]{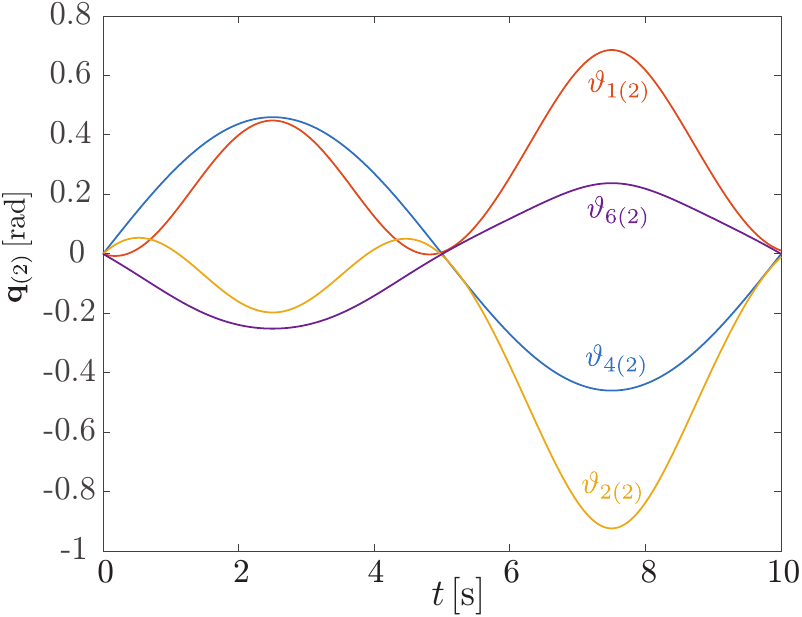}~~~~\includegraphics[height=6.5cm]{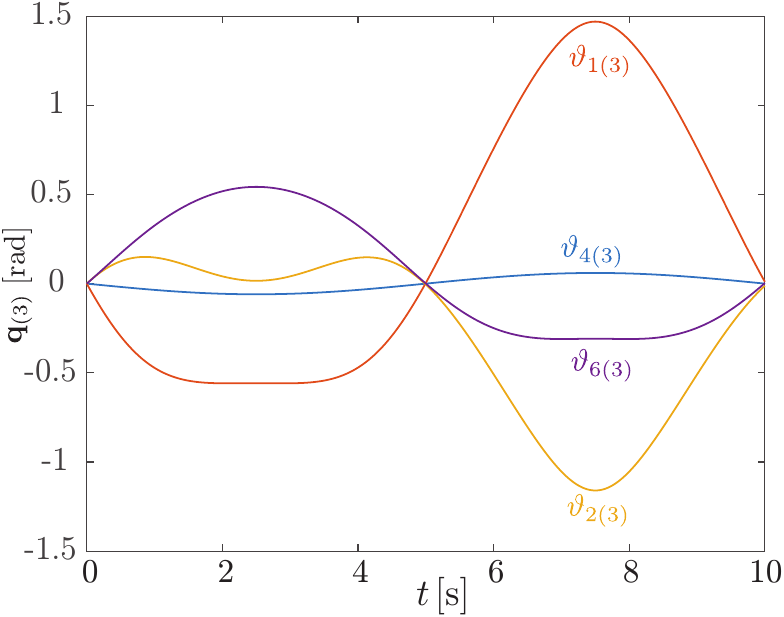}}
\caption{Time evolution of actuation torques $\mathbf{u}=\left(
u_{1},u_{2},u_{3}\right) ^{T}$ and generalized coordinates $\mathbf{q}%
_{\left( l\right) }=\left( \protect\vartheta _{4\left( l\right) },\protect%
\vartheta _{1\left( l\right) },\protect\vartheta _{2\left( l\right) },%
\protect\vartheta _{6\left( l\right) }\right) ^{T},l=1,2,3$ for the
straight-line platform motion of the Delta model.}
\label{figInvdyn}
\end{figure}

For the mere purpose of checking the correctness of the model and code
implementation, the inverse dynamics solution is computed for the
straight-line motion of the platform prescribed by $\mathbf{r}\left(
t\right) =\left( 0,0,-h\right) ^{T}+\left( 0.3,0.4,0.1\right) ^{T}\sin
\left( 2\pi t/T\right) ,t=0,\ldots ,T=10\,$s, with time step size of $\Delta
t=0.01\,$s. The time evolution of actuator joint coordinates $%
%TCIMACRO{\TeXButton{vartheta}{\mathbold{\vartheta}}}%
%BeginExpansion
\mathbold{\vartheta}%
%EndExpansion
_{\mathrm{act}}$ and the actuation torques $\mathbf{u}$ are shown in fig. %
\ref{figInvdyn}. The trajectories of all $N=18$ joints and the actuation
torques were validated against the commercial MBS dynamics simulation
software Alaska. The numerical results match up to the prescribed accuracy.
Figure \ref{figAlaska} shows the 3D-view of the model with the geometric
primitives. The results are identical up to the computational accuracy. 
\clearpage

\paragraph{Performance Assessment}

The overall performance of an inverse dynamics evaluation is dictated by the
time required for solving the geometric inverse kinematics problems for the $%
L$ limbs and for the subsequent evaluation of the inverse dynamics model.
This was determined for the closed form inverse dynamics model (\ref{InvDyn}%
) and the parallel implementation. The following numerical experiments were
conducted on a standard PC with an Intel i7-8700 CPU with 6 cores clocked at
3.2 GHz running MS-Windows 10 operating system. The following results, which
are listed in table \ref{tabTiming}, were obtained by averaging over $10^{6}$
evaluation runs. For parallel computation the Matlab Parallel Computing
Toolbox is used.

\begin{enumerate}
\item Computation times for computing the inverse dynamics solution, i.e.
for evaluating the symbolic expression of the monolithic form of (\ref%
{InvDyn}), were determined. The geometric inverse kinematics is solved by
executing two iterations of (\ref{GeomInvLimb}) for all limbs. The average
time needed for solving the IK and computing the inverse dynamics solution
was 2.74$\,\mu s$ as listed in the first row of table \ref{tabTiming}.

\item Evaluating the inverse dynamics solution (\ref{InvDyn}) involves
evaluation of the EOM (\ref{EOMTask}) and the subsequent multiplication with 
$\mathbf{J}_{\mathrm{IK}}^{-T}$, which is known symbolically. It is
instructive to quantify the time spent for evaluating (\ref{EOMTask}) alone.
The left-hand side of (\ref{EOMTask}) was evaluated, which delivers the term 
$\varphi _{\mathrm{t}}-\mathbf{W}_{\mathrm{t}}^{\mathrm{EE}}$ in (\ref%
{InvDyn2}). The average time needed for (\ref{EOMTask}) along with the IK
solution is shown in table \ref{tabTiming} as 2.75\thinspace $\mu s$. This
confirms that the time for evaluating $\mathbf{J}_{\mathrm{IK}}^{-T}$ is
negligible. Moreover, the closed form (\ref{InvDyn}), with $\varphi _{%
\mathrm{t}}-\mathbf{W}_{\mathrm{t}}^{\mathrm{EE}}$ being premultiplied with $%
\mathbf{J}_{\mathrm{IK}}^{-T}$, allows the compiler to optimize the code.

\item To investigate the potential of a distributed computation of the EOM,
the computation times spent for separate evaluation of the EOM of the limbs
is determined. The equations for an individual limb $l$ are evaluated at one
computation node: The geometric inverse kinematics problems $%
%TCIMACRO{\TeXButton{red}{}}%
%BeginExpansion
%
%EndExpansion
%TCIMACRO{\TeXButton{vartheta}{\mathbold{\vartheta}}}%
%BeginExpansion
\mathbold{\vartheta}%
%EndExpansion
%TCIMACRO{\TeXButton{black}{\color{black}}}%
%BeginExpansion
\color{black}%
%EndExpansion
_{\left( l\right) }=\psi _{\mathrm{IK}\left( l\right) }\left( \mathbf{x}%
\right) $ is solved performing two iterations of (\ref{qetaIteration}), and
the contribution of limb $l$ to the sum in (\ref{EOMSum}), i.e. the
expression%
\begin{equation}
\bar{\mathbf{W}}_{\left( l\right) }=\bar{\mathbf{F}}_{\left( l\right) }^{T}%
\bar{\mathbf{H}}_{\left( l\right) }^{T}\varphi _{\left( l\right) }(%
%TCIMACRO{\TeXButton{red}{}}%
%BeginExpansion
%
%EndExpansion
\bar{%
%TCIMACRO{\TeXButton{vartheta}{\mathbold{\vartheta}}}%
%BeginExpansion
\mathbold{\vartheta}%
%EndExpansion
}%
%TCIMACRO{\TeXButton{black}{\color{black}}}%
%BeginExpansion
\color{black}%
%EndExpansion
_{\left( l\right) },%
%TCIMACRO{\TeXButton{red}{}}%
%BeginExpansion
%
%EndExpansion
\dot{\bar{%
%TCIMACRO{\TeXButton{vartheta}{\mathbold{\vartheta}}}%
%BeginExpansion
\mathbold{\vartheta}%
%EndExpansion
}}%
%TCIMACRO{\TeXButton{black}{\color{black}}}%
%BeginExpansion
\color{black}%
%EndExpansion
_{\left( l\right) },%
%TCIMACRO{\TeXButton{red}{}}%
%BeginExpansion
%
%EndExpansion
\ddot{\bar{%
%TCIMACRO{\TeXButton{vartheta}{\mathbold{\vartheta}}}%
%BeginExpansion
\mathbold{\vartheta}%
%EndExpansion
}}%
%TCIMACRO{\TeXButton{black}{\color{black}}}%
%BeginExpansion
\color{black}%
%EndExpansion
_{\left( l\right) }),  \label{Fphil}
\end{equation}%
is evaluated on one computation node, while $L=3$ computation nodes are
running in parallel. This corresponds to the dashed box in fig. \ref%
{figScheme}, except that the platform EOM are not investigated as they are
computationally trivial. Table \ref{tabTiming} shows the average necessary
computation time. Notice that the reported times only account for the
evaluations of the EOM in order to compute the auxiliary wrenches $\bar{%
\mathbf{W}}_{\left( l\right) }$ and Jacobians $\mathbf{F}_{\left( l\right) }$%
, but they do not included the time needed for exchanging these quantities.
Thus, these results are indicative only as the communication overhead to
send them to the block on the right-hand side in fig. \ref{figScheme}
strongly depends on the parallel computing framework. For instance, the
communication overhead of the used Matlab Parallel Computing Toolbox (which
is not intended for massive parallel computation) by far exceeds the
computation time for the models. Therefore no timing is reported here. A
parallel implementation will have to ensure minimal communication overhead,
which is not the topic of this paper.

\item[3.$l$] It is also interesting to brake down the overall time to that
needed for parallel evaluation of the individual limbs. The time spent for
evaluating the inverse kinematics and EOM for limb $l$ (running on node $l$)
is reported as experiment $3.l$ in table \ref{tabTiming}. The different
mathematical expressions for the different limbs leads to slightly different
times. Clearly, the overall time is dictated by the maximal time spend for a
limb.
\end{enumerate}

It should be mentioned that the parallel implementation may further benefit
from employing a recursive $O\left( n\right) $ formulation for the inverse
dynamic evaluation of the EOM of the limbs. A Lie group formulation of a
recursive $O\left( n\right) $ inverse dynamics algorithm, using the same
geometric description as in appendix \ref{AppendixEOMTree}, was reported in 
\cite{ICRA2017,RAL2020}. The Matlab implementation can be found as media
attachment to \cite{RAL2020} and in the GitHub repository \texttt{%
shivesh1210/nth\_order\_eom\_time\_derivatives}.

%TCIMACRO{\TeXButton{B}{\begin{table}[tbp] \centering}}%
%BeginExpansion
\begin{table}[tbp] \centering%
%EndExpansion
\begin{tabular}{|l|c|l|l|c|}
\hline
\textbf{Exp.} & \textbf{\# Nodes} & \textbf{Equations} & \textbf{Computation
Results} & \textbf{Eval. time in }$\mu s$ \\ \hline
1 & 1 & IK (\ref{qetaIteration}) \& ID (\ref{InvDyn}) & $%
%TCIMACRO{\TeXButton{red}{}}%
%BeginExpansion
%
%EndExpansion
%TCIMACRO{\TeXButton{vartheta}{\mathbold{\vartheta}}}%
%BeginExpansion
\mathbold{\vartheta}%
%EndExpansion
%TCIMACRO{\TeXButton{black}{\color{black}}}%
%BeginExpansion
\color{black}%
%EndExpansion
\left( t\right) ,\mathbf{u}\left( t\right) $ & 2.74 \\ 
2 & 1 & IK (\ref{qetaIteration}) for all limbs \& EOM (\ref{EOMTask}) & $%
%TCIMACRO{\TeXButton{red}{}}%
%BeginExpansion
%
%EndExpansion
%TCIMACRO{\TeXButton{vartheta}{\mathbold{\vartheta}}}%
%BeginExpansion
\mathbold{\vartheta}%
%EndExpansion
%TCIMACRO{\TeXButton{black}{\color{black}}}%
%BeginExpansion
\color{black}%
%EndExpansion
\left( t\right) ,\mathbf{F}_{\left( l\right) },l=1,2,3$, $\varphi _{\mathrm{t%
}}-\mathbf{W}_{\mathrm{t}}^{\mathrm{EE}}$ in (\ref{InvDyn}) & 2.75 \\ 
3 & 3 & IK (\ref{qetaIteration}) \& (\ref{Fphil}) of all limbs in parallel & 
$%
%TCIMACRO{\TeXButton{red}{}}%
%BeginExpansion
%
%EndExpansion
%TCIMACRO{\TeXButton{vartheta}{\mathbold{\vartheta}}}%
%BeginExpansion
\mathbold{\vartheta}%
%EndExpansion
%TCIMACRO{\TeXButton{black}{\color{black}}}%
%BeginExpansion
\color{black}%
%EndExpansion
\left( t\right) ,\mathbf{F}_{\left( l\right) },\bar{\mathbf{F}}_{\left(
l\right) }^{T}\bar{\mathbf{H}}_{\left( l\right) }^{T}\varphi _{\left(
l\right) },l=1,2,3$ & 0.97 \\ 
3.1 & 3 & IK (\ref{qetaIteration}) \& (\ref{Fphil}) for limb $l=1$ & $%
%TCIMACRO{\TeXButton{red}{}}%
%BeginExpansion
%
%EndExpansion
%TCIMACRO{\TeXButton{vartheta}{\mathbold{\vartheta}}}%
%BeginExpansion
\mathbold{\vartheta}%
%EndExpansion
%TCIMACRO{\TeXButton{black}{\color{black}}}%
%BeginExpansion
\color{black}%
%EndExpansion
_{\left( 1\right) }\left( t\right) ,\mathbf{F}_{\left( 1\right) },\bar{%
\mathbf{F}}_{\left( 1\right) }^{T}\bar{\mathbf{H}}_{\left( 1\right)
}^{T}\varphi _{\left( 1\right) }$ & 0.97 \\ 
3.2 & 3 & IK (\ref{qetaIteration}) \& (\ref{Fphil}) for limb $l=2$ & $%
%TCIMACRO{\TeXButton{red}{}}%
%BeginExpansion
%
%EndExpansion
%TCIMACRO{\TeXButton{vartheta}{\mathbold{\vartheta}}}%
%BeginExpansion
\mathbold{\vartheta}%
%EndExpansion
%TCIMACRO{\TeXButton{black}{\color{black}}}%
%BeginExpansion
\color{black}%
%EndExpansion
_{\left( 2\right) }\left( t\right) ,\mathbf{F}_{\left( 2\right) },\bar{%
\mathbf{F}}_{\left( 2\right) }^{T}\bar{\mathbf{H}}_{\left( 2\right)
}^{T}\varphi _{\left( 2\right) }$ & 0.83 \\ 
3.3 & 3 & IK (\ref{qetaIteration}) \& (\ref{Fphil}) for limb $l=3$ & $%
%TCIMACRO{\TeXButton{red}{}}%
%BeginExpansion
%
%EndExpansion
%TCIMACRO{\TeXButton{vartheta}{\mathbold{\vartheta}}}%
%BeginExpansion
\mathbold{\vartheta}%
%EndExpansion
%TCIMACRO{\TeXButton{black}{\color{black}}}%
%BeginExpansion
\color{black}%
%EndExpansion
_{\left( 3\right) }\left( t\right) ,\mathbf{F}_{\left( 3\right) },\bar{%
\mathbf{F}}_{\left( 3\right) }^{T}\bar{\mathbf{H}}_{\left( 3\right)
}^{T}\varphi _{\left( 3\right) }$ & 0.83 \\ \hline
\end{tabular}%
\caption{Experimentally determined computation times for serial and parallel evaluation of different
expressions. Exp. 1: Total time for solving the inverse kinematics (IK) problem and for
evaluating the inverse dynamics (ID) solution (\ref{InvDyn}) on a single
processing node. Exp. 2: Time elapsed for solving the IK problem and for
evaluating the overall EOM (\ref{EOMTask}), which yields the term $\varphi _{\mathrm{t}}-\mathbf{W}_{\mathrm{t}}^{\mathrm{EE}}$ in (\ref{InvDyn}), on a single computation node. Exp.
3: Computation time needed for solving the IK problem and the term $\bar{\mathbf{F}}_{\left( l\right) }^{T}\varphi _{\left( l\right) }$ in (\ref{Fphil}) for all limbs $l=1,2,3$ in parallel (three parallel computation
nodes are used). Exp. $3.l$ shows the time consumed by node $l$ for
evaluating limb $l$.
}\label{tabTiming}%
%TCIMACRO{\TeXButton{E}{\end{table}}}%
%BeginExpansion
\end{table}%
%EndExpansion
\newpage

\section{Conclusion%
%TCIMACRO{\TeXButton{secConclusion}{\label{secConclusion}}}%
%BeginExpansion
\label{secConclusion}%
%EndExpansion
}

Utilizing the dynamic capabilities of PKM to their full extend necessitates
appropriately accurate dynamics models. Such dynamics models have been
reported in the past for PKM with simple limbs. In this paper a systematic
modeling approach for rigid body PKM with complex hybrid limbs and ideal
joints is presented so that now modular modeling approaches are available
for the full spectrum of practically relevant types of PKM. The method is
also general in terms of the formulation used for dynamics modeling the
limbs. A Lie group formulation, respectively the geometric formulation in
terms of joint screw coordinates, is used in this paper. Since PKM with
simple limbs are included as special case, the paper can also be read as a
guide to the modeling of PKM in general. The approach rests on the concept
of constraint embedding, were the loop constraints within the limbs are
resolved, so that the aggregated submechanism of a FC functions
kinematically as a compound joint contributing its own dynamics. The method
is demonstrated in detail for the 3\underline{R}R[2RR]R Delta robot. For
this example initial results on the computational performance of the
presented parallel implementation are reported, which are indicative only as
the actual performance gained by a distributed computation highly depends on
the implementation.

Any dynamics modeling starts with an appropriate kinematic model, and shall
hence account for the particular kinematic topology of PKM, which will
especially exploit the modularity of PKM. 'Classical PKM', i.e. those that
are currently used in industry, can be regarded as rigid body system
interconnected either by ideal kinematic joints or by elastic elements
giving rise to 'lumped parameter' models. This is contrasted by the recent
trend of continuum robots consisting of inherently flexible elements. These
flexible elements are usually rods or slender beams, which allows for a
application of Cosserat beam models giving rise to closed form quasistatic
inverse kinematics solutions \cite{BrysonRucker2014}. The presented modular
modeling approach will be adopted to continuum parallel robots by replacing
the representative limb with a continuum model.

\section*{Acknowledgment}

The author acknowledges support by the LCM-K2 Center within the framework of
the Austrian COMET-K2 program.

\appendix

\section{Kinematics of Lower-Pair Mechanisms with Tree-Topology%
%TCIMACRO{\TeXButton{AppendixRigMotion}{\label{AppendixRigMotion}}}%
%BeginExpansion
\label{AppendixRigMotion}%
%EndExpansion
}

A body-fixed reference frame $\mathcal{F}_{k}$ is attached at body $k$. The
pose of body $k$ relative to the IFR $\mathcal{F}_{0}$ is represented by the
homogenous transformation matrix 
\begin{equation}
\mathbf{C}_{k}=\left( 
\begin{array}{cc}
\mathbf{R}_{k} & \mathbf{r}_{k} \\ 
\mathbf{0} & 1%
\end{array}%
\right) \in SE\left( 3\right)  \label{Ck}
\end{equation}%
which describes the transformation from $\mathcal{F}_{k}$ to $\mathcal{F}%
_{0} $, where $\mathbf{R}_{k}\in SO\left( 3\right) $ is the corresponding
rotation matrix, and $\mathbf{r}_{k}\in {\mathbb{R}}^{3}$ is the position
vector of the origin of $\mathcal{F}_{k}$ measured in $\mathcal{F}_{0}$.
Coordinate vectors resolved in frame $\mathcal{F}_{k}$ are denoted with ${%
^{k}\mathbf{p}}$, where the superscript is omitted when it refers to the IFR
($k=0$). In particular, the platform pose is described by the transformation 
$\mathbf{C}_{\mathrm{p}}$ from platform frame $\mathcal{F}_{\mathrm{p}}$ to
inertia frame $\mathcal{F}_{0}$.

The fact that frame transformations form the Lie group $SE\left( 3\right) $
gives rise to Lie group formulations of kinematics and dynamics of MBS,
which have become an established approach in robotics \cite{ModernRobotics}
and are increasingly used for MBS dynamics noticing its algorithmic
equivalence to the matrix and operator algebra methods \cite%
{JainBook,Featherstone2008}. In this paper the Lie group formulation and
notation from \cite{MUBOScrew1,MUBOScrew2} is used. One important feature of
these methods is the compact and flexible description of the kinematic of
kinematic chains by means of the so-called product of exponentials (POE). In
the following, it is assumed again, for the sake of simplicity, that all
joints are 1-DOF joints, so that $n_{l}=\mathfrak{n}_{l}$ and $\mathfrak{N}%
_{l}=N_{l}$.

First consider a single kinematic chain with $n$ 1-DOF lower pair joints,
with joint variables $%
%TCIMACRO{\TeXButton{vartheta}{\mathbold{\vartheta}}}%
%BeginExpansion
\mathbold{\vartheta}%
%EndExpansion
=\left( \vartheta _{1},\ldots ,\vartheta _{n}\right) ^{T}$ (rotation angles
or translation coordinates). The configuration $\mathbf{C}_{k}\in SE\left(
3\right) $ of body $k$ of this chain is determined as%
\begin{equation}
\mathbf{C}_{k}\left( 
%TCIMACRO{\TeXButton{vartheta}{\mathbold{\vartheta}}}%
%BeginExpansion
\mathbold{\vartheta}%
%EndExpansion
\right) =f_{k}\left( \vartheta _{1},\ldots ,\vartheta _{k}\right) \mathbf{A}%
_{k}  \label{POE}
\end{equation}%
with the product of exponentials%
\begin{equation}
f_{k}(%
%TCIMACRO{\TeXButton{vartheta}{\mathbold{\vartheta}}}%
%BeginExpansion
\mathbold{\vartheta}%
%EndExpansion
)=\exp \left( \mathbf{Y}_{1}\vartheta _{1}\right) \exp \left( \mathbf{Y}%
_{2}\vartheta _{2}\right) \cdot \ldots \cdot \exp \left( \mathbf{Y}%
_{k}\vartheta _{k}\right)
\end{equation}%
where $\mathbf{A}_{k}:=\mathbf{C}_{k}\left( \mathbf{0}\right) $ is the zero
reference configuration, and ${\mathbf{Y}}_{i}$ is the screw coordinate
vector of joint $i$ in spatial representation. The latter is determined as%
\begin{equation}
{\mathbf{Y}}_{i}=\left( 
\begin{array}{c}
{\mathbf{e}}_{i} \\ 
{\mathbf{y}}_{i}\times {\mathbf{e}}_{i}+h_{i}{\mathbf{e}}_{i}%
\end{array}%
\right)  \label{XY}
\end{equation}%
where $\mathbf{e}_{i}\in {\mathbb{R}}^{3}$ is a unit vector along the joint
axis, and $\mathbf{y}_{i}\in {\mathbb{R}}^{3}$ is the vector to a point on
the axis, both resolved in IFR $\mathcal{F}_{0}$. The scalar $h_{i}\in {%
\mathbb{R}}$ is the pitch of the joint. For the particular case of revolute (%
$h=0$) and prismatic joints ($h=\infty $), the screw coordinates are%
\begin{equation}
\mathrm{Revolute:\ }\mathbf{Y}_{i}=\left( 
\begin{array}{c}
{\mathbf{e}}_{i} \\ 
{\mathbf{y}}_{i}\times {\mathbf{e}}_{i}%
\end{array}%
\right) ,\ \ \ \mathrm{Prismatic:\ }{\mathbf{Y}}_{i}=\left( 
\begin{array}{c}
\mathbf{0} \\ 
\mathbf{e}_{i}%
\end{array}%
\right) .
\end{equation}%
It is at times beneficial to represent the screw coordinates of joint $i$ in
the body-fixed frame $\mathcal{F}_{i}$ at body $i$. This is then denoted
with ${^{i}\mathbf{X}}_{i}$, and determined as%
\begin{equation}
{^{i}\mathbf{X}}_{i}=\left( 
\begin{array}{c}
{^{i}}\mathbf{e}_{i} \\ 
\ {^{i}}\mathbf{x}_{i}\times {^{i}}\mathbf{e}_{i}+{^{i}}\mathbf{e}_{i}h_{i}%
\end{array}%
\right)
\end{equation}%
with a unit vector ${^{i}}\mathbf{e}_{i}\in {\mathbb{R}}^{3}$ along the
joint axis, and ${^{i}}\mathbf{x}_{i}\in {\mathbb{R}}^{3}$ being the vector
to a point on this axis, where now both are represented in $\mathcal{F}_{i}$%
. The body-fixed and spatial representation are related by 
\begin{equation}
{\mathbf{Y}}_{i}=\mathbf{Ad}_{\mathbf{C}_{i}}{^{i}\mathbf{X}}_{i},\ \ {^{i}%
\mathbf{X}}_{i}=\mathbf{Ad}_{\mathbf{A}_{i}}^{-1}{\mathbf{Y}}_{i}.
\label{XtoY}
\end{equation}%
Now for a general tree-topology system, the ordering is defined by the
spanning tree and the configuration of body $k$ is determined by (\ref{Ckl})
in terms of the tree-joint variables $%
%TCIMACRO{\TeXButton{eta}{\mathbold{\eta}}}%
%BeginExpansion
\mathbold{\eta}%
%EndExpansion
$.

The twist of body $k$ represented in the body-fixed frame $\mathcal{F}_{k}$
is%
\begin{equation}
\mathbf{V}_{k}=\left( 
\begin{array}{c}
{^{k}}%
%TCIMACRO{\TeXButton{w}{\bm{\omega}}}%
%BeginExpansion
\bm{\omega}%
%EndExpansion
_{k} \\ 
{^{k}}\mathbf{v}_{k}%
\end{array}%
\right)  \label{VkDef}
\end{equation}%
consists of the translation velocity ${^{k}}\mathbf{v}_{k}$ and angular
velocity ${^{k}}%
%TCIMACRO{\TeXButton{w}{\bm{\omega}}}%
%BeginExpansion
\bm{\omega}%
%EndExpansion
_{k}$ of the body-fixed frame $\mathcal{F}_{k}$, relative to the world frame 
$\mathcal{F}_{0}$, both resolved in $\mathcal{F}_{k}$. It is given in terms
of the joint rates $\dot{\vartheta}_{i}$ as 
\begin{eqnarray}
\mathbf{V}_{k} &=&\mathbf{J}_{k,1}\dot{\vartheta}_{1}+\mathbf{J}_{k,}\dot{%
\vartheta}_{2}+\ldots +\mathbf{J}_{k,k-1}\dot{\vartheta}_{k-1}+\mathbf{J}%
_{k,k}\dot{\vartheta}_{k}  \label{VkJ} \\
&=&\mathbf{J}_{k}\dot{%
%TCIMACRO{\TeXButton{vartheta}{\mathbold{\vartheta}}}%
%BeginExpansion
\mathbold{\vartheta}%
%EndExpansion
}  \notag
\end{eqnarray}%
with the geometric Jacobian of body $k$ 
\begin{equation}
\mathbf{J}_{k}=\left( \mathbf{J}_{k,1},\mathbf{J}_{k,2},\cdots ,\mathbf{J}%
_{k,k-1},\mathbf{J}_{k,k},\mathbf{0},\ldots \mathbf{0}\right) .
\end{equation}%
Therein, $\mathbf{J}_{k,i}$ is the instantaneous joint screw coordinate
vector of joint $i$ represented in $\mathcal{F}_{k}$, which is given
explicitly as%
\begin{equation}
\mathbf{J}_{k,i}=\left( 
\begin{array}{c}
{^{k}}\mathbf{e}_{i} \\ 
{^{k}}\mathbf{b}_{i}\times {^{k}}\mathbf{e}_{i}+{^{k}}\mathbf{e}_{i}h_{i}%
\end{array}%
\right)  \label{Jki}
\end{equation}%
where ${^{k}}\mathbf{e}_{i}(%
%TCIMACRO{\TeXButton{vartheta}{\mathbold{\vartheta}}}%
%BeginExpansion
\mathbold{\vartheta}%
%EndExpansion
)$ is a unit vector along the axis of joint $i$ and ${^{k}}\mathbf{b}_{i}(%
%TCIMACRO{\TeXButton{vartheta}{\mathbold{\vartheta}}}%
%BeginExpansion
\mathbold{\vartheta}%
%EndExpansion
)$ is the vector to a point on that axis, both measured and resolved in
frame $\mathcal{F}_{k}$ at body $k$, and $h_{i}$ is the pitch of the joint.
An efficient way to compute $\mathbf{J}_{k,i}$, follows by observing that
they can be determined by a frame transformation of the screw coordinates ${%
^{i}\mathbf{X}}_{i}$ current configuration of $\mathcal{F}_{k}$, and can
thus be calculated as \cite{MUBOScrew1}%
\begin{equation}
\mathbf{J}_{k,i}=\mathbf{Ad}_{\mathbf{C}_{k,i}}{^{i}\mathbf{X}}%
_{i},i=1,\ldots ,k
\end{equation}%
where the adjoint matrix matrix, which described the transformation of screw
coordinates according to the frame transformation $\mathbf{C}\in SE\left(
3\right) $, is%
\begin{equation}
\mathbf{Ad}_{\mathbf{C}}=\left( 
\begin{array}{cc}
\mathbf{R} & \ \mathbf{0} \\ 
\tilde{\mathbf{r}}\mathbf{R} & \mathbf{R}%
\end{array}%
\right) ,\ \ \mathrm{with}\ \mathbf{C}=\left( 
\begin{array}{cc}
\mathbf{R} & \mathbf{r} \\ 
\mathbf{0} & 1%
\end{array}%
\right)
\end{equation}%
and $\mathbf{C}_{k,i}=\mathbf{C}_{k}^{-1}\mathbf{C}_{i}$ is the
configuration of body $i$ relative to body $k$. Now for a general
tree-topology system, the Jacobian of body $k$ is given by (\ref{Jk}), which
determines the body-fixed twist as in (\ref{Vk}) in terms of $%
%TCIMACRO{\TeXButton{eta}{\mathbold{\eta}}}%
%BeginExpansion
\mathbold{\eta}%
%EndExpansion
,\dot{%
%TCIMACRO{\TeXButton{eta}{\mathbold{\eta}}}%
%BeginExpansion
\mathbold{\eta}%
%EndExpansion
}$. The relation (\ref{Jki}) indeed applies to any joint, i.e. tree- and
cut-joints of the limb.

Consider now the tree-topology system of limb $l$ without platform
comprising $n_{l}=\mathfrak{n}_{l}$ bodies. The \emph{system twist vector}
is expressed as%
\begin{equation}
\mathsf{V}_{\left( l\right) }:=\left( 
\begin{array}{c}
\mathbf{V}_{1} \\ 
\vdots \\ 
\mathbf{V}_{n_{l}}%
\end{array}%
\right) _{%
%TCIMACRO{\TeXButton{-1ex}{\hspace{-1ex}}}%
%BeginExpansion
\hspace{-1ex}%
%EndExpansion
\left( l\right) }=\mathsf{J}_{\left( l\right) }%
%TCIMACRO{\TeXButton{red}{}}%
%BeginExpansion
%
%EndExpansion
\dot{%
%TCIMACRO{\TeXButton{vartheta}{\mathbold{\vartheta}}}%
%BeginExpansion
\mathbold{\vartheta}%
%EndExpansion
}%
%TCIMACRO{\TeXButton{black}{\color{black}}}%
%BeginExpansion
\color{black}%
%EndExpansion
_{\left( l\right) \left( l\right) }
\end{equation}%
in terms of the \emph{geometric system Jacobian} 
\begin{equation}
\mathsf{J}_{\left( l\right) }=\left( 
\begin{array}{c}
\mathbf{J}_{1} \\ 
\vdots \\ 
\mathbf{J}_{n_{l}}%
\end{array}%
\right) _{%
%TCIMACRO{\TeXButton{-1ex}{\hspace{-1ex}}}%
%BeginExpansion
\hspace{-1ex}%
%EndExpansion
\left( l\right) }.  \label{JbSys2}
\end{equation}%
The latter possesses the factorization%
\begin{equation}
\mathsf{J}_{\left( l\right) }=\mathsf{A_{\left( l\right) }X}_{\left(
l\right) }.  \label{JbSys}
\end{equation}%
Assuming a canonical directed spanning tree $\vec{G}_{\left( l\right) }$
(assumption \ref{assCanonicalTree}), matrix $\mathsf{A}_{\left( l\right)
}\left( 
%TCIMACRO{\TeXButton{red}{}}%
%BeginExpansion
%
%EndExpansion
%TCIMACRO{\TeXButton{vartheta}{\mathbold{\vartheta}}}%
%BeginExpansion
\mathbold{\vartheta}%
%EndExpansion
%TCIMACRO{\TeXButton{black}{\color{black}}}%
%BeginExpansion
\color{black}%
%EndExpansion
\right) $ is the block-triangular, and $\mathsf{X}_{\left( l\right) }$ is
the block-diagonal matrix%
\begin{equation}
\mathsf{A}_{\left( l\right) }=\left( 
\begin{array}{ccccc}
\mathbf{I} & \mathbf{0} & \mathbf{0} & \cdots & \mathbf{0} \\ 
& \mathbf{I} & \mathbf{0} & \cdots & \mathbf{0} \\ 
&  &  & \ddots & \vdots \\ 
& \mathbf{Ad}_{\mathbf{C}_{i,j}} &  &  & \mathbf{0} \\ 
&  &  &  & \mathbf{I}%
\end{array}%
\right) _{%
%TCIMACRO{\TeXButton{-1ex}{\hspace{-1ex}}}%
%BeginExpansion
\hspace{-1ex}%
%EndExpansion
\left( l\right) },\ \ \mathsf{X}_{\left( l\right) }=\left( 
\begin{array}{ccccc}
{^{1}\mathbf{X}}_{1} & \mathbf{0} & \mathbf{0} &  & \mathbf{0} \\ 
\mathbf{0} & {^{2}\mathbf{X}}_{2} & \mathbf{0} & \cdots & \mathbf{0} \\ 
\mathbf{0} & \mathbf{0} & {^{3}\mathbf{X}}_{3} &  & \mathbf{0} \\ 
\vdots & \vdots & \ddots & \ddots &  \\ 
\mathbf{0} & \mathbf{0} & \cdots & \mathbf{0} & {^{n_{l}}}\mathbf{X}_{n_{l}}%
\end{array}%
\right) _{%
%TCIMACRO{\TeXButton{-1ex}{\hspace{-1ex}}}%
%BeginExpansion
\hspace{-1ex}%
%EndExpansion
\left( l\right) }  \label{Ab}
\end{equation}%
where $\mathbf{Ad}_{\mathbf{C}_{i,j}}=\mathbf{0}$ if $j\npreceq i$ (body $j$
is not a predecessor of body $i$), and $i-1$ is the predecessor relation
relative to the root-directed tree.

The derivatives of the Jacobian are frequently needed, e.g. for acceleration
forward/inverse kinematics. The partial derivatives and the time derivative
of the columns of $\mathbf{J}_{i}$, i.e. the instantaneous joint screw
coordinates in body-fixed representation, can be expressed in closed form by
simple vector operation. The non-zero terms are \cite%
{MUBOScrew2,Mueller_MMT2019}%
\begin{eqnarray}
\frac{\partial }{\partial \vartheta _{k}}\mathbf{J}_{i,j} &=&\mathbf{ad}_{%
\mathbf{J}_{i,j}}\mathbf{J}_{i,k},~\mathrm{if}\ j\prec k\preceq i
\label{Jder} \\
\dot{\mathbf{J}}_{i,j} &=&\sum_{j\prec k\preceq i}\mathbf{ad}_{\mathbf{J}%
_{i,j}}\mathbf{J}_{i,k}\dot{\vartheta}_{k}  \label{Jdot} \\
&=&-\mathbf{ad}_{\Delta {^{i}}\mathbf{V}_{i,j}}\mathbf{J}_{i,j}  \notag
\end{eqnarray}%
where $\Delta {^{i}}\mathbf{V}_{i,j}:=\mathbf{V}_{i}-\mathbf{Ad}_{\mathbf{C}%
_{i,j}}\mathbf{V}_{j}$ is the relative twist of body $i$ and $j$ represented
in reference frame at body $i$. The time derivative of the system Jacobian
can be expressed as (omitting subscript $\left( l\right) $)%
\begin{equation}
\dot{\mathsf{J}}(%
%TCIMACRO{\TeXButton{red}{}}%
%BeginExpansion
%
%EndExpansion
%TCIMACRO{\TeXButton{vartheta}{\mathbold{\vartheta}}}%
%BeginExpansion
\mathbold{\vartheta}%
%EndExpansion
%TCIMACRO{\TeXButton{black}{\color{black}}}%
%BeginExpansion
\color{black}%
%EndExpansion
,%
%TCIMACRO{\TeXButton{red}{}}%
%BeginExpansion
%
%EndExpansion
\dot{%
%TCIMACRO{\TeXButton{vartheta}{\mathbold{\vartheta}}}%
%BeginExpansion
\mathbold{\vartheta}%
%EndExpansion
}%
%TCIMACRO{\TeXButton{black}{\color{black}}}%
%BeginExpansion
\color{black}%
%EndExpansion
)=-\mathsf{A}\left( 
%TCIMACRO{\TeXButton{red}{}}%
%BeginExpansion
%
%EndExpansion
%TCIMACRO{\TeXButton{vartheta}{\mathbold{\vartheta}}}%
%BeginExpansion
\mathbold{\vartheta}%
%EndExpansion
%TCIMACRO{\TeXButton{black}{\color{black}}}%
%BeginExpansion
\color{black}%
%EndExpansion
\right) \mathsf{a(}%
%TCIMACRO{\TeXButton{red}{}}%
%BeginExpansion
%
%EndExpansion
\dot{%
%TCIMACRO{\TeXButton{vartheta}{\mathbold{\vartheta}}}%
%BeginExpansion
\mathbold{\vartheta}%
%EndExpansion
}%
%TCIMACRO{\TeXButton{black}{\color{black}}}%
%BeginExpansion
\color{black}%
%EndExpansion
)\mathsf{J}\left( 
%TCIMACRO{\TeXButton{red}{}}%
%BeginExpansion
%
%EndExpansion
%TCIMACRO{\TeXButton{vartheta}{\mathbold{\vartheta}}}%
%BeginExpansion
\mathbold{\vartheta}%
%EndExpansion
%TCIMACRO{\TeXButton{black}{\color{black}}}%
%BeginExpansion
\color{black}%
%EndExpansion
\right)
\end{equation}%
with%
\begin{equation}
\mathsf{a}(%
%TCIMACRO{\TeXButton{red}{}}%
%BeginExpansion
%
%EndExpansion
\dot{%
%TCIMACRO{\TeXButton{vartheta}{\mathbold{\vartheta}}}%
%BeginExpansion
\mathbold{\vartheta}%
%EndExpansion
}%
%TCIMACRO{\TeXButton{black}{\color{black}}}%
%BeginExpansion
\color{black}%
%EndExpansion
):=\mathrm{diag}~(\dot{\vartheta}_{1}\mathbf{ad}_{{{^{1}}\mathbf{X}_{1}}%
},\ldots ,\dot{\vartheta}_{n_{l}}\mathbf{ad}_{{{^{n}}\mathbf{X}_{n}}}).
\label{a}
\end{equation}%
The adjoint operator matrix, which produces the Lie bracket (screw product),
is%
\begin{equation}
\mathbf{ad}_{{^{i}\mathbf{X}}_{i}}=\left( 
\begin{array}{cc}
{^{i}}\bm{\tilde{\xi}}_{i} & \ \mathbf{0} \\ 
{^{i}}\tilde{\bm{\eta}}_{i} & {^{i}}\bm{\tilde{\xi}}_{i}%
\end{array}%
\right) ,\ \ \mathrm{with}\ {^{i}\mathbf{X}}_{i}=\left( 
\begin{array}{c}
{^{i}}\bm{\xi}_{i} \\ 
{^{i}}\bm{\eta}_{i}%
\end{array}%
\right)  \label{adSE3}
\end{equation}%
where $\widetilde{\mathbf{x}}\in so\left( 3\right) $ is the skew symmetric
matrix associated to vector $\mathbf{x}\in {\mathbb{R}}^{3}$.{}

\section{EOM of Tree-Topology System%
%TCIMACRO{\TeXButton{AppendixEOMTree}{\label{AppendixEOMTree}}}%
%BeginExpansion
\label{AppendixEOMTree}%
%EndExpansion
}

The dynamics of a rigid body is governed by the NE-equations. They are
expressed in compact form as the Euler-Poincar\'{e} equations of the rigid
body w.r.t. an arbitrary body-fixed reference frame $\mathcal{F}_{\mathsf{b}%
} $ 
\begin{equation}
\mathbf{M}\dot{\mathbf{V}}+\mathbf{G}\left( \mathbf{V}\right) \mathbf{MV}+%
\mathbf{W}^{\mathrm{grav}}=\mathbf{W}  \label{NEBodyFixedMatrix}
\end{equation}%
where $\mathbf{W}=(%
%TCIMACRO{\TeXButton{tau}{\bm{\tau}}}%
%BeginExpansion
\bm{\tau}%
%EndExpansion
{,}\mathbf{f})^{T}\in se^{\ast }\left( 3\right) $ is the body-fixed
representation of the wrench acting on the body. The constant $6\times 6$
inertia matrix $\mathbf{M}$ and gyroscopic matrix $\mathbf{G}$ is,
respectively,%
\begin{equation}
\mathbf{M}=\left( 
\begin{array}{cc}
%TCIMACRO{\TeXButton{Theta}{\bm{\Theta}} }%
%BeginExpansion
\bm{\Theta}
%EndExpansion
& m\widetilde{\mathbf{d}} \\ 
-m\widetilde{\mathbf{d}}\ \ \  & m\mathbf{I}%
\end{array}%
\right) ,\ \mathbf{G}\left( \mathbf{V}\right) :=-\mathbf{ad}_{\mathbf{V}%
}^{T}=\left( 
\begin{array}{cc}
\widetilde{%
%TCIMACRO{\TeXButton{w}{\bm{\omega}}}%
%BeginExpansion
\bm{\omega}%
%EndExpansion
} & \widetilde{\mathbf{v}} \\ 
\mathbf{0} & \widetilde{%
%TCIMACRO{\TeXButton{w}{\bm{\omega}}}%
%BeginExpansion
\bm{\omega}%
%EndExpansion
}%
\end{array}%
\right)
\end{equation}%
with the body-fixed inertia tensor $%
%TCIMACRO{\TeXButton{Theta}{\bm{\Theta}}}%
%BeginExpansion
\bm{\Theta}%
%EndExpansion
$ w.r.t. $\mathcal{F}_{\mathsf{b}}$, and $\mathbf{d}$ is the position vector
of the COM represented in $\mathcal{F}_{\mathsf{b}}$. The gravity wrench is
given by

\begin{equation}
\mathbf{W}_{\mathrm{p}}^{\mathrm{grav}}=-\mathbf{M}_{\mathrm{p}}\mathbf{Ad}_{%
\mathbf{C}_{\mathrm{p}}}^{-1}\left( 
\begin{array}{c}
\mathbf{0} \\ 
{\mathbf{g}}%
\end{array}%
\right)
\end{equation}%
where ${\mathbf{g}}$ is the gravity vector resolved in the IFR $\mathcal{F}%
_{0}$. The classical separated form of the NE-equations is obtained when
writing (\ref{NEBodyFixedMatrix}) separately for translation and rotation

\begin{eqnarray}
%TCIMACRO{\TeXButton{Theta}{\bm{\Theta}}}%
%BeginExpansion
\bm{\Theta}%
%EndExpansion
\dot{%
%TCIMACRO{\TeXButton{w}{\bm{\omega}}}%
%BeginExpansion
\bm{\omega}%
%EndExpansion
}+\widetilde{%
%TCIMACRO{\TeXButton{w}{\bm{\omega}}}%
%BeginExpansion
\bm{\omega}%
%EndExpansion
}%
%TCIMACRO{\TeXButton{Theta}{\bm{\Theta}}}%
%BeginExpansion
\bm{\Theta}%
%EndExpansion
%TCIMACRO{\TeXButton{w}{\bm{\omega}}}%
%BeginExpansion
\bm{\omega}%
%EndExpansion
+m\widetilde{\mathbf{d}}(\dot{\mathbf{v}}{+}\widetilde{%
%TCIMACRO{\TeXButton{w}{\bm{\omega}}}%
%BeginExpansion
\bm{\omega}%
%EndExpansion
}\mathbf{v}) &=&%
%TCIMACRO{\TeXButton{tau}{\bm{\tau}} }%
%BeginExpansion
\bm{\tau}
%EndExpansion
\notag \\
m%
%TCIMACRO{\TeXButton{big}{\big}}%
%BeginExpansion
\big%
%EndExpansion
(\dot{\mathbf{v}}+\widetilde{%
%TCIMACRO{\TeXButton{w}{\bm{\omega}}}%
%BeginExpansion
\bm{\omega}%
%EndExpansion
}\mathbf{v}+(\dot{\widetilde{%
%TCIMACRO{\TeXButton{w}{\bm{\omega}}}%
%BeginExpansion
\bm{\omega}%
%EndExpansion
}}+\widetilde{%
%TCIMACRO{\TeXButton{w}{\bm{\omega}}}%
%BeginExpansion
\bm{\omega}%
%EndExpansion
}\widetilde{%
%TCIMACRO{\TeXButton{w}{\bm{\omega}}}%
%BeginExpansion
\bm{\omega}%
%EndExpansion
})\mathbf{d}%
%TCIMACRO{\TeXButton{big}{\big}}%
%BeginExpansion
\big%
%EndExpansion
) &=&\mathbf{f}{.}  \label{BHBodyFixedExplicit}
\end{eqnarray}%
They clearly simplify if the body-fixed reference frame is located at the
COM, i.e. if $\mathbf{d}=\mathbf{0}$.

The dynamic EOM of the tree-topology system comprising $n_{l}=\mathfrak{n}%
_{l}$ rigid bodies can be expressed in closed form as in (\ref{EOMLimb}) by
means of simple matrix operations \cite{MUBOScrew2}. Denote with $\mathbf{M}%
_{i}$ the body-fixed mass matrix of body $i$. The $n_{l}\times n_{l}$
generalized matrix $\bar{\mathbf{M}}_{\left( l\right) }$ and the matrix $%
\bar{\mathbf{C}}_{\left( l\right) }(\bar{%
%TCIMACRO{\TeXButton{eta}{\mathbold{\eta}}}%
%BeginExpansion
\mathbold{\eta}%
%EndExpansion
}_{\left( l\right) },\dot{\bar{%
%TCIMACRO{\TeXButton{eta}{\mathbold{\eta}}}%
%BeginExpansion
\mathbold{\eta}%
%EndExpansion
}}_{\left( l\right) })$, which determines the generalized Coriolis and
centrifugal forces, are%
\begin{eqnarray}
\bar{\mathbf{M}}_{\left( l\right) } &=&\mathsf{J}_{\left( l\right) }^{T}%
\mathsf{M}_{\left( l\right) }\mathsf{J}_{\left( l\right) }  \label{Ml} \\
\bar{\mathbf{C}}_{\left( l\right) } &=&-\mathsf{J}_{\left( l\right) }^{T}(%
\mathsf{M}_{\left( l\right) }\mathsf{A}_{\left( l\right) }\mathsf{a}_{\left(
l\right) }+\mathsf{b}_{\left( l\right) }^{T}\mathsf{M}_{\left( l\right) })%
\mathsf{J}_{\left( l\right) }.  \label{Coriolisl}
\end{eqnarray}%
where%
\begin{align}
\mathsf{M}_{\left( l\right) }& :=\mathrm{diag}\,(\ldots ,\mathbf{M}_{i-2},%
\mathbf{M}_{i-1},\mathbf{M}_{i},\ldots )_{\left( l\right) }  \notag \\
\mathsf{a}_{\left( l\right) }(%
%TCIMACRO{\TeXButton{red}{}}%
%BeginExpansion
%
%EndExpansion
\dot{\bar{%
%TCIMACRO{\TeXButton{vartheta}{\mathbold{\vartheta}}}%
%BeginExpansion
\mathbold{\vartheta}%
%EndExpansion
}}%
%TCIMACRO{\TeXButton{black}{\color{black}}}%
%BeginExpansion
\color{black}%
%EndExpansion
_{\left( l\right) })& :=%
%TCIMACRO{\TeXButton{red}{}}%
%BeginExpansion
%
%EndExpansion
\mathrm{diag}~(\ldots ,\dot{\vartheta}_{i-2}\mathbf{ad}_{{\mathbf{X}_{i-2}}},%
\dot{\vartheta}_{i-1}\mathbf{ad}_{{\mathbf{X}_{i-1}}},\dot{\vartheta}_{i}%
\mathbf{ad}_{{\mathbf{X}_{i}}}\ldots )_{\left( l\right) }%
%TCIMACRO{\TeXButton{black}{\color{black}} }%
%BeginExpansion
\color{black}
%EndExpansion
\\
\mathsf{b}_{\left( l\right) }(\mathsf{V}_{\left( l\right) })& :=\mathrm{diag}%
~(\ldots ,\mathbf{ad}_{\mathbf{V}_{i-2}},\mathbf{ad}_{\mathbf{V}_{i-1}},%
\mathbf{ad}_{\mathbf{V}_{i}},\ldots )_{\left( l\right) }  \notag
\end{align}%
with $\mathbf{ad}_{{\mathbf{X}_{i}}}$ in (\ref{adSE3}). The ordering of
joint variables is according to the root-directed spanning tree $\vec{G}%
_{\left( l\right) }$. The generalized gravity forces are

\begin{equation}
\mathbf{Q}_{\left( l\right) }^{\mathrm{grav}}=-\mathsf{J}_{\left( l\right)
}^{T}\mathsf{M}_{\left( l\right) }\mathsf{U}_{\left( l\right) }\left( 
\begin{array}{c}
\mathbf{0} \\ 
^{0}\mathbf{g}%
\end{array}%
\right) =-\mathsf{J}_{\left( l\right) }^{T}\left( 
\begin{array}{c}
\mathbf{M}_{1}\mathbf{Ad}_{\mathbf{C}_{1}}^{-1} \\ 
\mathbf{M}_{2}\mathbf{Ad}_{\mathbf{C}_{2}}^{-1} \\ 
\vdots \\ 
\mathbf{M}_{n_{l}}\mathbf{Ad}_{\mathbf{C}_{n_{l}}}^{-1}%
\end{array}%
\right) _{%
%TCIMACRO{\TeXButton{-1ex}{\hspace{-1ex}}}%
%BeginExpansion
\hspace{-1ex}%
%EndExpansion
\left( l\right) }\left( 
\begin{array}{c}
\mathbf{0} \\ 
^{0}\mathbf{g}%
\end{array}%
\right)  \label{Qgravl}
\end{equation}%
with%
\begin{equation}
\mathsf{U}_{\left( l\right) }\left( \mathbf{q}\right) =\left( 
\begin{array}{c}
\mathbf{Ad}_{\mathbf{C}_{1}}^{-1} \\ 
\mathbf{Ad}_{\mathbf{C}_{2}}^{-1} \\ 
\vdots \\ 
\mathbf{Ad}_{\mathbf{C}_{n_{l}}}^{-1}%
\end{array}%
\right)
\end{equation}%
Notice that the topology is entirely encoded in matrix $\mathsf{A}_{\left(
l\right) }$, and thus in $\mathsf{J}_{\left( l\right) }$ via (\ref{JbSys}).

\section{List of Symbols and Abbreviations%
%TCIMACRO{\TeXButton{secSymbols}{\label{secSymbols}}}%
%BeginExpansion
\label{secSymbols}%
%EndExpansion
}

An index $\left( l\right) $ on a matrix or vector indicates that all
elements are referring to limb $l$

IFR - Inertial Frame

RL - Representative Limb

FC - Fundamental Cycle (topologically independent loop)

EE - End-Effector

$\mathfrak{N}_{l}$ - total number of joints in limb $l$

$\mathfrak{n}_{l}$ - number of tree-joints in limb $l$

$L$ - number of limbs

$\Gamma $ - topological graph, $\Gamma _{\left( l\right) }$ - topological
graph of limb $l=1,\ldots ,L$

$\gamma _{l}$ - number of FCs of limb $l$, i.e. of $\Gamma _{\left( l\right)
}$

$\Lambda _{\lambda \left( l\right) }$ - FC $\lambda =1,\ldots ,\gamma _{l}$
of limb $l$

$N_{\lambda ,l}$ - number of joint variables associated to FC $\lambda $ of
limb $l$

$N_{l}$ - total number of joint variables in limb $l$, $N_{l}=N_{1,l}+\ldots
+N_{\gamma _{l},l}$

$N_{\mathrm{act}}$ - number of actuated joint coordinates, $N_{\mathrm{act}%
\left( l\right) }$ - number of actuated joint coordinates in limb $l$

$n$ - total number of joint variables of the tree-topology system according
to $\vec{G}$

$n_{l}$ - number of tree-joint variables of limb $l$

$\bar{n}_{l}$ - number of tree-joint variables of limb $l$ when the platform
is removed

$\bar{n}$ - total number of variables of the tree-topology system when the
platform is removed, $\bar{n}:=\bar{n}_{1}+\ldots +\bar{n}_{L}$

$\underline{k}$ - index of the last body in the path from body $k$ to the
ground, i.e. $0=\underline{k}-1$

$\underline{\lambda }$ - cut-edge of $\Lambda _{l}$. It serves as start edge
when traversing the FC.

$\bar{\lambda}$ - last edge when traversing the FC starting from $\underline{%
\lambda }$

$\delta _{0,l}$ - number of joint variables of limb $l$ that are not part of
a FC of $\Gamma _{\left( l\right) }$

$\delta _{\lambda ,l}$ - DOF of FC $\lambda $ of limb $l$ when disconnected
from the platform

$\sigma _{\left( \lambda ,l\right) }$ - entries of the fundamental cycle
matrix

$m_{\lambda ,l}$ - number of generically independent loop constraints of FC $%
\Lambda _{\lambda \left( l\right) }$ of limb $l$

$%
%TCIMACRO{\TeXButton{red}{}}%
%BeginExpansion
%
%EndExpansion
%TCIMACRO{\TeXButton{eta}{\mathbold{\eta}}}%
%BeginExpansion
\mathbold{\eta}%
%EndExpansion
%TCIMACRO{\TeXButton{black}{\color{black}}}%
%BeginExpansion
\color{black}%
%EndExpansion
_{\left( l\right) }\in {\mathbb{V}}^{N_{l}}$ - overall vector of joint
variables $\vartheta _{1},\ldots ,\vartheta _{N_{l}}$ of limb $l$ (when
connected to platform)

$%
%TCIMACRO{\TeXButton{red}{}}%
%BeginExpansion
%
%EndExpansion
%TCIMACRO{\TeXButton{vartheta}{\mathbold{\vartheta}}}%
%BeginExpansion
\mathbold{\vartheta}%
%EndExpansion
%TCIMACRO{\TeXButton{black}{\color{black}}}%
%BeginExpansion
\color{black}%
%EndExpansion
_{\left( l\right) }\in {\mathbb{V}}^{n_{l}}$ - vector of tree-joint
variables $\vartheta _{1},\ldots ,\vartheta _{n_{l}}$ of limb $l$ (when
connected to platform)

$%
%TCIMACRO{\TeXButton{red}{}}%
%BeginExpansion
%
%EndExpansion
\bar{%
%TCIMACRO{\TeXButton{vartheta}{\mathbold{\vartheta}}}%
%BeginExpansion
\mathbold{\vartheta}%
%EndExpansion
}%
%TCIMACRO{\TeXButton{black}{\color{black}}}%
%BeginExpansion
\color{black}%
%EndExpansion
_{\left( l\right) }\in {\mathbb{V}}^{\bar{n}_{l}}$ - vector of tree-joint
variables $\vartheta _{1},\ldots ,\vartheta _{\bar{n}_{l}}$ of limb $l$
(when disconnected from platform)

$%
%TCIMACRO{\TeXButton{vartheta}{\mathbold{\vartheta}}}%
%BeginExpansion
\mathbold{\vartheta}%
%EndExpansion
_{\mathrm{act}}$ - vector of actuated joint coordinates

$\mathbf{q}_{\left( \lambda ,l\right) }$ - vector of $\delta _{\lambda ,l}$
independent variables in terms of which the loop constraints for $\Lambda
_{\left( \lambda ,l\right) }$ can be expressed

$\mathbf{q}_{\left( l\right) }$ - vector of $\delta _{l}$ generalized
coordinates of limb $l$ when separated from PKM

$\mathbf{q}$ - vector of $\delta $ generalized coordinates of the PKM

$\mathbf{y}_{\left( \lambda ,l\right) }$ - vector of $m_{\lambda ,l}$
dependent joint variables of FC $\Lambda _{\left( \lambda ,l\right) }$ of
limb $l$

$\mathbf{y}_{\left( l\right) }$ - vector of $m_{l}$ dependent joint
variables of limb $l$ when separated from PKM

$\mathbf{L}_{k\left( l\right) }$ - compound geometric Jacobian of body $k$
limb $l$ so that $\mathbf{V}_{k\left( l\right) }=\mathbf{L}_{k\left(
l\right) }\dot{\mathbf{q}}_{\left( l\right) }$

$\mathbf{L}_{\mathrm{p}\left( l\right) }$ - forward kinematics Jacobian of
limb $l$ so that $\mathbf{V}_{\mathrm{p}\left( l\right) }=\mathbf{L}_{%
\mathrm{p}\left( l\right) }\dot{\mathbf{q}}_{\left( l\right) }$

$\mathbf{J}_{\mathrm{IK}}$ - inverse kinematics Jacobian of the PKM so that $%
\dot{%
%TCIMACRO{\TeXButton{vartheta}{\mathbold{\vartheta}}}%
%BeginExpansion
\mathbold{\vartheta}%
%EndExpansion
}_{\left( l\right) \mathrm{act}}=\mathbf{J}_{\mathrm{IK}}(%
%TCIMACRO{\TeXButton{eta}{\mathbold{\eta}}}%
%BeginExpansion
\mathbold{\eta}%
%EndExpansion
_{\left( l\right) })\mathbf{V}_{\mathrm{t}}$

$\psi _{\mathrm{IK}\left( l\right) }$ - inverse kinematic map of limb $l$ so
that $%
%TCIMACRO{\TeXButton{eta}{\mathbold{\eta}}}%
%BeginExpansion
\mathbold{\eta}%
%EndExpansion
_{\left( l\right) }=\psi _{\mathrm{IK}}(\mathbf{x})$

$\psi _{\mathrm{FK}}$ - forward kinematic map of mechanism so that $%
%TCIMACRO{\TeXButton{eta}{\mathbold{\eta}}}%
%BeginExpansion
\mathbold{\eta}%
%EndExpansion
=\psi _{\mathrm{FK}}(%
%TCIMACRO{\TeXButton{vartheta}{\mathbold{\vartheta}}}%
%BeginExpansion
\mathbold{\vartheta}%
%EndExpansion
_{\mathrm{act}})$

$f_{\mathrm{IK}}$ - inverse kinematic map of PKM so that $%
%TCIMACRO{\TeXButton{vartheta}{\mathbold{\vartheta}}}%
%BeginExpansion
\mathbold{\vartheta}%
%EndExpansion
_{\mathrm{act}}=f_{\mathrm{IK}}(\mathbf{x})$

$\varphi _{\left( l\right) }$ - implicit form of the ODE defining the EOM of
limb $l$. This serves as inverse dynamics solution.

$\varphi _{\mathrm{t}}$ - left-hand side of the taskspace formulation of EOM
for the PKM

\bibliographystyle{IEEEtran}
\bibliography{RobHandbook}

\end{document}